\documentclass{article}

\usepackage{hyperref}

\usepackage{caption}

\usepackage{times}
\usepackage{booktabs}

\input{math_commands.tex}
\usepackage{float}
\usepackage{url}
\usepackage{microtype}
\usepackage{graphicx}
\usepackage{grffile}
\usepackage{subfigure}
\usepackage{subfiles}
\usepackage{wrapfig}
\usepackage{verbatim}
\usepackage{booktabs} %
\usepackage[rightcaption]{sidecap}
\newcommand{\update}[1]{\textcolor{black}{#1}}

\usepackage[accepted]{icml2021}
\icmltitlerunning{On Proximal Policy Optimization's Heavy-tailed Gradients}

\begin{document}

\twocolumn[
\icmltitle{On Proximal Policy Optimization's Heavy-tailed Gradients}

\icmlsetsymbol{equal}{*}

\begin{icmlauthorlist}

\icmlauthor{Saurabh Garg}{mld}
\icmlauthor{Joshua Zhanson}{csd}
\icmlauthor{Emilio Parisotto}{mld}
\icmlauthor{Adarsh Prasad}{mld}
\icmlauthor{J. Zico Kolter}{csd}
\icmlauthor{Zachary C. Lipton}{mld}
\icmlauthor{Sivaraman Balakrishnan}{stats}
\icmlauthor{Ruslan Salakhutdinov}{mld}
\icmlauthor{Pradeep Ravikumar}{mld}
\end{icmlauthorlist}

\icmlaffiliation{mld}{Machine Learning Department, Carnegie Mellon University}
\icmlaffiliation{csd}{Computer Science Department, Carnegie Mellon University}
\icmlaffiliation{stats}{Department of Statistics and Data Science, Carnegie Mellon University}

\icmlcorrespondingauthor{Saurabh Garg}{sgarg2@andrew.cmu.edu}
\icmlkeywords{Heavy-tailed Gradients, Proximal Policy Optimization, Robust Estimation, Deep Reinforcement Learning}

\vskip 0.3in
]

\printAffiliationsAndNotice{}  %

\begin{abstract}
Modern policy gradient algorithms
such as Proximal Policy Optimization (PPO)
rely on an arsenal of heuristics,
including loss clipping and gradient clipping, 
to ensure successful learning. 
These heuristics are reminiscent of
techniques from robust statistics,
commonly used for estimation
in outlier-rich (``heavy-tailed'') regimes.
In this paper, we present a detailed empirical study 
to characterize the heavy-tailed nature 
of the gradients of the PPO surrogate reward function. 
We demonstrate that the gradients,
especially for the \emph{actor} network, 
exhibit pronounced heavy-tailedness
and that it increases as the agent's policy 
diverges from the behavioral policy
(i.e., as the agent goes further off policy).
Further examination implicates 
the likelihood ratios and advantages
in the surrogate reward 
as the main sources 
of the observed heavy-tailedness.
We then highlight issues arising 
due to the heavy-tailed nature of the gradients.
In this light, we study the effects 
of the standard PPO \emph{clipping heuristics},
demonstrating that these tricks primarily serve 
to offset heavy-tailedness in gradients. 
Thus motivated, we propose incorporating \textsc{Gmom}, 
a high-dimensional robust estimator, 
into PPO as a substitute for three clipping tricks. 
Despite requiring less hyperparameter tuning,
our method matches the performance of PPO 
(with all heuristics enabled)
on a battery of MuJoCo continuous control tasks. 

\end{abstract}

\section{Introduction}
\newcommand{\green}[1]{\textcolor[rgb]{.1,.6,.1}{#1}}
\newcommand{\blue}[1]{\textcolor{blue}{#1}}
\newcommand{\sbcomment}[1]{{\bf{{\rev{{SB --- #1}}}}}}
\newcommand{\rev}[1]{{\color{blue}{#1}}}

As Deep Reinforcement Learning (DRL) methods 
have made strides on such diverse tasks
as game playing and continuous control
\citep{berner2019dota,silver2017mastering,mnih2015human},
policy gradient methods
\citep{williams1992, sutton2000policy, mnih2016asynchronous}
have risen as a popular alternative 
to dynamic programming approaches.
Since \citet{mnih2016asynchronous}'s 
breakthrough results demonstrated 
the applicability of policy gradients in DRL,
a number of popular variants have emerged
\citep{schulman2017proximal,espeholt2018impala}. 
Proximal Policy Optimization (PPO) \citep{schulman2017proximal}---one 
of the most popular policy gradient methods---introduced 
the clipped importance sampling update,
an effective heuristic for off-policy learning.
However, while their stated motivation for clipping
draws upon trust-region enforcement, 
the updates in practice tend to deviate
from such trust regions~\citep{ilyas2018closer} 
and exhibit sensitivity to implementation details 
such as random seeds and hyperparameter choices
\citep{engstrom2019implementation}.
This brittleness characterizes not just PPO,
but policy gradient methods more generally
\citep{ilyas2018closer, henderson2017deep,henderson2018did,islam2017reproducibility},
raising a broader concern 
about our understanding of these methods.

In this work, we take a step
towards understanding the workings of PPO, 
the most prominent and widely used 
deep policy gradient method. 
Noting that the heuristics implemented in PPO 
are evocative of estimation techniques 
from robust statistics 
in \emph{outlier-rich} 
and \emph{heavy-tailed} settings,
we conjecture that the heavy-tailed distribution of gradients 
is the main obstacle addressed by these heuristics.
We perform a rigorous empirical study 
to confirm the existence
of heavy-tailedness in PPO gradients
and to investigate its causes and consequences.

Our first contribution is to analyze 
the role played by each component 
of the PPO objective in the 
heavy-tailedness of the gradients.
We observe that as training proceeds, 
gradients of both the actor and the critic 
loss grow more heavy-tailed.
Our findings show that during \emph{on-policy} gradient steps 
the advantage estimates are the primary contributors
to the heavy-tailed nature of the gradients.
Moreover, as \emph{off-policyness} increases during training
(i.e. as the behavioral and actor policy diverge),
the likelihood ratios that appear 
in the surrogate objective
exacerbate the heavy-tailedness.

Second, we highlight the consequences 
of the heavy-tailedness of PPO's gradients.
Empirically, we find 
that heavy-tailedness in likelihood ratios 
induced during off-policy training 
can be a significant factor 
causing optimization instability 
leading to low average rewards. 
Moreover, we also show that removing heavy-tailedness 
in advantage estimates can 
enable agents to achieve superior performance.
Subsequently, we demonstrate 
that the clipping heuristics
present in standard PPO implementations
(i.e., gradient clipping, actor objective clipping, 
and value loss clipping) 
significantly counteract the heavy-tailedness 
induced by off-policy training.

Finally, motivated by this analysis, 
we present an algorithm that uses 
Geometric Median-of-Means (\textsc{Gmom}),
a high-dimensional robust aggregation method 
adapted from the statistics literature. 
Without using any of the objective clipping 
or gradient clipping heuristics implemented in PPO, 
the \textsc{Gmom} algorithm nearly matches 
PPO's performance on MuJoCo~\citep{todorov2012mujoco} tasks, 
which strengthens our conjecture 
that heavy-tailedness is a critical concern 
facing policy gradient methods, 
and that the benefits of PPO's clipping heuristics
come primarily from addressing this problem.

\section{Preliminaries}\label{sec:prelim} %

We define a Markov Decision Process (MDP) 
as a tuple $(\calS, \calA, R, \gamma, P )$, 
where $\calS$ represents the set of environments states, 
$\calA$ represents the set of agent actions, 
$R : \calS \times \calA \to \real$ is the reward function, 
$\gamma$ is the discount factor, and 
$P: \calS \times \calA \times \calS \to \real$ 
is the state transition probability distribution. 
The goal in reinforcement learning is to learn a policy \update{
$\pi : \calS \times \calA \to \real_{+}$} such that 
the expected cumulative discounted reward 
(known as returns) is maximized. 
Formally,  
$\allowbreak \pi^* \defeq$ 
$\argmax_\pi \Exp_{a_t \sim \pi(\cdot \vert s_t), s_{t+1} \sim P(\cdot \vert s_t, a_t)}\left[ \sum_{t=0}^\infty \gamma^t R(s_t, a_t) \right]$.

Policy gradient methods directly %
parameterize the policy 
(also known as \emph{actor} network),
i.e.,  
they define a policy $\pi_\theta$, parameterized by $\theta$.  %
Since directly optimizing the cumulative rewards
can be challenging, 
modern policy gradient algorithms 
typically optimize a surrogate reward function
which includes a likelihood ratio 
in order to 
re-use stale (off-policy) trajectories via
importance sampling.
For example, \citet{schulman2015trust} iteratively optimize: 
\begin{align*}
    \max_{\theta_t} \, \Exp_{(s_t, a_t) \sim \pi_{\theta_{t-1}}}\left[ \frac{\pi_{\theta_t}(a_t \vert s_t)}{\pi_{\theta_{t-1}}(a_t \vert s_t)}  A_{\pi_{\theta_{t-1}}}(s_t, a_t) \right] \,, \numberthis \label{eq:ppo-no-clip}
\end{align*}
where  $A_{\pi_{\theta_{t}}} = Q_{{\theta_{t}}}(s_t,a_t) - V_{{\theta_{t}}}(s_t)$. Here, the Q-function
$Q_{{\theta_{t}}}(s,a)$ is the expected discounted reward after taking an action $a$ at state $s$ and following $\pi_{\theta_{t}}$ afterwards and $V_{{\theta_{t}}}(s)$ is the value estimate (implemented with a \emph{critic} network). 

\update{However, the surrogate is indicative 
of the true reward function 
only when $\pi_{\theta_{t}}$ and $\pi_{\theta_{t-1}}$
are close in distribution. 
Different policy gradient methods~\citep{schulman2015trust,schulman2017proximal,kakade2002natural}
attempt to enforce the closeness in different ways.  
In Natural Policy Gradients~\citep{kakade2002natural} 
and Trust Region Policy Optimization (TRPO)~\citep{schulman2015trust}, 
authors utilize a conservative policy iteration 
with an explicit divergence constraint 
which provides provable lower bounds guarantees 
on the improvements of the parameterized policy.
On the other hand, PPO~\citep{schulman2017proximal} 
implements a clipping heuristic on the likelihood ratio 
to avoid excessively large policy
updates.}
Specifically, PPO optimizes the following objective: 
 \begin{align*}
    \max_{\theta_t} \, & \Exp_{(s_t, a_t) \sim \pi_{\theta_{t-1}}} \left[ \min \left(\rho_t \hat A_{\pi_{\theta_{t-1}}}(s_t, a_t) \right. \right., \\ & \qquad \left. \left. \clip (\rho_t, 1-\epsilon, 1+\epsilon) \hat A_{\pi_{\theta_{t-1}}}(s_t, a_t) \right) \right] \,, \numberthis \label{eq:ppo}
\end{align*}
where $\rho_t \defeq \frac{\pi_{\theta_t}(a_t, s_t)}{\pi_{\theta_{t-1}}(a_t, s_t)}$ and $\text{clip}(x,1-\epsilon, 1+\eps )$ clips $x$ to stay between $1+\eps$ and $1-\eps$.  We refer to $\rho_t$ as \emph{likelihood-ratios}. \update{Due to a  minimum with the unclipped surrogate reward, the PPO objective acts as a pessimistic bound on the true surrogate reward.}
As in standard PPO implementation, we use Generalized Advantage Estimation (GAE)~\citep{schulman2015high}.
Instead of fitting the value network
via regression to target values (denoted by $V_{trg}$), via
\begin{align*}
    \min_{{\theta_t}} \, \Exp_{s_t \sim \pi_{\theta_{t-1}}} \left[ (V_{\theta_t} (s_t) - V_{trg} (s_t)) ^2 \right], \numberthis \label{eq:value-no-clip}
\end{align*} 

standard implementations fit the value network with a PPO-like objective: 
\begin{align*}
    \min_{\theta_t}& \,\Exp_{s_t \sim \pi_{\theta_{t-1}}} \max\left\{\left(V_{\theta_t}(s_t) - V_{trg} (s_t)\right)^2 ,\left(\text{clip}\left(V_{\theta_t} (s_t), \right. \right. \right. \\ 
    & \left. \left. \left. V_{\theta_{t-1}} (s_t)-\varepsilon,
	  V_{\theta_{t-1}}(s_t) + \varepsilon\right) - V_{trg}(s_t)\right)^2\right\}%
	  \numberthis 
	  \label{eq:value} \,,
\end{align*}
where $\epsilon$ is the same value used to clip 
probability ratios in PPO's loss function (\Eqref{eq:ppo}). 
PPO uses the following training procedure: 
At any iteration $t$, the agent creates
a clone of the current policy $\pi_{\theta_t}$ 
which interacts with the environment to collect rollouts $\mathcal{B}$ 
(i.e., state-action pairs $\{(s_i,a_i)\}_{i=1}^N$). 
Then the algorithm optimizes the policy $\pi_\theta$ 
and value function $V_\theta$ for a fixed $K$ gradient steps 
on the sampled data $\mathcal{B}$. 
Since at every iteration the first gradient step 
is taken on the same policy 
from which the data was sampled, 
we refer to these gradient updates as \emph{on-policy} steps. 
And as for the remaining $K-1$ steps,
the sampling policy differs from the current agent,
we refer to these updates as \emph{off-policy} steps.

Throughout the paper, we consider a stripped-down variant of PPO (denoted PPO-\textsc{NoClip}) 
that consists of policy gradient with importance weighting,
but has been simplified as follows: 
(i) no likelihood-ratio clipping (\Eqref{eq:ppo-no-clip}), i.e., no \emph{objective function clipping} ; 
(ii) value network optimized via regression 
to target values (\Eqref{eq:value-no-clip})
without \emph{value function clipping}; 
and (iii) no \emph{gradient clipping}. \update{Overall PPO-\textsc{NoClip} uses the objective summarized in App. \ref{sec:App_prelim}.  One may argue that since PPO-\textsc{NoClip} removes the clipping heuristic from PPO, the unconstrained maximization of \Eqref{eq:ppo-no-clip} may lead to excessively large policy updates. In App.~\ref{sec:App-KL}, we empirically justify the use of \Eqref{eq:ppo-no-clip} by showing that with the small learning rate used in our experiments (tuned hyperparameters in Table~\ref{table:hyperparameters}), PPO-\textsc{NoClip} maintains a KL-based trust region like PPO throughout the training.}

\subsection{Framework for estimating Heavy-Tailedness} 
\label{subsec:heavy-tailed}
We now formalize our setup
for studying the distribution of gradients. 
Throughout the paper, we use the following 
definition of the heavy-tailed property:

\begin{definition}[\citet{resnick2007heavy}] \label{def:ht}
A non-negative random variable $w$ is called \emph{heavy-tailed}
if its tail probability $\smash{F_w(t) \defeq P(w\ge t)}$ 
is asymptotically equivalent to $t^{-\alpha^*}$ as $t \to \infty$ 
for some positive number $\alpha^*$. 
Here $\alpha^*$ (known as the tail index of $w$) determines the heavy-tailedness.
\end{definition} 

For a heavy-tailed distribution with index $\alpha^*$,
its $\alpha$-th moment exists only if $\alpha < \alpha^*$,
i.e., $\Exp[w^\alpha] < \infty $ iff $\alpha < \alpha^*$.
A value of $\alpha^* = 1.0$ corresponds to a Cauchy distribution 
and $\alpha^* = \infty$ (i.e., all moments exist) 
corresponds to a Gaussian distribution. 
Intuitively, as $\alpha^*$ decreases, 
the central peak of the distribution gets higher, 
the valley before the central peak gets deeper,
and the tails get heavier. 
In other words, the lower the tail-index,
the more heavy-tailed the distribution. 
However, in the finite sample setting, 
estimating the tail index is notoriously challenging
\citep{simsekli2019tail, danielsson2016tail,hill1975simple}.

In this study, we explore three estimators 
as heuristic measures to understand heavy tails 
and non-Gaussianity of gradients (refer to \Appref{sec:App_estimators} for details): 
\emph{(i)} \emph{Alpha-index estimator} which measures alpha-index for symmetric $\alpha$-stable distributions; 
\emph{(ii)} \emph{Anderson-Darling test}~\citep{anderson1954test}
on random projections of stochastic Gradient Noise (GN) to perform Gaussianity testing~\citep{panigrahi2019non}.
To our knowledge, %
the deep learning literature has only explored these two estimators
for analyzing 
the heavy-tailed nature of gradients. 
Finally, in our work, we propose using %
\emph{(iii)} \emph{Kurtosis}. 
To quantify the heavy-tailedness relative to a normal distribution, 
we measure kurtosis (fourth standardized moment) of the gradient norms.
Given samples $\{X_i\}_{i=1}^N$, the kurtosis $\kappa$ is given by: 
    $\kappa = \frac{\sum_{i=1}^N (X_i - \bar{X})^4/N}{\left( \sum_{i=1}^N (X_i - \bar X)^2/N \right)^2} \,$
where $\bar X$ is the empirical mean of the samples. 
With a slight breach of notation,
we use kurtosis to denote $\kappa^{1/4}$.
In \Appref{sec:App_estimators}, 
we show behavior of kurtosis 
on finite samples from Gaussian and Pareto distributions.  
It is well known that for a Pareto distribution with shape $\alpha \ge 4$, the lower the tail-index (shape parameter $\alpha$) the higher the kurtosis. For $\alpha<4$, since the fourth moment is non-existent, kurtosis is infinity.    
While for Gaussian distribution, the kurtosis value is approximately $1.31$. 
In \Appref{sec:App_estimators}, we 
discuss limitations of $\alpha$-index estimator
and Anderson-Darling test 
when used as heuristics to understand 
heavy tails.  
Hence, in the main paper, we include results with 
Kurtosis and relegate results with the other estimators.

\section{Heavy-Tailedness in Policy-Gradients: A Case Study on PPO} \label{sec:analysis}
We now examine the distribution of gradients in PPO.
To start, we examine the behavior 
of gradients at only on-policy steps.
We fix the policy at the beginning of every training iteration 
and just consider the gradients for the first step
(see \Appref{sec:App_exp} for details). 
As the training proceeds, the gradients clearly become 
more heavy-tailed (Fig.~\ref{fig:intro}(a)).
To thoroughly understand this behavior and the contributing factors,
we separately analyze the contributions 
from different components in the loss function. 
We also 
separate out 
the contributions 
coming from actor and critic networks.

To decouple the behavior of na\"ive policy gradients 
from PPO optimizations, 
we consider a variant of PPO which 
we call PPO-\textsc{NoClip} 
as described in Section~\ref{sec:prelim}. 
Recall that in a nutshell
PPO-\textsc{NoClip} implements 
policy gradient with just importance sampling. 
In what follows, we perform a fine-grained 
analysis of PPO at on-policy iterations.  

\vspace{-5pt}
\subsection{Heavy-tailedness in on-policy training} \label{subsec:on-policy}

\begin{figure*}[t!] 
        \subfigure[]{\includegraphics[width=0.32\linewidth]{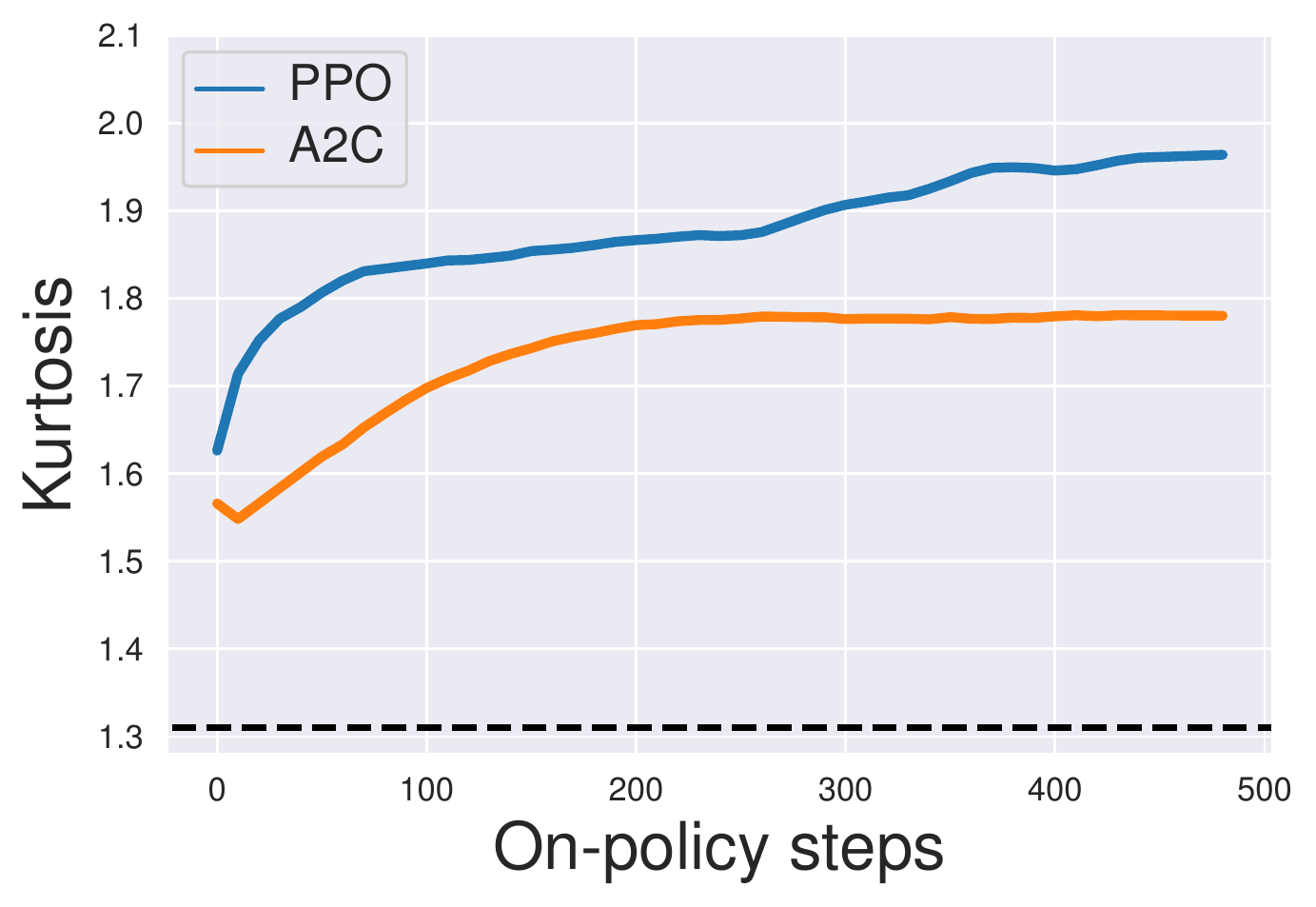}} 
        \subfigure[]{\includegraphics[width=0.32\linewidth]{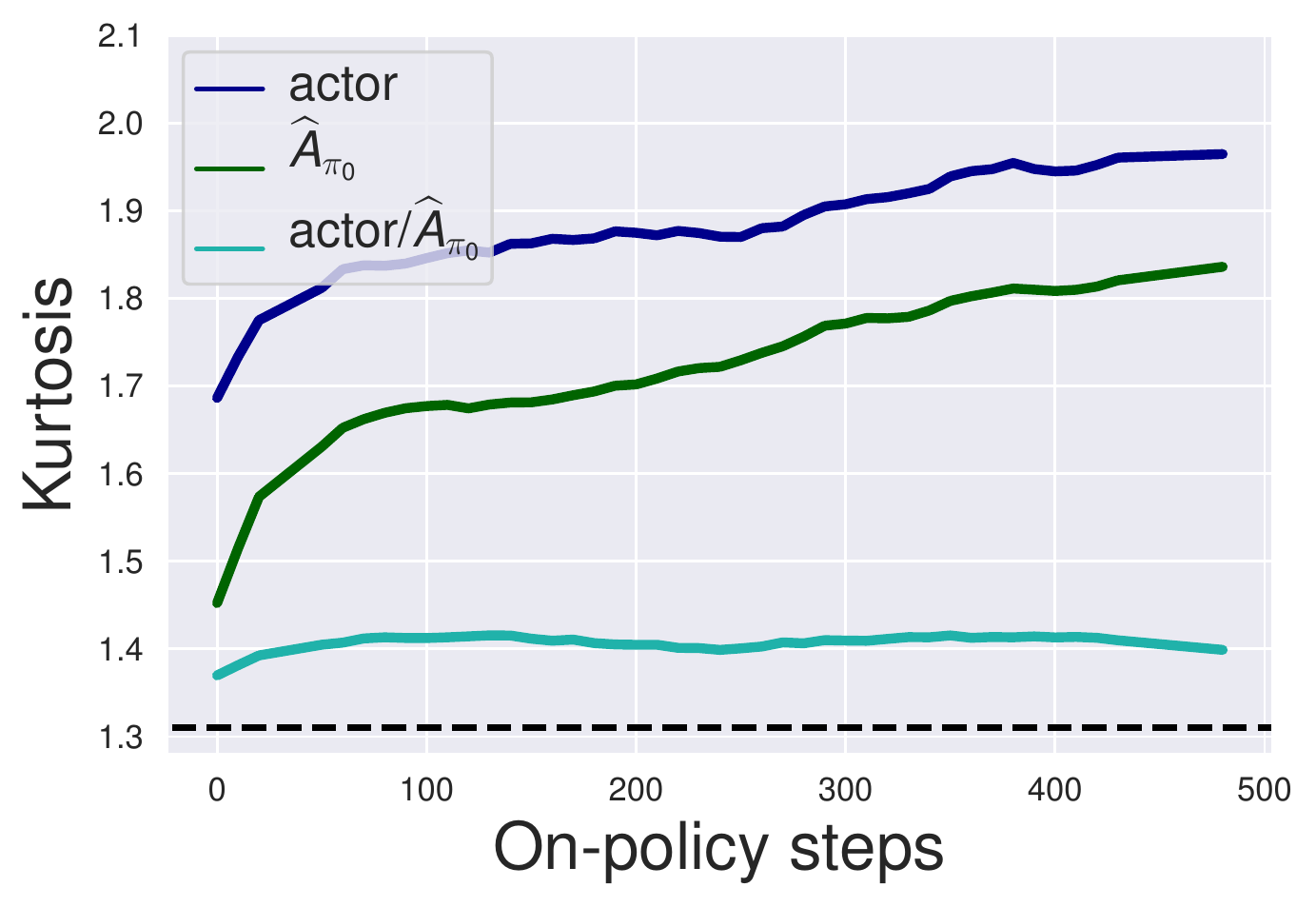}} 
        \vspace{-5pt}
        \subfigure[]{\includegraphics[width=0.32\linewidth]{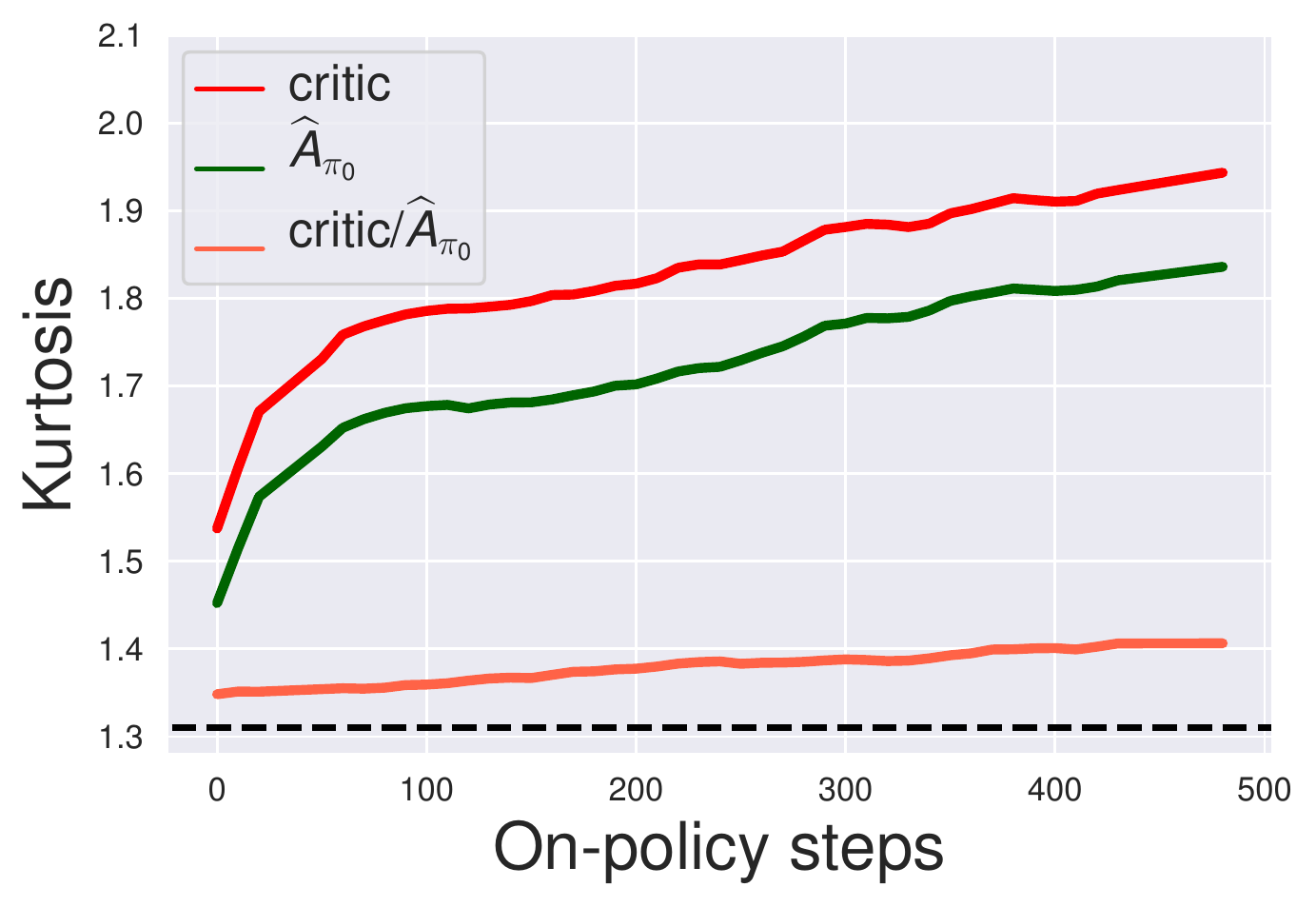}}
    \caption{ \textbf{Heavy-tailedness in PPO during on-policy iterations}. 
    All plots show mean kurtosis aggregated
    over 8 MuJoCo environments.
    For other estimators, see \Appref{Appsec:estimator}.
    For individual environments \update{with error bars},  see \Appref{Appsec:individual}. 
    Increases in Kurtosis implies an increase in heavy-tailedness. 
    Dotted line represents the Kurtosis value for a Gaussian distribution.   
    (a) Kurtosis vs on-policy iterations for A2C and PPO. 
    Evidently, as training proceeds, 
    the gradients become more heavy-tailed for both the methods. 
    (b) Kurtosis vs on-policy iterations for actor networks in PPO. 
    (c) Kurtosis vs on-policy iterations for critic networks in PPO. 
    Both critic and actor gradients become more heavy-tailed 
    as the agent is trained. 
    Note that as the gradients become more heavy-tailed, we observe a corresponding
    increase of heavy-tailedness in the advantage estimates ($\hat A_{\pi_0}$). 
    However, ``actor/$\hat A_{\pi_0}$''
    and  ``critic/$\hat A_{\pi_0}$''
    (i.e., actor or critic gradient norm 
    divided by advantage)  
    remain light-tailed throughout the training. 
    In \Appref{Appsec:negativeAdv}, we perform ablation tests 
    to highlight the reason 
    for heavy-tailed behavior of advantages. 
    }\label{fig:intro}
\end{figure*}

Given the trend of increasing heavy-tailedness in on-policy gradients, 
we first separately 
analyze the contributions 
of the actor and critic networks. 
On both these component network gradients, we observe similar trends, 
with the heavy-tailedness in the actor gradients
being marginally higher than the critic network (Fig.~\ref{fig:intro}). 
Note that during on-policy steps, since  
the likelihood-ratios are just 1, the 
gradient of actor network is given by 
$\nabla_\theta \log\left( \pi_\theta(a_t, s_t)\right)\hat A_{\pi_0}(s_t, a_t)$ 
and the gradient of the critic network is given by $\nabla_\theta  V_\theta \hat A_{\pi_0}(s_t, a_t)$ 
where $\pi_0$ is the behavioral policy. 
To explain the rising heavy-tailed behavior, 
we separately plot the advantages 
$\smash{\hat A_{\pi_0}}$ 
and the advantage divided gradients 
(i.e, $\smash{\nabla \log(\pi_\theta (a_t\vert s_t))}$ 
and $\nabla_\theta  V_\theta$).
Strikingly, we observe that while the advantage divided gradients 
are not heavy-tailed for both value and policy network,
the heavy-tailedness in advantage estimates increases as training proceeds. 
This elucidates that during on-policy updates,
outliers in advantage estimates are 
the only source of heavy-tailedness 
in actor and critic networks. 

To understand the reasons behind the observed behavior of advantages, 
we plot value estimates as computed by the critic network
and the discounted returns
used to calculate advantages
(Fig.~\ref{fig:app_adv1} in \Appref{Appsec:negativeAdv})
\update{We don't observe any discernible heavy-tailedness 
trends in value estimates 
and a slight increase in returns.} 
However, remarkably, we notice a very similar course 
of an increase in heavy-tailedness with negative advantages
(whereas positive advantages remained light-tailed) as training proceeds. 
In \Appref{Appsec:negativeAdv2}, 
we also provide evidence to this observation by showing the trends 
of increasing heavy-tailed behavior with the histograms of 
$\log(\abs{A_{\pi_{\theta}}})$ 
grouped by their sign as training proceeds 
for one MuJoCo environment (HalfCheetah-v2). 
\update{This observation highlights that, at least
in MuJoCo control environments, 
there is a positive bias of the learned value estimate 
for actions with negative advantages.}
In addition, our experiments also suggest 
that the outliers in advantages
(primarily, in negative advantages) 
are the root cause of observed heavy-tailed behavior 
in the actor and critic gradients.

We also analyze the gradients of A2C~\citep{mnih2016asynchronous}---an 
on-policy RL algorithm---and observe similar trends (Fig.~\ref{fig:intro}(a)), 
but at a relatively smaller degree of heavy-tailedness.  
Although they start at a similar magnitude, 
the heavy-tailed nature escalates 
at a higher rate in PPO\footnote{\update{In Appendix \ref{sec:App-A2C}, 
we show a corresponding trend in the heavy-tailedness of advantage estimates.}}. 
This observation may lead us to ask: 
What is the cause of heightened heavy-tailedness in PPO 
(when compared with A2C)? 
Next, we demonstrate that off-policy training 
can exacerbate the heavy-tailed behavior. 

\vspace{-5pt}
\subsection{Offpolicyness escalte heavytailness in gradients}\label{subsec:offpolicy}

\begin{figure*}[bt!] 
    \centering
        \subfigure{\includegraphics[width=0.32\linewidth]{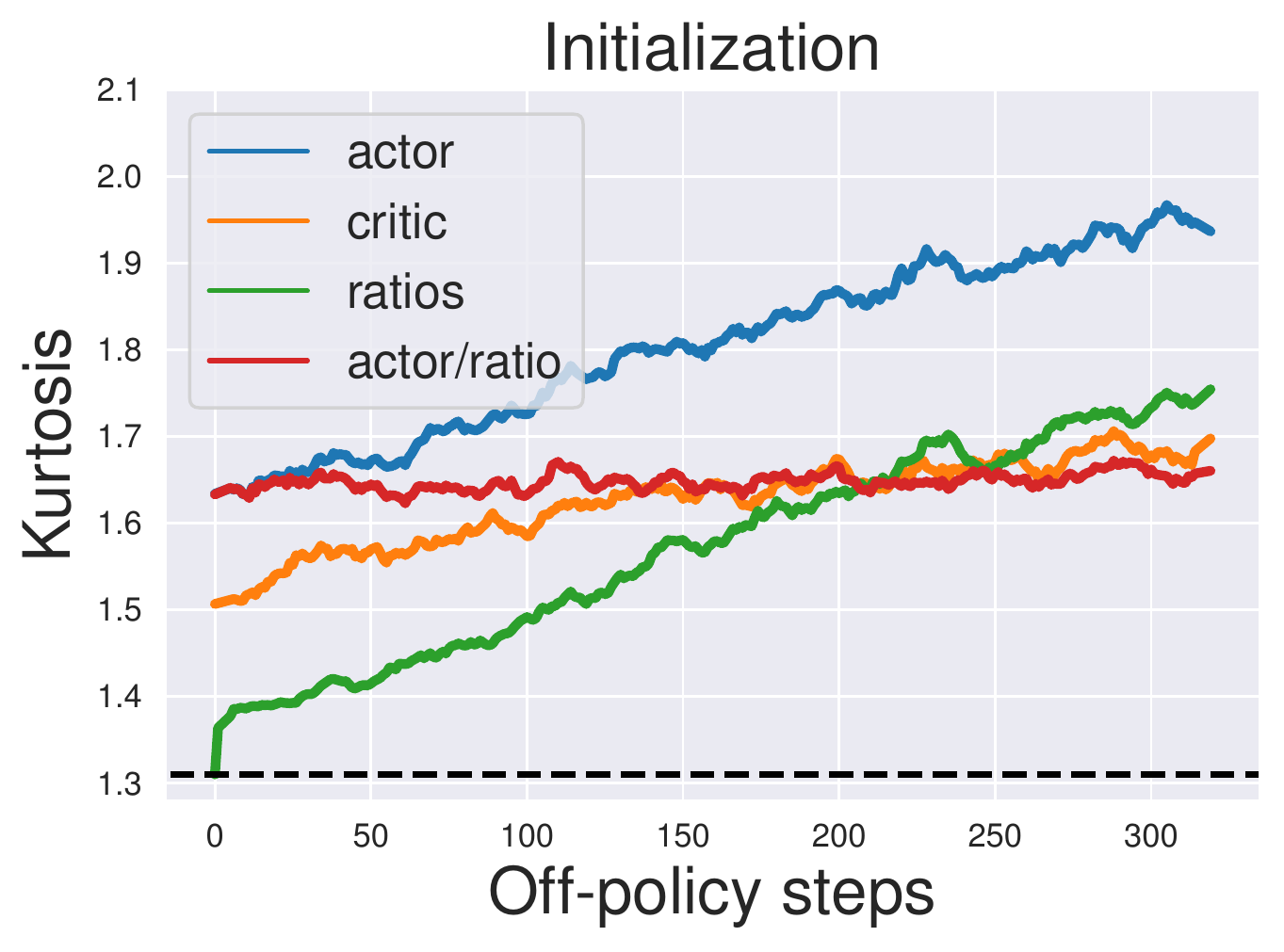}}\hfil
        \subfigure{\includegraphics[width=0.32\linewidth]{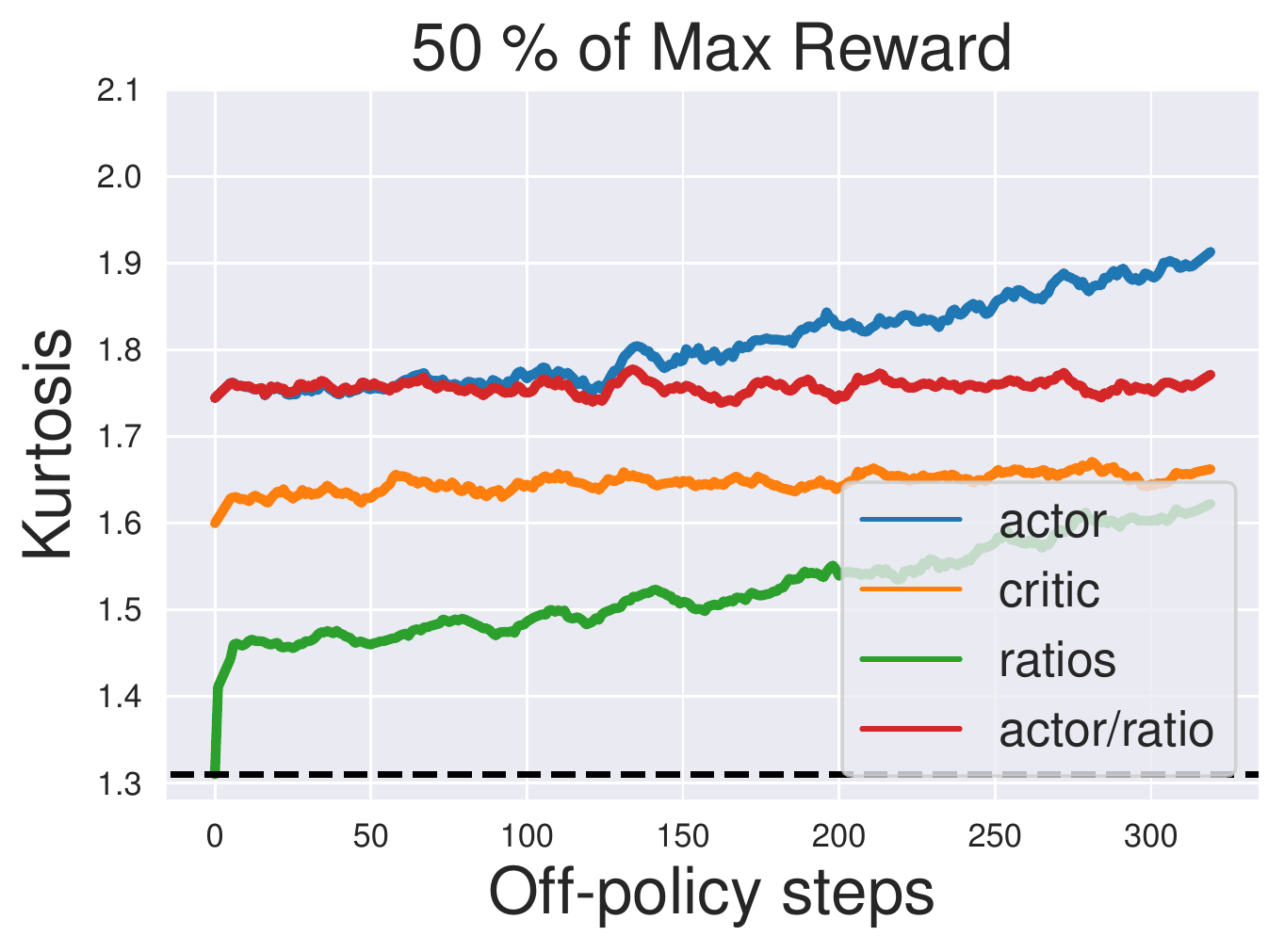}}\hfil
        \subfigure{ \includegraphics[width=0.32\linewidth]{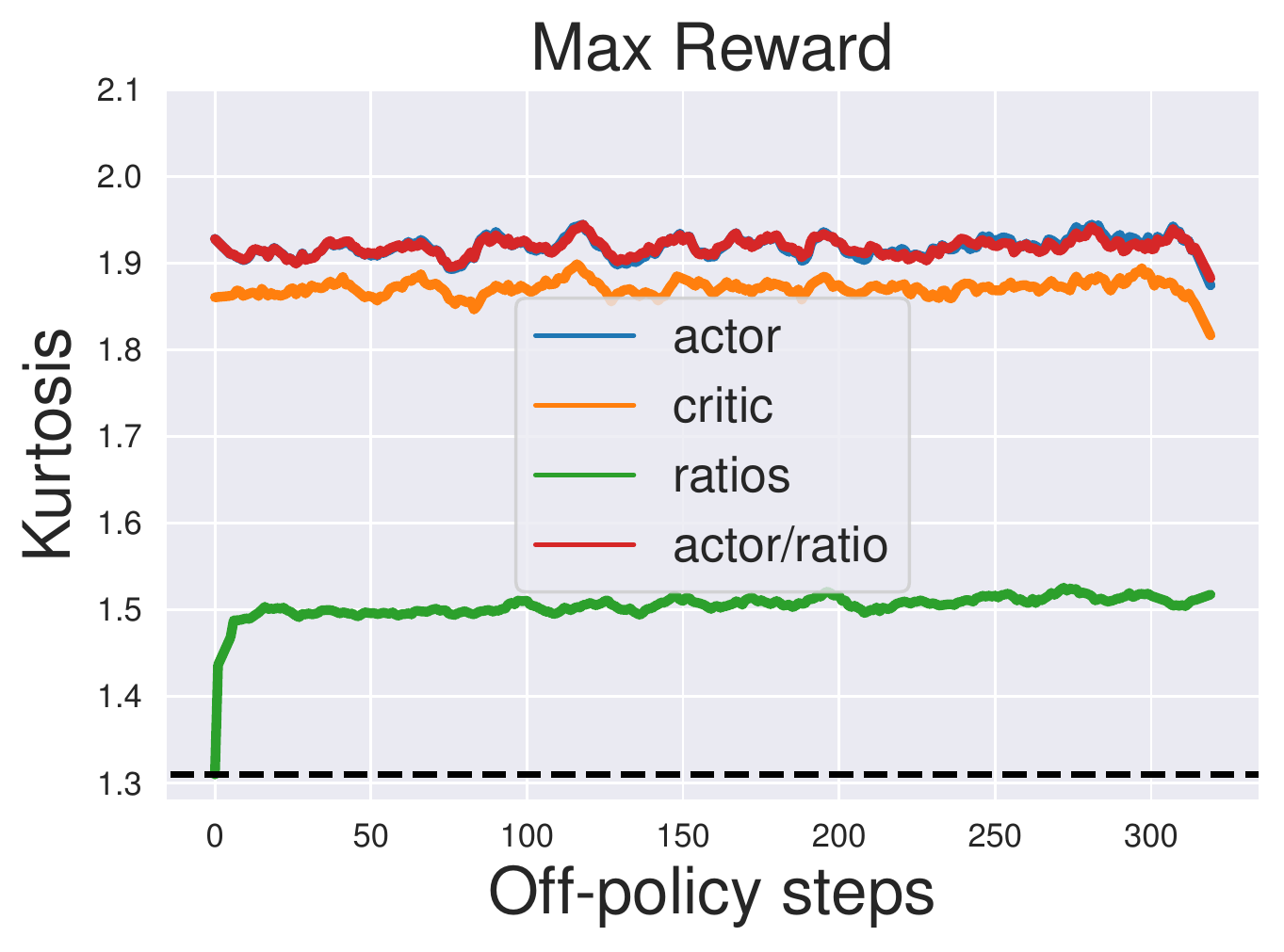}}\hfil
        \par\medskip
    \caption{ \textbf{Heavy-tailedness in PPO-\textsc{NoClip} during off-policy steps}
    at various stages of training iterations in MuJoCo environments. 
    All plots show mean kurtosis aggregated over 8 Mujoco environments.
    Plots for other estimators can be found in App.~\ref{Appsec:estimator}.  
    We also show trends with these estimators (\update{with error bars})
    on individual environments in App~\ref{Appsec:individual}. 
    Increases in Kurtosis implies an increase in heavy-tailedness.  
    Dotted line represents the Kurtosis value for a Gaussian distribution. 
    \update{Note that the analysis is done with gradients taken on a fixed batch of data within a single iteration.}  
    As off-policyness increases, the actor gradients get substantially heavy-tailed. 
    This trend is corroborated by the increase of heavy-tailedness in ratios. 
    Moreover, consistently we observe that the heavy-tailedness 
    in ``actor/ratios'' stays constant. 
    While initially during training, the heavy-tailedness 
    in the ratio's increases substantially, 
    during later stages the increase tapers off. 
    The overall increase across training iterations 
    is due to the induced heavy-tailedness in the advantage estimates 
    (cf. Sec.~\ref{subsec:on-policy}). 
     }\label{fig:ppo-offpolicy-paper} 
\end{figure*}

To analyze the gradients at off-policy steps, 
we perform the following experiment:
At various stages of training 
(i.e., at initialization, 50\% of maximum reward, and maximum reward), 
we fix the actor and the critic network 
at each gradient step during off-policy training 
and analyze the collected gradients 
(see \Appref{sec:App_exp} for details). 
First, in the early stages of training, 
as the off-policyness increases, 
the heavy-tailedness in gradients 
(both actor and critic) increases. 
However, unlike with on-policy steps, 
actor gradients 
are the major contributing factor 
to the overall heavy-tailedness 
of the gradient distribution. 
In other words, the increase in heavy-tailedness 
of actor gradients due to off-policy training
is substantially greater than for critic gradients
(Fig.~\ref{fig:ppo-offpolicy-paper}).  
Moreover, the
increase 
lessens in later stages of training 
as the agent approaches its peak performance.

Now we turn our attention to explaining
the possible causes for such a profound increase. 
The strong increase in heavy-tailedness 
of the actor gradients during off-policy training 
coincides
with a increase of heavy-tailedness in the distribution 
of likelihood ratios $\rho$, 
given by $\pi_\theta(a_t, s_t) / \pi_0(a_t, s_t)$. 
The corresponding increase in 
heavy-tailedness in ratios 
can be 
explained
theoretically.  
In continuous control RL tasks, 
the actor network often implements the policy 
with a Gaussian distribution,
where the policy parameters estimate the mean 
and the (diagonal) covariance. 
With a simple example, 
we highlight the heavy-tailed behavior 
of such likelihood-ratios of Gaussian density function. 
This example highlights how even a minor increase 
in the standard deviation of the distribution of the current policy
(as compared to behavior policy) can induce heavy-tails.    %

\textbf{Example 1}~\citep{wang2018variational}. 
Assume $\pi_1(x) = \calN\left(x; 0, \sigma_1^2 \right)$ and $\pi_2(x) = \calN\left(x; 0, \sigma_2^2\right)$. Let $\rho = \pi_1(x)/\pi_2(x)$ at a sample $x\sim \pi_2$. If $\sigma_1 \le \sigma_2$, then likelihood ratio $\rho$ is bounded and its distribution is not heavy-tailed. However, when $\sigma_1 > \sigma_2$, then $w$ has a heavy-tailed distribution with the tail-index (Definition~\ref{def:ht}) $\alpha^* = \sigma_1^2/(\sigma^2_1 - \sigma^2_2)$. 

During off-policy training, 
to understand the heavy-tailedness of actor gradients 
beyond the contributions from likelihood ratios, 
we inspect the actor gradients normalized by likelihood-ratios, i.e., 
\begin{align*}
    \frac{\nabla_\theta \pi_\theta(a_t, s_t)/ \pi_0(a_t, s_t)}{\pi_\theta(a_t, s_t)/ \pi_0(a_t, s_t)}& \hat A_{\pi_0}(s_t, a_t) = \\ & \nabla_\theta  \log\left( \pi_\theta(a_t, s_t)\right) \hat A_{\pi_0}(s_t, a_t) \,.
\end{align*}
Note that this gradient expression 
is similar to on-policy actor gradients. 
Since we observe an increasing trend in heavy-tailedness 
of the actor gradients even during on-policy training, 
one might ask: 
does these gradients' heavy-tailedness increase
during off-policy gradient updates?

Recall that in PPO, we fix the value function 
at the beginning of off-policy training 
and pre-compute advantage estimates that will later be
used throughout the training. 
Since the advantages were the primary factor 
dictating the increase during on-policy training, 
ideally, we should not observe 
any increase in the heavy-tailed behavior.
Confirming this hypothesis, 
we show that the heavy-tailedness in this quantity 
indeed stays constant during the off-policy training 
(Fig.~\ref{fig:ppo-offpolicy-paper}), i.e., 
$\nabla_\theta \log\left( \pi_\theta(a_t, s_t)\right)A_{\pi_0}(s_t, a_t)$ 
doesn't cause the increased heavy-tailed nature as long as $\pi_0$ is fixed.

Our findings from off-policy analysis
strongly suggest that when the behavioral policy is held fixed, 
heavy-tailedness in the importance ratios is the fundamental cause.  
In addition, in Sec.~\ref{subsec:on-policy}, 
we showed that when importance-ratio's are 1 
(i.e., the data on which the gradient step is taken is on-policy),
advantages induce heavy-tailedness.  
With these two observations, we conclude that the scalars 
(either the likelihood-ratios or the advantage estimates) in the 
objective 
are the primary causes 
of 
the underlying heavy-tailedness in the gradients.

\section{\update{How do Heavy-Tailed Policy-Gradients affect Training?}} \label{sec:abalation}
\begin{figure*}[t]
        \centering 
        \subfigure[]{\includegraphics[width=.32\textwidth]{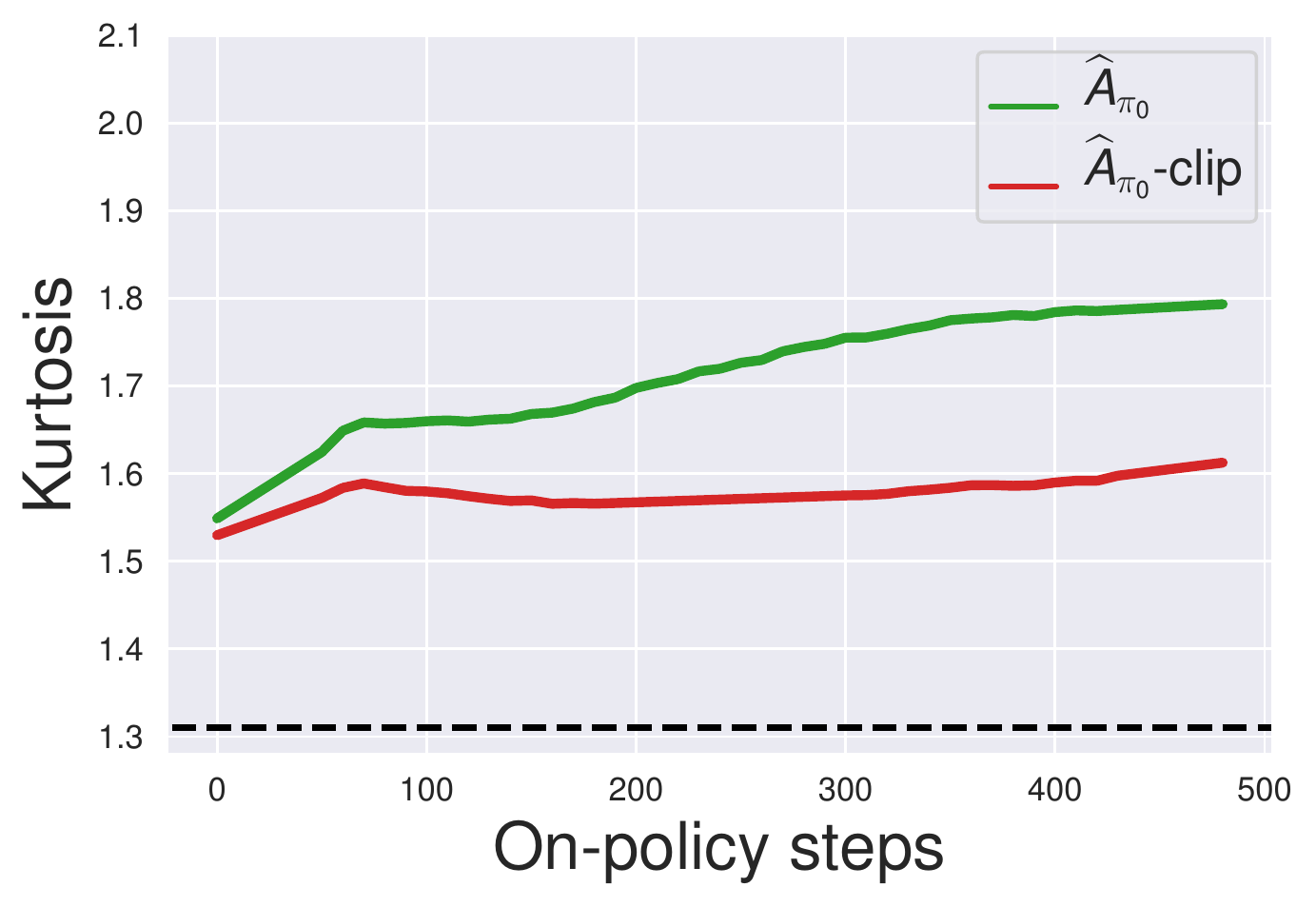} }\hfil
        \subfigure[]{\includegraphics[width=0.32\textwidth]{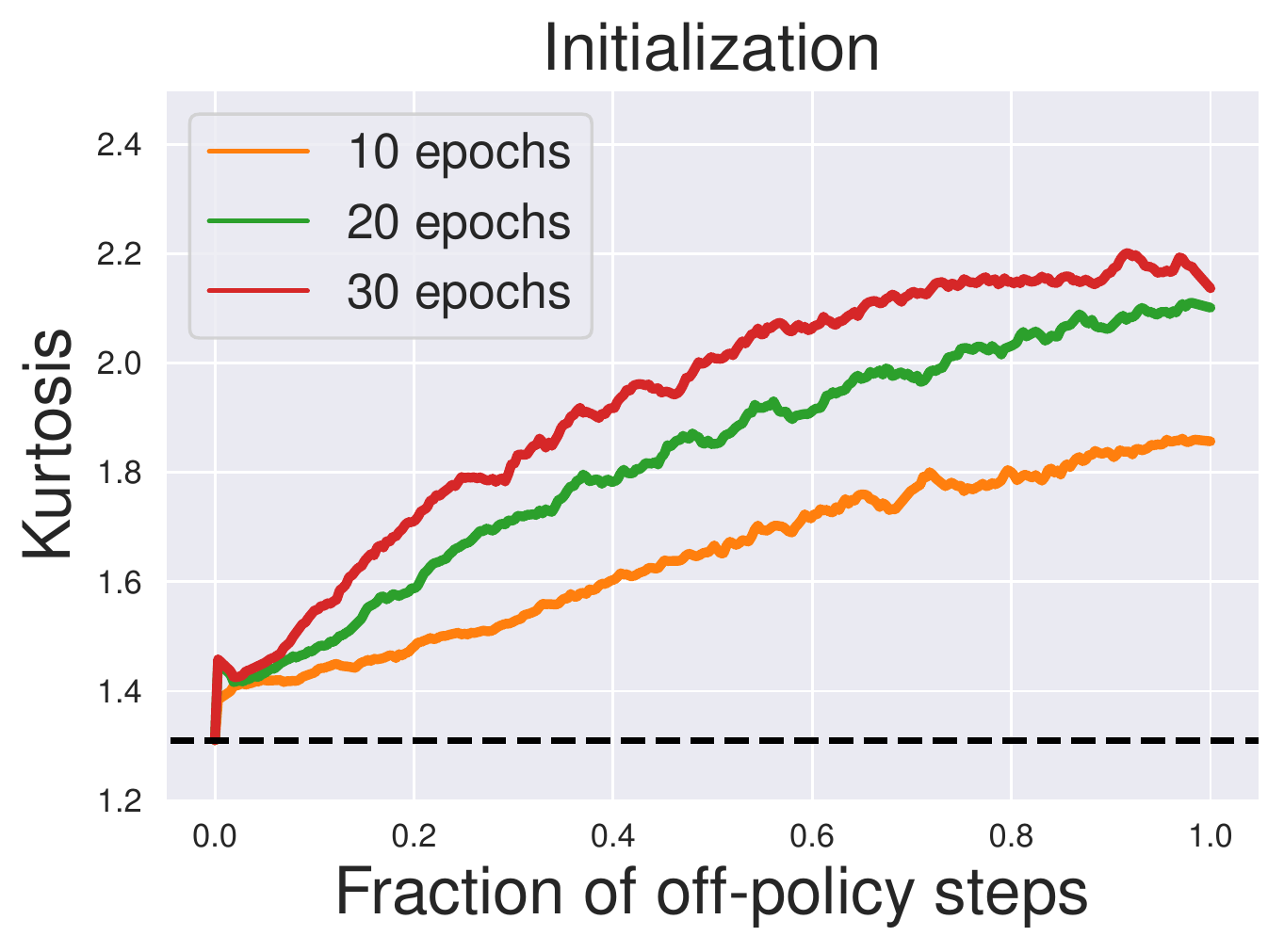}}\hfil
        \subfigure[]{\includegraphics[width=0.33\textwidth]{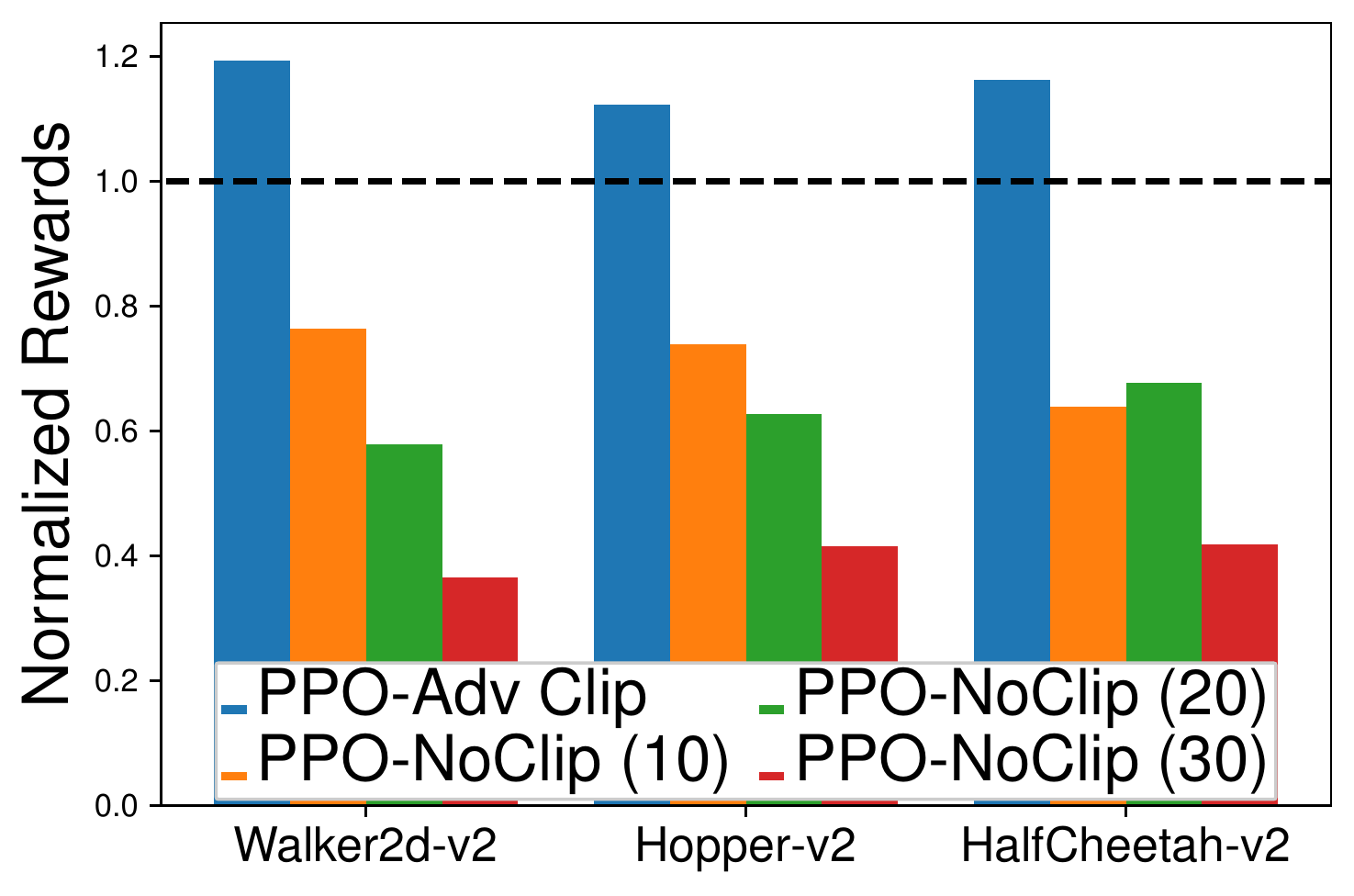}} 
        \caption{\update{ (a) \textbf{Heavy-tailedness in PPO advantages with per-environment tuned advantage clipping threshold} and (b) \textbf{Heavy-tailedness in PPO-\textsc{NoClip} likelihood-ratios as the degree of off-policyness is varied} in MuJoCo environments. All plots show mean kurtosis aggregated over 8 Mujoco environments. 
        With clipping advantages at appropriate thresholds (tuned per environment), we observe that the heavy-tailedness in advantages remains almost constant with training.
        For (b), we plot kurtosis vs the fraction of off-policy steps (i.e. number of steps taken normalized by the total number of gradients steps in one epoch). As the number of off-policy epochs increase, the heavy-tailedness in ratios increases substantially.
        (c) \textbf{Normalized rewards for PPO-AdvClip and for PPO-\textsc{NoClip} as the degree of off-policyness is varied} (number of off-policy steps in parenthesis).  
        Normalized w.r.t. the max reward obtained with PPO  (with all heuristics enabled) and performance of a random agent. 
        Evidently, as off-policy training increases, the max reward achieved drops. With advantage clipping (tuned per environment), we observe improved performance of the agent.      
        (See App~\ref{sec:App_ablation} for reward curves on individual environments.) }
        }\label{fig:affect-heavytailedness}

\end{figure*}

In the previous section, 
we investigated into the root cause 
of the heavy-tailed behaviour.  
That apparent heavy-tailed nature of PPO's gradients
may lead us to ask: 
\emph{how do heavy-tailed gradients affect agents' performance?}
In this section, we show that heavy-tailed gradients 
harm the performance of the underlying agent. 
Subsequently, we investigate into PPO heuristics and 
demonstrate how these heuristics 
alleviate for
the heavy-tailed nature of the gradient distribution.

\subsection{\update{Effect of heavy-tailedness in advantages}} \label{subsec:adv_affect}

\update{Analysis in Sec.~\ref{subsec:on-policy} shows 
that multiplicative advantage estimate 
in the PPO loss is a significant 
contributing factor to the observed heavy-tailedness. 
Motivated by this, we now study 
the impacts of \emph{clipping advantages} 
on the underlying agent. 
In particular, we clip negative advantages 
which are the primary contributors to the induced heavy-tailedness.}

\update{Depending on the observed heavy-tailedness, 
we tune a per-environment clipping threshold 
for advantages to maximize the performance of the agent 
trained with PPO. 
Intuitively, we expect that 
clipping should improve optimization 
and hence should lead to an improved performance. 
Corroborating this intuition, 
we observe significant improvements (Fig.~\ref{fig:affect-heavytailedness} (c)). 
We also plot the trend of heavy-tailedness 
in clipped advantage estimates during training. 
As we clip negative advantages below the obtained threshold, 
we observe that the induced heavy-tailedness stays constant throughout training (Fig.~\ref{fig:affect-heavytailedness} (a)).
}
\update{ Our experiment unearths an intriguing finding.
Since the advantage estimates significantly 
contribute to the observed heavy-tailed behavior, 
we show that clipping outlier advantages stabilizes the training 
and improves agents' performance on 5 out of 8 MuJoCo tasks 
(per environment rewards in App~\ref{sec:App_ablation}). 
While tuning a clipping threshold per environment may not be practical, 
the primary purpose of this study is to illustrate that heavy-tailedness 
in advantages can actually hurt the optimization process, 
and clipping advantages leads to 
improvements in the performance of agent. }

\subsection{\update{Effect of heavy-tailedness in likelihood-ratios}}
\label{subsec:ratios_affect}

In Sec.~\ref{subsec:offpolicy}, 
we demonstrated 
the heavy-tailed behavior of gradients 
during off-policy training which 
increases with off-policy gradient steps in PPO-\textsc{NoClip}. 
Moreover, we observe a corresponding increase 
in the heavy-tailedness of likelihood ratios. 
Motivated by this connection, we train agents 
with increased off-policy gradient steps to 
understand the effect of the off-policy 
induced heavy-tailedness on the performance of the agent. 
With PPO-\textsc{NoClip}, we train agents for $20$ and $30$ 
offline epochs (instead of $10$ in Table~\ref{table:hyperparameters})\footnote{
\update{Note that even with $20$ and $30$ offline epochs 
the agent maintains a KL based trust-region throughout training 
(Fig. ~\ref{fig:offpolicy-kl-reward} in App.~\ref{sec:App_ablation}). 
Beyond $30$ offline steps, 
successive policies often diverge---failing 
to maintain a KL based trust region.}}  
and analyze its performance.

\update{First, as expected, 
we observe an increase in heavy-tailedness 
in the likelihood ratios with escalated offline training 
(Fig.~\ref{fig:affect-heavytailedness}(b)). 
Moreover, the heavy-tailedness in advantages 
remains unaffected with an increase in the 
number of offline epochs (Fig.~\ref{fig:offpolicy-traininig-adv} 
in App.~\ref{sec:App_ablation}) 
confirming that the observed behavior 
is primarily due to heightened heavy-tailedness in likelihood ratios. 
We conjecture that induced heavy-tailedness
can make the optimization process harder. 
Corroborating this hypothesis, we observe that 
as the number of offline epochs increases, 
the performance of agent trained with PPO-\textsc{NoClip} deteriorates, 
and the training becomes unstable (Fig.~\ref{fig:affect-heavytailedness} (c)).} 
\update{Findings from this experiment 
clearly highlight issues due to induced heavy-tailedness
in likelihood ratios during off-policy training.
While offline training enables sample efficient training, 
restricting the number of off-policy epochs allows 
effective tackling of optimization issues induced 
due to the heavy-tailed nature 
which are beyond just trust-region enforcement.}

\vspace{-5pt}
\subsection{Explaining roles of various PPO objective optimizations}\label{subsec:explain}

Motivated from our results from the previous sections, 
we now take a deeper look at how 
the core idea of likelihood-ratio clipping 
and auxiliary optimizations implemented in PPO and understand how they  
affect the heavy-tailedness during training. 
First, we make a key observation. 
Note that the PPO-clipping heuristics 
don't get triggered for the first gradient step taken 
(when a new batch of data is sampled). 
But rather these heuristics may alter the loss only 
when behavior policy is different from the policy that is being optimized.  
Hence, in order to understand the effects of clipping heuristics, 
we perform the following analysis 
on the off-policy gradients of the PPO-\textsc{NoClip}: 
At each update step on the agent trained with PPO-\textsc{NoClip}, 
we compute the gradients while progressively 
including optimizations from the standard PPO objective.

\begin{figure}[t!] 
    \centering
       
        \subfigure{ \includegraphics[width=0.48\linewidth]{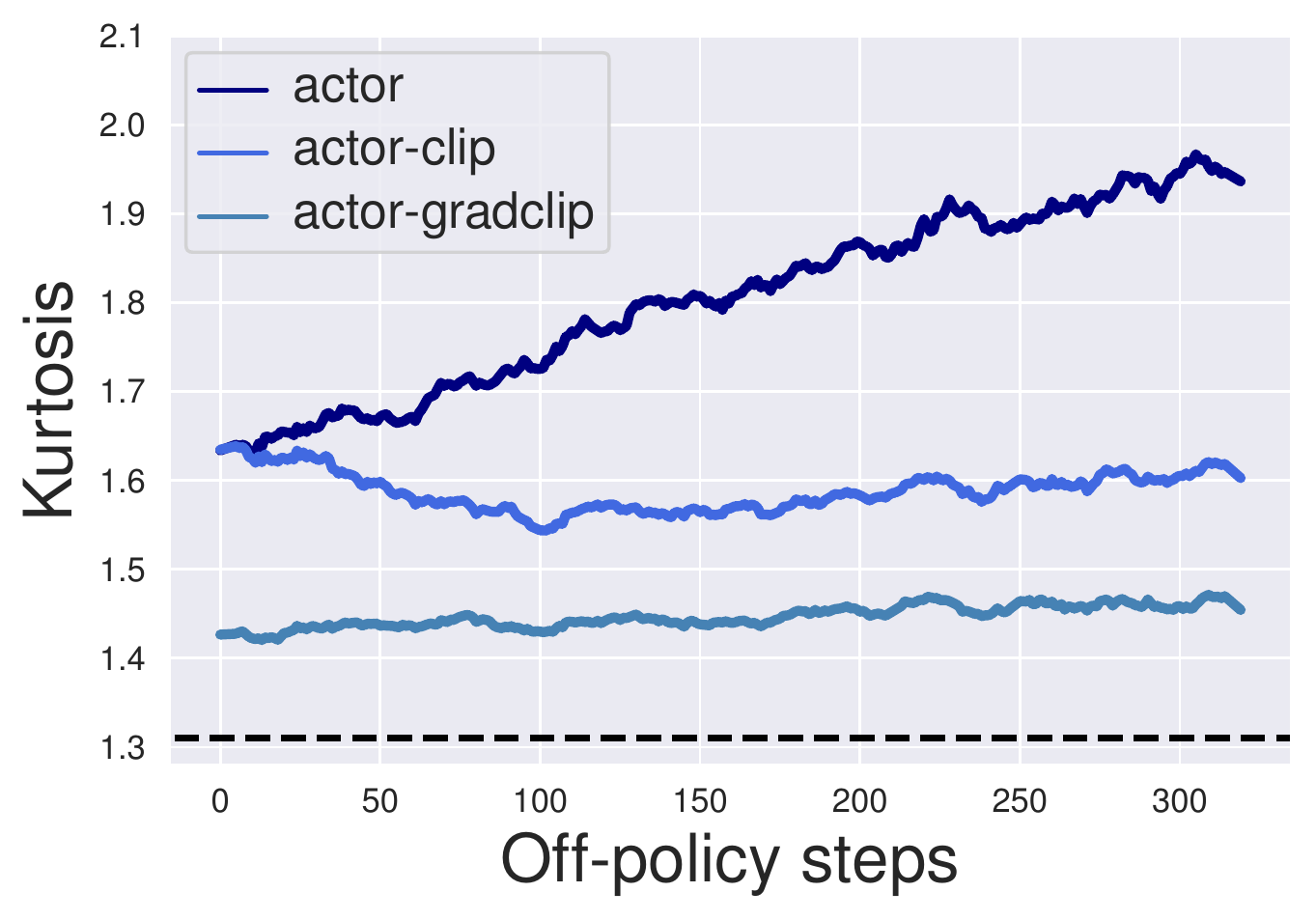}}\hfil
        \subfigure{\includegraphics[width=0.48\linewidth]{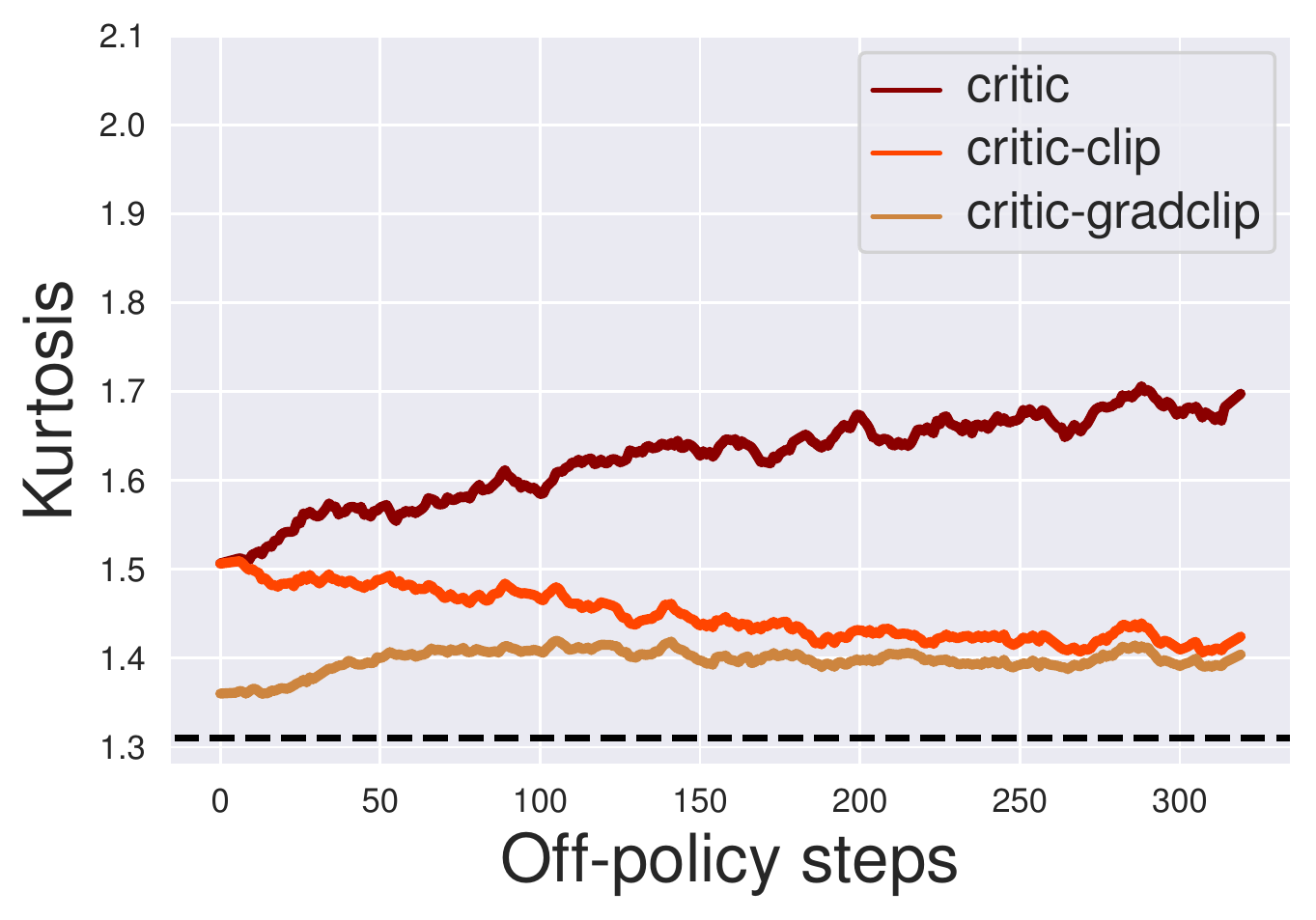}}\hfil
       
    \caption{%
        \textbf{Heavy-tailedness in PPO-\textsc{NoClip} with PPO-heuristics} applied progressively during off-policy steps,
        with kurtosis aggregated across 8 MuJoCo environments.
        For other estimators, see App.~\ref{Appsec:estimator}.  
        Dotted line represents the Kurtosis value for a Gaussian distribution.   
        ``-clip'' denotes loss clipping on corresponding networks. ``-gradclip'' denotes both gradient clipping and loss clipping.  Increases in Kurtosis implies an increase in heavy-tailedness.    As training progresses during off-policy steps, the increased heavy-tailedness in actor and critic gradients is mitigated by PPO-heuristics.
        }\label{fig:ppo-offpolicy-2} 
\end{figure}

Our results demonstrate that both the likelihood-ratio clipping 
and value-function clipping in loss during training 
offset the enormous heavy-tailedness induced due to 
off-policy training (Fig.~\ref{fig:ppo-offpolicy-2}). 
Recall that by clipping the likelihood ratios and the value function, 
the PPO objective is discarding samples 
(i.e., replacing them with zero when) 
used for gradient aggregation. 
Since heavy-tailedness in the distribution of likelihood ratios 
is the central contributing factor during off-policy training, 
by truncating likelihood-ratios $\rho_t$ 
which lie outside $(1-\epsilon, 1+\epsilon)$ interval, 
PPO is primarily mitigating heavy-tailedness in actor gradients.
Similarly, by rejecting samples from the value function loss 
which lie outside an $\epsilon$ boundary of a fixed \emph{target} estimate, 
the heuristics alleviate the slight heavy-tailed nature 
induced with off-policy training in the critic network. 

While PPO heuristics alleviate the heavy-tailedness 
induced with off-policy training, 
these heuristics don't alter heavy-tailed nature of advantage estimates. 
Since none of these heuristics directly target the outliers present 
in the advantage estimates, 
we believe that our findings can guide a development 
of fundamentally stable RL algorithms by targeting the outliers present 
in the advantage estimates 
(the primary cause of increasing heavy-tailedness throughout training).

\section{Mitigating Heavy-Tailedness with Robust Gradient Estimation} \label{sec:gmom}
Motivated by our analysis showing 
that the gradients in PPO-\textsc{NoClip} exhibit heavy-tailedness
that increases during off-policy training,
we propose an alternate method of gradient aggregation---using 
the gradient estimation framework from ~\citet{prasad2018robust}---that 
is better suited to the heavy-tailed estimation paradigm than the sample mean. 
To support our hypothesis that addressing 
the primary benefit of PPO's various clipping heuristics
lies in mitigating this heavy-tailedness,
we aim to show that equipped with our robust estimator,
PPO-\textsc{NoClip} can achieve comparable results 
to state-of-the-art PPO implementations, 
even with the clipping heuristics turned off. 

We now consider robustifying PPO-\textsc{NoClip}
(policy gradient with just importance sampling).
Informally, for gradient distributions 
which do not enjoy Gaussian-like concentration, 
the empirical-expectation-based estimates of the gradient 
do not necessarily point in the right descent direction,
leading to bad solutions. 
To this end, we leverage a robust mean aggregation technique 
called Geometric Median-Of-Means (\textsc{Gmom}) 
due to \citet{minsker2015geometric}. 
We first split the samples 
into non-overlapping sub-samples 
and estimate the sample mean of each.
The \textsc{Gmom} estimator is then given 
by the geometric median-of-means of the sub-samples.
Formally, let $\{x_1, \ldots, x_n\}\in R$ 
be $n$ i.i.d. random variables 
sampled from a distribution $\calD$. 
Then the \textsc{Gmom} estimator 
for estimating the mean 
can be described as follows: 
Partition the $n$ samples 
into b blocks $B_1, \ldots, B_b$, 
each of size $\lfloor n/b\rfloor$. 
Compute sample means in each block, 
i.e., $\{\hat \mu_1, \ldots, \mu_b\}$,
where $\hat \mu_i = \sum_{x_j\in B_i} x_j/\abs{B_i}$. 
Then the \textsc{Gmom} estimator $\hat \mu_{\textsc{Gmom}}$ 
is given by the \emph{geometric median} 
of $\{\hat \mu_1, \ldots, \mu_b \}$ defined as follows: 
    $\hat \mu_{\textsc{Gmom}} = \argmin_\mu \sum_{i=1}^b \norm{\mu - \hat \mu_i }{2}$.
We present $\gmom$ algorithm along with the Weiszfeld’s algorithm 
used for computing the approximate geometric median in App.~\ref{sec:App_GMOM}.

\begin{figure}[t!]
    \centering 
    \begin{minipage}{\linewidth}
    \begin{algorithm}[H]
        \centering
        \captionof{algorithm}{\textsc{Block-Gmom}}
        \label{alg:block-gmom}
    \begin{algorithmic}[1]
        \INPUT: Samples $S = \{x_1, \ldots, x_n\}$, number of blocks $b$, Model optimizer $\calO_G$, $b$ block optimizers $\calO_B$,  network $f_\theta$, loss $\ell$ 
        \STATE Partition S into b blocks $B_1, \ldots B_b$ of equal size.  
        \FOR{$i$ in $1\ldots b$}  
            \STATE $\hat \mu_i = \calO^{(i)}_B\left(\sum_{x_j\in B_i} \nabla_\theta \ell(f_\theta, x_j)/\abs{B_i} \right)$
        \ENDFOR
        \STATE $\hat \mu_{\gmom} = \calO_G\left(\textsc{Weiszfeld}(\hat \mu_1, \ldots, \hat \mu_b)\right)$.
        \OUTPUT: Gradient estimate $\hat \mu_{\gmom}$ 
    \end{algorithmic}
    \end{algorithm}
    \end{minipage}
\end{figure}

\textsc{Gmom} has been shown to have several favorable properties 
when used for statistical estimation in heavy-tailed settings. 
Intuitively, \textsc{Gmom} reduces the effect of outliers on a mean estimate 
by taking a intermediate mean of blocks of samples 
and then computing the geometric median of those block means. 
The robustness comes from the additional geometric median step 
where a small number of samples with large norms 
would not affect a \textsc{Gmom} estimate 
as much as they would a sample mean. 
Formally, given $n$ samples from a heavy-tailed distribution, 
the \textsc{Gmom} estimate concentrates better around the true mean 
than the sample mean which satisfies the following: 

\begin{theorem}[\citet{minsker2015geometric}]\label{thm1} 
Suppose we are given $n$ samples $\{x_i\}_{i=1}^n$ 
from a distribution with mean $\mu$ and covariance $\Sigma$. 
Assume $\delta >0$. 
Choose the number of blocks $b = 1+\lfloor 3.5 \log (1/\delta)\rfloor$. 
Then, with probability at least $1-\delta$, 
    $\norm{\mu_{\textsc{Gmom}} - \mu}{2} \lesssim \sqrt{\frac{\text{trace}(\Sigma)\log(1/\delta)}{n}}$  and
    $\norm{\frac{1}{n}\sum_{i=1}^n x_i - \mu}{2} \gtrsim \sqrt{\frac{\text{trace}(\Sigma)}{n\delta}}$.
 \end{theorem}

When applying stochastic gradient descent
or its variants
in deep learning, 
one typically
backpropagates the mean loss,
avoiding computing per-sample gradients.
However, computing $\gmom$
requires per-sample gradients. 
Consequently, we propose a simple (but novel)
variant of $\gmom$ called \textsc{Block-Gmom} 
which avoids the extra sample-size dependent computational penalty 
of calculating sample-wise gradients. 
Notice that by Theorem~\ref{thm1}, 
the number of blocks required to compute $\gmom$
is independent of the sample size 
to obtain the guarantee with high probability.  
To achieve this, instead of calculating sample-wise gradients, 
we compute block-wise gradients by backpropagating 
on sample-mean aggregated loss for each block. 
Moreover, such an implementation not only increases efficiency 
but also allows incorporating adaptive optimizers for individual blocks. 
Algorithm~\ref{alg:block-gmom} presents the overall \textsc{Block-Gmom}.

\begin{figure}[t!]
    \centering 
    \includegraphics[width=0.8\linewidth]{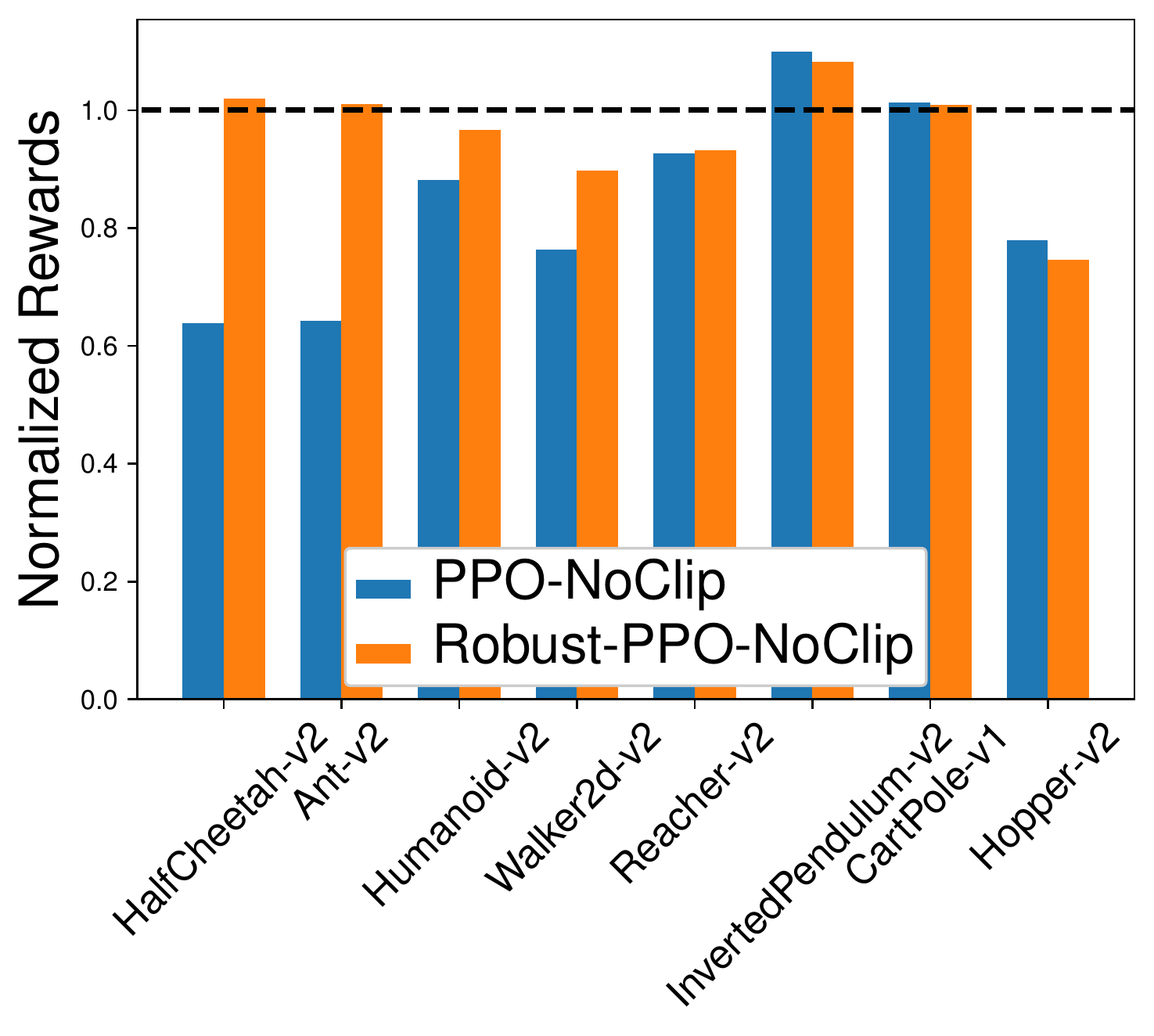}
    \vspace{-5pt}

    \caption{ \textbf{Normalized rewards for \textsc{Robust}-PPO-\textsc{NoClip} and PPO-\textsc{NoClip}}.
    Normalized w.r.t. %
    the max reward obtained with PPO 
    (with all heuristics enabled) 
    and performance of a random agent. 
    (See App~\ref{Appsec:rewards} for reward curves on individual environment.)
    }\label{fig:normalized-rewards}
\end{figure}

\vspace{-5pt}
\subsection{Results on MuJoCo environment}

We perform experiments on 8 MuJoCo~\citep{todorov2012mujoco} 
control tasks.
To use \textsc{Block-Gmom} aggregation with PPO-\textsc{NoClip}, 
we extract actor-network and critic-network gradients at each step 
and separately run the Algorithm~\ref{alg:block-gmom} on both the networks.
For our experiments, we use SGD as $\calO_B$ and Adam as $\calO_G$ 
and refer to this variant of PPO-\textsc{NoClip} 
as \textsc{Robust}-PPO-\textsc{NoClip}. 
We compare the performances of PPO, 
PPO-\textsc{NoClip}, and \textsc{Robust}-PPO-\textsc{NoClip},
using hyperparameters that are
tuned individually for each method
but held fixed across all tasks
(Table~\ref{table:hyperparameters}). 

For 7 tasks, we observe significant improvements 
with \textsc{Robust}-PPO-\textsc{NoClip} over PPO-\textsc{NoClip} 
and performance close to that achieved by PPO 
(with all clipping heuristics enabled) 
(Fig.~\ref{fig:normalized-rewards}). 
Although we do not observe improvements over PPO, 
we believe that this result corroborates our conjecture that
PPO heuristics primarily aim to offset the heavy-tailedness induced with training.

\section{Related Work}
Studying the behavior of SGD, 
\citet{simsekli2019tail} questioned 
the Gaussianity of SGD noise, 
highlighting its \emph{heavy-tailed} nature.  
Subsequently, there has been a growing interest
in understanding the nature of SGD noise
in different deep learning tasks 
with a specific focus on its influence on generalization performance 
versus induced optimization difficulties ~\citep{csimcsekli2020fractional, zhang2019adam, panigrahi2019non}. 
In particular,~\citet{zhang2019adam} studied the nature of stochastic gradients 
in natural language processing (e.g., BERT-pretraining) 
and highlighted the effectiveness of adaptive methods (e.g. Adam and gradient clipping).
Some recent work has also made progress 
towards understanding the effectiveness 
of gradient clipping in convergence~\citep{zhang2019adam,zhang2019gradient,csimcsekli2020fractional} 
in presence of heavy-tailed noise. 
On the other hand, \citet{simsekli2019tail} 
highlighted \emph{the benefits of heavy-tailed noise} 
in achieving wider minima with better generalization, by analyzing SGD as an SDE driven by Levy motion 
(whose increments are $\alpha-$stable heavy-tailed noise).

On the RL side, 
 \citet{bubeck2013bandits} studied 
the stochastic multi-armed bandit problem 
when the reward distribution is heavy-tailed. 
The authors designed a robust version 
of the classical Upper Confidence Bound algorithm 
by  replacing the empirical average 
of observed rewards 
with robust estimates obtained 
via
the univariate median-of-means estimator~\citep{nemirovski1983problem} 
on the observed sequence of rewards. 
\citet{medina2016no} extended this approach 
to the problem of linear bandits 
under heavy-tailed noise. 
There is also a long line of work 
in deep RL which focuses on 
reducing the variance 
of stochastic policy gradients~\citep{gu2016q,wu2018variance, metelli2018policy,cheng2020trajectory, metelli2020importance}.  
On the flip side, \citet{chung2020beyond} 
highlighted the beneficial impacts of stochasticity of policy gradients on the optimization process.
In simple MDPs, authors showed that larger higher moments 
with fixed variance leads to improved exploration. 
This aligns with the conjecture of \citet{simsekli2019tail} in the context of supervised learning that heavy-tailedness in gradients can improve generalization. \citet{chung2020beyond} thus pointed out the importance of a 
careful analysis of stochasticity in gradients to better understand policy gradient algorithms.

We consider our work a stepping stone 
towards analyzing stochastic gradients beyond just their variance. 
We hypothesize that in deep RL where the optimization 
process is known to be brittle 
\citep{henderson2018did,henderson2017deep,engstrom2019implementation, ilyas2018closer}, perhaps due to the flexibility of the neural representation, heavy-tailedness can cause heightened instability rather than help in efficient exploration. 
This perspective aligns with one line of work~\citep{zhang2019adam} 
where authors demonstrate that heavy-tailedness 
can cause instability in the learning process in deep models. 
Indeed with ablation experiments in Sec.~\ref{sec:abalation}, 
we show that increasing heavy-tailedness in likelihood ratios 
hurts the agent performance, and mitigating heavy-tailedness 
in advantage estimates improves the agent performance.

\section{Conclusion}
In this paper, we empirically characterized
PPO's gradients,
demonstrating that they
become more heavy-tailed as training proceeds.
Our detailed analysis
showed that at on-policy steps, 
the heavy-tailed nature of the gradients
is primarily attributable to the 
multiplicative advantage estimates. 
On the other hand, we observed that 
during off-policy training, 
the heavy-tailedness of the likelihood ratios 
of the surrogate reward function 
exacerbates the observed heavy-tailedness. 

Subsequently, we examined issues 
due to heavy-tailed nature of gradients. 
We demonstrated that PPO's clipping heuristics
primarily serve
to offset the heavy-tailedness induced 
by off-policy training. 
With this motivation, 
we showed that a robust estimation technique 
could effectively replace 
all three of PPO's clipping heuristics:
likelihood-ratio clipping, 
value loss clipping, 
and gradient clipping.

In future work, we plan to conduct 
similar analysis on gradients 
for other RL algorithms 
such as deep Q-learning.
Moreover, we believe that our findings on 
heavy-tailed nature of advantage estimates   
can significantly impact 
algorithm development 
for policy gradient algorithms.

\section*{Acknowledgements}
We acknowledge the support of Lockheed Martin, DARPA via HR00112020006, and NSF via IIS-1909816, OAC-1934584.
\bibliography{iclr2021_conference}

\begin{thebibliography}{40}
\providecommand{\natexlab}[1]{#1}
\providecommand{\url}[1]{\texttt{#1}}
\expandafter\ifx\csname urlstyle\endcsname\relax
  \providecommand{\doi}[1]{doi: #1}\else
  \providecommand{\doi}{doi: \begingroup \urlstyle{rm}\Url}\fi

\bibitem[Anderson \& Darling(1954)Anderson and Darling]{anderson1954test}
Anderson, T.~W. and Darling, D.~A.
\newblock A test of goodness of fit.
\newblock \emph{Journal of the American statistical association}, 49\penalty0
  (268):\penalty0 765--769, 1954.

\bibitem[Berner et~al.(2019)Berner, Brockman, Chan, Cheung, D{k{e}}biak,
  Dennison, Farhi, Fischer, Hashme, Hesse, et~al.]{berner2019dota}
Berner, C., Brockman, G., Chan, B., Cheung, V., D{k{e}}biak, P., Dennison, C.,
  Farhi, D., Fischer, Q., Hashme, S., Hesse, C., et~al.
\newblock Dota 2 with large scale deep reinforcement learning.
\newblock \emph{arXiv preprint arXiv:1912.06680}, 2019.

\bibitem[Bubeck et~al.(2013)Bubeck, Cesa-Bianchi, and
  Lugosi]{bubeck2013bandits}
Bubeck, S., Cesa-Bianchi, N., and Lugosi, G.
\newblock Bandits with heavy tail.
\newblock \emph{IEEE Transactions on Information Theory}, 59\penalty0
  (11):\penalty0 7711--7717, 2013.

\bibitem[Cheng et~al.(2020)Cheng, Yan, and Boots]{cheng2020trajectory}
Cheng, C.-A., Yan, X., and Boots, B.
\newblock Trajectory-wise control variates for variance reduction in policy
  gradient methods.
\newblock In \emph{Conference on Robot Learning}, pp.\  1379--1394. PMLR, 2020.

\bibitem[Chung et~al.(2020)Chung, Thomas, Machado, and Roux]{chung2020beyond}
Chung, W., Thomas, V., Machado, M.~C., and Roux, N.~L.
\newblock Beyond variance reduction: Understanding the true impact of baselines
  on policy optimization.
\newblock \emph{arXiv preprint arXiv:2008.13773}, 2020.

\bibitem[Danielsson et~al.(2016)Danielsson, Ergun, de~Haan, and
  de~Vries]{danielsson2016tail}
Danielsson, J., Ergun, L.~M., de~Haan, L., and de~Vries, C.~G.
\newblock Tail index estimation: Quantile driven threshold selection.
\newblock \emph{Available at SSRN 2717478}, 2016.

\bibitem[Engstrom et~al.(2019)Engstrom, Ilyas, Santurkar, Tsipras, Janoos,
  Rudolph, and Madry]{engstrom2019implementation}
Engstrom, L., Ilyas, A., Santurkar, S., Tsipras, D., Janoos, F., Rudolph, L.,
  and Madry, A.
\newblock Implementation matters in deep rl: A case study on ppo and trpo.
\newblock In \emph{International Conference on Learning Representations}, 2019.

\bibitem[Espeholt et~al.(2018)Espeholt, Soyer, Munos, Simonyan, Mnih, Ward,
  Doron, Firoiu, Harley, Dunning, et~al.]{espeholt2018impala}
Espeholt, L., Soyer, H., Munos, R., Simonyan, K., Mnih, V., Ward, T., Doron,
  Y., Firoiu, V., Harley, T., Dunning, I., et~al.
\newblock Impala: Scalable distributed deep-rl with importance weighted
  actor-learner architectures.
\newblock \emph{arXiv preprint arXiv:1802.01561}, 2018.

\bibitem[Gu et~al.(2016)Gu, Lillicrap, Ghahramani, Turner, and Levine]{gu2016q}
Gu, S., Lillicrap, T., Ghahramani, Z., Turner, R.~E., and Levine, S.
\newblock Q-prop: Sample-efficient policy gradient with an off-policy critic.
\newblock \emph{arXiv preprint arXiv:1611.02247}, 2016.

\bibitem[Henderson et~al.(2017)Henderson, Islam, Bachman, Pineau, Precup, and
  Meger]{henderson2017deep}
Henderson, P., Islam, R., Bachman, P., Pineau, J., Precup, D., and Meger, D.
\newblock Deep reinforcement learning that matters.
\newblock \emph{arXiv preprint arXiv:1709.06560}, 2017.

\bibitem[Henderson et~al.(2018)Henderson, Romoff, and Pineau]{henderson2018did}
Henderson, P., Romoff, J., and Pineau, J.
\newblock Where did my optimum go?: An empirical analysis of gradient descent
  optimization in policy gradient methods.
\newblock \emph{arXiv preprint arXiv:1810.02525}, 2018.

\bibitem[Hill(1975)]{hill1975simple}
Hill, B.~M.
\newblock A simple general approach to inference about the tail of a
  distribution.
\newblock \emph{The annals of statistics}, pp.\  1163--1174, 1975.

\bibitem[Ilyas et~al.(2018)Ilyas, Engstrom, Santurkar, Tsipras, Janoos,
  Rudolph, and Madry]{ilyas2018closer}
Ilyas, A., Engstrom, L., Santurkar, S., Tsipras, D., Janoos, F., Rudolph, L.,
  and Madry, A.
\newblock A closer look at deep policy gradients.
\newblock \emph{arXiv preprint arXiv:1811.02553}, 2018.

\bibitem[Islam et~al.(2017)Islam, Henderson, Gomrokchi, and
  Precup]{islam2017reproducibility}
Islam, R., Henderson, P., Gomrokchi, M., and Precup, D.
\newblock Reproducibility of benchmarked deep reinforcement learning tasks for
  continuous control.
\newblock \emph{arXiv preprint arXiv:1708.04133}, 2017.

\bibitem[Kakade(2002)]{kakade2002natural}
Kakade, S.~M.
\newblock A natural policy gradient.
\newblock In \emph{Advances in neural information processing systems}, pp.\
  1531--1538, 2002.

\bibitem[Medina \& Yang(2016)Medina and Yang]{medina2016no}
Medina, A.~M. and Yang, S.
\newblock No-regret algorithms for heavy-tailed linear bandits.
\newblock In \emph{International Conference on Machine Learning}, pp.\
  1642--1650, 2016.

\bibitem[Metelli et~al.(2018)Metelli, Papini, Faccio, and
  Restelli]{metelli2018policy}
Metelli, A.~M., Papini, M., Faccio, F., and Restelli, M.
\newblock Policy optimization via importance sampling.
\newblock \emph{arXiv preprint arXiv:1809.06098}, 2018.

\bibitem[Metelli et~al.(2020)Metelli, Papini, Montali, and
  Restelli]{metelli2020importance}
Metelli, A.~M., Papini, M., Montali, N., and Restelli, M.
\newblock Importance sampling techniques for policy optimization.
\newblock \emph{Journal of Machine Learning Research}, 21\penalty0
  (141):\penalty0 1--75, 2020.

\bibitem[Minsker et~al.(2015)]{minsker2015geometric}
Minsker, S. et~al.
\newblock Geometric median and robust estimation in banach spaces.
\newblock \emph{Bernoulli}, 21\penalty0 (4):\penalty0 2308--2335, 2015.

\bibitem[Mnih et~al.(2015)Mnih, Kavukcuoglu, Silver, Rusu, Veness, Bellemare,
  Graves, Riedmiller, Fidjeland, Ostrovski, et~al.]{mnih2015human}
Mnih, V., Kavukcuoglu, K., Silver, D., Rusu, A.~A., Veness, J., Bellemare,
  M.~G., Graves, A., Riedmiller, M., Fidjeland, A.~K., Ostrovski, G., et~al.
\newblock Human-level control through deep reinforcement learning.
\newblock \emph{nature}, 518\penalty0 (7540):\penalty0 529--533, 2015.

\bibitem[Mnih et~al.(2016)Mnih, Badia, Mirza, Graves, Lillicrap, Harley,
  Silver, and Kavukcuoglu]{mnih2016asynchronous}
Mnih, V., Badia, A.~P., Mirza, M., Graves, A., Lillicrap, T., Harley, T.,
  Silver, D., and Kavukcuoglu, K.
\newblock Asynchronous methods for deep reinforcement learning.
\newblock In \emph{International conference on machine learning}, pp.\
  1928--1937, 2016.

\bibitem[Mohammadi et~al.(2015)Mohammadi, Mohammadpour, and
  Ogata]{mohammadi2015estimating}
Mohammadi, M., Mohammadpour, A., and Ogata, H.
\newblock On estimating the tail index and the spectral measure of multivariate
  $\alpha $-stable distributions.
\newblock \emph{Metrika}, 78\penalty0 (5):\penalty0 549--561, 2015.

\bibitem[Nemirovski \& Yudin(1983)Nemirovski and Yudin]{nemirovski1983problem}
Nemirovski, A. and Yudin, D.
\newblock \emph{Problem Complexity and Method Efficiency in Optimization}.
\newblock A Wiley-Interscience publication. Wiley, 1983.

\bibitem[Panigrahi et~al.(2019)Panigrahi, Somani, Goyal, and
  Netrapalli]{panigrahi2019non}
Panigrahi, A., Somani, R., Goyal, N., and Netrapalli, P.
\newblock Non-gaussianity of stochastic gradient noise.
\newblock \emph{arXiv preprint arXiv:1910.09626}, 2019.

\bibitem[Prasad et~al.(2018)Prasad, Suggala, Balakrishnan, and
  Ravikumar]{prasad2018robust}
Prasad, A., Suggala, A.~S., Balakrishnan, S., and Ravikumar, P.
\newblock Robust estimation via robust gradient estimation.
\newblock \emph{arXiv preprint arXiv:1802.06485}, 2018.

\bibitem[Resnick(2007)]{resnick2007heavy}
Resnick, S.~I.
\newblock \emph{Heavy-tail phenomena: probabilistic and statistical modeling}.
\newblock Springer Science \& Business Media, 2007.

\bibitem[Schulman et~al.(2015{\natexlab{a}})Schulman, Levine, Abbeel, Jordan,
  and Moritz]{schulman2015trust}
Schulman, J., Levine, S., Abbeel, P., Jordan, M., and Moritz, P.
\newblock Trust region policy optimization.
\newblock In \emph{International conference on machine learning}, pp.\
  1889--1897, 2015{\natexlab{a}}.

\bibitem[Schulman et~al.(2015{\natexlab{b}})Schulman, Moritz, Levine, Jordan,
  and Abbeel]{schulman2015high}
Schulman, J., Moritz, P., Levine, S., Jordan, M., and Abbeel, P.
\newblock High-dimensional continuous control using generalized advantage
  estimation.
\newblock \emph{arXiv preprint arXiv:1506.02438}, 2015{\natexlab{b}}.

\bibitem[Schulman et~al.(2017)Schulman, Wolski, Dhariwal, Radford, and
  Klimov]{schulman2017proximal}
Schulman, J., Wolski, F., Dhariwal, P., Radford, A., and Klimov, O.
\newblock Proximal policy optimization algorithms.
\newblock \emph{arXiv preprint arXiv:1707.06347}, 2017.

\bibitem[Silver et~al.(2017)Silver, Schrittwieser, Simonyan, Antonoglou, Huang,
  Guez, Hubert, Baker, Lai, Bolton, et~al.]{silver2017mastering}
Silver, D., Schrittwieser, J., Simonyan, K., Antonoglou, I., Huang, A., Guez,
  A., Hubert, T., Baker, L., Lai, M., Bolton, A., et~al.
\newblock Mastering the game of go without human knowledge.
\newblock \emph{nature}, 550\penalty0 (7676):\penalty0 354--359, 2017.

\bibitem[Simsekli et~al.(2019)Simsekli, Sagun, and
  Gurbuzbalaban]{simsekli2019tail}
Simsekli, U., Sagun, L., and Gurbuzbalaban, M.
\newblock A tail-index analysis of stochastic gradient noise in deep neural
  networks.
\newblock \emph{arXiv preprint arXiv:1901.06053}, 2019.

\bibitem[{\c{S}}im{\c{s}}ekli et~al.(2020){\c{S}}im{\c{s}}ekli, Zhu, Teh, and
  G{\"u}rb{\"u}zbalaban]{csimcsekli2020fractional}
{\c{S}}im{\c{s}}ekli, U., Zhu, L., Teh, Y.~W., and G{\"u}rb{\"u}zbalaban, M.
\newblock Fractional underdamped langevin dynamics: Retargeting sgd with
  momentum under heavy-tailed gradient noise.
\newblock \emph{arXiv preprint arXiv:2002.05685}, 2020.

\bibitem[Sutton et~al.(2000)Sutton, McAllester, Singh, and
  Mansour]{sutton2000policy}
Sutton, R.~S., McAllester, D.~A., Singh, S.~P., and Mansour, Y.
\newblock Policy gradient methods for reinforcement learning with function
  approximation.
\newblock In \emph{Advances in neural information processing systems}, pp.\
  1057--1063, 2000.

\bibitem[Todorov et~al.(2012)Todorov, Erez, and Tassa]{todorov2012mujoco}
Todorov, E., Erez, T., and Tassa, Y.
\newblock Mujoco: A physics engine for model-based control.
\newblock In \emph{2012 IEEE/RSJ International Conference on Intelligent Robots
  and Systems}, pp.\  5026--5033. IEEE, 2012.

\bibitem[Wang et~al.(2018)Wang, Liu, and Liu]{wang2018variational}
Wang, D., Liu, H., and Liu, Q.
\newblock Variational inference with tail-adaptive f-divergence.
\newblock In \emph{Advances in Neural Information Processing Systems}, pp.\
  5737--5747, 2018.

\bibitem[Williams(1992)]{williams1992}
Williams, R.~J.
\newblock Simple statistical gradient-following algorithms for connectionist
  reinforcement learning.
\newblock \emph{Mach. Learn.}, 8\penalty0 (3–4):\penalty0 229–256, May
  1992.
\newblock ISSN 0885-6125.
\newblock \doi{10.1007/BF00992696}.
\newblock URL \url{https://doi.org/10.1007/BF00992696}.

\bibitem[Wu et~al.(2018)Wu, Rajeswaran, Duan, Kumar, Bayen, Kakade, Mordatch,
  and Abbeel]{wu2018variance}
Wu, C., Rajeswaran, A., Duan, Y., Kumar, V., Bayen, A.~M., Kakade, S.,
  Mordatch, I., and Abbeel, P.
\newblock Variance reduction for policy gradient with action-dependent
  factorized baselines.
\newblock \emph{arXiv preprint arXiv:1803.07246}, 2018.

\bibitem[Xie et~al.(2021)Xie, Sato, and Sugiyama]{xie2021a}
Xie, Z., Sato, I., and Sugiyama, M.
\newblock A diffusion theory for deep learning dynamics: Stochastic gradient
  descent exponentially favors flat minima.
\newblock In \emph{International Conference on Learning Representations}, 2021.
\newblock URL \url{https://openreview.net/forum?id=wXgk_iCiYGo}.

\bibitem[Zhang et~al.(2019{\natexlab{a}})Zhang, He, Sra, and
  Jadbabaie]{zhang2019gradient}
Zhang, J., He, T., Sra, S., and Jadbabaie, A.
\newblock Why gradient clipping accelerates training: A theoretical
  justification for adaptivity.
\newblock \emph{arXiv preprint arXiv:1905.11881}, 2019{\natexlab{a}}.

\bibitem[Zhang et~al.(2019{\natexlab{b}})Zhang, Karimireddy, Veit, Kim, Reddi,
  Kumar, and Sra]{zhang2019adam}
Zhang, J., Karimireddy, S.~P., Veit, A., Kim, S., Reddi, S.~J., Kumar, S., and
  Sra, S.
\newblock Why adam beats sgd for attention models.
\newblock \emph{arXiv preprint arXiv:1912.03194}, 2019{\natexlab{b}}.

\end{thebibliography}
\bibliographystyle{icml2021}

\onecolumn
\appendix

\section{Detailed backgroud} \label{sec:App_prelim}

We define a Markov Decision Process (MDP) 
as a tuple $(\calS, \calA, R, \gamma, P )$, 
where $\calS$ represent the set of environments states, 
$\calA$ represent the set of agent actions, 
$R : \calS \times \calA \to \real$ is the reward function, 
$\gamma$ is the discount factor, and 
$P: \calS \times \calA \times \calS \to \real$ 
is the state transition probability distribution. 
The goal in reinforcement learning is to learn a policy 
$\pi_\theta : \calS \times \calA \to [0,1]$, 
parameterized by $\theta$, such that 
the expected cumulative discounted reward 
(known as returns) is maximized. 
Formally,  
\begin{align*}
    \pi^* = \argmax_\pi \Exp_{ a_t \sim \pi(\cdot \vert s_t), s_{t+1} \sim P(\cdot \vert s_t, a_t)}\left[ \sum_{t=0}^\infty \gamma^t R(s_t, a_t) \right] \,. \numberthis \label{eq:arl}
\end{align*}
Policy gradient methods directly optimize
a paraterized policy function 
(also known as \emph{actor network}). %
The central idea behind policy gradient methods 
is to perform stochastic gradient ascent on 
expected return (\Eqref{eq:arl}) 
to learn parameters $\theta$. 
Under mild conditions~\citep{sutton2000policy}, 
the gradient of the \Eqref{eq:arl} 
can be written as 
 \begin{align*}
     \nabla_\theta J(\theta) = \Exp_{ \tau \sim \pi_\theta}\left[  \sum_{t=0}^\infty \gamma^t R(s_t, a_t)\nabla_\theta \log(\pi_\theta(a_t\vert s_t)) \right] \,,
 \end{align*}
 where $\tau \sim \pi_\theta$ are trajectories
 sampled according to $\pi_\theta (\tau)$ 
 and $J(\theta)$ is the objective maximised in \Eqref{eq:arl}. 
 With the observation that action $a_t$ only affects 
 the reward from time $t$ onwards, 
 we re-write the objective $J(\theta)$,
 replacing returns using the Q-function,
 i.e., the expected discounted reward 
 after taking an action $a$ at state $s$ 
 and following $\pi_\theta$ afterwards. 
 Mathematically, 
 $Q_{\pi_\theta}(s,a) = \Exp_{ \tau \sim \pi_\theta} \left[ \sum_{k=0}^\infty \gamma^k R(s_{t+k}, a_{t+k}) \vert a_t= a, s_t = s \right]$. 
 Using the Q-function, we can write 
 the gradient of the objective function as
 \begin{align*}
     \nabla_\theta J(\theta) = \Exp_{ \tau \sim \pi_\theta}\left[  \sum_{t=0}^\infty Q_{\pi_\theta}(s_t,a_t) \nabla_\theta \log(\pi_\theta(a_t\vert s_t)) \right] \,.
 \end{align*}
 
However, the variance in the above expectation can be large,
which raises difficulties for 
estimating the expectation empirically. 
To reduce the variance of this estimate, 
a baseline is subtracted from the Q-function---often 
the value function or expected cumulative discounted reward 
starting at a certain state and following a given policy
i.e., $V_{\pi_\theta}(s) = \Exp_{ \tau \sim \pi_\theta} \left[ \sum_{k=0}^\infty \gamma^k R(s_{t+k}, a_{t+k}) \vert s_t = s \right]$. 
The network that estimates the value function 
is often referred to as \emph{critic}. 
Define $A_{\pi_\theta}(s_t, a_t) = Q_{\pi_\theta}(s_t,a_t) - V_{\pi_\theta}(s_t)$ 
as the \emph{advantage} of performing action $a_t$ at state $s_t$. 
Incorporating an advantage function, 
the gradient of the objective function can be written:
  \begin{align*}
     \nabla_\theta J(\theta) = \Exp_{ \tau \sim \pi_\theta}\left[  \sum_{t=0}^\infty A_{\pi_\theta}(s_t,a_t) \nabla_\theta \log(\pi_\theta(a_t\vert s_t)) \right] \,. \numberthis \label{eq:aA2C}
 \end{align*}

\Eqref{eq:aA2C} is the n\"aive actor-critic objective and is used by A2C. 
  
\textbf{Trust region methods and PPO.}
Since directly optimizing the cumulative rewards can be challenging, 
modern policy gradient optimization algorithms 
often optimize a surrogate reward function
in place of the true reward. 
Most commonly, the surrogate reward objective 
includes a likelihood ratio to allow
importance sampling from a behavior policy $\pi_0$ 
while optimizing policy $\pi_\theta$, 
such as the surrogate reward used by \citet{schulman2015trust}: 
\begin{align*}
    \max_\theta \, \Exp_{(s_t, a_t) \sim \pi_0}\left[ \frac{\pi_\theta(a_t, s_t)}{\pi_0(a_t, s_t)} \hat A_{\pi_0}(s_t, a_t) \right] \,, \numberthis \label{eq:appo-no-clip}
\end{align*}
where $\hat A_{\pi} = \frac{A_{\pi} - \mu(A_\pi)}{\sigma(A_\pi)}$ 
(we refer to this as the \emph{normalized advantages}). 
\update{However, the surrogate is indicative 
of the true reward function 
only when $\pi_\theta$ and $\pi_0$
are close in distribution. 
Different policy gradient methods~\citep{schulman2015trust,schulman2017proximal,kakade2002natural}
attempt to enforce the closeness in different ways.  
In Natural Policy Gradients~\citep{kakade2002natural} and Trust Region Policy Optimization (TRPO)~\citep{schulman2015trust}, authors utilize a conservation policy iteration with an explicit divergence constraint which provides provable lower bounds guarantee on the improvements of the parameterized policy.
On the other hand, PPO~\citep{schulman2017proximal} implements a clipping heuristic on the likelihood ratio 
of the surrogate reward function to avoid excessively large policy
updates.}
Specifically, PPO optimizes the following objective: 
 \begin{align*}
    \max_{\theta_t} \, & \Exp_{(s_t, a_t) \sim \pi_{\theta_{t-1}}} \left[ \min \left(\rho_t \hat A_{\pi_{\theta_{t-1}}}(s_t, a_t) \clip (\rho_t, 1-\epsilon, 1+\epsilon) \hat A_{\pi_{\theta_{t-1}}}(s_t, a_t) \right) \right] \,, \numberthis \label{eq:appo}
\end{align*}
where $\rho_t \defeq \frac{\pi_{\theta_t}(a_t, s_t)}{\pi_{\theta_{t-1}}(a_t, s_t)}$ and $\text{clip}(x,1-\epsilon, 1+\eps )$ clips $x$ to stay between $1+\eps$ and $1-\eps$.  We refer to $\rho_t$ as \emph{likelihood-ratios}. \update{Due to a  minimum with the unclipped surrogate reward, the PPO objective acts as a pessimistic bound on the true surrogate reward.}
As in standard PPO implementation, we use Generalized Advantage Estimation (GAE)~\citep{schulman2015high}.
Moreover, instead of fitting the value network
via regression to target values (denoted by $V_{trg}$):  
\begin{align*}
    \min_{{\theta_t}} \, \Exp_{s_t \sim \pi_{\theta_{t-1}}} \left[ (V_{\theta_t} (s_t) - V_{trg} (s_t)) ^2 \right], \numberthis \label{eq:avalue-no-clip}
\end{align*} 

standard implementations fit the value network with a PPO-like objective: 
\begin{align*}
    \min_{\theta_t} \, \Exp_{s_t \sim \pi_{\theta_{t-1}}} \max\left\{\left(V_{\theta_t}(s_t) - V_{trg} (s_t)\right)^2 ,\left(\text{clip}\left(V_{\theta_t} (s_t), V_{\theta_{t-1}} (s_t)-\varepsilon,
	  V_{\theta_{t-1}}(s_t) + \varepsilon\right) - V_{trg}(s_t)\right)^2\right\}%
	  \numberthis 
	  \label{eq:avalue} \,,
\end{align*}
where $\epsilon$ is the same value used to clip 
probability ratios in PPO's loss function (\Eqref{eq:appo}).
PPO uses the following training procedure: 
At any iteration $t$, the agent creates
a clone of the current policy $\pi_{\theta_t}$ 
which interacts with the environment to collects rollouts $\calS$ 
(i.e., state-action pair $\{(s_i,a_i)\}_{i=1}^N$). 
Then the algorithm optimizes the policy $\pi_\theta$ 
and value function for a fixed $K$ gradient steps 
on the sampled data $\calS$. 
Since at every iteration the first gradient step 
is taken on the same policy 
from which the data was sampled, 
we refer to these gradient updates as \emph{on-policy} steps. 
And as for the remaining $K-1$ steps,
the sampling policy differs from the current agent,
we refer to these updates as \emph{off-policy} steps. 

Throughout the paper, we consider a stripped-down variant of PPO (denoted PPO-\textsc{NoClip})
that consists of policy gradient with importance weighting (\Eqref{eq:appo-no-clip}),
but has been simplified as follows: 
i) no likelihood-ratio clipping, i.e., no \emph{objective function clipping}; 
ii) value network optimized via regression 
to target values (\Eqref{eq:avalue-no-clip})
without \emph{value function clipping}; 
and iii) no \emph{gradient clipping}. 
\update{Overall PPO-\textsc{NoClip} uses the following objective: 
\begin{align*}
    \max_\theta \, \Exp_{(s_t, a_t) \sim \pi_0}\left[ \frac{\pi_\theta(a_t, s_t)}{\pi_0(a_t, s_t)} \hat A_{\pi_0}(s_t, a_t)  
    - c (V_{\theta_t} - V_{targ})^2 \right] \,.
\end{align*} 
where $c$ is a coefficient of the value function loss (tune as a hyperparameter).   
Moreover, no gradient clipping is incorporated in PPO-\textsc{NoClip}.  
One may argue that since PPO-\textsc{NoClip} removes the clipping heuristic from PPO, the unconstrained maximization of \Eqref{eq:ppo-no-clip} may lead to excessively large policy updates. In App.~\ref{sec:App-KL}, we empirically justify the use of \Eqref{eq:ppo-no-clip} by showing that with the small learning rate used in our experiments (optimal hyperparameters in Table~\ref{table:hyperparameters}), PPO-\textsc{NoClip} maintains a KL based trust-region like PPO throughout the training. We elaborate this in App.~\ref{sec:App-KL}. }

\section{Details on  estimators} \label{sec:App_estimators}

We now formalize our setup
for studying the distribution of gradients. 
Throughout the paper, we use the following 
definition of the heavy-tailed property:

\begin{definition}[\citet{resnick2007heavy}] \label{def:ht_app}
A non-negative random variable $w$ is called \emph{heavy-tailed}
if its tail probability $\smash{F_w(t) \defeq P(w\ge t)}$ 
is asymptotically equivalent to $t^{-\alpha^*}$ as $t \to \infty$ 
for some positive number $\alpha^*$. 
Here $\alpha^*$ determines the heavy-tailedness 
and $\alpha^*$ is called tail index of $w$.     
\end{definition}

For a heavy-tailed distribution with index $\alpha^*$,
its $\alpha$-th moment exist only if $\alpha < \alpha^*$,
i.e., $\Exp[w^\alpha] < \infty $ iff $\alpha < \alpha^*$.
A value of $\alpha^* = 1.0$ corresponds to a Cauchy distribution 
and $\alpha^* = \infty$ (i.e., all moments exist) 
corresponds to a Gaussian distribution. 
Intuitively, as $\alpha^*$ decreases, 
the central peak of the distribution gets higher, 
the valley before the central peak gets deeper,
and the tails get heavier. 
In other words, the lower the tail-index,
the more heavy-tailed the distribution. 
However, in the finite sample setting, 
estimating the tail index is notoriously challenging
\citep{simsekli2019tail, danielsson2016tail,hill1975simple}.

In this study, we explore three estimators 
as heuristic measures to understand heavy tails 
and non-Gaussianity of gradients.
\begin{itemize}
    \item \emph{Alpha-index estimator}. 
This estimator was proposed in \cite{mohammadi2015estimating} 
for symmeteric $\alpha$-stable distributions 
and was used by \cite{simsekli2019tail} 
to understand the noise behavior of SGD. 
This estimator is derived under the (strong)
assumption that the stochastic Gradient Noise (GN) vectors 
are coordinate-wise independent
and follow a symmetric alpha-stable distribution. Formally, let $\{X_i\}_{i=1}^N$ be a collection of $N=mn$ (centered) random variables. Define $Y_i = \sum_{j=1}^m X_{j+(i-1)m}$ for $i \in [n]$. Then, the estimator is given by \begin{align*}
    \frac{1}{\alpha} \defeq \frac{1}{\log m}\left( \frac{1}{n} \sum_{i=1}^n \log\abs{Y_i} - \frac{1}{n}\sum_{i=1}{N}\log\abs{X_i} \right) \,. 
\end{align*}

Instead of treating each co-ordinate of gradient noise as an independent scalar, we use these estimators on gradient norms as discussed in ~\citet{zhang2019adam}.
With alpha-index estimator, smaller alpha-index value signify higher degree of heavy-tailedness.

    \item \emph{Anderson-Darling test}~\citep{anderson1954test}
on random projections of GN to perform Gaussianity testing.
\citet{panigrahi2019non} proposed the Gaussianity test 
on the projections of GN along $1000$ random directions. 
Their estimate is then the fraction of directions 
accepted by the Anderson Darling test. 
While this estimator is informative about the Gaussian behavior,
it is not useful to quantify and understand 
the trends of heavy-tailedness 
if the predictor nature is non-Gaussian. 

    \item To our knowledge, %
the deep learning literature has only exploredthese two estimators
for analyzing 
the heavy-tailed nature of gradients. 
\emph{(iii)} Finally, in our work, we propose using kurtosis \emph{Kurtosis.} 
To quantify the heavy-tailedness relative to a normal distribution, 
we measure kurtosis (fourth standardized moment) of the gradient norms.
Given samples $\{X_i\}_{i=1}^N$, the kurtosis $\kappa$ is given by 
\begin{align*}
    \kappa = \frac{\sum_{i=1}^N (X_i - \bar{X})^4/N}{\left( \sum_{i=1}^N (X_i - \bar X)^2/N \right)^2} \,,
\end{align*}
where $\bar X$ is the empirical mean of the samples.
\end{itemize}
 
Note that both $\alpha$-index and Anderson-Darling need very strong assumptions to be valid.  (i) $\alpha$-index requires that the true distribution is symmetric and $\alpha$-stable. In the multivariate setting, the test proposed by \citet{simsekli2019tail} relies on the covariance being isotropic. 
These theoretical limitations also lead to practical consequences. Specifically, we found that for low-rank Gaussian distributions (for which ideally $\alpha$ = 2), the existing estimators report an $\alpha$ = 1.1, wrongly suggesting heavy-tailendess. Similar limitations were pointed out in recent works~\citep{zhang2019adam,xie2021a}.
(ii) Anderson-Darling tests for Gaussianity and is not useful in quantifying the degree of heavy-tailedness. Moreover, the test fails for sub-Gaussian distributions such as uniform distribution.

On the other hand, our proposed estimator doesn't require symmetry or Gaussianity, and works well in the aforementioned pathological situations arising in practice. Moreover, well-known fat tailed distributions such as Student's t-distribution, exponential distribution, etc., have higher Kurtosis than normal distrbution. Even for distributions with less than $4$ moments, empirical Kurtosis can be used to understand the ``relative'' trends in tail behavior for different distributions at fixed sample sizes (\figref{fig:E3}). 

\subsection{Synthetic study}

\begin{figure*}[bt!] 
    \centering
        \subfigure[]{\includegraphics[width=0.48\linewidth]{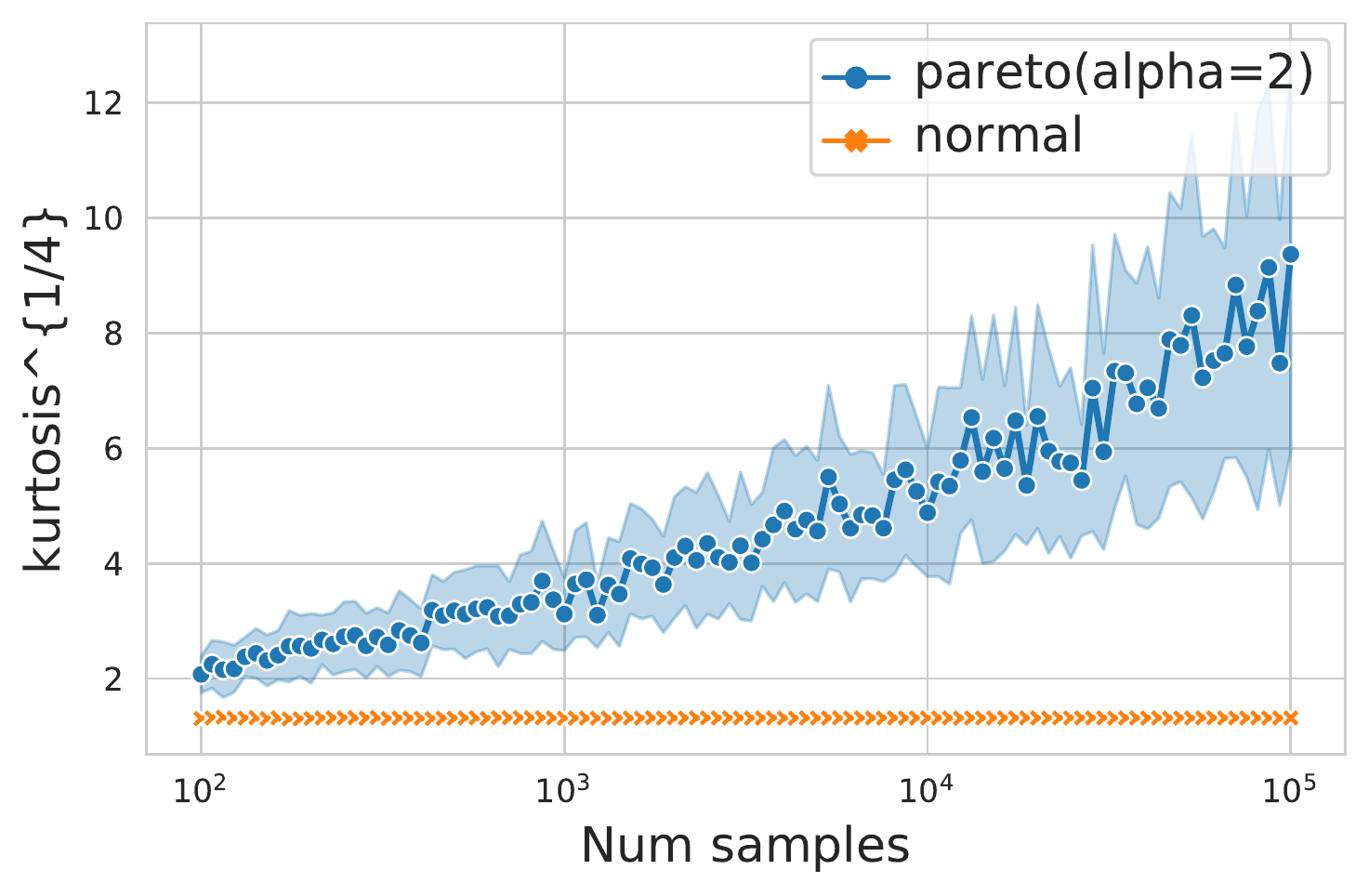}}\hfil
        \subfigure[]{ \includegraphics[width=0.48\linewidth]{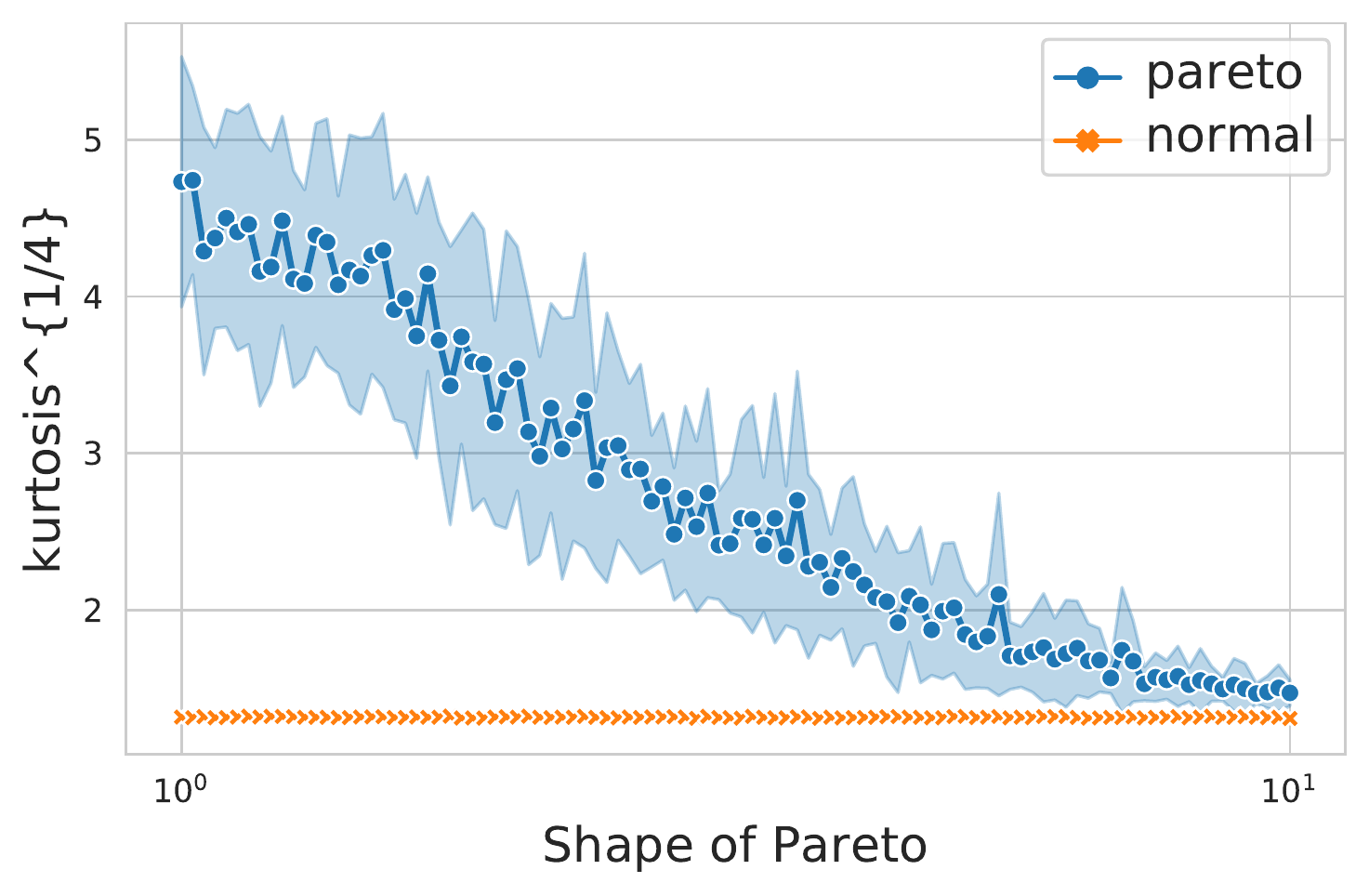}}\hfil
        \par\medskip

    \caption{Kurtosis plots. Analysis on norms of  $100$-dimensional vectors such that each coordinate is sampled iid from Pareto distribution or normal distribution. (a) Variation in kurtosis ($\kappa^{1/4}$) as the sample size is varied for samples from normal distribution and Pareto with tail index 2 (i.e, $\alpha = 2$). (b) Variation in kurtosis ($\kappa^{1/4}$) as the shape of Pareto is varied at fix sample size.  }\label{fig:E3}
    \end{figure*}

In Figure~\ref{fig:E3}, we show the trends with varying tail index and sample sizes. Clearly as the tail-index increases, i.e., the shape parameter increases, the kurtosis decreases (signifying its correlation to capture tail-index). 
Although for tail-index smaller than 4 the kurtosis is not defined, we plot empirical kurtosis and show its increasing trend sample size. 
We fix the tail index of Pareto at  $2$ and plot  finite sample kurtosis and observe that it increases almost exponentially with the sample size. 
These two observations together hint that kurtosis is a neat surrogate measure for heavy-tailedness. 

\newpage

\section{GMOM Algorithm} \label{sec:App_GMOM}

\begin{minipage}{0.48\textwidth}
    \begin{algorithm}[H]
        \caption{\textsc{GMOM}}
        \label{alg:gmom}
    \begin{algorithmic}[1]
        \INPUT: Samples $S = \{x_1, \ldots, x_n\}$, number of blocks $b$ 
        \STATE $m = \lfloor n/b\rfloor$.       
        \FOR{$i$ in $1\ldots b$}  
            \STATE $\hat \mu_i = \sum_{j=0}^m x_{j+i*m}$/ m.
        \ENDFOR
        \STATE $\hat \mu_{\gmom} = \textsc{Weiszfeld}(\hat \mu_1, \ldots, \hat \mu_b)$.
        \OUTPUT: Estimate $\hat \mu_{\gmom}$ 
    \end{algorithmic}
    \end{algorithm}
\end{minipage}
\hfill 
\begin{minipage}{0.48\textwidth}
    \begin{algorithm}[H]
        \caption{\textsc{Weiszfeld}}
        \label{alg:weiszfeld}
    \begin{algorithmic}[1]
        \INPUT: Samples $S = \{\mu_1, \ldots, \mu_b\}$, number of blocks $b$
        \STATE Initialize $\mu$ arbitrarily. 
        \FOR {iteration $\leftarrow 1, \ldots, n$ }  
                \STATE $d_j \defeq \frac{1}{\norm{\mu - \mu_j }{2}}$ for \, {$j$ in $1, \ldots, b$}. 
                \STATE $\mu \defeq \left(\sum_{j=1}^b \mu_j d_j \right)/\left(\sum_{j=1}^b d_j \right)$
        \ENDFOR
        \OUTPUT: Estimate $\mu$
    \end{algorithmic}
    \end{algorithm}
\end{minipage}

\section{Experimental setup for gradient distribution study}\label{sec:App_exp}

Recall that PPO uses the following training procedure: 
At any iteration $t$, the agent creates
a clone of the current policy $\pi_{\theta_t}$ 
which interacts with the environment to collects rollouts $\calS$ 
(i.e., state-action pair $\{(s_i,a_i)\}_{i=1}^N$). 
Then the algorithm optimizes the policy $\pi_\theta$ 
and value function for a fixed $K$ gradient steps 
on the sampled data $\calS$. 
Since at every iteration the first gradient step 
is taken on the same policy 
from which the data was sampled, 
we refer to these gradient updates as \textbf{on-policy steps}. 
And as for the remaining $K-1$ steps,
the sampling policy differs from the current agent,
we refer to these updates as \textbf{off-policy steps}.  For all experiments, we aggregate our estimators across \update{30 seeds and 8 environments}. We do this by first computing the estimators for individual experiments and then taking the sample mean across all runs.  We now describe the exact experimental details. 

In all of our experiments, for each gradient update, we have a batch size of 64.
Hence for an individual estimate, we aggregate over 64 samples (batch size in experiments) to compute our estimators. For Anderson Darling test, we use 100 random directions to understand the behavior of stochastic gradient noise.

\textbf{On-policy heavy-tailed estimation.} At every on-policy gradient step \update{(i.e. first step on newly sampled data)}, we freeze the policy and value network, and save the sample-wise gradients of the actor and critic objective. The estimators are calculated at every tenth on-policy update throughout the training.  

\textbf{Off-policy heavy-tailed estimation} At every off-policy gradient step \update{(i.e. the gradient updates made on a fixed batch of data when the sampling policy differs from the policy being optimized)}, we freeze the policy and value network, and save the sample-wise gradients of the actor and critic objective. Then at various stages of training, i.e., initialization, $50\%$ max reward and max reward \update{(which corresponds to different batches of sampled data)}, we fix the collected trajectories and collect sample-wise gradients for the 320 steps taken. \update{We now elaborate the exact setup with one instance, at $50\%$ of the maximum reward. First, we find the training iteration where the agent achieves approximately $50\%$ of the maximum reward individually for each environment. Then at this training iteration, we freeze the policy and value network and save the sample-wise gradients of the actor and critic objective for off-policy steps.}  

\textbf{Analysis of PPO-\textsc{NoClip} with progressively applying PPO heuristics}.  
We compute the gradients for the off-policy steps taken with the PPO-\textsc{NoClip} objective as explained above. Then at each gradient step, we progressively add heuristics from PPO and re-compute the gradients for analysis. 
Note that we still always update the value and policy network with PPO-\textsc{NoClip} objective gradients. 

\section{\update{Mean KL divergence between current and previous policy}} \label{sec:App-KL}

\begin{figure}[t] 
    \centering
        \subfigure{\includegraphics[width=0.24\linewidth]{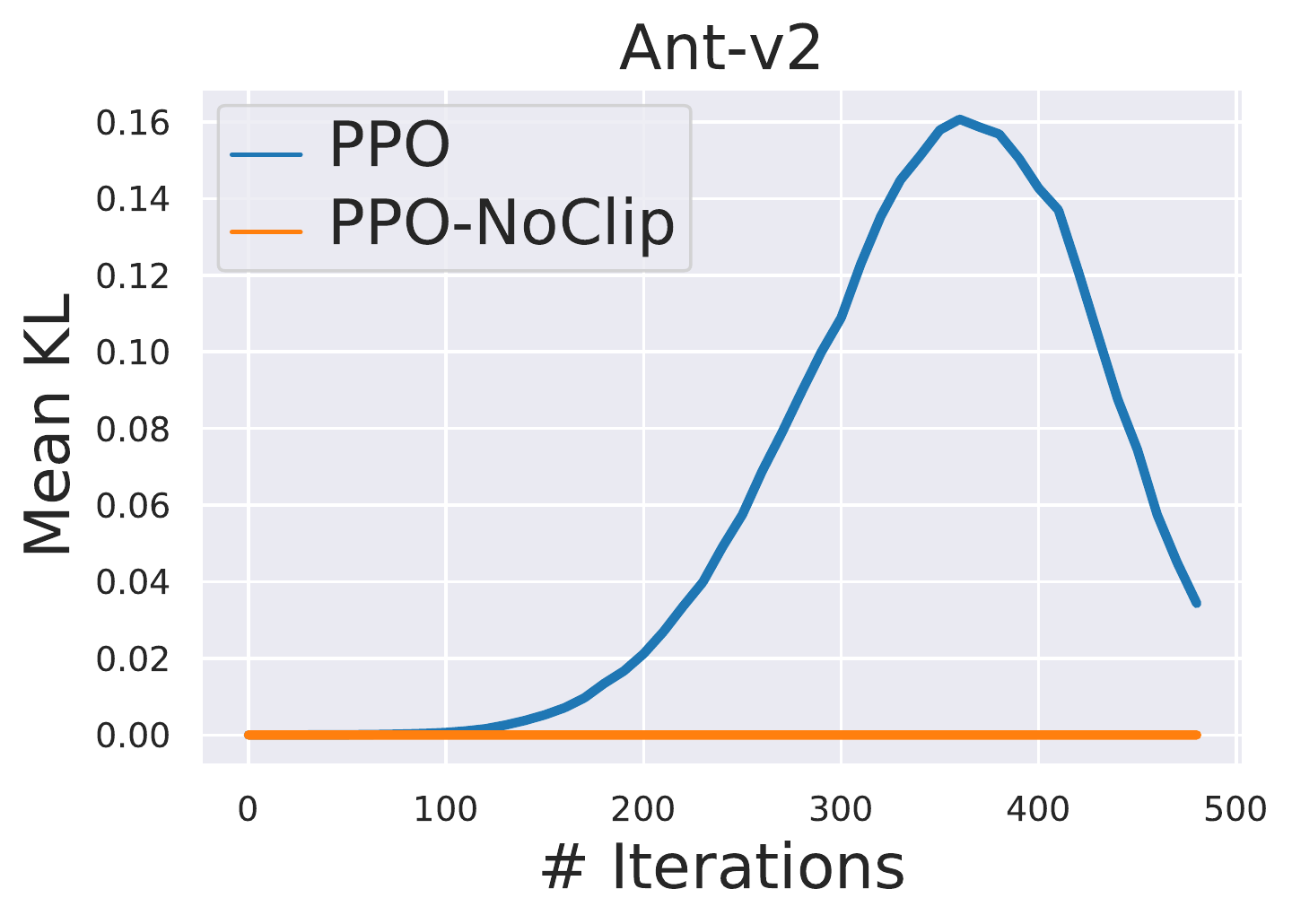}}\hfil
        \subfigure{\includegraphics[width=0.24\linewidth]{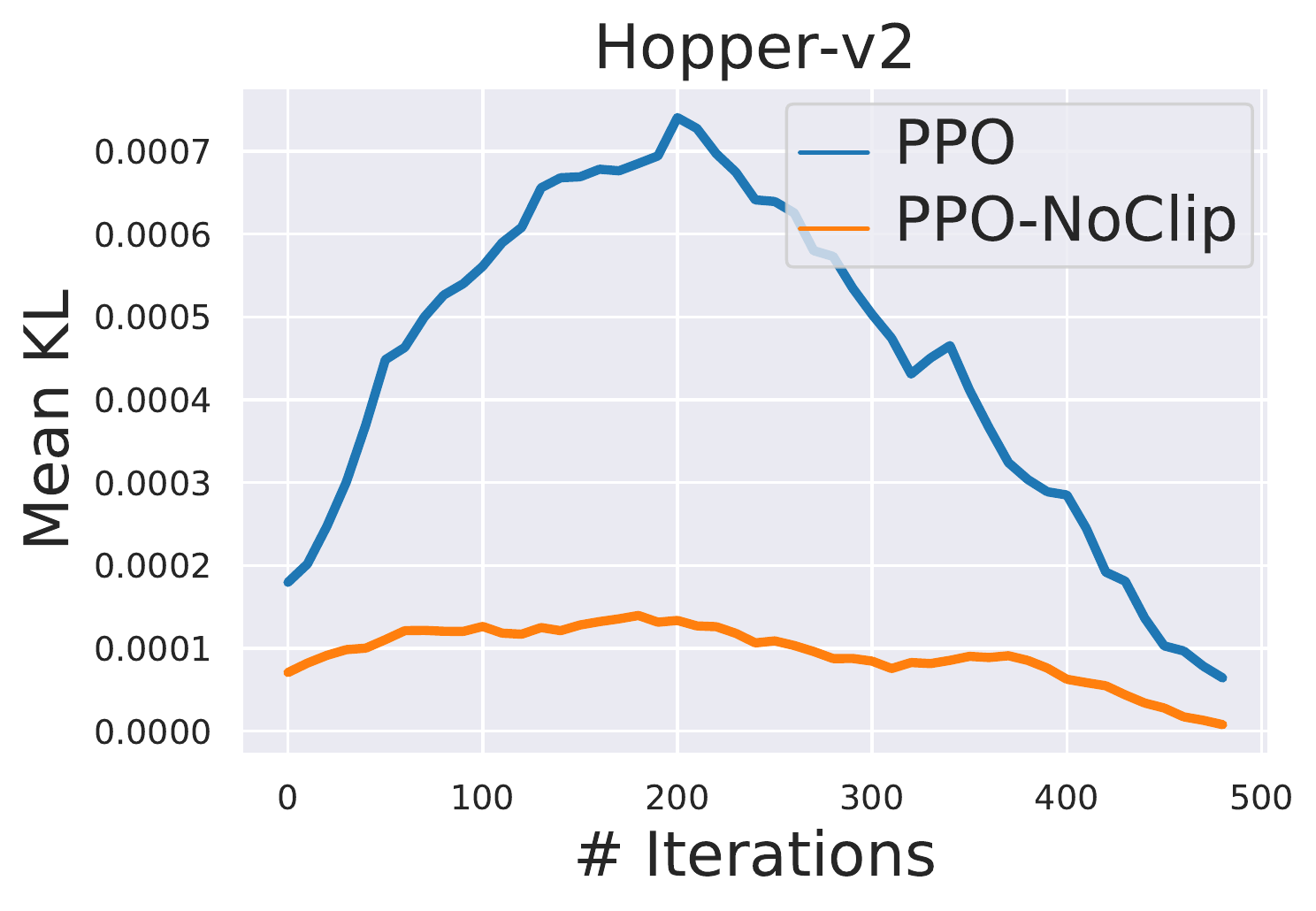}}\hfil
        \subfigure{ \includegraphics[width=0.24\linewidth]{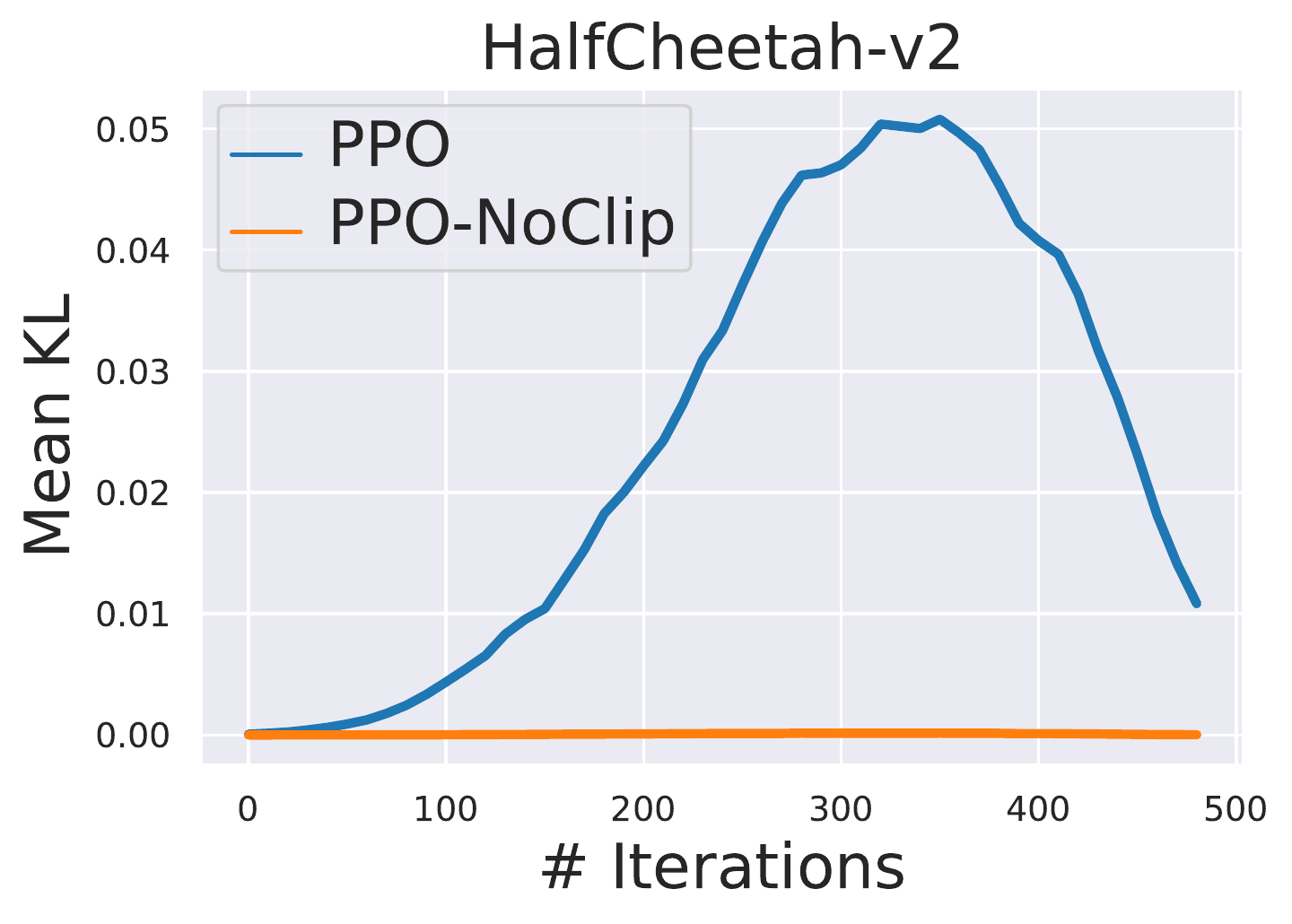}}\hfil
        \subfigure{\includegraphics[width=0.24\linewidth]{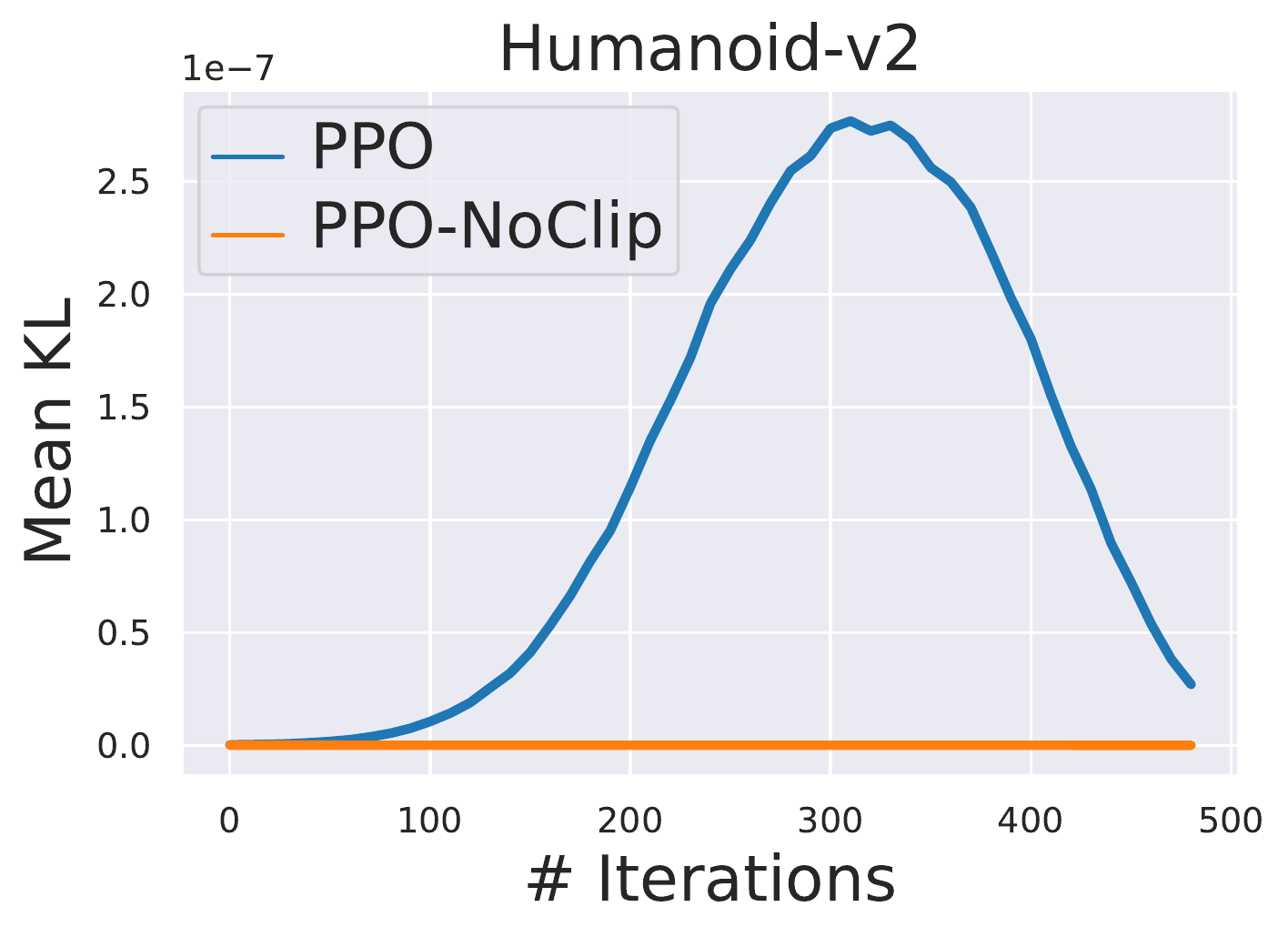}}\hfil
        \par\medskip
        \subfigure{\includegraphics[width=0.24\linewidth]{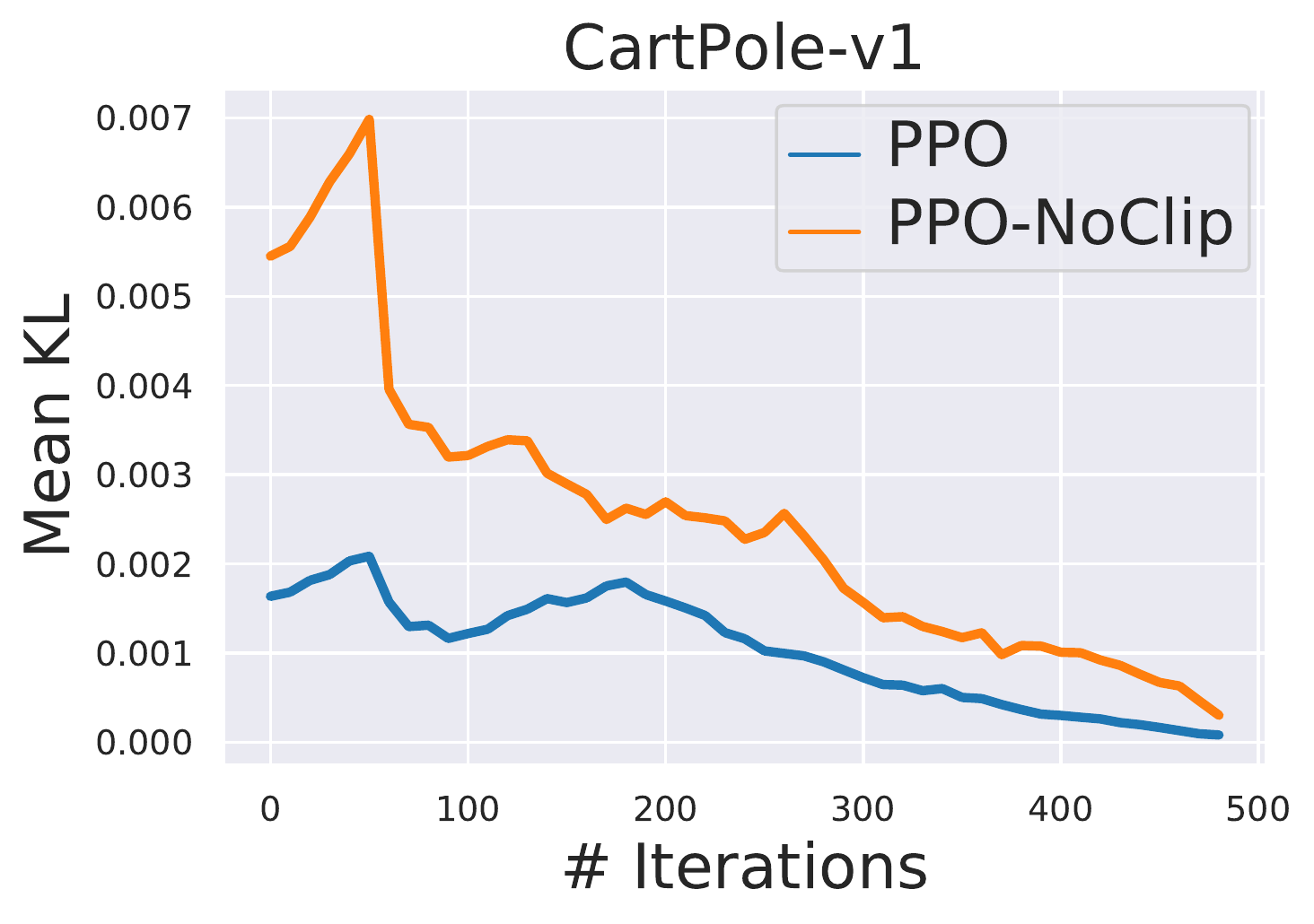}}\hfil
        \subfigure{ \includegraphics[width=0.24\linewidth]{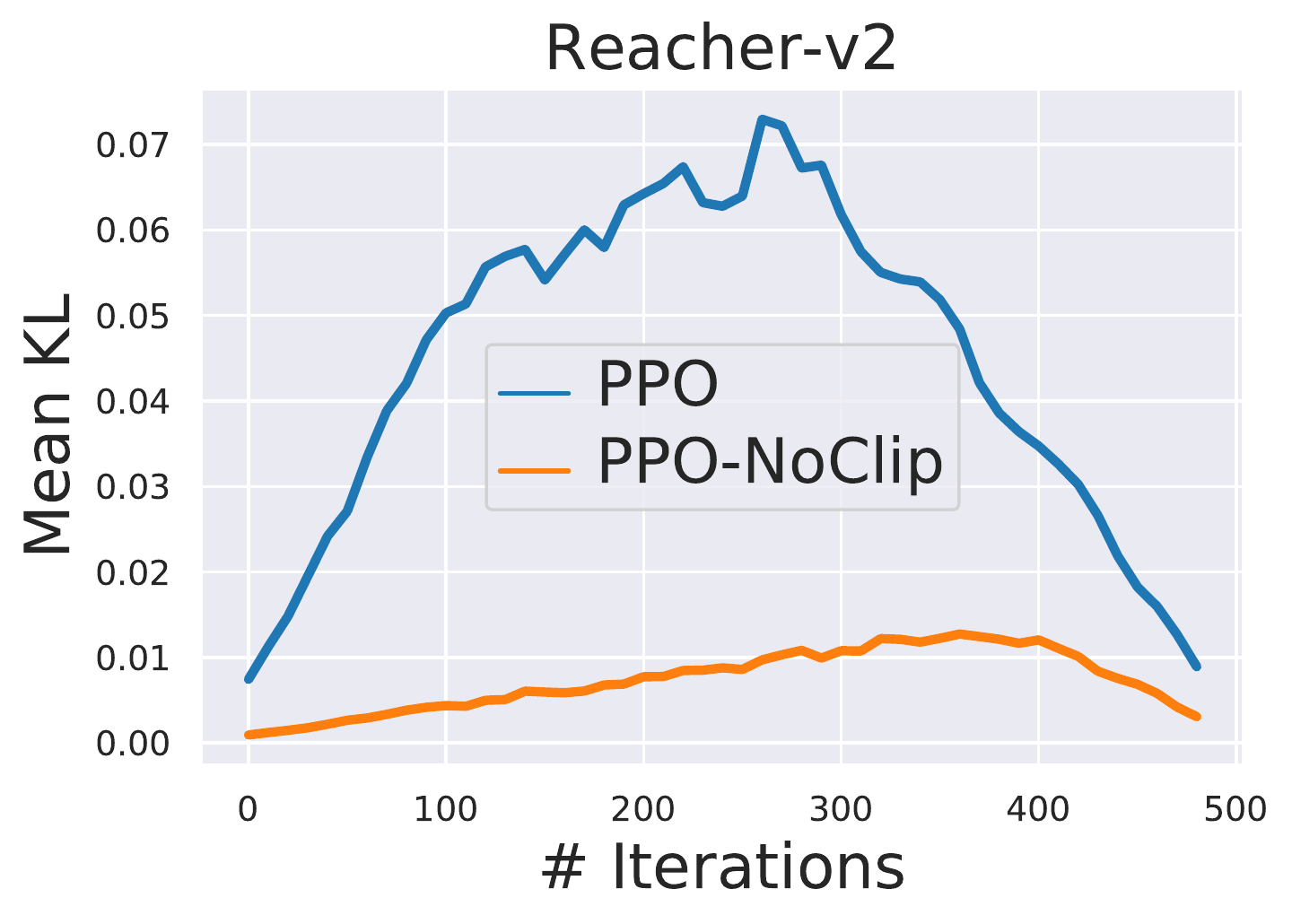}}\hfil
        \subfigure{\includegraphics[width=0.24\linewidth]{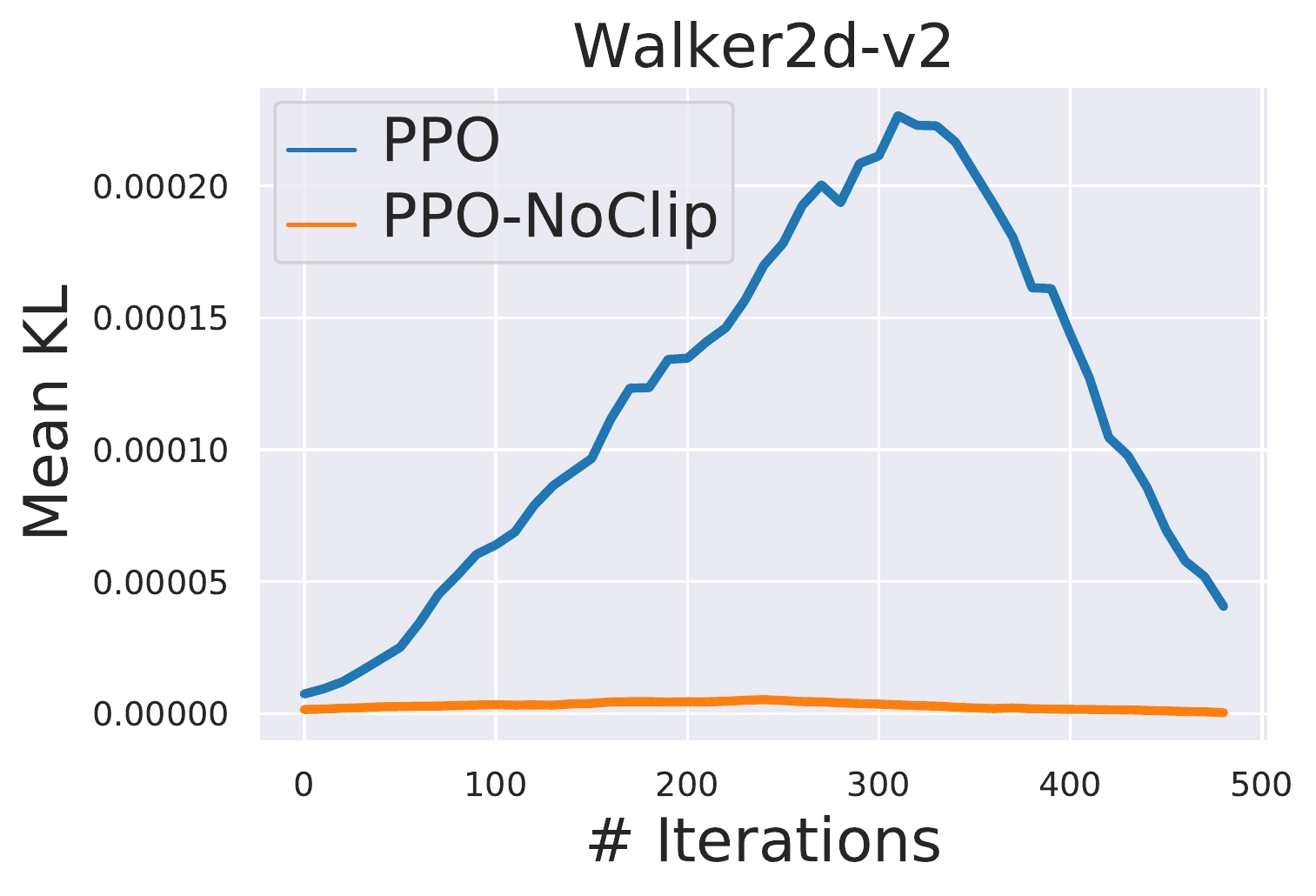}}\hfil
        \subfigure{\includegraphics[width=0.24\linewidth]{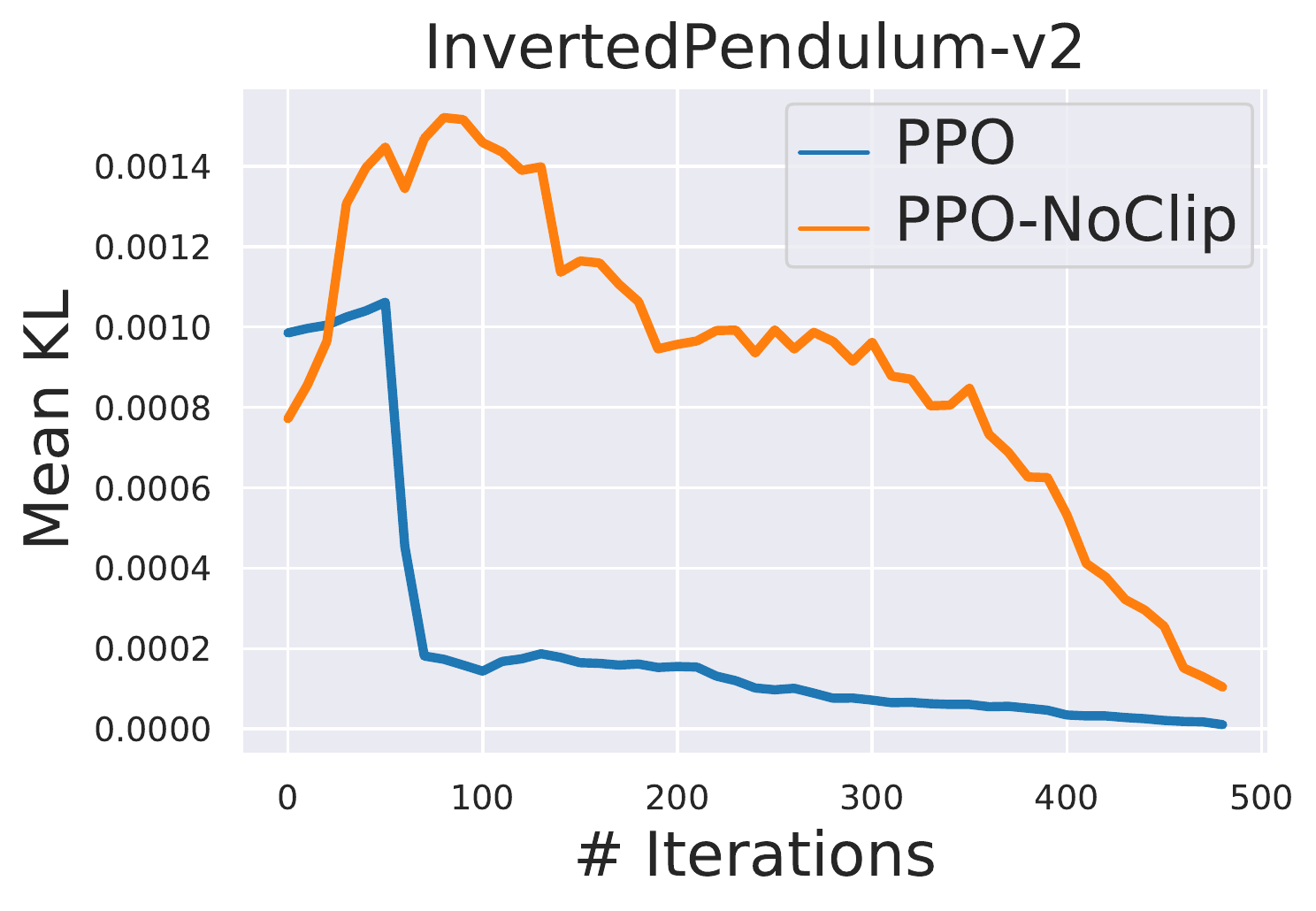}}\hfil
        \par\medskip
    \caption{ \update{\textbf{KL divergence between current and previous policies with the optimal hyperparameters (parameters in Table~\ref{table:hyperparameters})}. We measure the mean empirical KL divergence between the policy obtained at the end of off-policy training (after every 320 gradient steps) and the sampling policy at the beginning of every training iteration. The quantities are measured over the state-action pairs collected in the training step (\citet{engstrom2019implementation} observed similar results with both unseen data and training data). We observe that both the algorithms maintain a KL based trust region. The trend with KL divergence in PPO matches with the observations made in \citet{engstrom2019implementation} where they also observed that it peeks in halfway in training. } }\label{fig:kl-div} 
     
\end{figure}

\update{
Enforcing a trust region is a core algorithmic property of PPO and TRPO. While the trust-region enforcement is not directly clear from the reward curves or heavy-tailed analysis, inspired by \citet{engstrom2019implementation}, we perform an additional experiment to understand how this algorithmic property varies with PPO and our variant PPO-\textsc{NoClip} with optimal hyperparameters.}
\update{In Fig~\ref{fig:kl-div}, we measure mean KL divergence between successive policies of the agent while training with PPO and PPO-\textsc{NoClip}. Recall that while PPO implements a clipping heuristics in the likelihood ratios (as a surrogate to approximate the KL constraint of TRPO), we remove that clipping heuristics in PPO-\textsc{NoClip}.  
} 

\update{\citet{engstrom2019implementation} pointed out that trust-region enforced in PPO is heavily dependent on the method with which the clipped PPO objective is optimized, rather than on the objective itself. Corroborating their findings, we indeed observe that with optimal parameters (namely small learning rate used in our experiments), PPO-\textsc{NoClip} indeed manages to maintain a trust region with mean KL metric (Fig~\ref{fig:kl-div}) on all 8 MuJoCo environments. This highlights that instead of the core algorithmic objective used for training, the size of the step taken determines the underlying objective landscape, and its constraints. }\update{On a related note, \citet{ilyas2018closer} also highlighted that the objective landscape of PPO algorithm in the typical sample-regime in which they operate can be very different from the true reward landscape.  
}

\section{Trends with advantages} \label{Appsec:negativeAdv}
\subsection{Kurtosis for returns, value estimate and advantages grouped with sign} \label{Appsec:negativeAdv1}
\begin{figure}[H]
    \centering
    \includegraphics[width=0.35\linewidth]{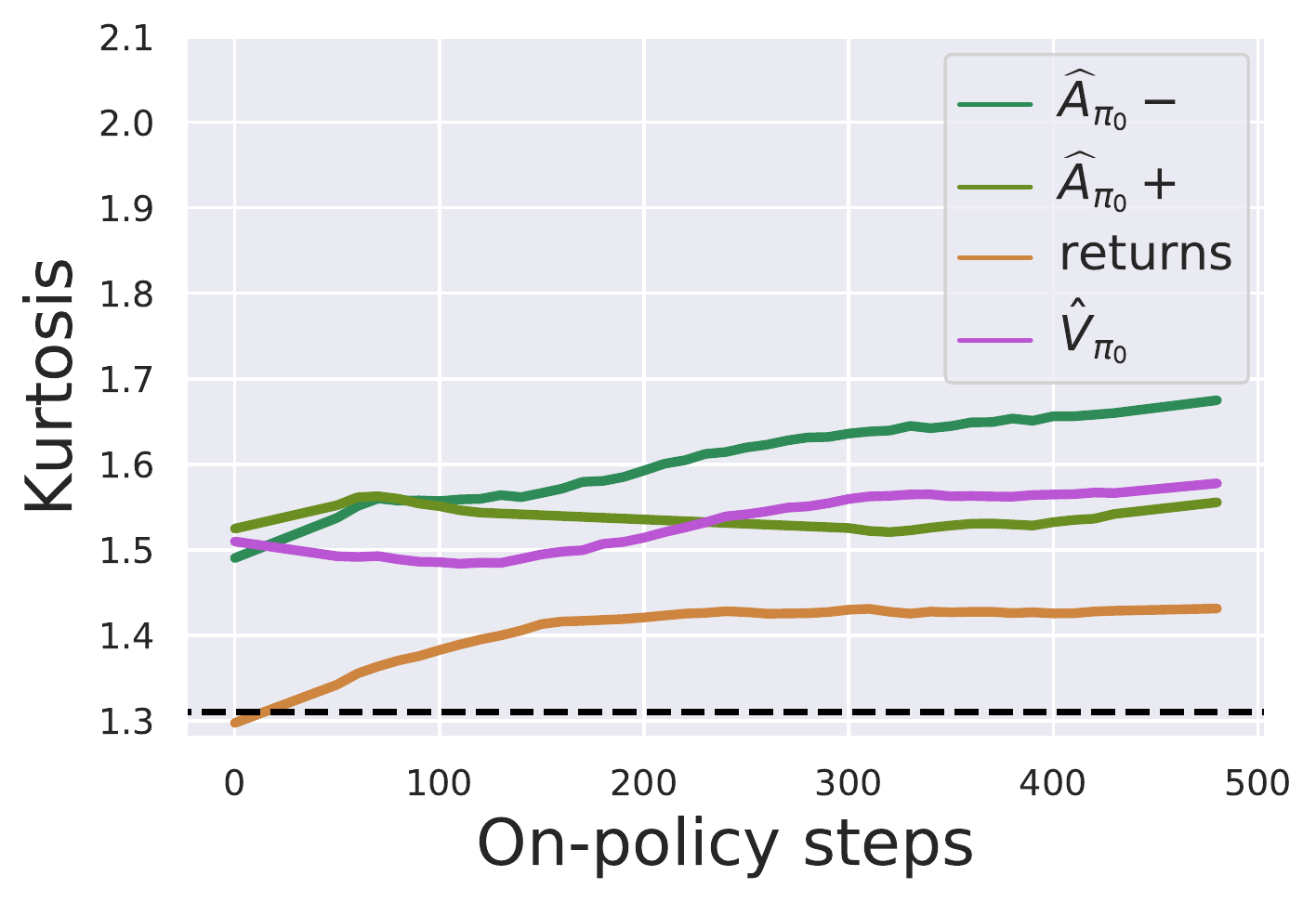}
    \caption{\update{ Heavy-tailedness in advantages grouped by their sign, rewards and value estimates. 
    Clearly, as the training progresses the negative advantages become heavy-tailed. For returns, we observe an initial slight increase in the heavy-tailedness which quickly plateaus to a small magnitude of heavytailedness. The heavytailedness in the value estimates and positive advantages remain almost constant throughout the training. }  
    }
\end{figure}

\subsection{\update{Heavy-tailedness in A2C and PPO in onpolicy iterations}} \label{sec:App-A2C}

\begin{figure}[H]
    \centering
    \includegraphics[width=0.35\linewidth]{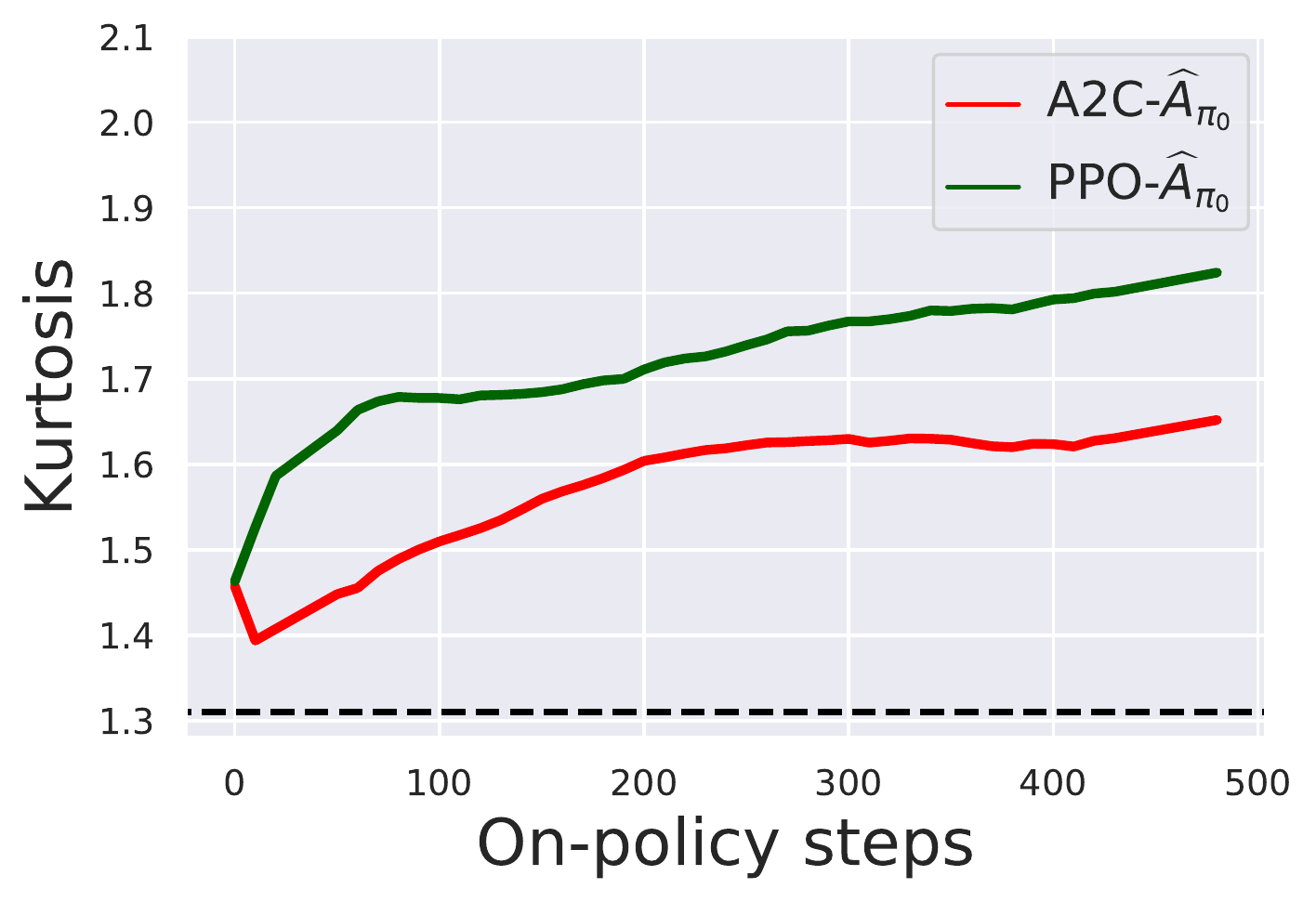}
    \caption{\update{ Heavy-tailedness in advantages for A2C and PPO during on-policy iterates. 
    Clearly, as the training progresses heavy-tailedness in PPO advantages increases rapidly when compared with A2C advantages. The observed behavior arises to the off-policy training of the agent in PPO. This explains why we observe heightened heavy-tailedness in PPO during onpolicy iterations in Fig~\ref{fig:intro}(a).}  
    }\label{fig:app_adv1}
\end{figure}

\subsection{Histograms of advantages on HalfCheetah over training iterations}\label{Appsec:negativeAdv2}

\begin{figure}[H]
    \centering
        \subfigure{\includegraphics[width=0.3\linewidth]{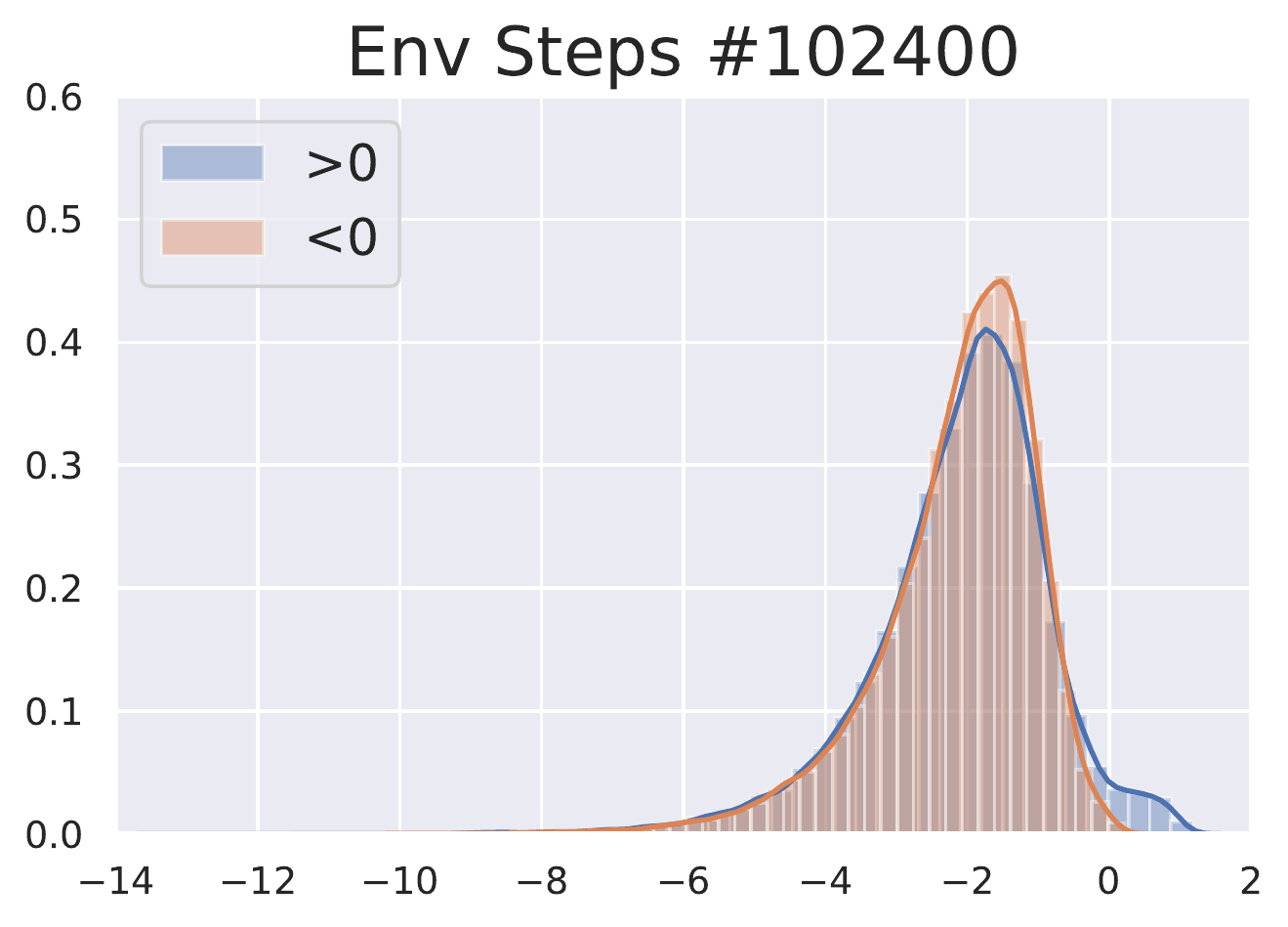}}
        \subfigure{\includegraphics[width=0.3\linewidth]{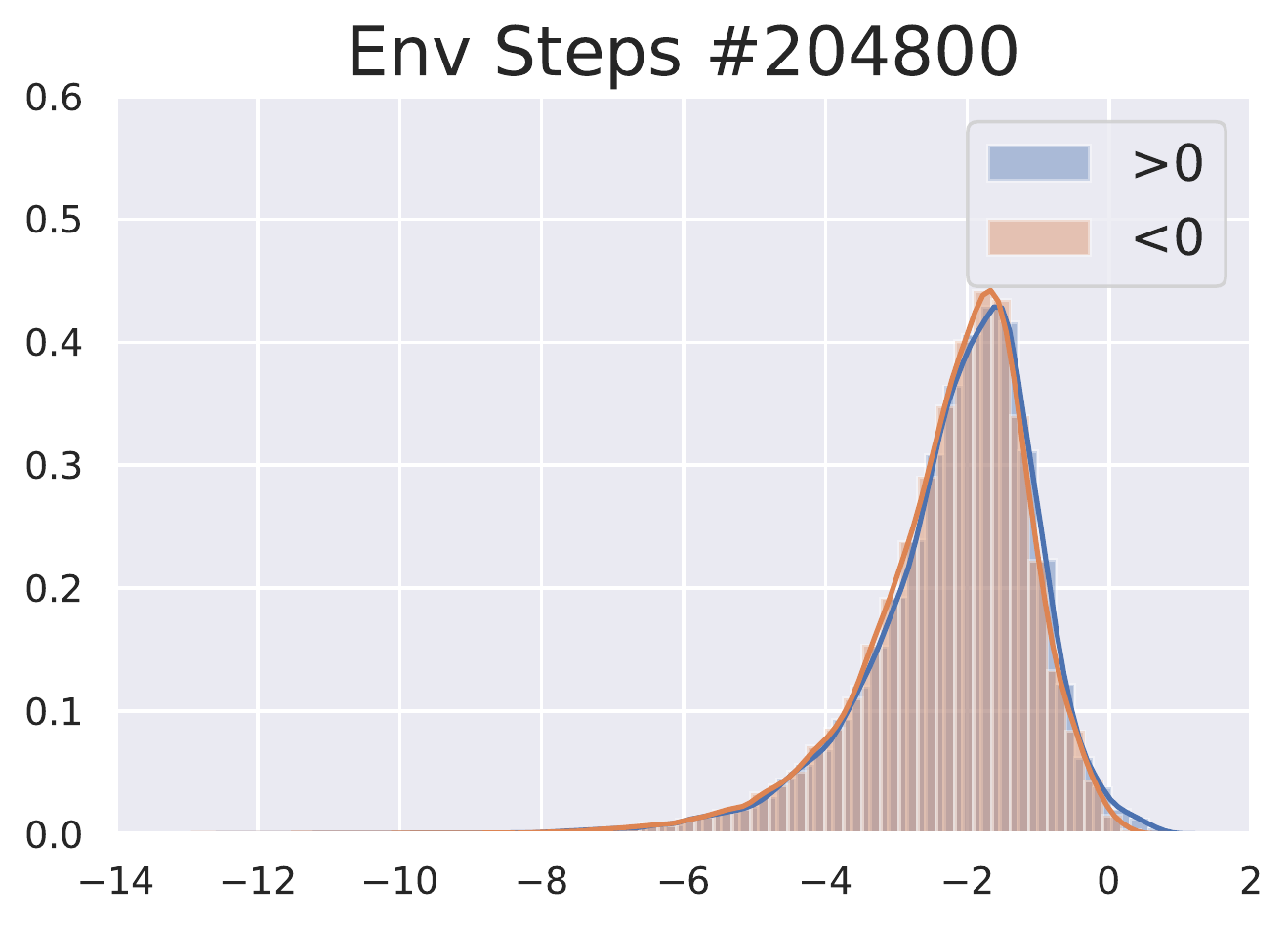}}
        \subfigure{\includegraphics[width=0.3\linewidth]{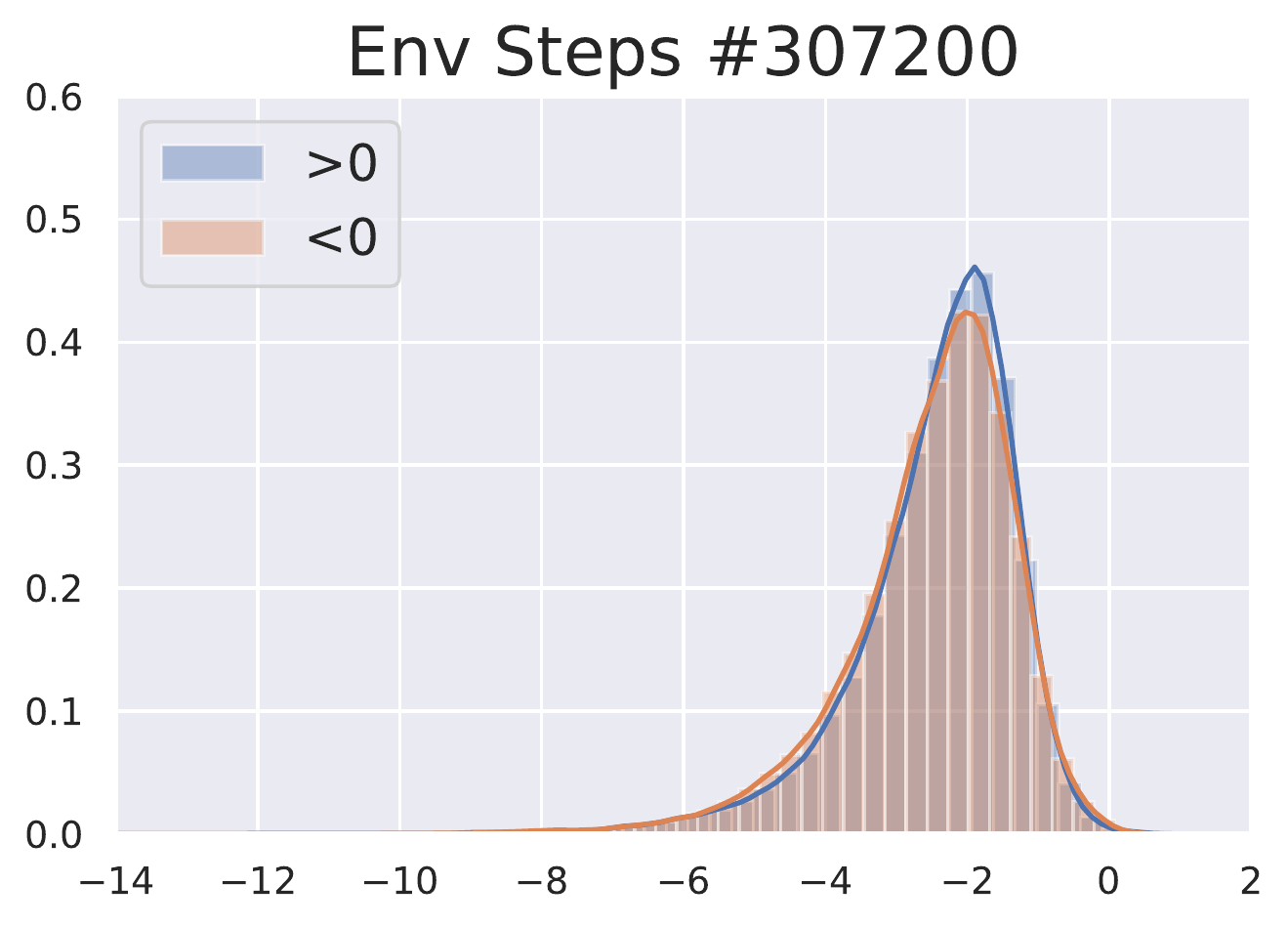}}
        \subfigure{\includegraphics[width=0.3\linewidth]{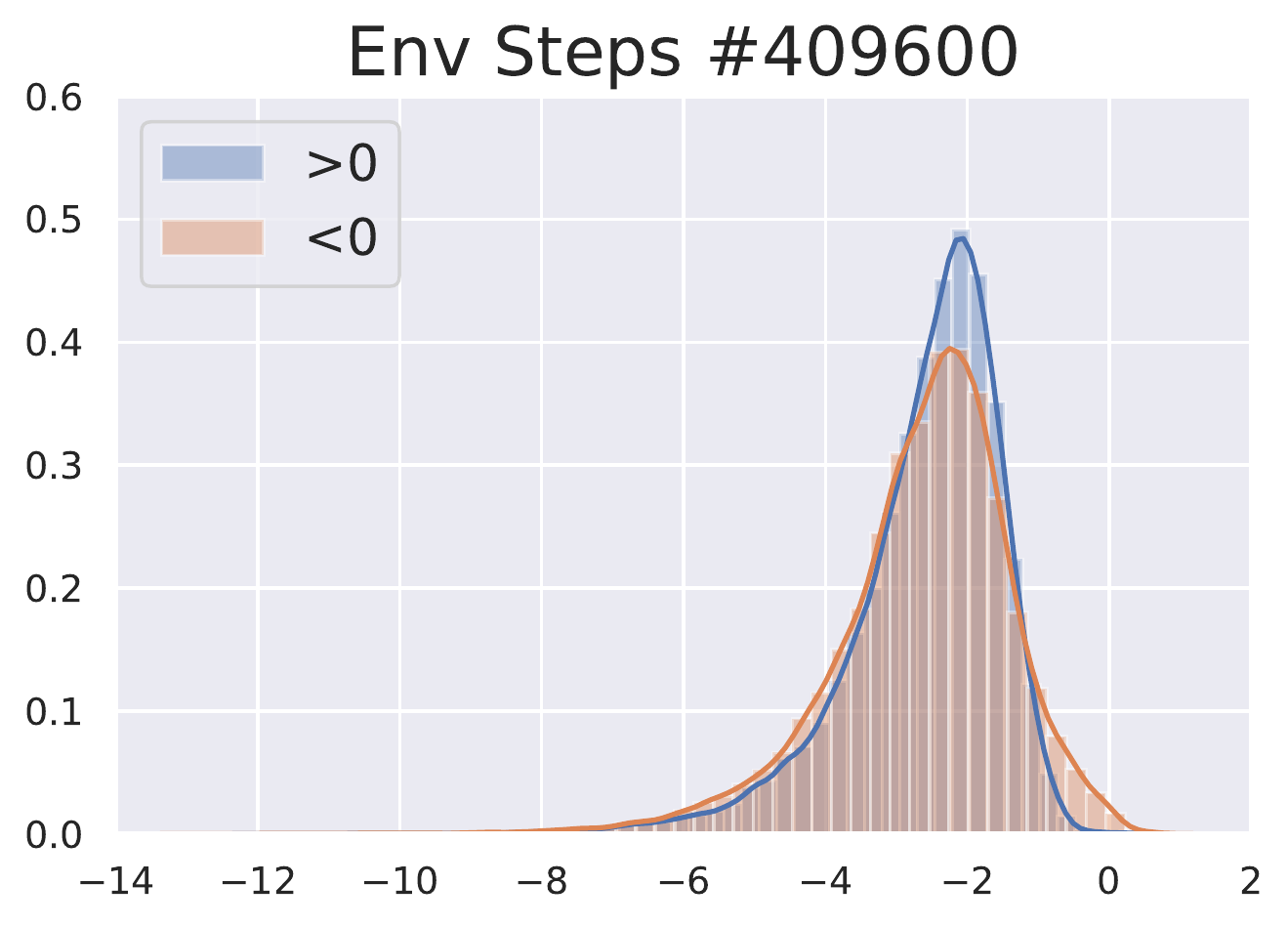}}
        \subfigure{\includegraphics[width=0.3\linewidth]{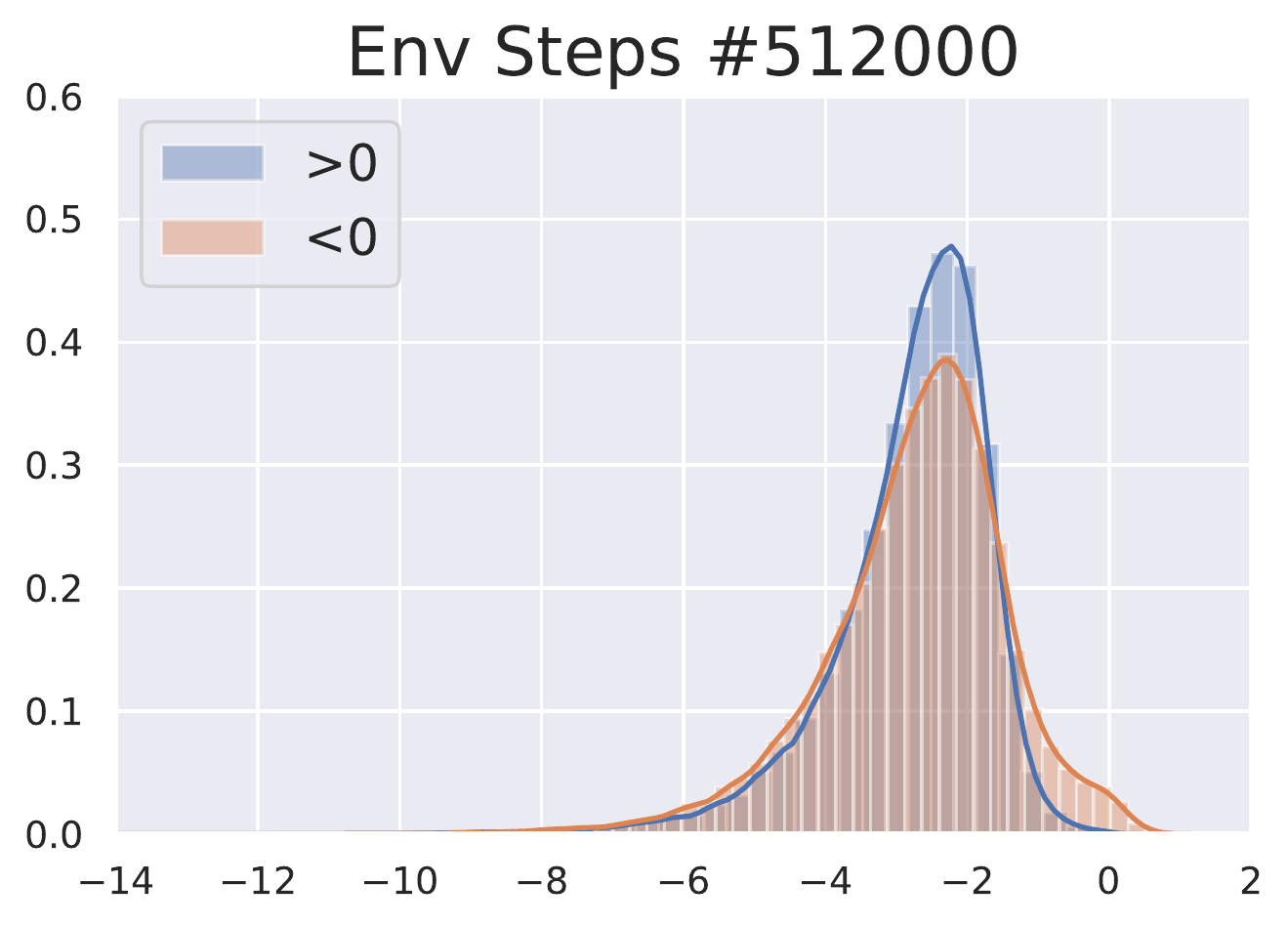}}
        \subfigure{\includegraphics[width=0.3\linewidth]{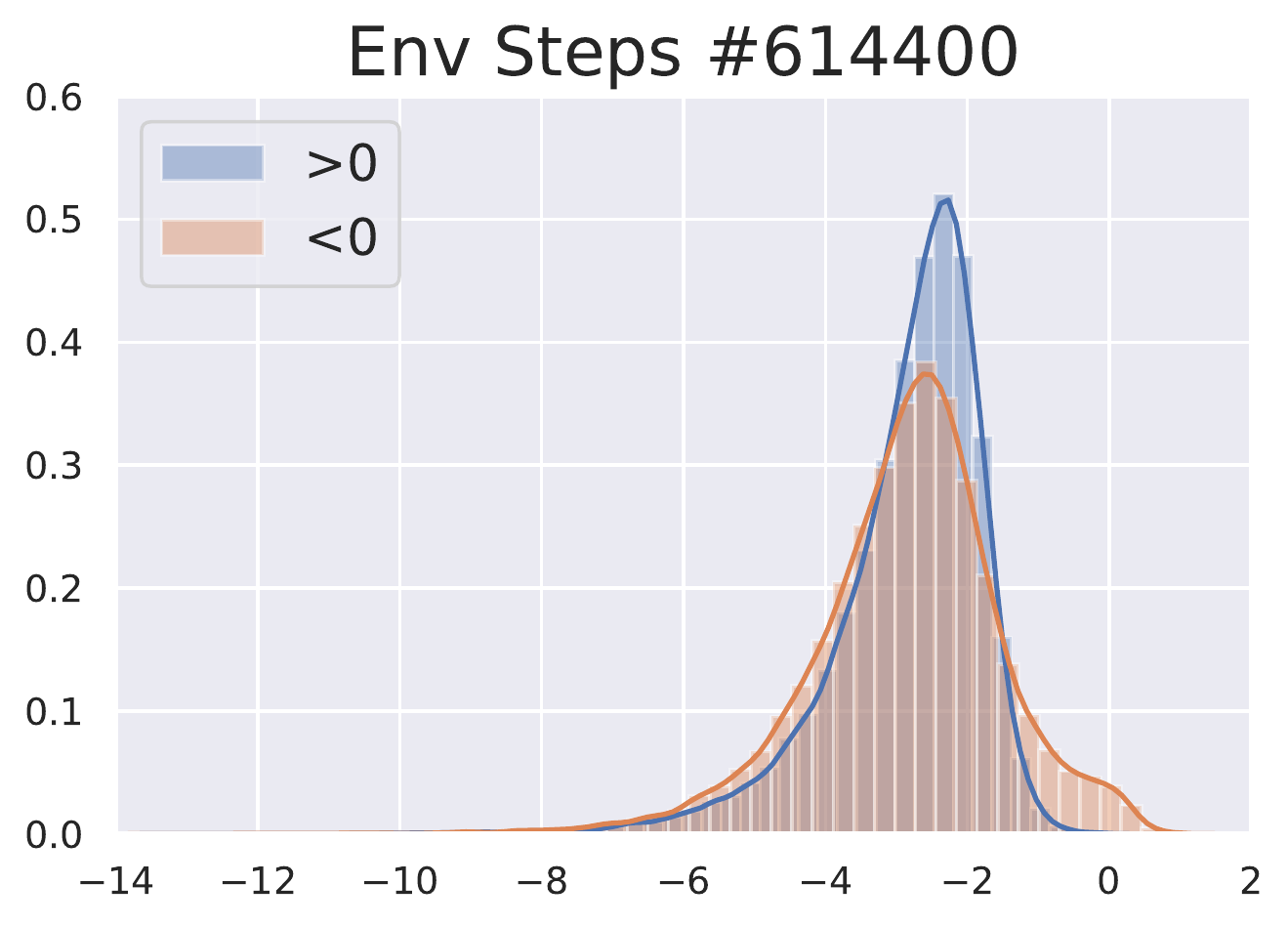}}
        \subfigure{\includegraphics[width=0.3\linewidth]{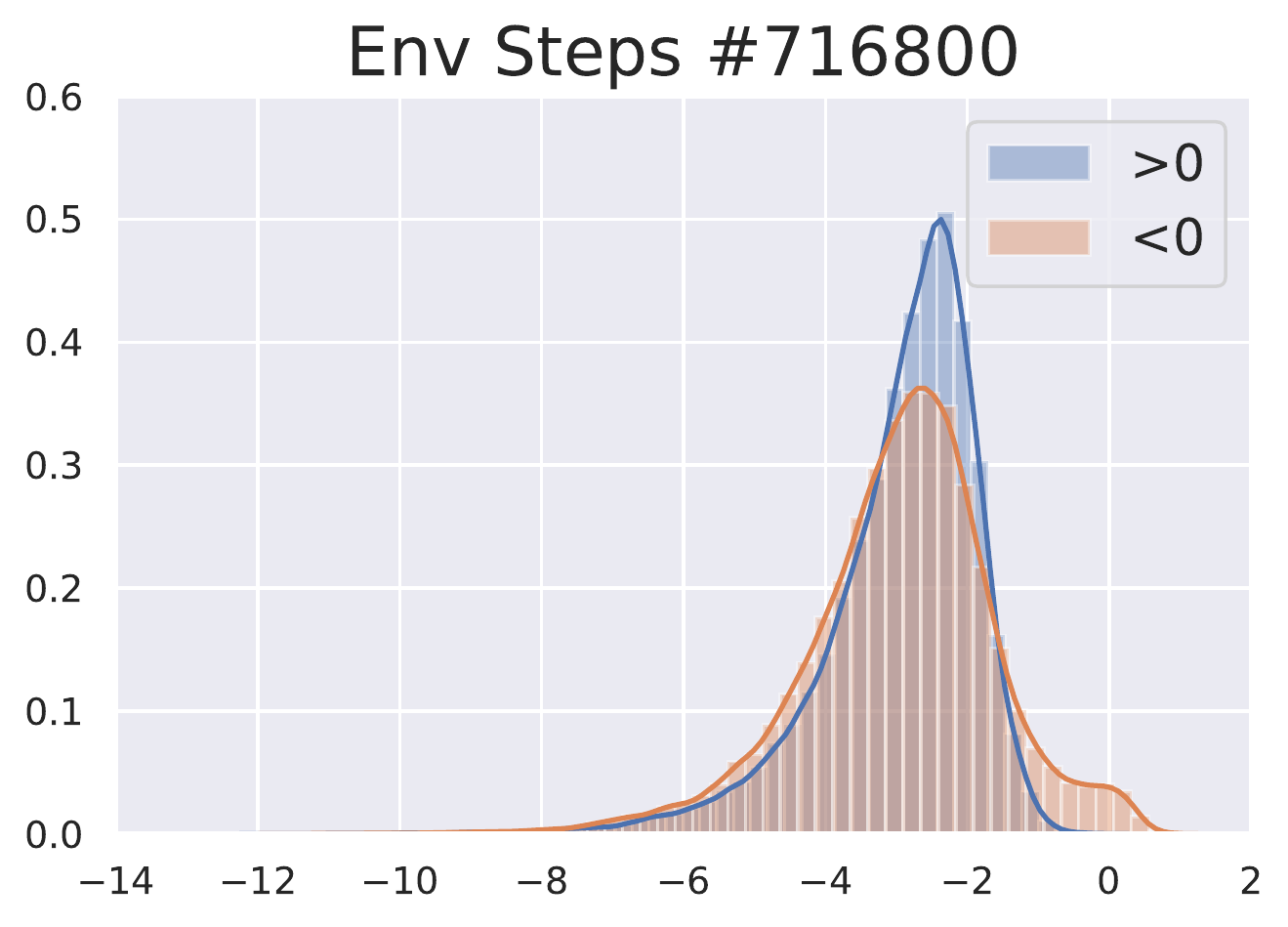}}
        \subfigure{\includegraphics[width=0.3\linewidth]{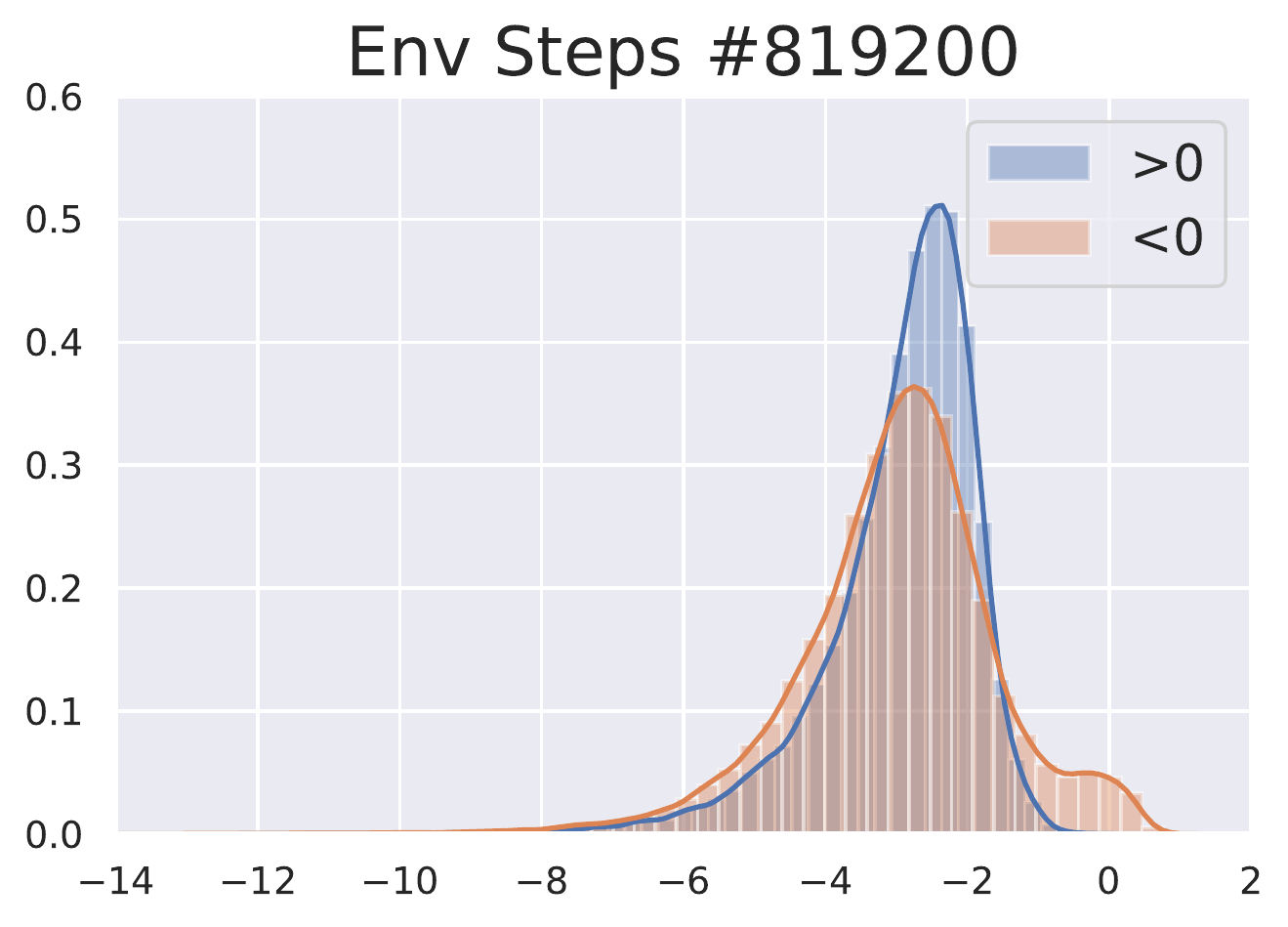}}
        \subfigure{\includegraphics[width=0.3\linewidth]{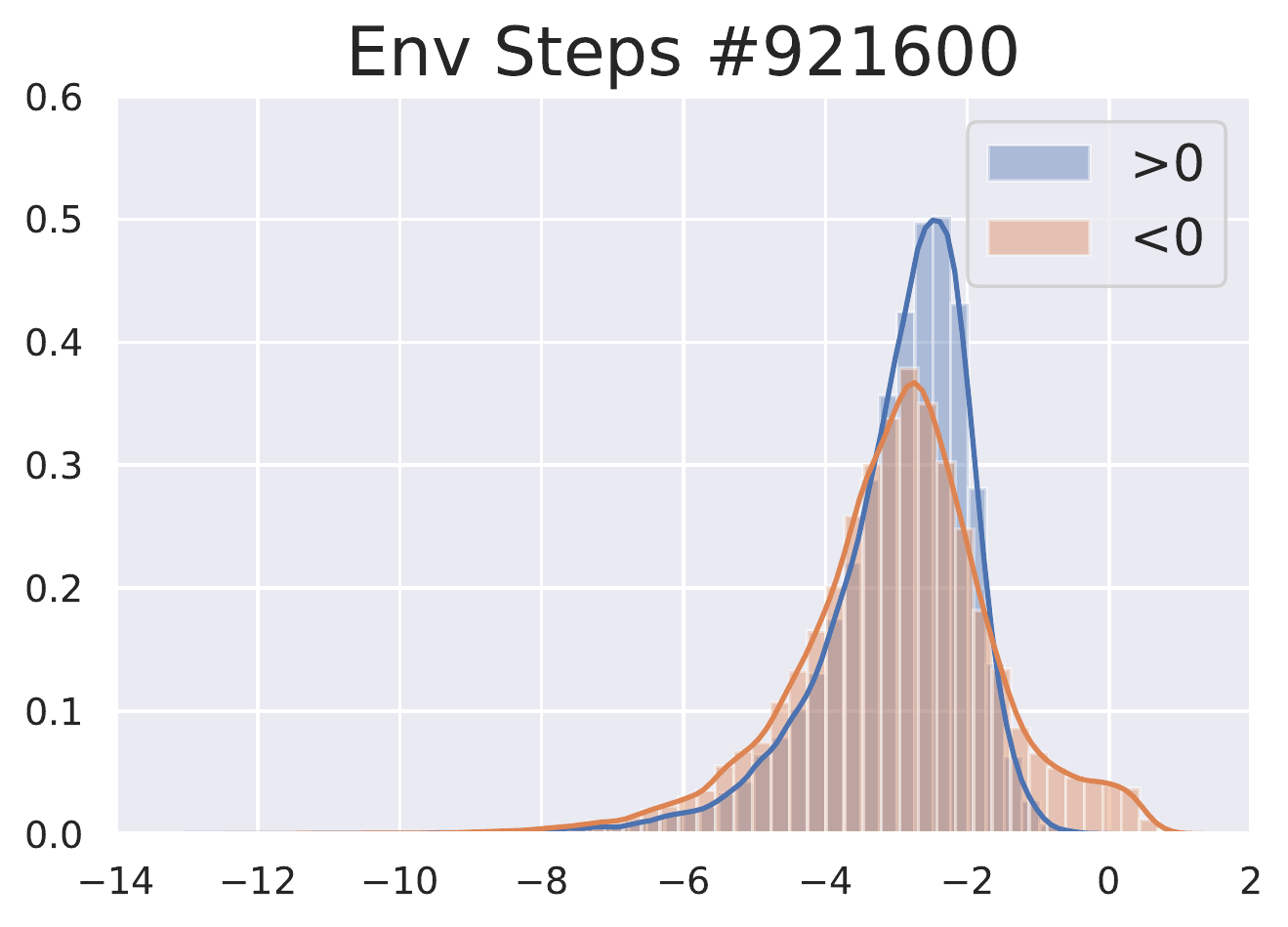}}
    \caption{ \update{Distribution of $\log(\abs{A_{\pi_\theta}})$  over training grouped by sign
of $\log(\abs{A_{\pi_\theta}})$ for HalfCheetah-v2 . To elaborate, we collect the advantages and separately plot the grouped advantages with their sign, i.e., we draw histograms separately for negative and positive advantages.   
As training proceeds, we clearly observe the increasing heavy-tailed behavior in negative advatanges as captured by the higher fraction of $\log(\abs{A_{\pi_\theta}})$ with large magnitude.  Moreover, the histograms for positive advantages (which resembel Gaussain pdf) stay almost the same throughout training. This highlights the particular heavy-tailed (outlier-rich) nature of negative advantages corroborating our experiments with kurtosis and tail-index estimators. }
    }\label{fig:App_adv2}
\end{figure}
\newpage

\section{Analysis with other estimators} \label{Appsec:estimator}

\subsection{On-policy gradient analysis}\label{subsec:App_on-policy}

\begin{figure}[H] 
\centering 
        \subfigure[]{\includegraphics[width=0.3\linewidth]{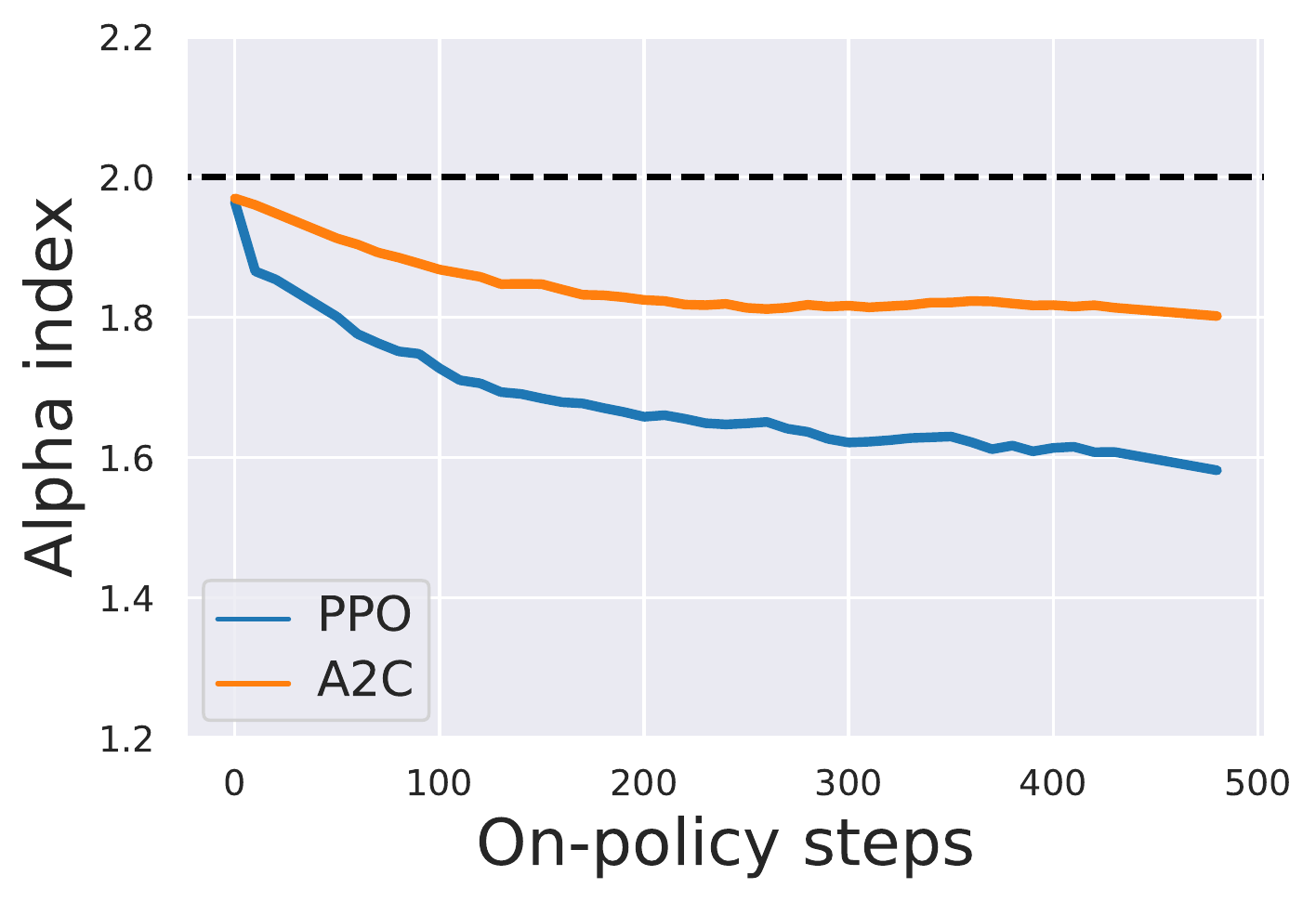}} \hfil
        \subfigure[]{\includegraphics[width=0.3\linewidth]{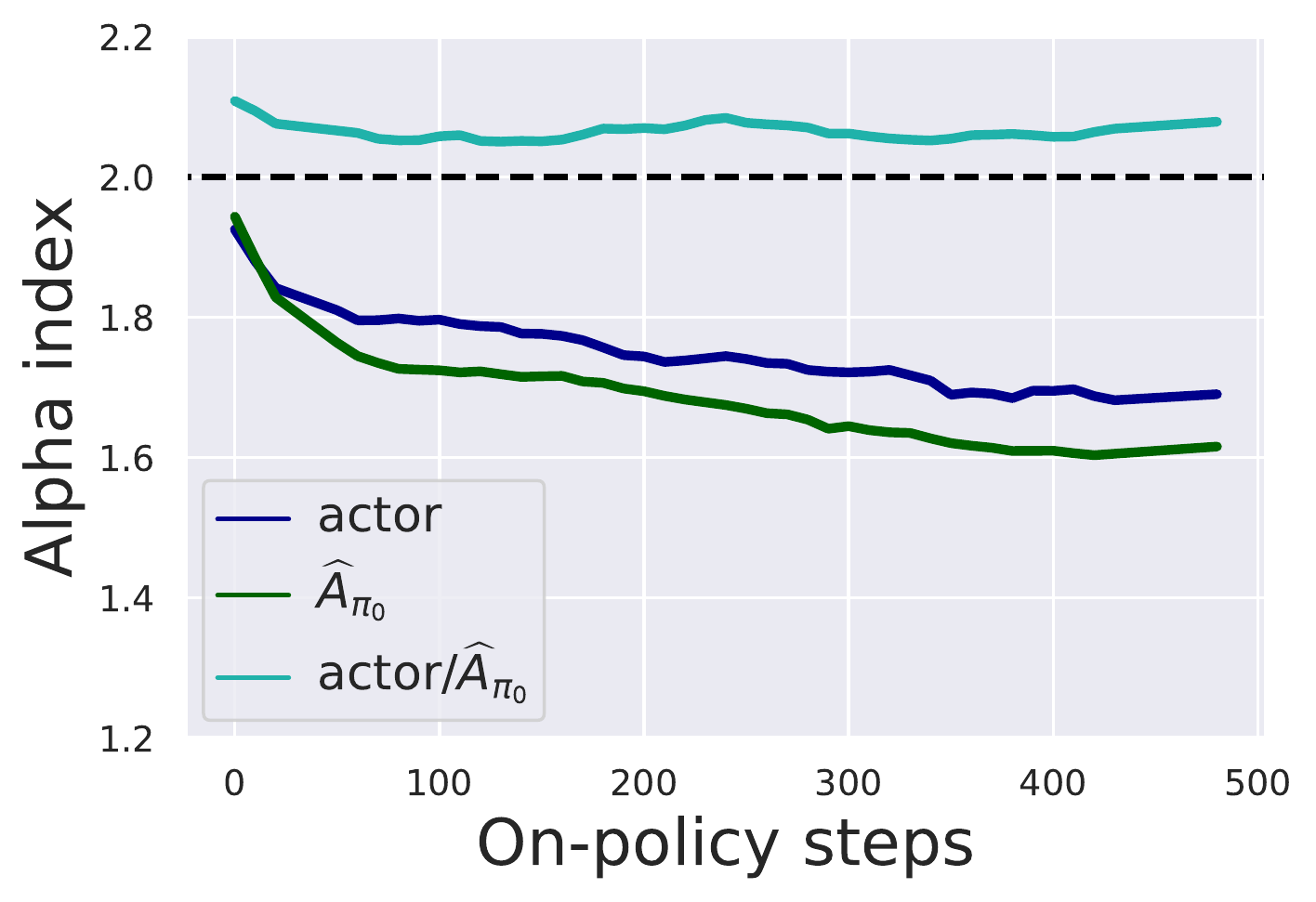}} \hfil 
        \subfigure[]{\includegraphics[width=0.3\linewidth]{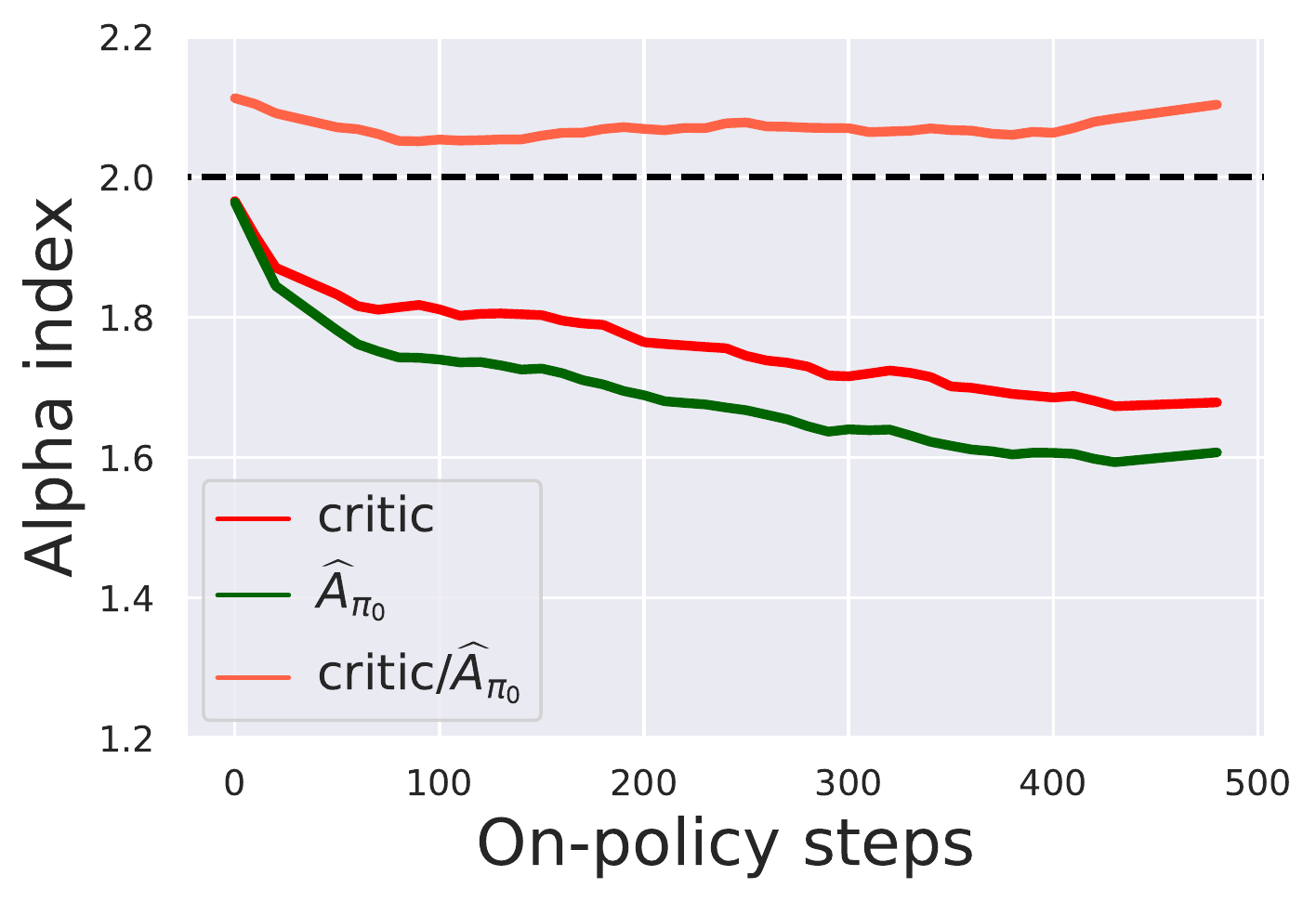}} \hfil
    \vspace{-6pt}
    \caption{ \textbf{Heavy-tailedness in PPO during on-policy iterations}. 
    All plots show mean alpha index aggregated over 8 MuJoCo environments.
    A decrease in alpha-index implies an increase in heavy-tailedness.  
    (a) Alpha index vs on-policy iterations for A2C and PPO. 
    Evidently, as training proceeds, 
    the gradients become more heavy-tailed for both the methods. 
    (b) Alpha index  vs on-policy iterations for actor networks in PPO. 
    (c) Alpha index  vs on-policy iterations for critic networks in PPO. 
    Both critic and actor gradients become more heavy-tailed 
    on-policy steps as the agent is trained. 
    Note that as the gradients become more heavy-tailed, we observe a corresponding
    increase of heavy-tailedness in the advantage estimates ($\hat A_{\pi_0}$ ) . 
    However, ``actor/$\hat A_{\pi_0}$'' and  ``critic/$\hat A_{\pi_0}$''
    (i.e., actor or critic gradient norm divided by GAE estimates)  
    remain light-tailed throughout the training. 
    }\label{fig:intro-alpha}
    \vspace{-15pt}
\end{figure}

\begin{figure}[H] 
    \centering 
        \subfigure[]{\includegraphics[width=0.3\linewidth]{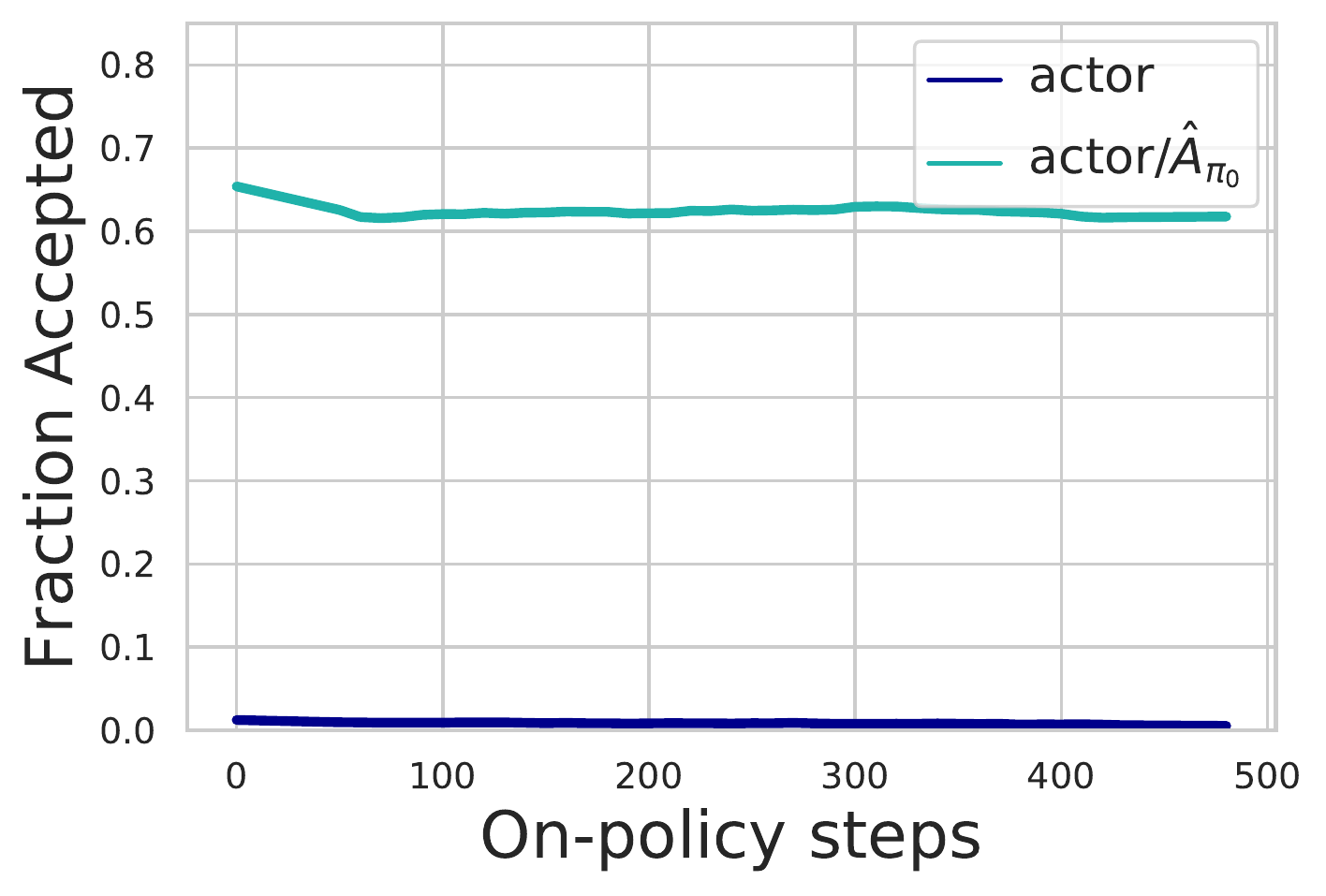}} \hfil
        \vspace{-5pt}
        \subfigure[]{\includegraphics[width=0.3\linewidth]{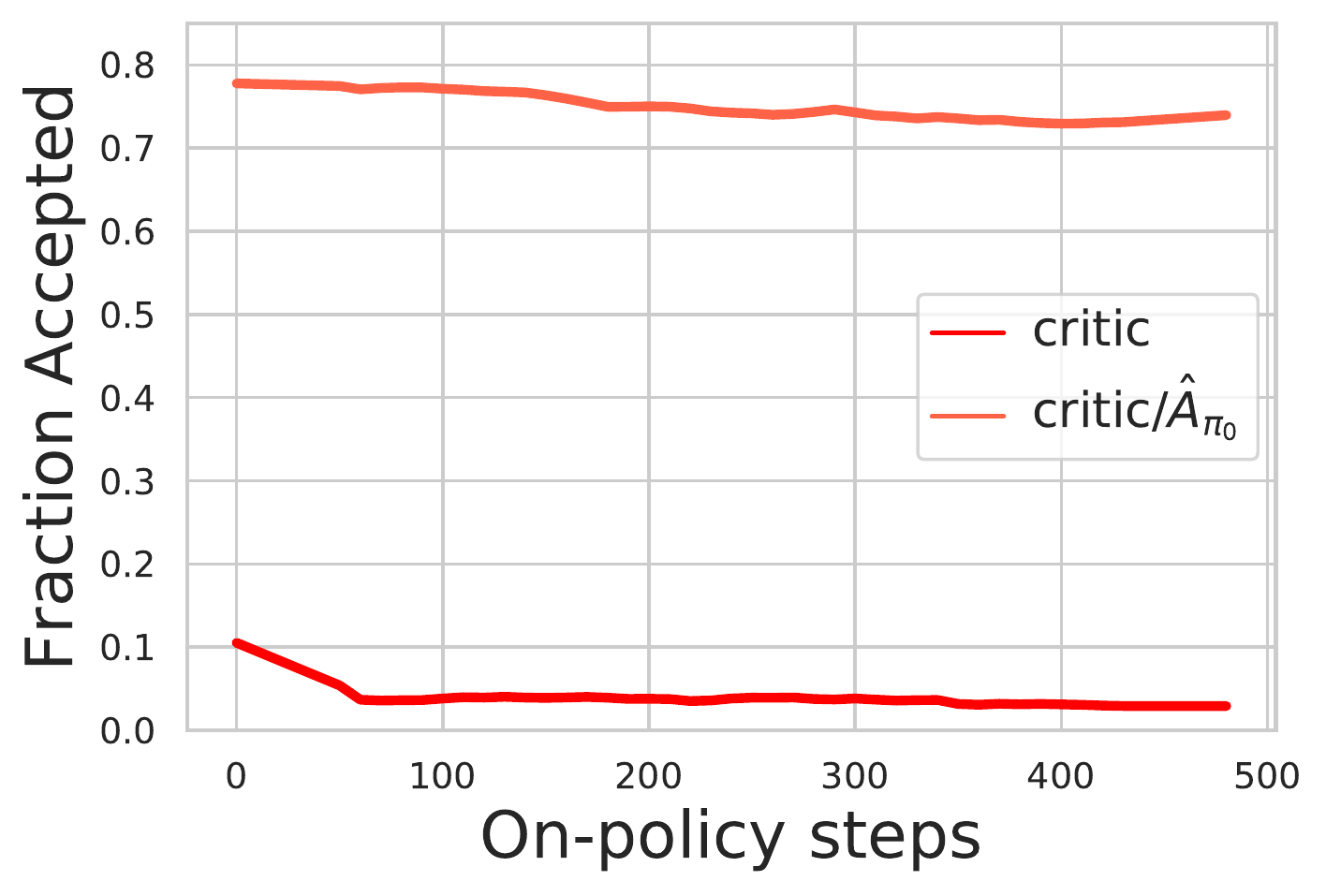}}
    \vspace{-6pt}
    \caption{ \textbf{Heavy-tailedness in PPO during on-policy iterations}. 
    All plots show mean fraction of directions accepted by Anderon-Darling test over 8 MuJoCo environments.
    A higher accepted fraction indicates a Gaussian behavior. 
    (b) Fraction accepted  vs on-policy iterations for actor networks in PPO. 
    (c) Fraction accepted vs on-policy iterations for critic networks in PPO. 
    Both critic and actor gradients remain non-Gaussian as the agent is trained. 
    However, ``actor/$\hat A_{\pi_0}$'' and  ``critic/$\hat A_{\pi_0}$''
    (i.e., actor or critic gradient norm divided by GAE estimates)  
    have fairly high fraction of directions accepted, hinting their Gaussian nature.   
    }\label{fig:intro-gaussian}
    \vspace{-5pt}
\end{figure}

\subsection{Off-policy gradient analysis}\label{subsec:App_off-policy}

\begin{figure}[H] 
    \centering
        \subfigure{\includegraphics[width=0.3\linewidth]{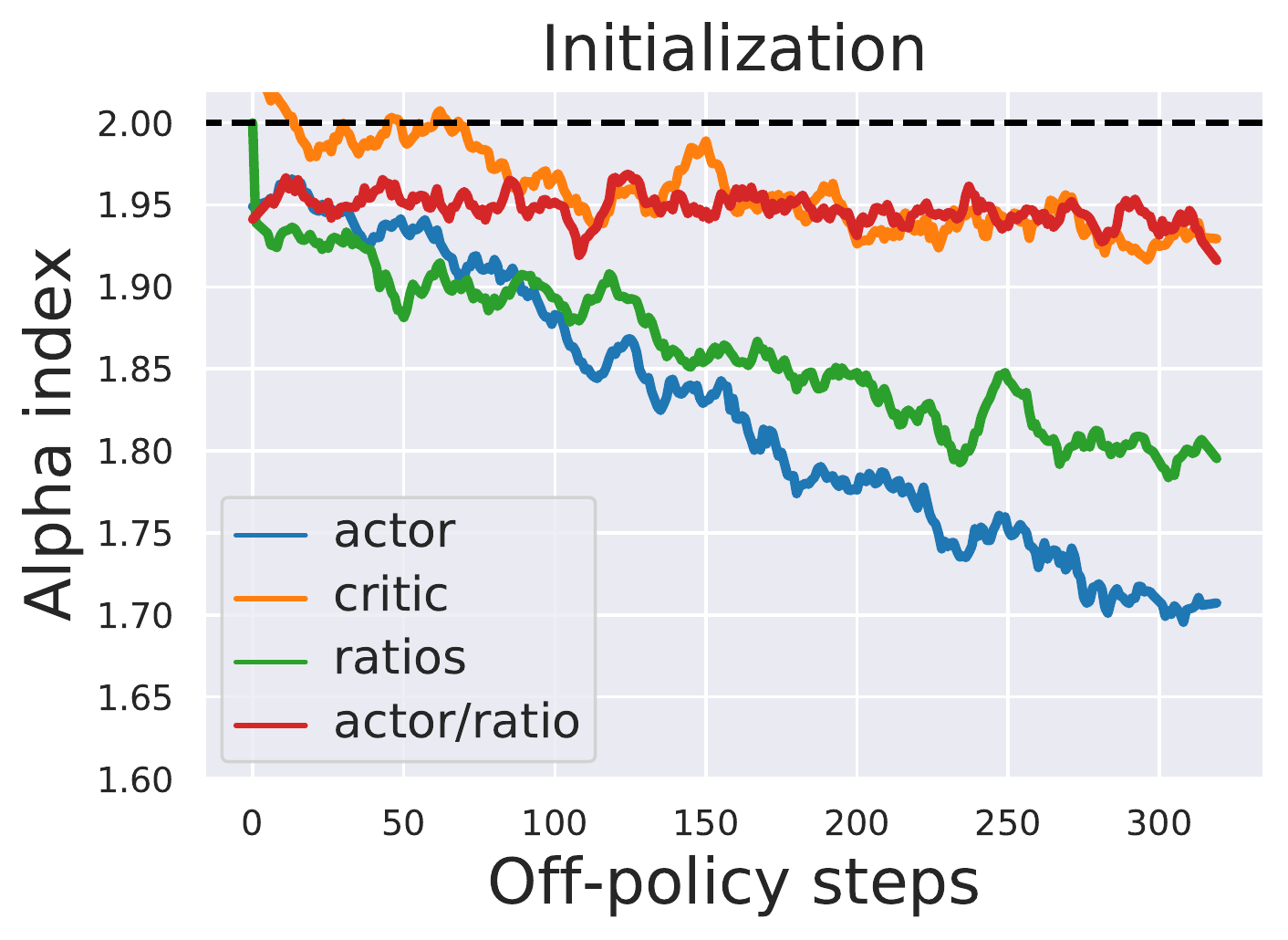}}\hfil
        \subfigure{\includegraphics[width=0.3\linewidth]{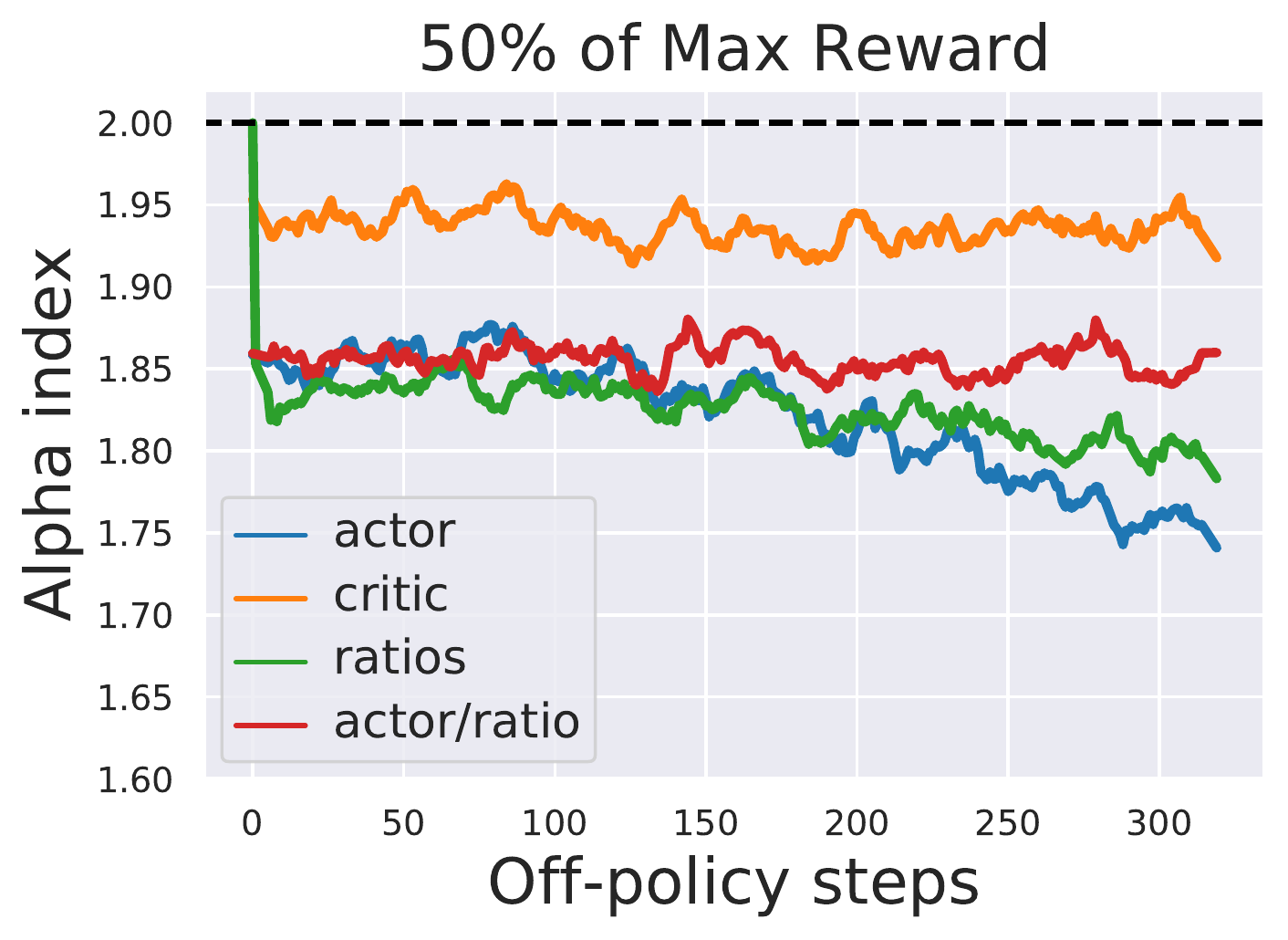}}\hfil
        \subfigure{ \includegraphics[width=0.3\linewidth]{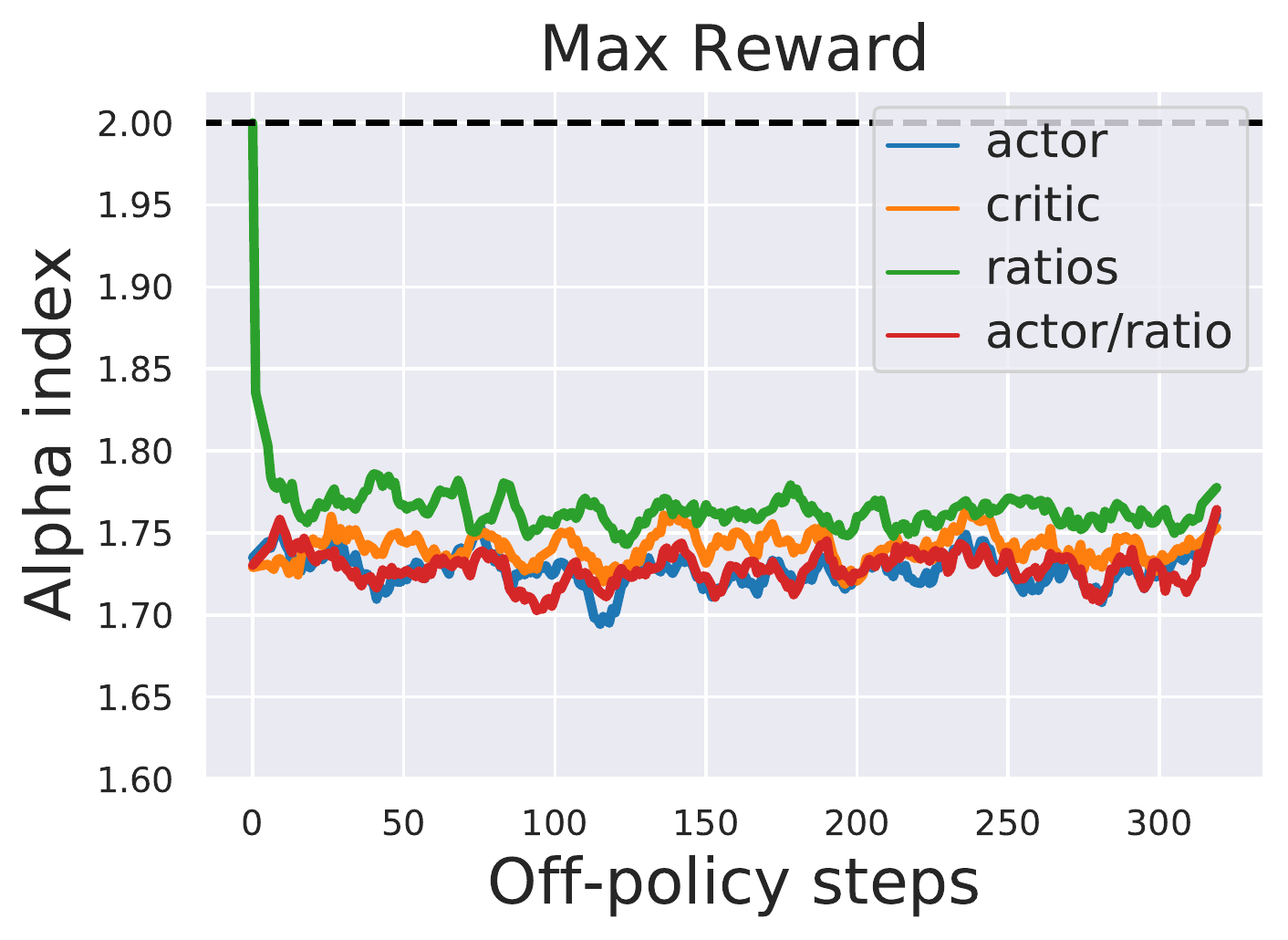}}\hfil
        \par\medskip
       
    \caption{ \textbf{Heavy-tailedness in PPO-\textsc{NoClip} during off-policy steps} at various stages of training iterations in MuJoCo environments. All plots show mean alpha index aggregated over 8 Mujoco environments. 
    A decrease in alpha index implies an increase in heavy-tailedness.
    As off-policyness increases, the actor gradients get substantially heavy-tailed. This trend is corroborated by the increase of heavy-tailedness in ratios. Moreover, consistently we observe that the heavy-tailedness in ``actor/ratios'' stays constant. 
    While initially during training, the heavy-tailedness in the ratio's increases substantially, during later stages the increase tapers off. The overall increase across training iterations is explained by the induced heavy-tailedness in the advantage estimates (cf. Sec.~\ref{subsec:on-policy}). 
     }\label{fig:ppo-offpolicy} 
     \vspace{-20pt}
\end{figure}

\newpage
\section{Hyperparameter settings and Rewards curves on individual enviornments} \label{Appsec:rewards}

\begin{table}[h]
\centering
\begin{tabular}{lrrr} 
  \toprule
  Hyperparameter & Values  \\ \midrule
  Steps per PPO iteration & 2048\\
  Number of minibatches & 32\\
  PPO learning rate & 0.0003 \\ 
  \textsc{Robust}-PPO-\textsc{NoClip} learning rate & 0.00008 \\
  PPO-\textsc{NoClip} learning rate & 0.00008 \\
  Discount factor $\gamma$ & 0.99\\
  GAE parameter $\lambda$ & 0.95\\ %
  Entropy loss coefficient & 0.0\\
  PPO value loss coefficient & 2.0 \\
  \textsc{Robust}-PPO-\textsc{NoClip} value loss coefficient & 2.0 \\
  PPO-\textsc{NoClip} value loss coefficient & 2.0 \\
  Max global L2 gradient norm (only for PPO) & 0.5\\
  Clipping coefficient (only for PPO) & 0.2\\
  Policy epochs & 10\\
  Value epochs & 10\\
  GMOM number of blocks & 8 \\
  GMOM Weiszfeld iterations & 100 \\
 \bottomrule
\end{tabular}
\caption{Hyperparameter settings. Sweeps were run over learning rates \{ 0.000025, 0.00005, 0.000075, 0.00008, 0.00009 , 0.0001, 0.0003, 0.0004 \} and value loss coefficient \{ 0.1, 0.5, 1.0, 2.0, 10.0\} with \update{30 random seeds} per learning rate.}
\label{table:hyperparameters}
\end{table}

\begin{figure}[H] 
    \centering
        
        \subfigure{ \includegraphics[width=0.24\linewidth]{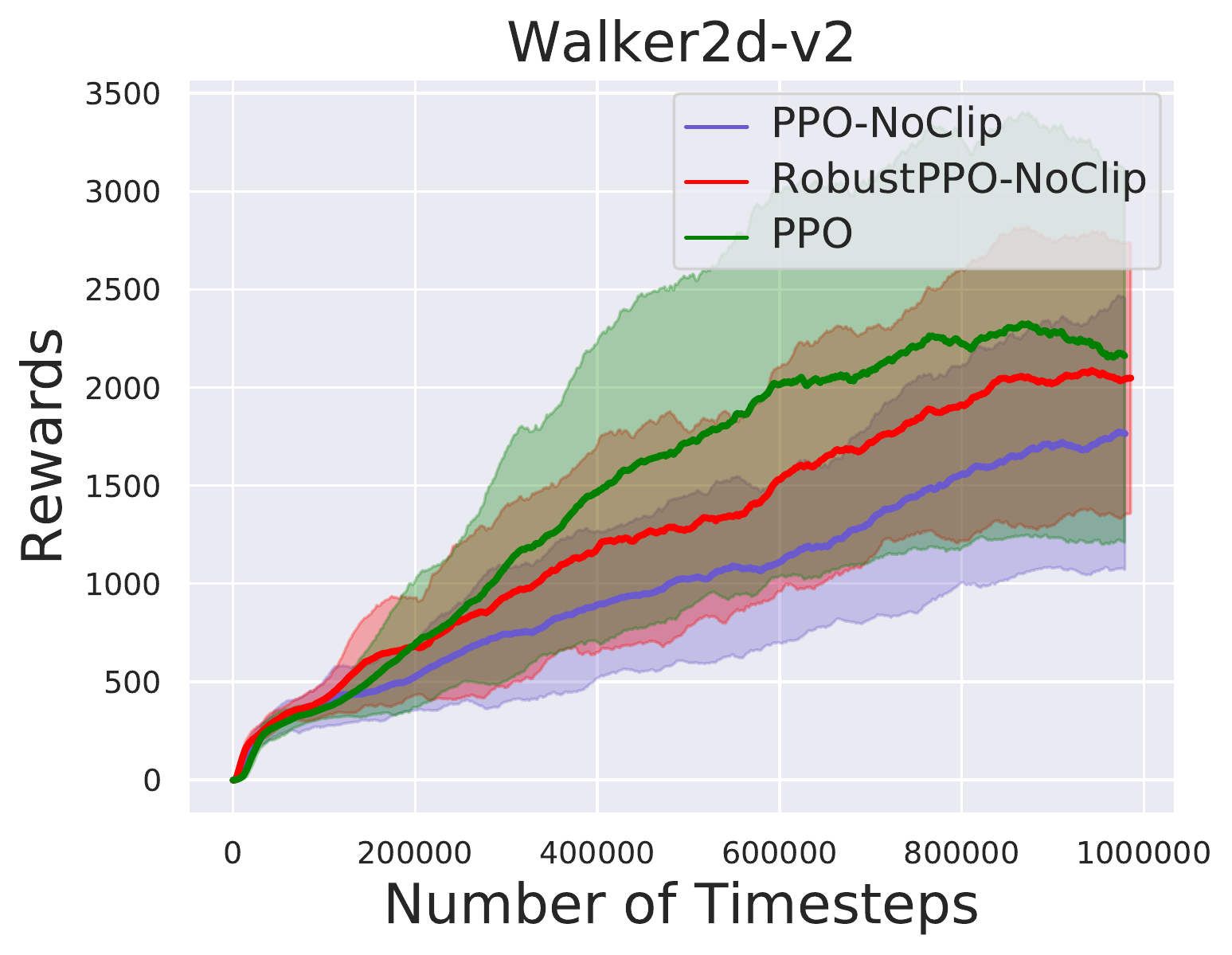}}\hfil
        \subfigure{ \includegraphics[width=0.24\linewidth]{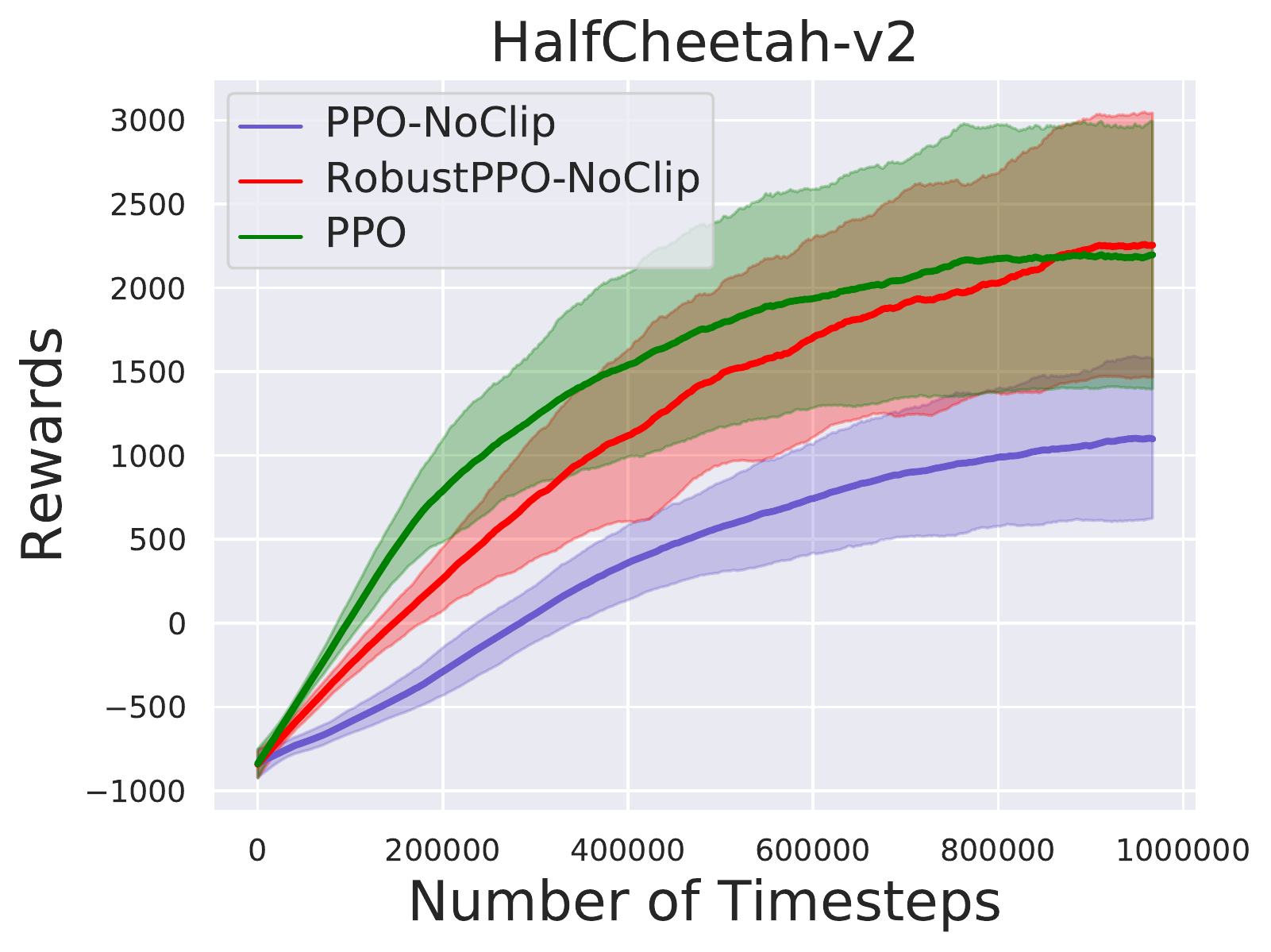}}\hfil
        \subfigure{\includegraphics[width=0.24\linewidth]{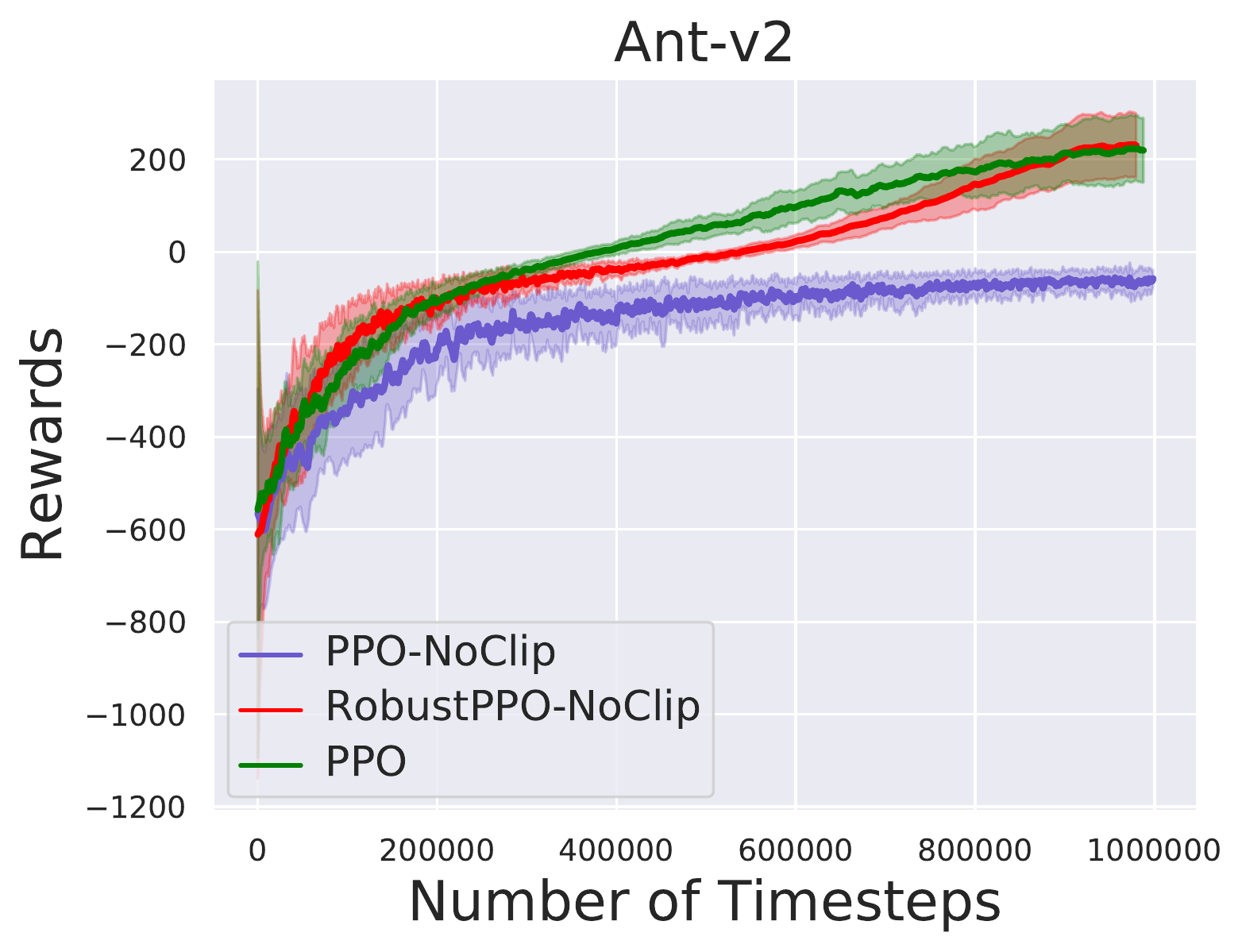}}\hfil
        \subfigure{ \includegraphics[width=0.24\linewidth]{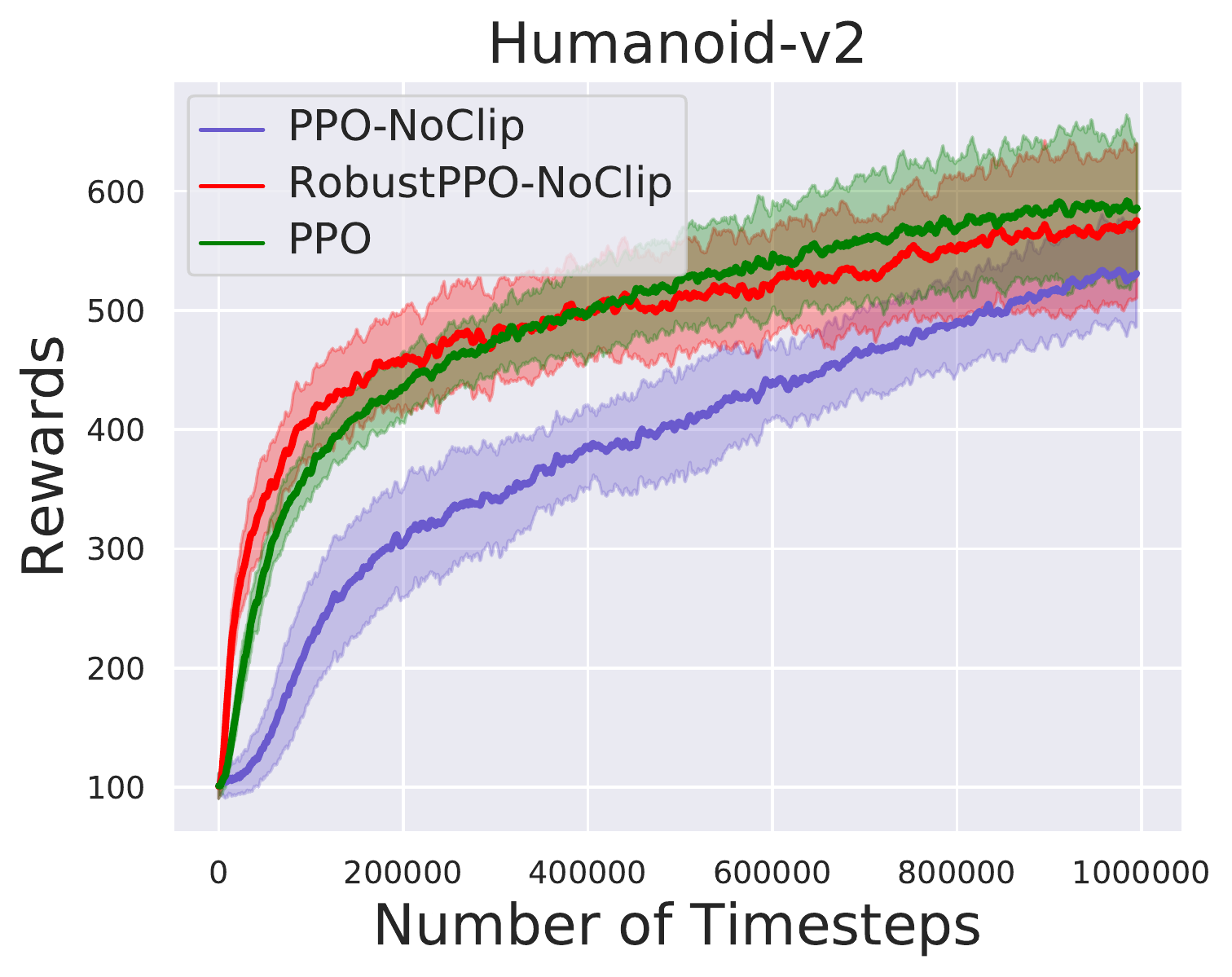}}\hfil
        \par\medskip
        
        \subfigure{ \includegraphics[width=0.24\linewidth]{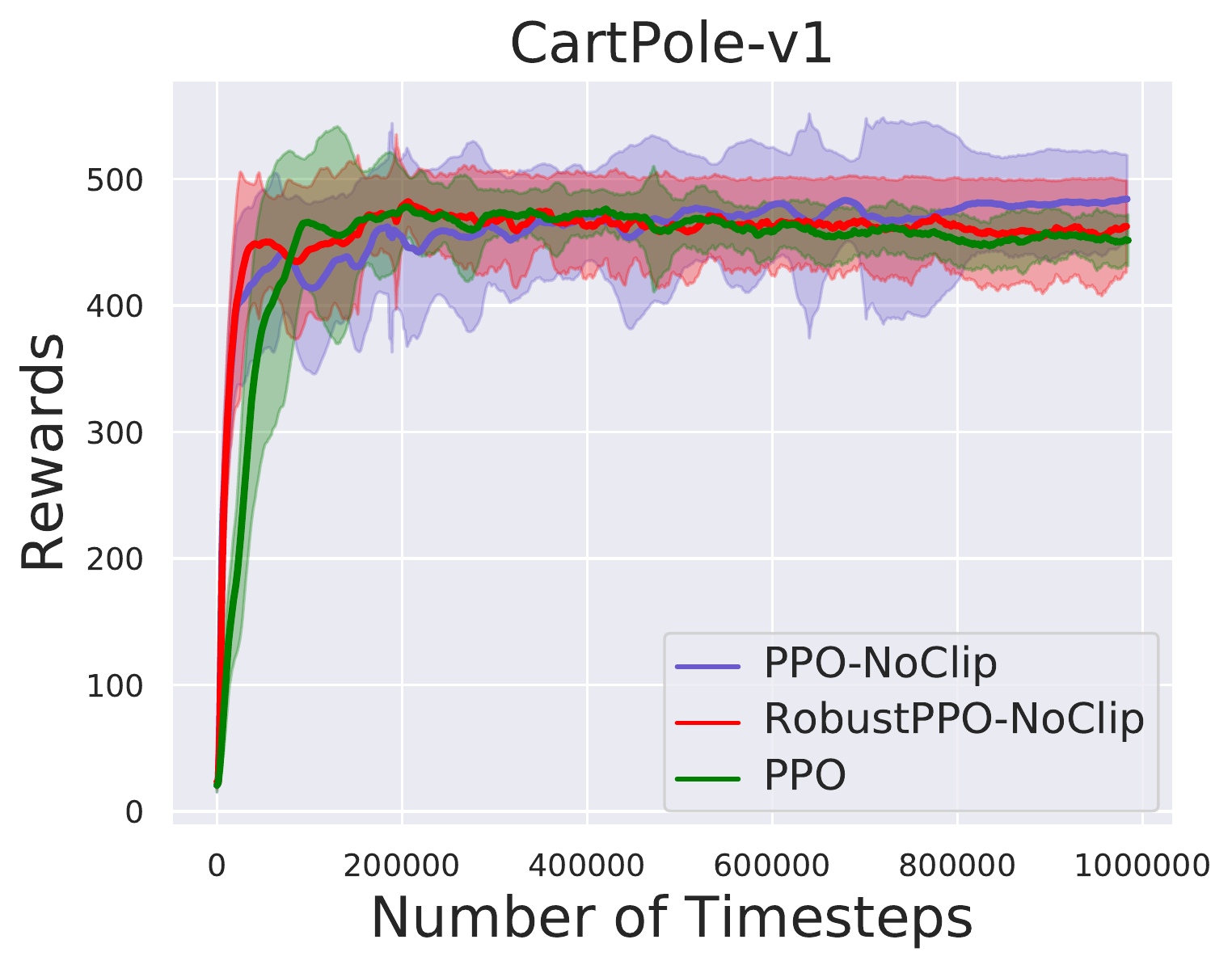}}\hfil
        \subfigure{\includegraphics[width=0.24\linewidth]{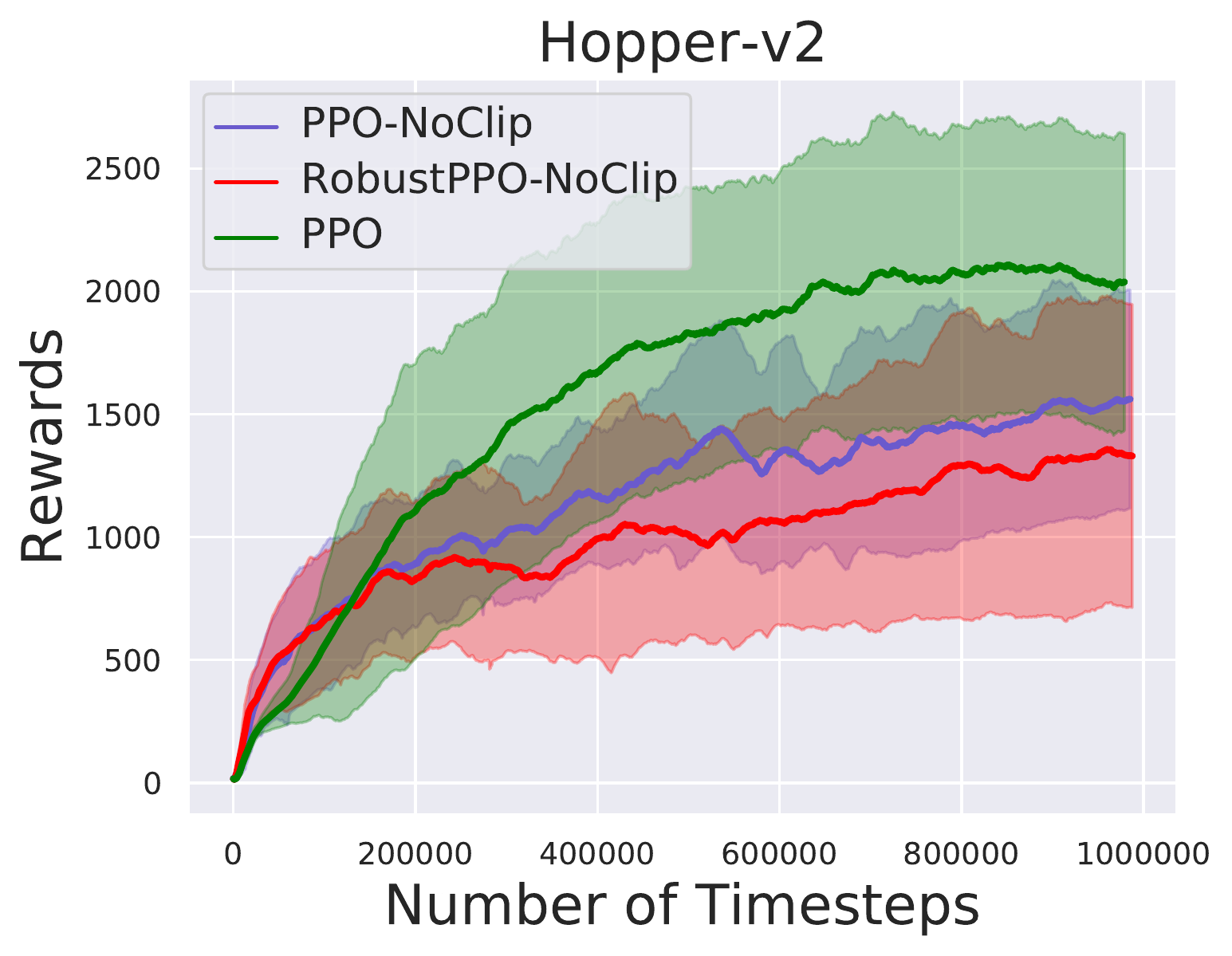}}\hfil
        \subfigure{ \includegraphics[width=0.24\linewidth]{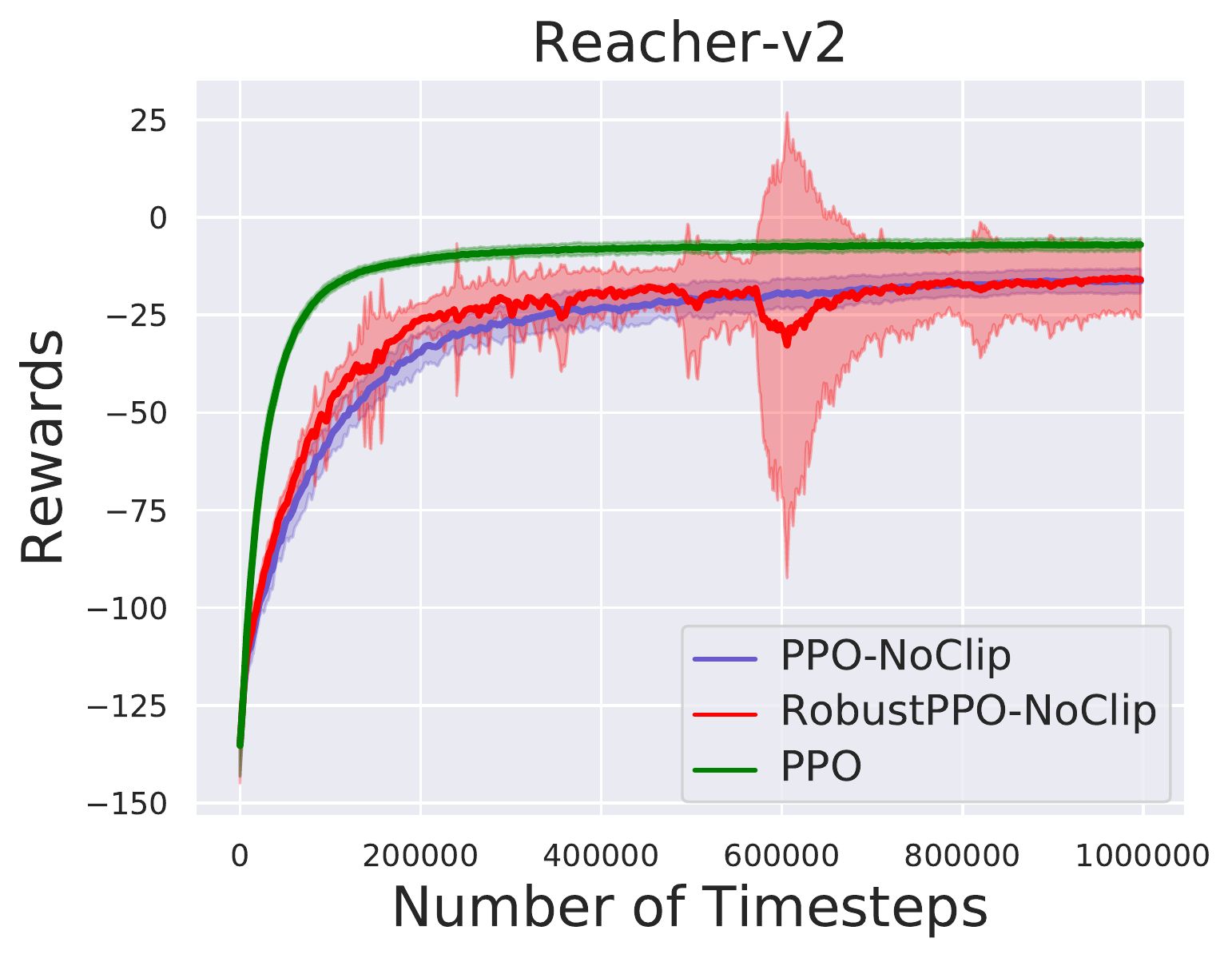}}\hfil
        \subfigure{ \includegraphics[width=0.24\linewidth]{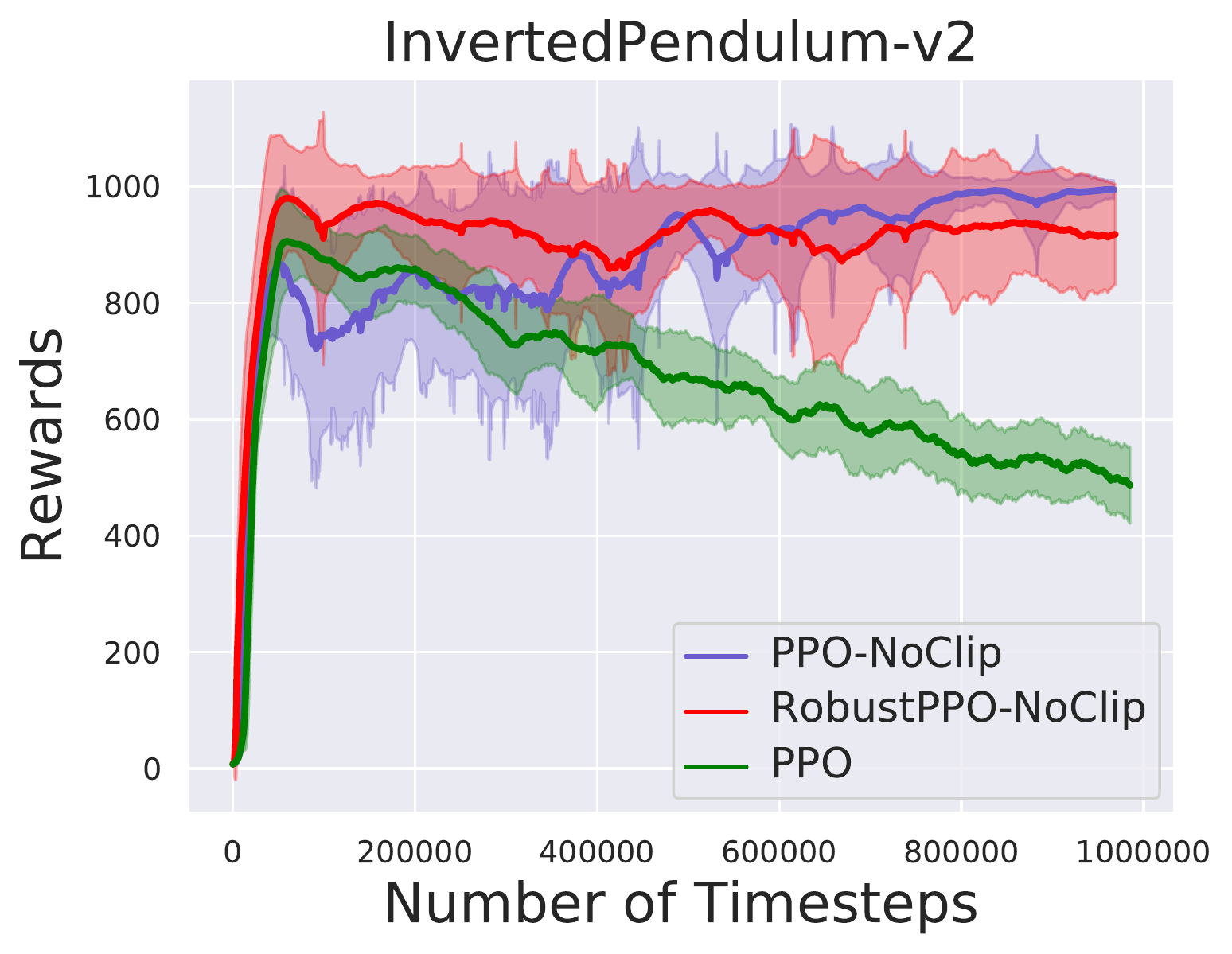}}\hfil
        \par\medskip

    \caption{  \update{\textbf{Reward curves as training progresses in 8 different Mujoco Environments} aggregated across 30 random seeds and for hyperparameter setting tabulated in Table~\ref{table:hyperparameters}. The shaded region denotes the one standard deviation across seeds. We observe that except in Hopper-v2 environment, the mean reward with \textsc{Robust}-PPO-\textsc{NoClip} is significantly better than PPO-\textsc{NoClip} and close to that achieved by PPO with optimal hyperparameters. Aggregated results shown in Fig.~\ref{fig:normalized-rewards}.}
    }\label{figure:mujoco-rewards} 
\end{figure}

\pagebreak
\section{ \update{Analysis on individual enviornments.} } \label{Appsec:individual}

\update{Overall, in the figures below, we show that the trends observed in aggregated plots in Section~\ref{sec:analysis} with Kurtosis hold true on individual environments. While the degree of heavy-tailedness varies in different environments, the trend of increase in heavy-tailedness remains the same.}

\begin{figure}[H] 
    \centering
        \subfigure{\includegraphics[width=0.24\linewidth]{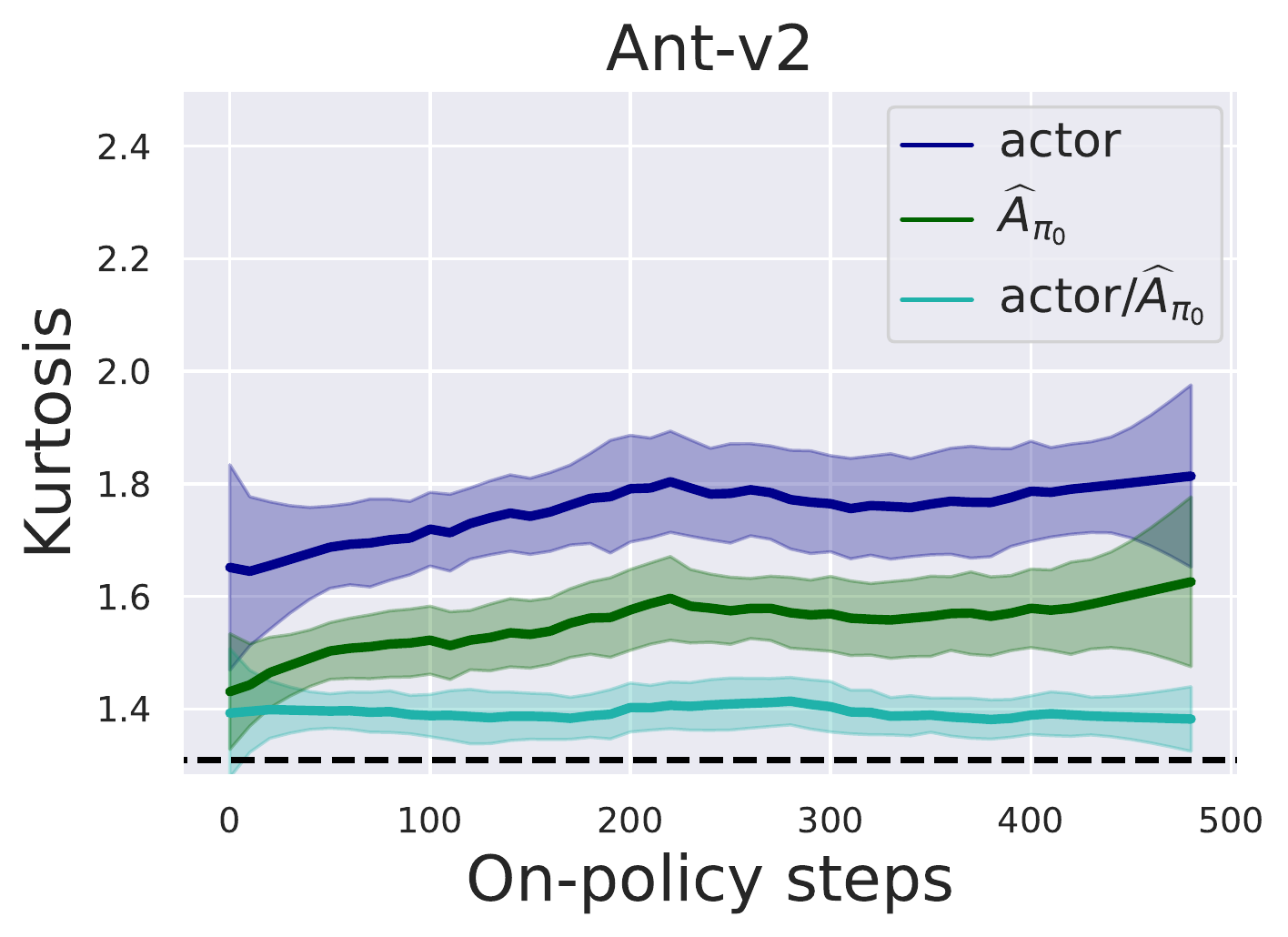}}\hfil
        \subfigure{\includegraphics[width=0.24\linewidth]{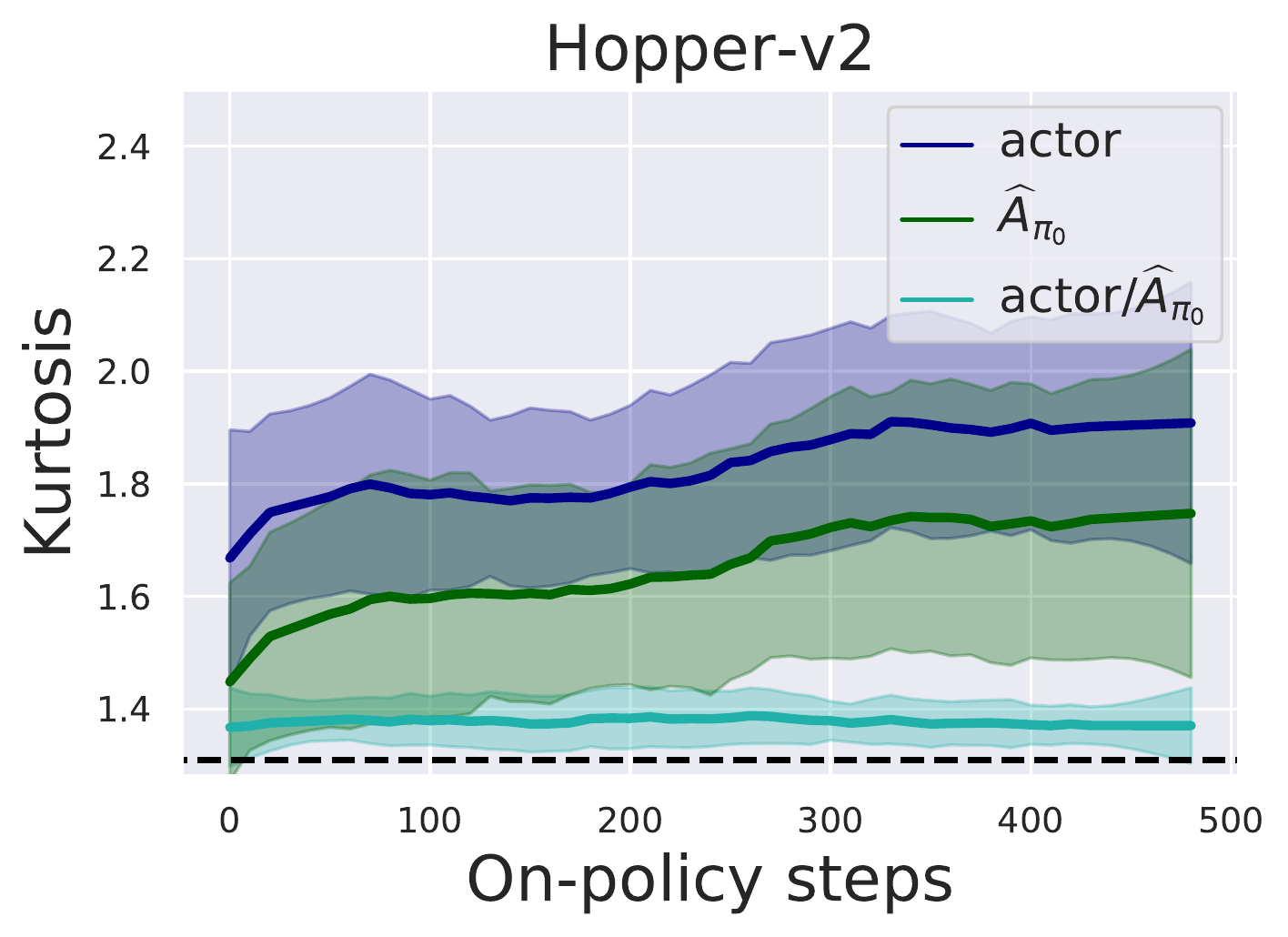}}\hfil
        \subfigure{ \includegraphics[width=0.24\linewidth]{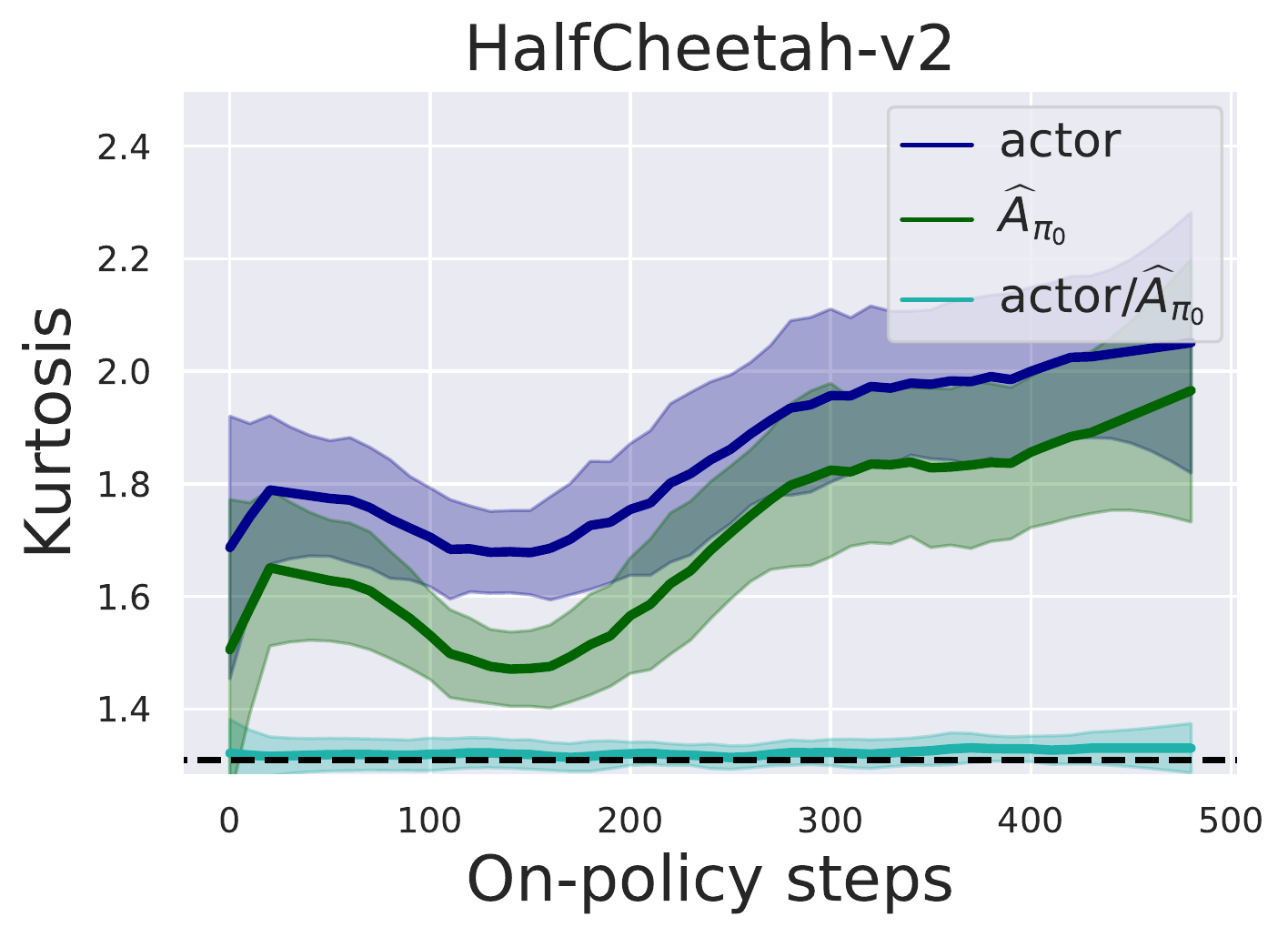}}\hfil
        \subfigure{\includegraphics[width=0.24\linewidth]{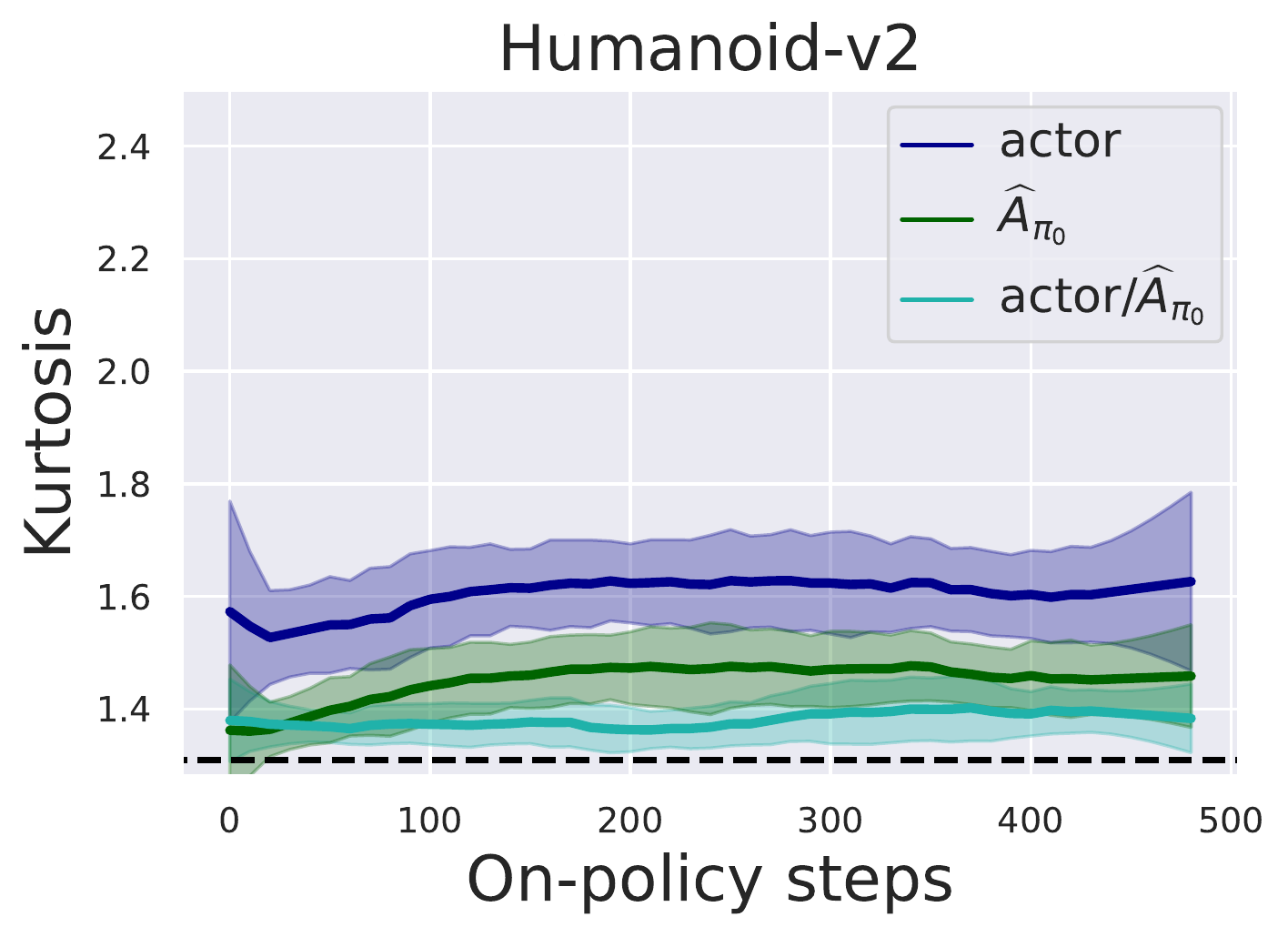}}\hfil
        \par\medskip

        \subfigure{\includegraphics[width=0.24\linewidth]{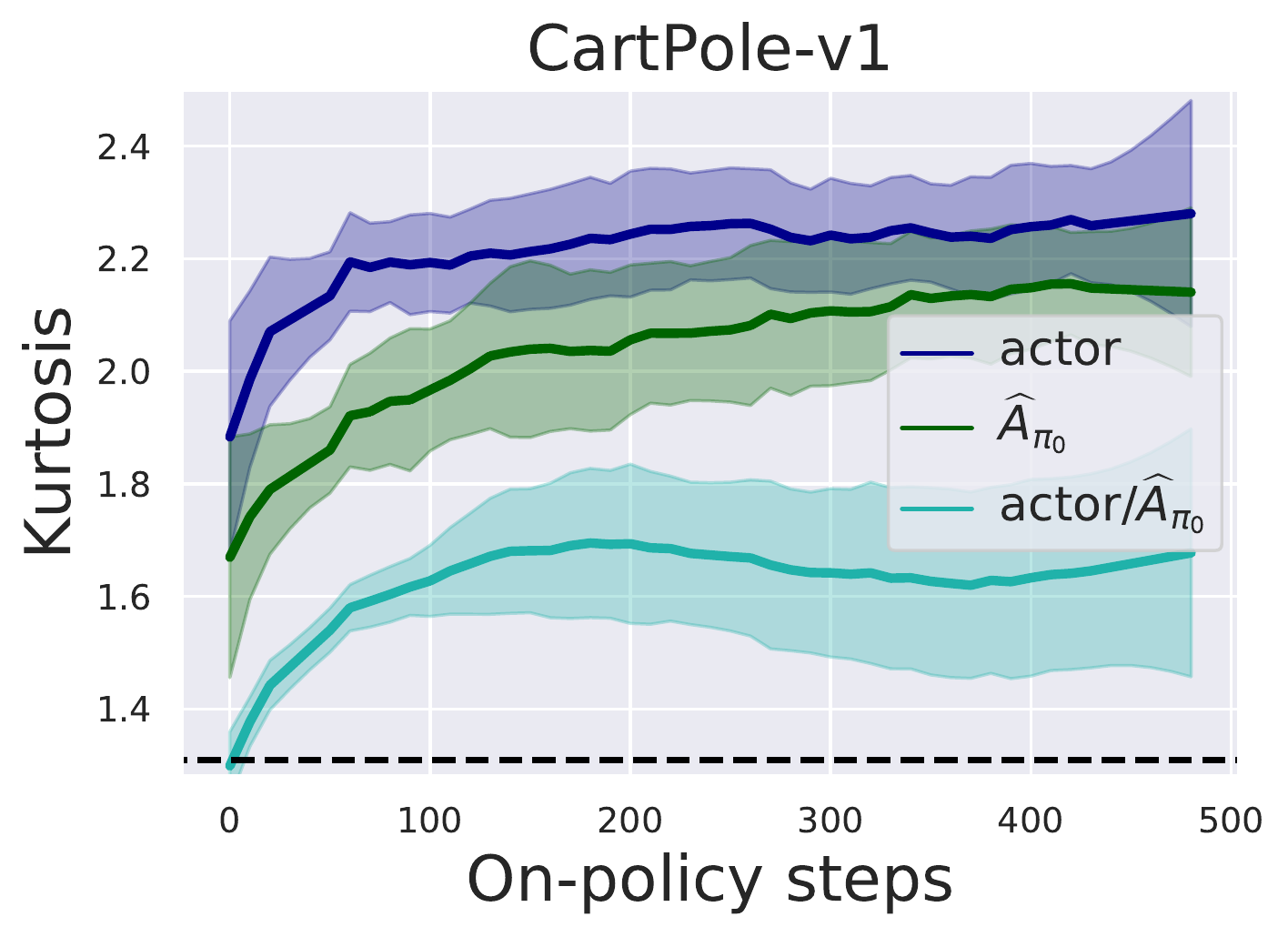}}\hfil
        \subfigure{ \includegraphics[width=0.24\linewidth]{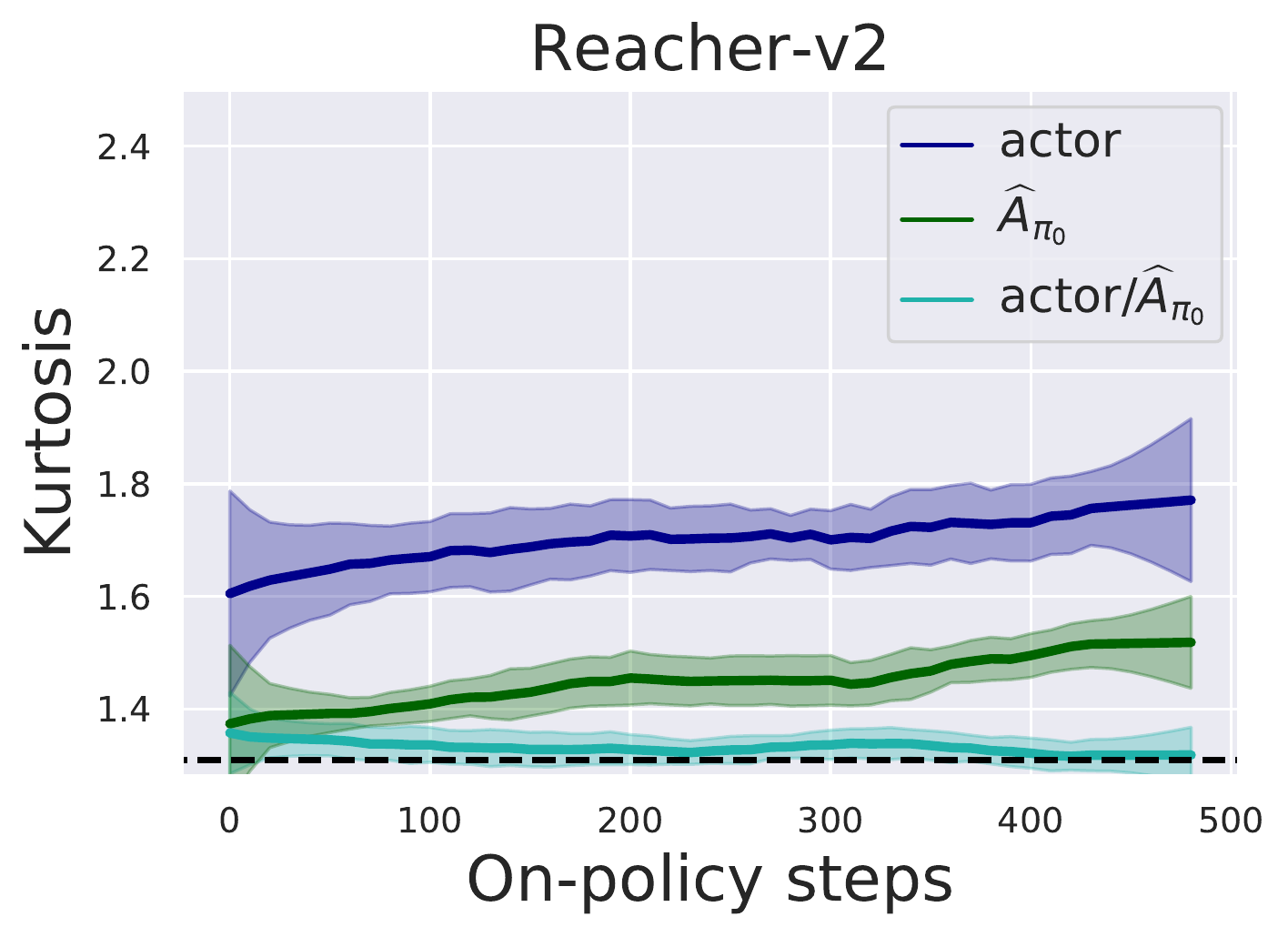}}\hfil
        \subfigure{\includegraphics[width=0.24\linewidth]{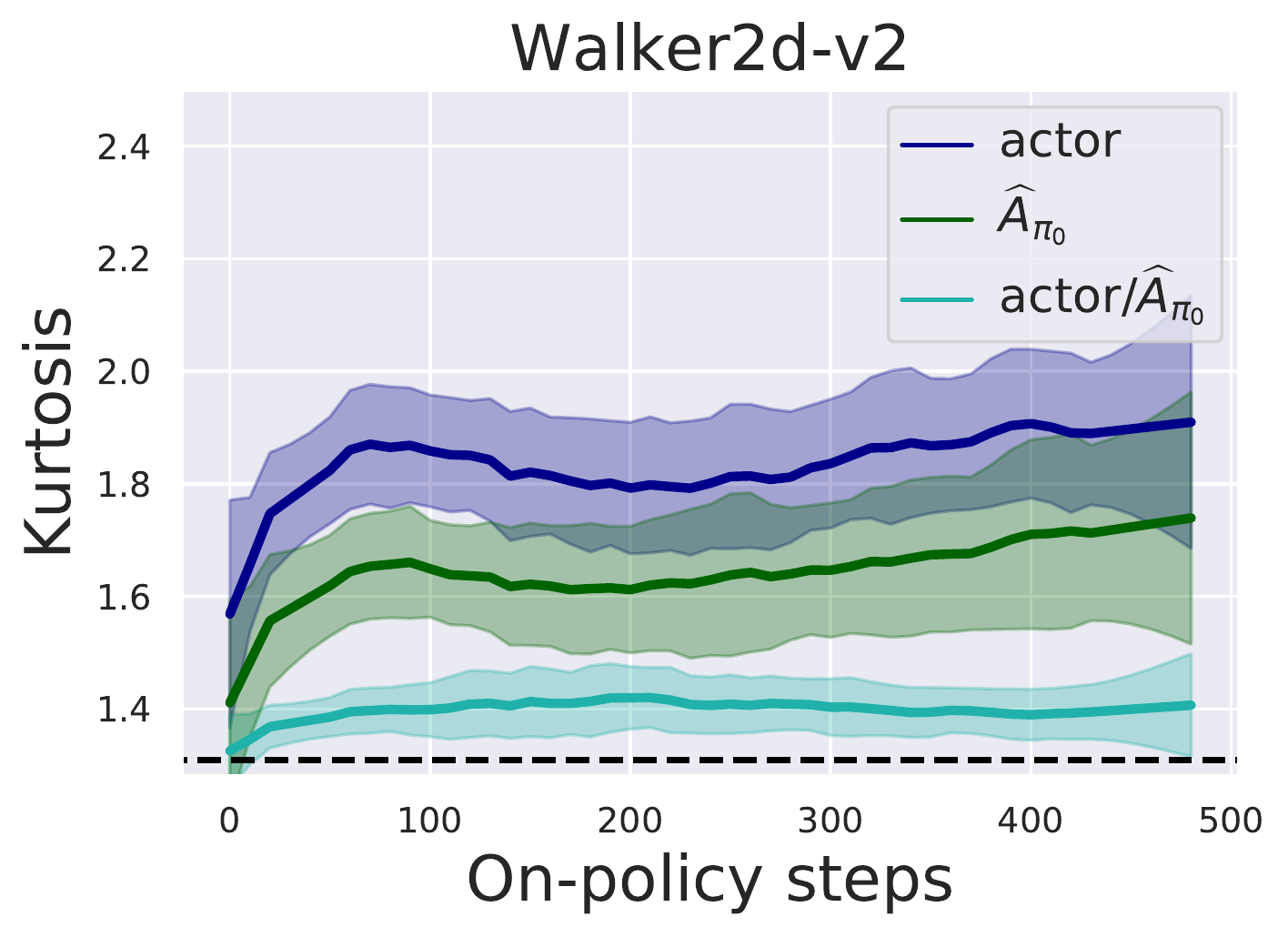}}\hfil
        \subfigure{ \includegraphics[width=0.24\linewidth]{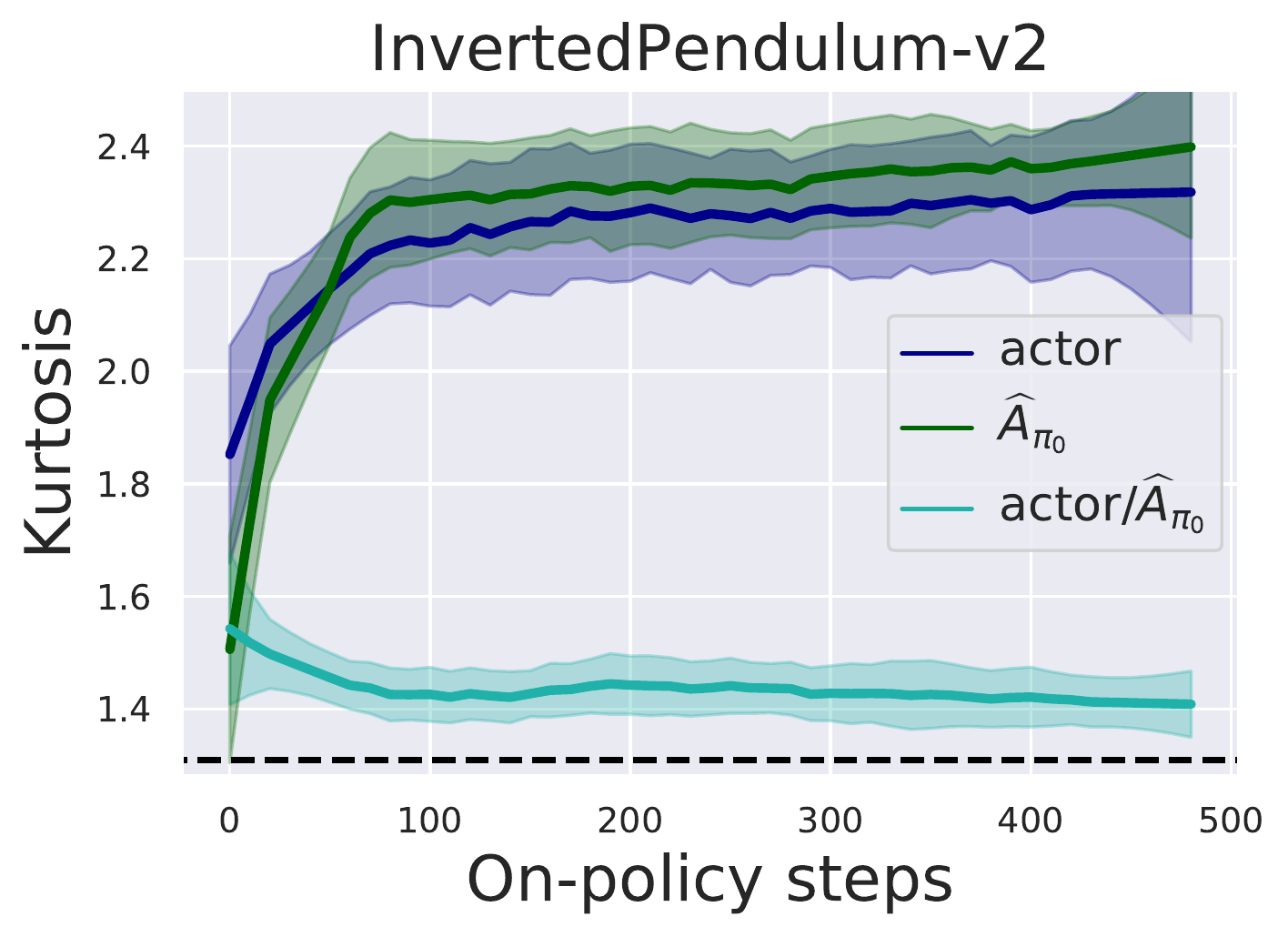}}\hfil
        \par\medskip
       
    \caption{\update{\textbf{Heavy-tailedness in  actor gradients for PPO during on-policy steps} for 8 MuJoCo environments. All plots show mean and std of kurtosis aggregated over 30 random seeds. As the agent is trained, actor gradients become more heavy-tailed. Note that as the gradients become more heavy-tailed, we observe a corresponding
    increase of heavy-tailedness in the advantage estimates ($\hat A_{\pi_0}$). 
    However, ``actor/$\hat A_{\pi_0}$''
    (i.e., actor gradient norm 
    divided by advantage)  
    remain light-tailed throughout the training.  } 
     }\label{fig:ppo-onpolicy-actor-env} 
     
\end{figure}

\begin{figure}[H] 
    \centering
        \setcounter{subfigure}{0}
        \subfigure{\includegraphics[width=0.24\linewidth]{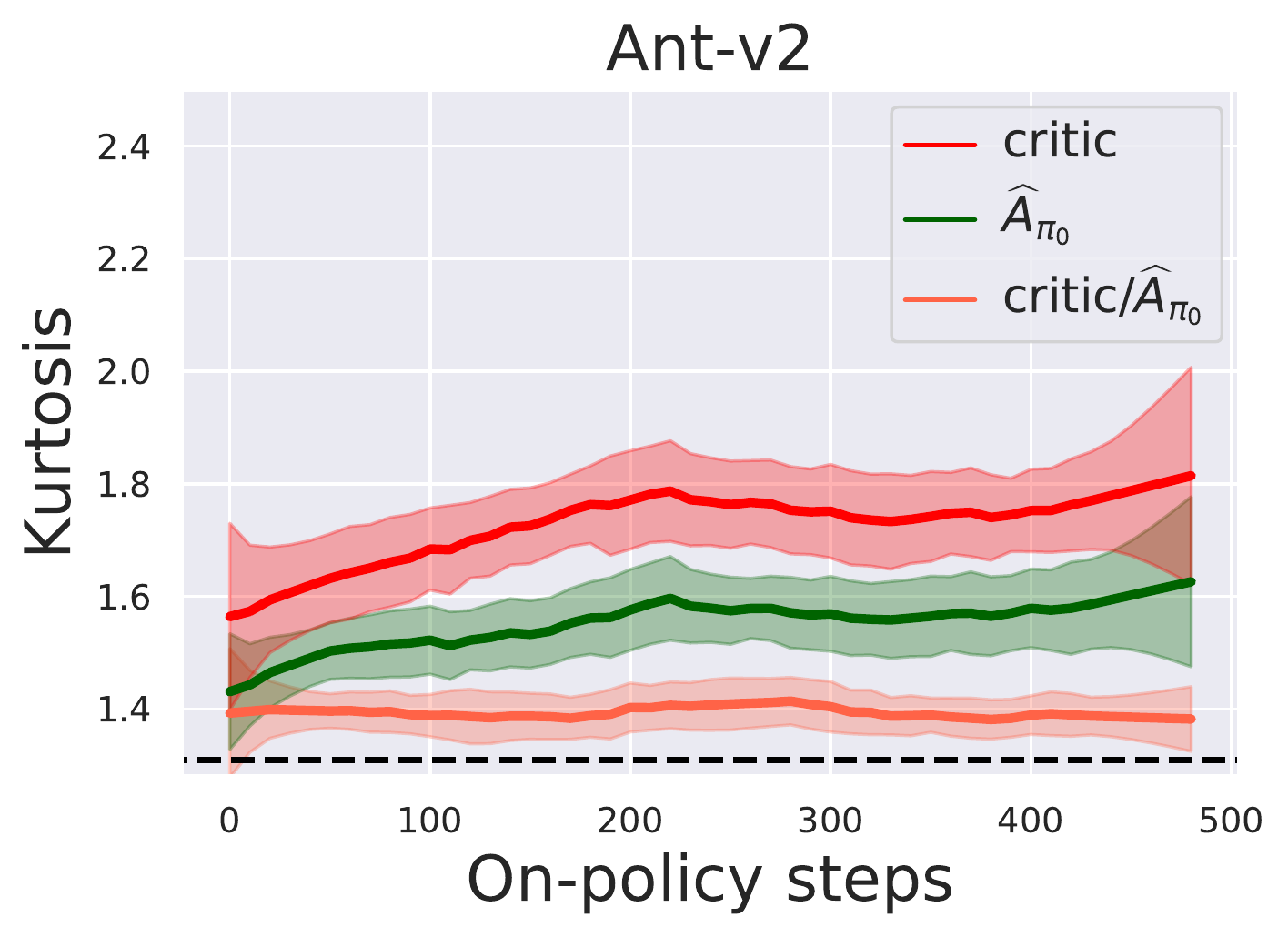}}\hfil
        \subfigure{\includegraphics[width=0.24\linewidth]{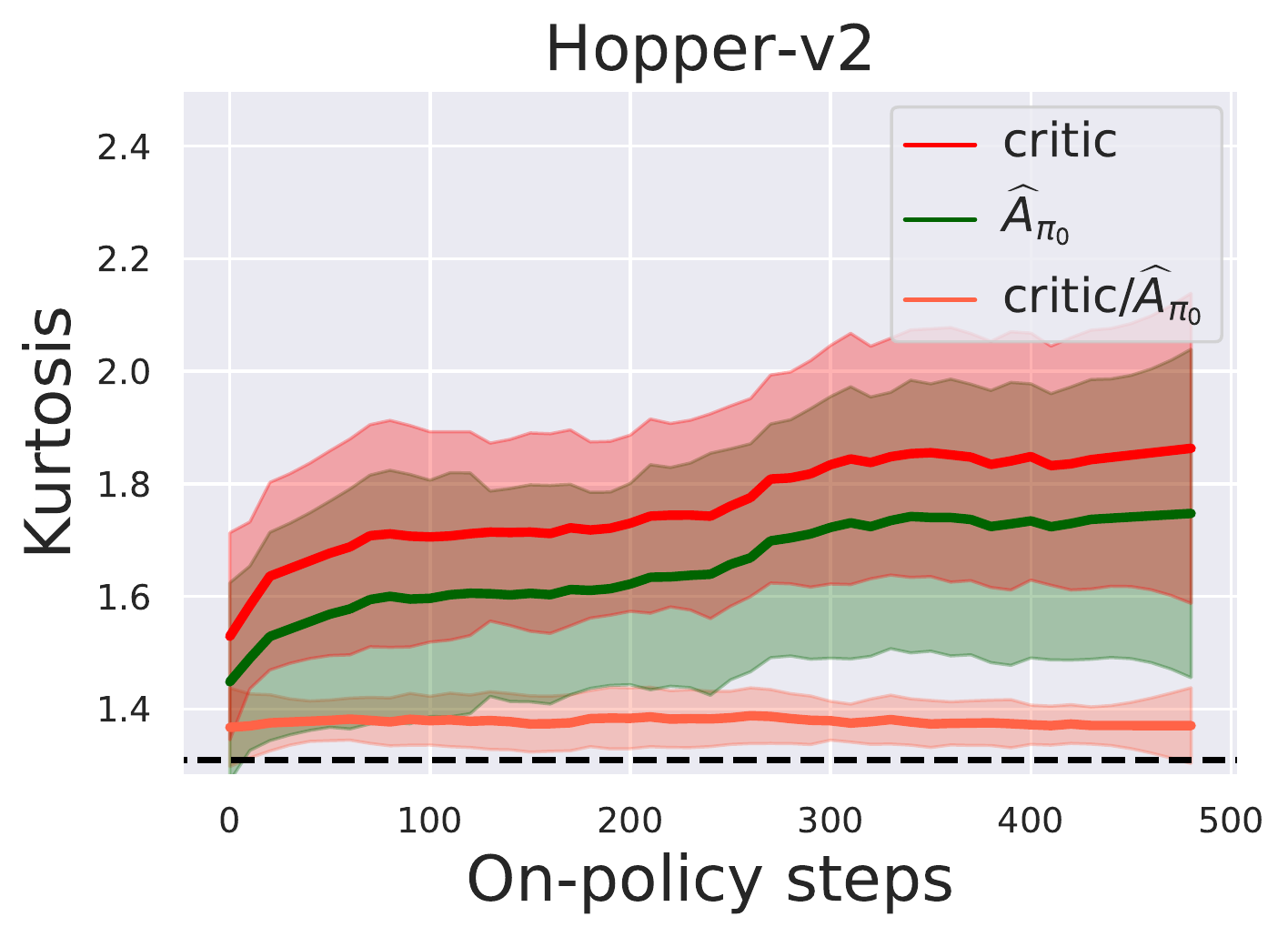}}\hfil
        \subfigure{ \includegraphics[width=0.24\linewidth]{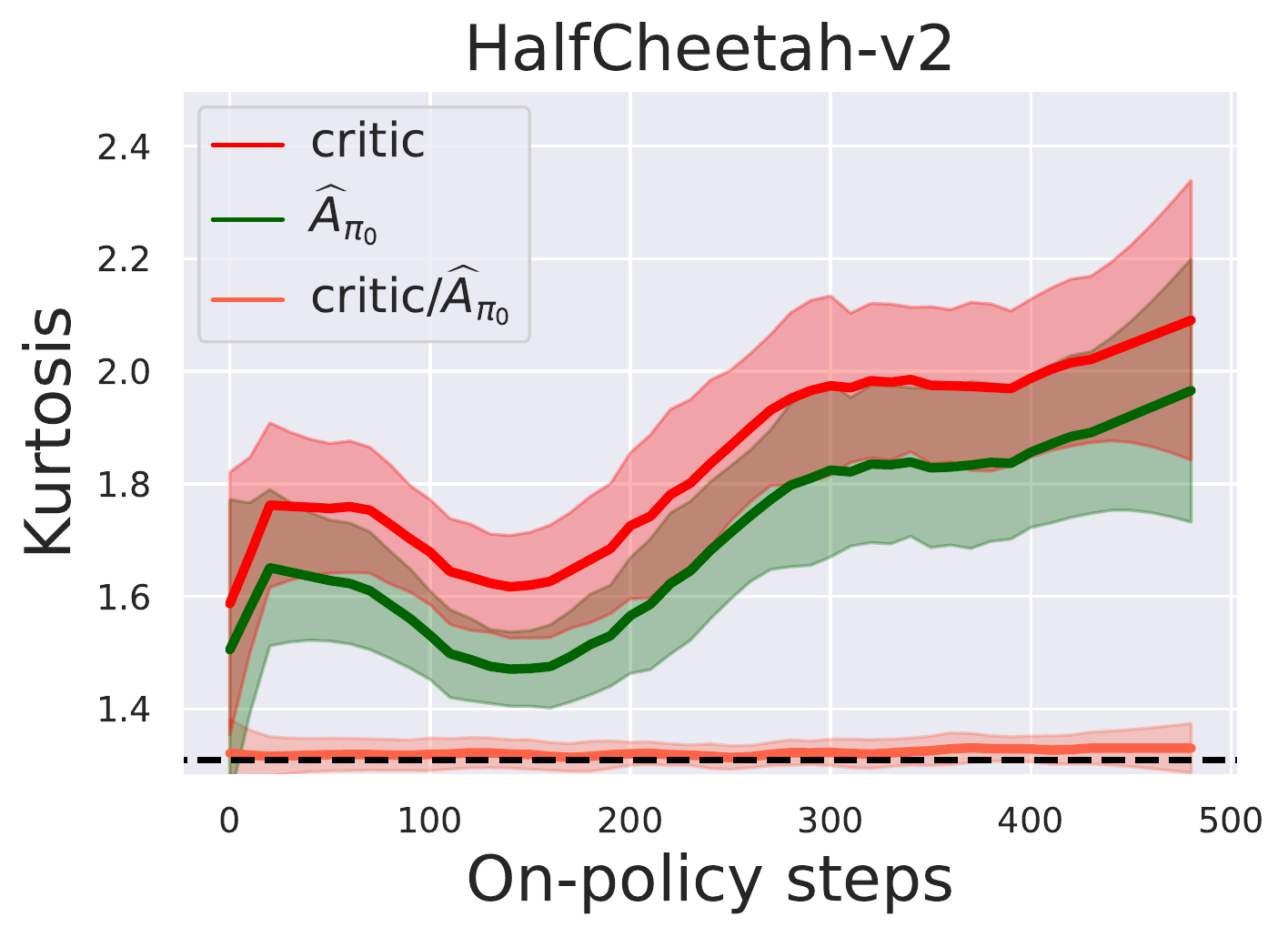}}\hfil
        \subfigure{\includegraphics[width=0.24\linewidth]{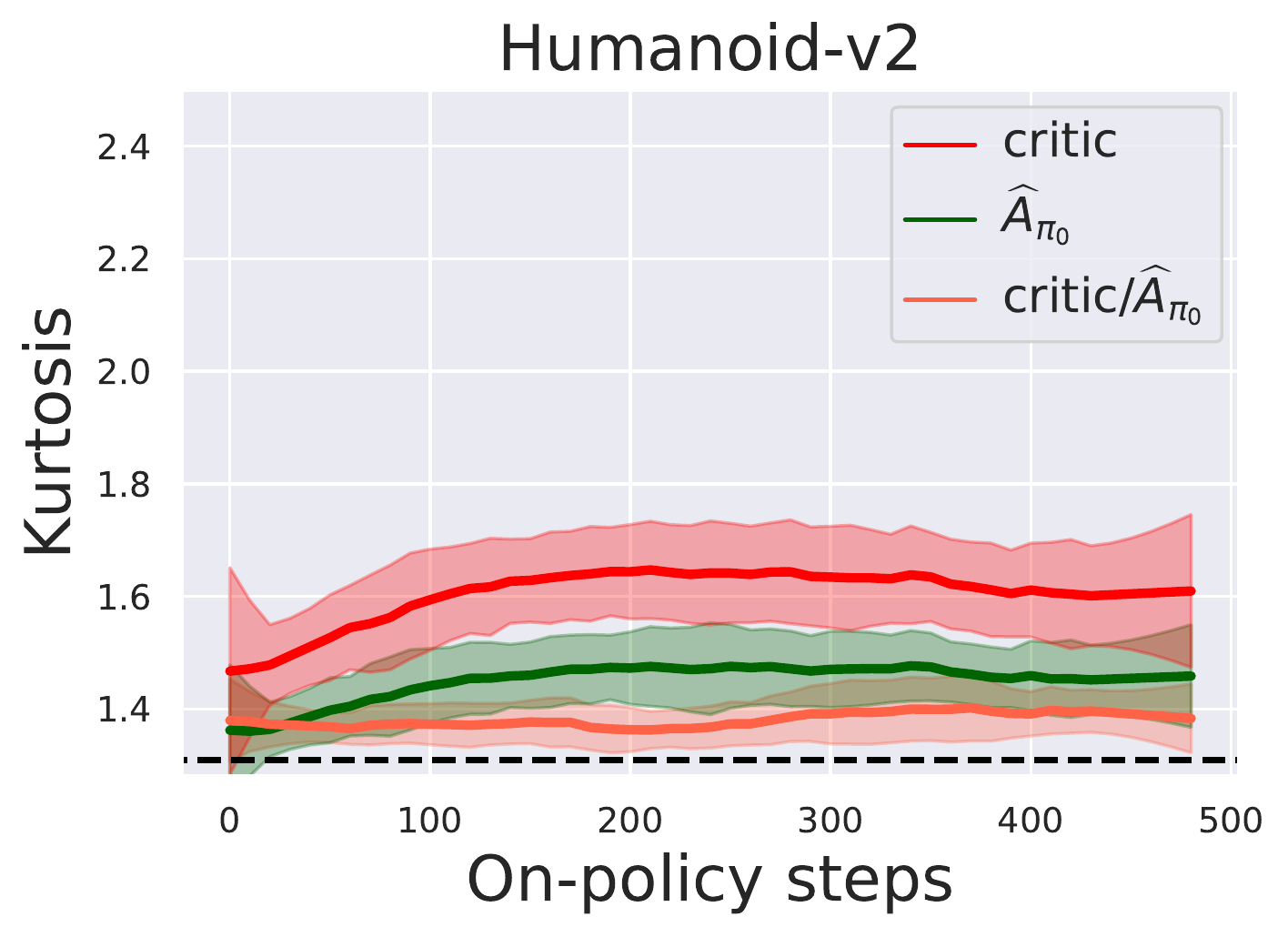}}\hfil
        \par\medskip
        
        \subfigure{\includegraphics[width=0.24\linewidth]{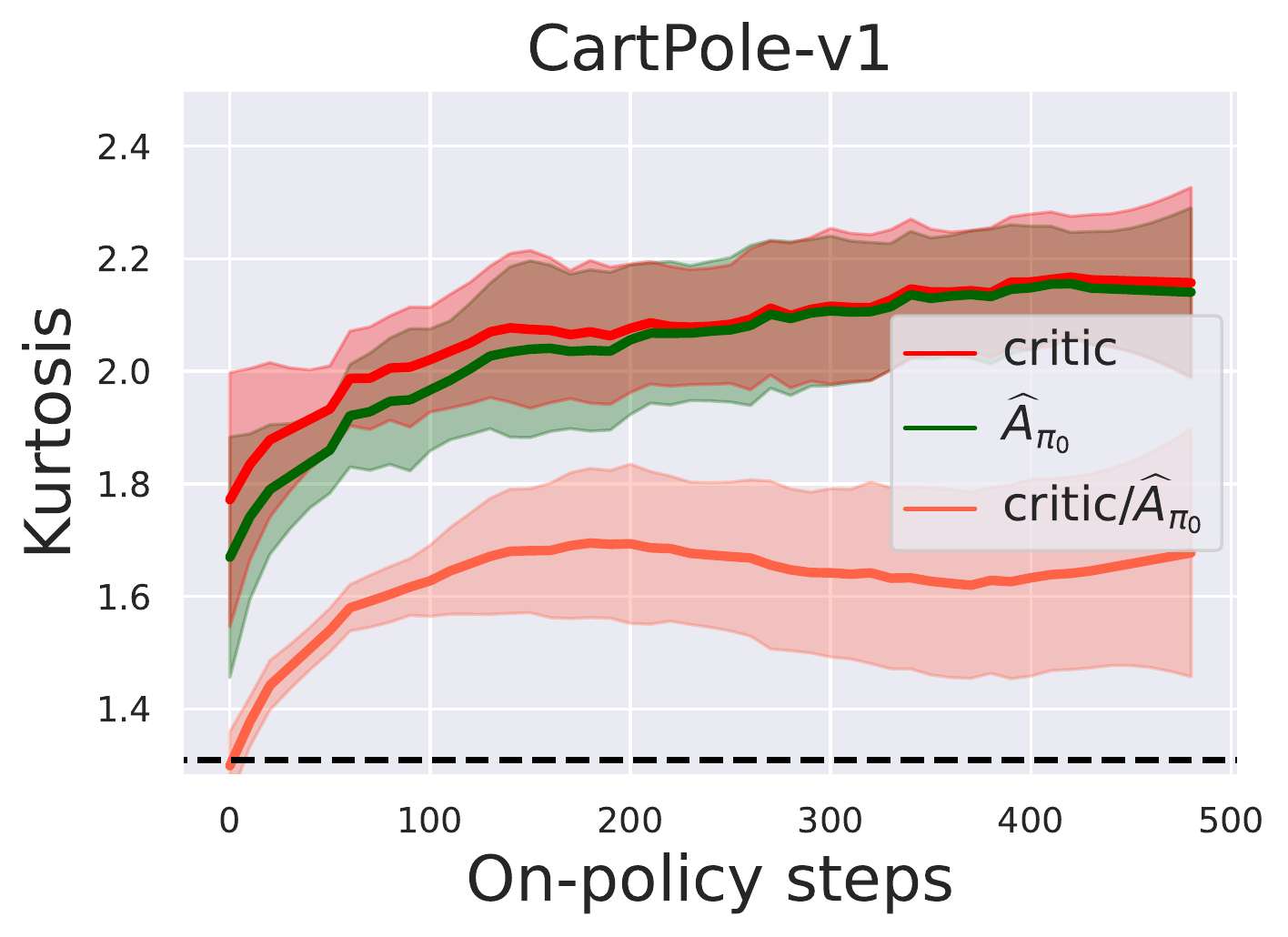}}\hfil
        \subfigure{ \includegraphics[width=0.24\linewidth]{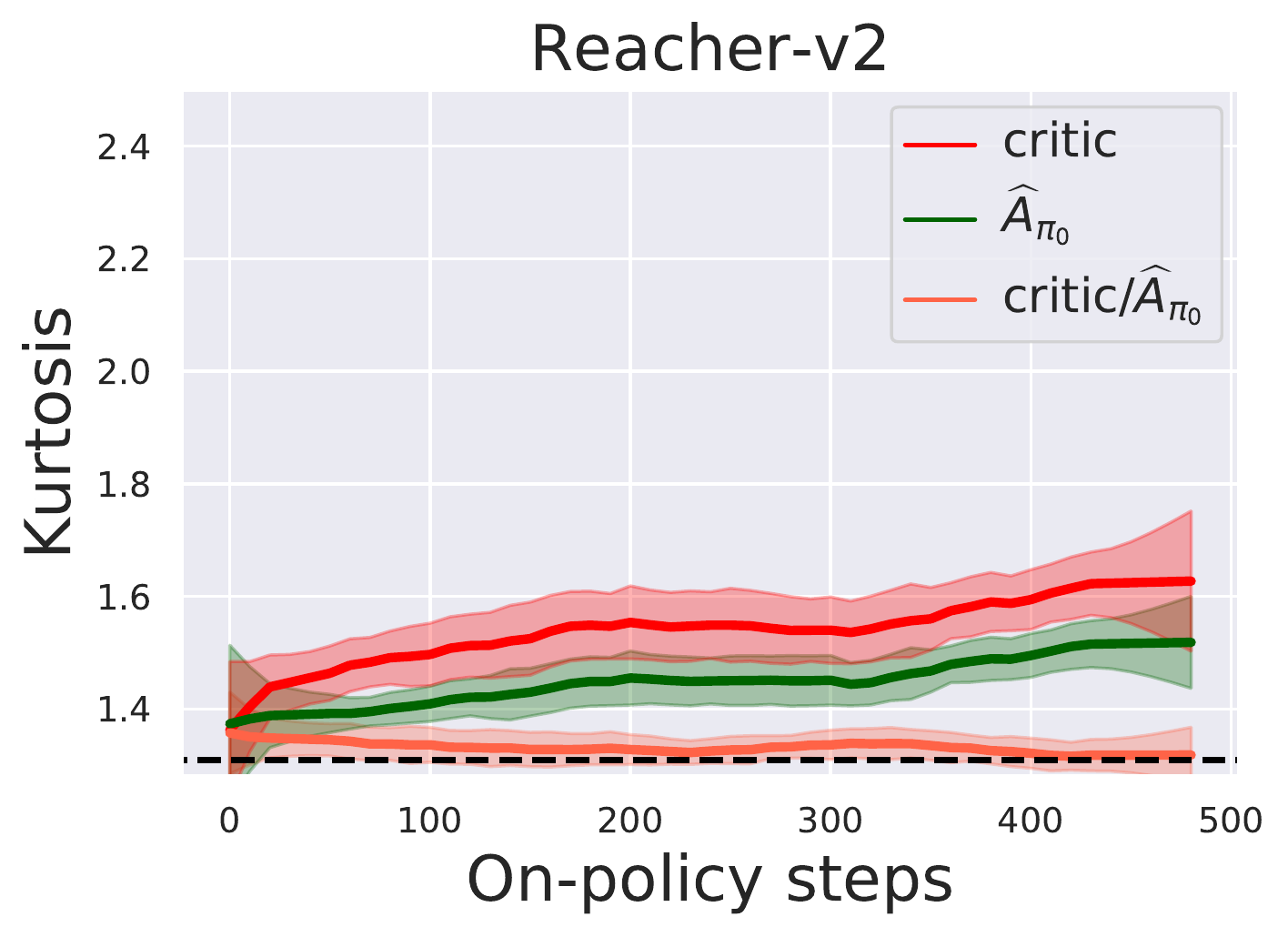}}\hfil
        \subfigure{\includegraphics[width=0.24\linewidth]{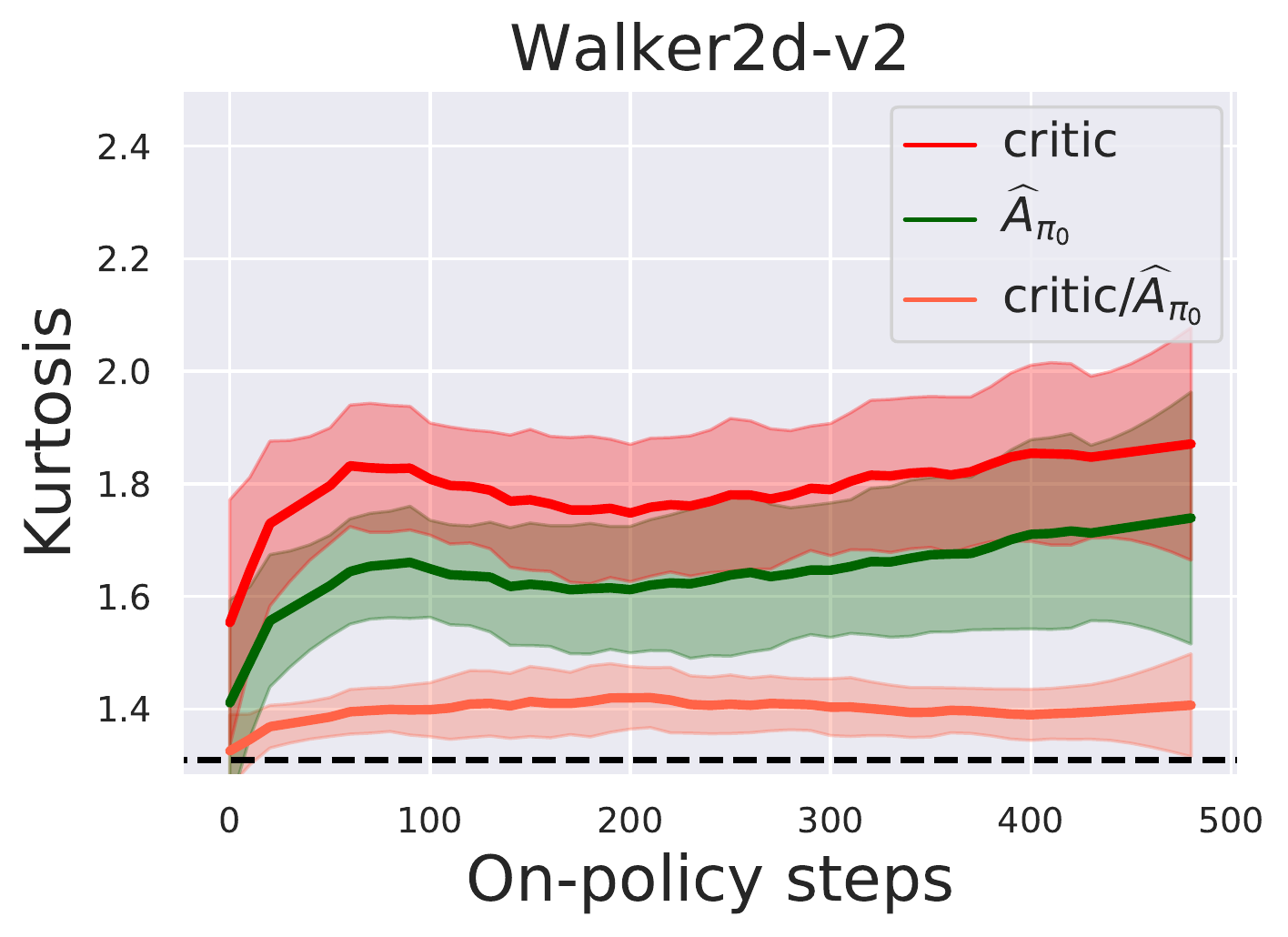}}\hfil
        \subfigure{\includegraphics[width=0.24\linewidth]{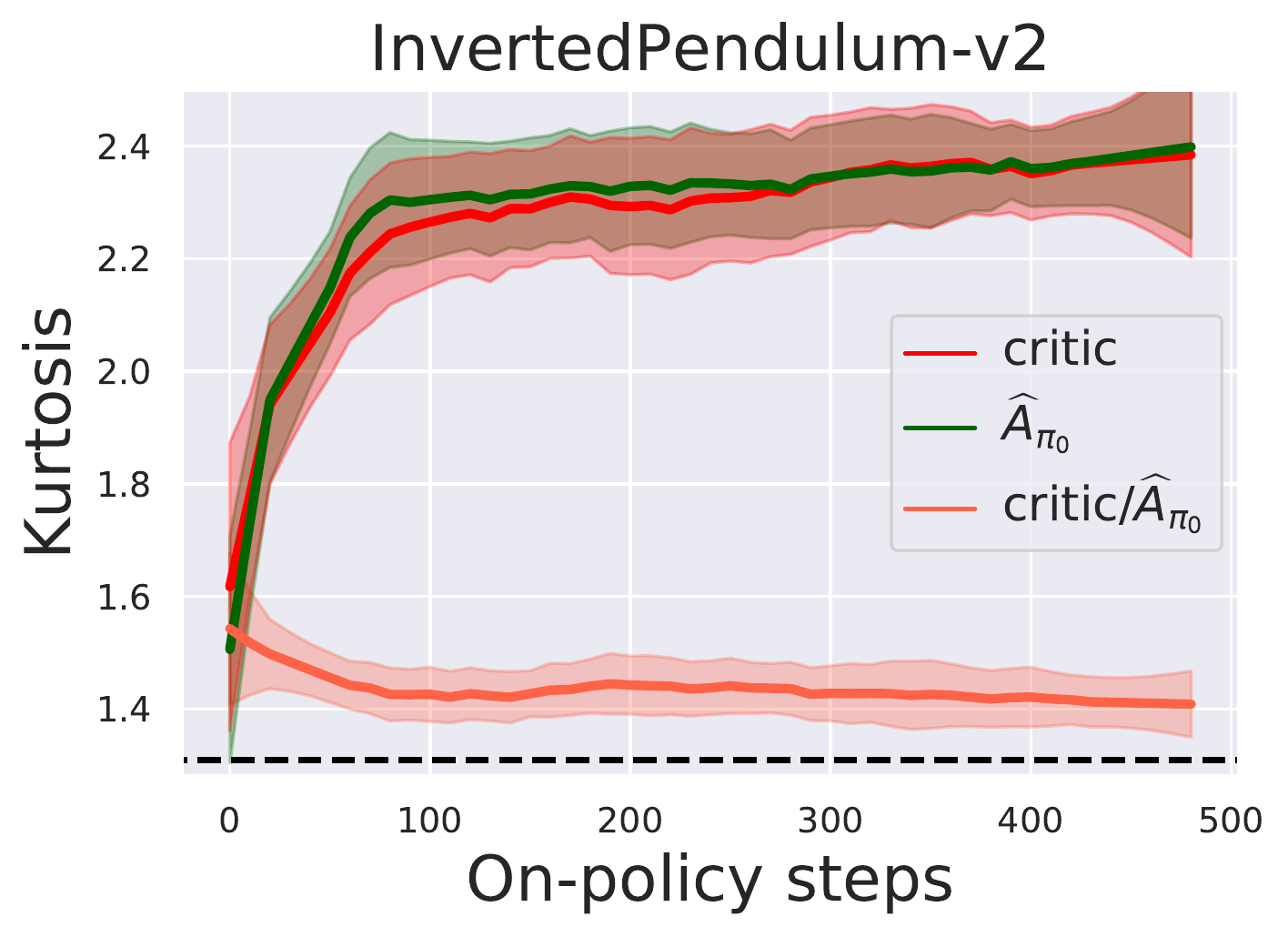}}\hfil
        \par\medskip
    \caption{ \update{\textbf{Heavy-tailedness in critic gradients for PPO during on-policy steps} for 8 MuJoCo environments. All plots show mean and std of kurtosis aggregated over 30 random seeds. As the agent is trained, critic gradients become more heavy-tailed. Note that as the gradients become more heavy-tailed, we observe a corresponding
    increase of heavy-tailedness in the advantage estimates ($\hat A_{\pi_0}$). 
    However, ``critic/$\hat A_{\pi_0}$''
    (i.e., critic gradient norm 
    divided by advantage)  
    remain light-tailed throughout the training. } }\label{fig:ppo-onpolicy-critic-envs} 
     
\end{figure}

\begin{figure}[t] 
    \centering
    \subfigure{\includegraphics[width=0.24\linewidth]{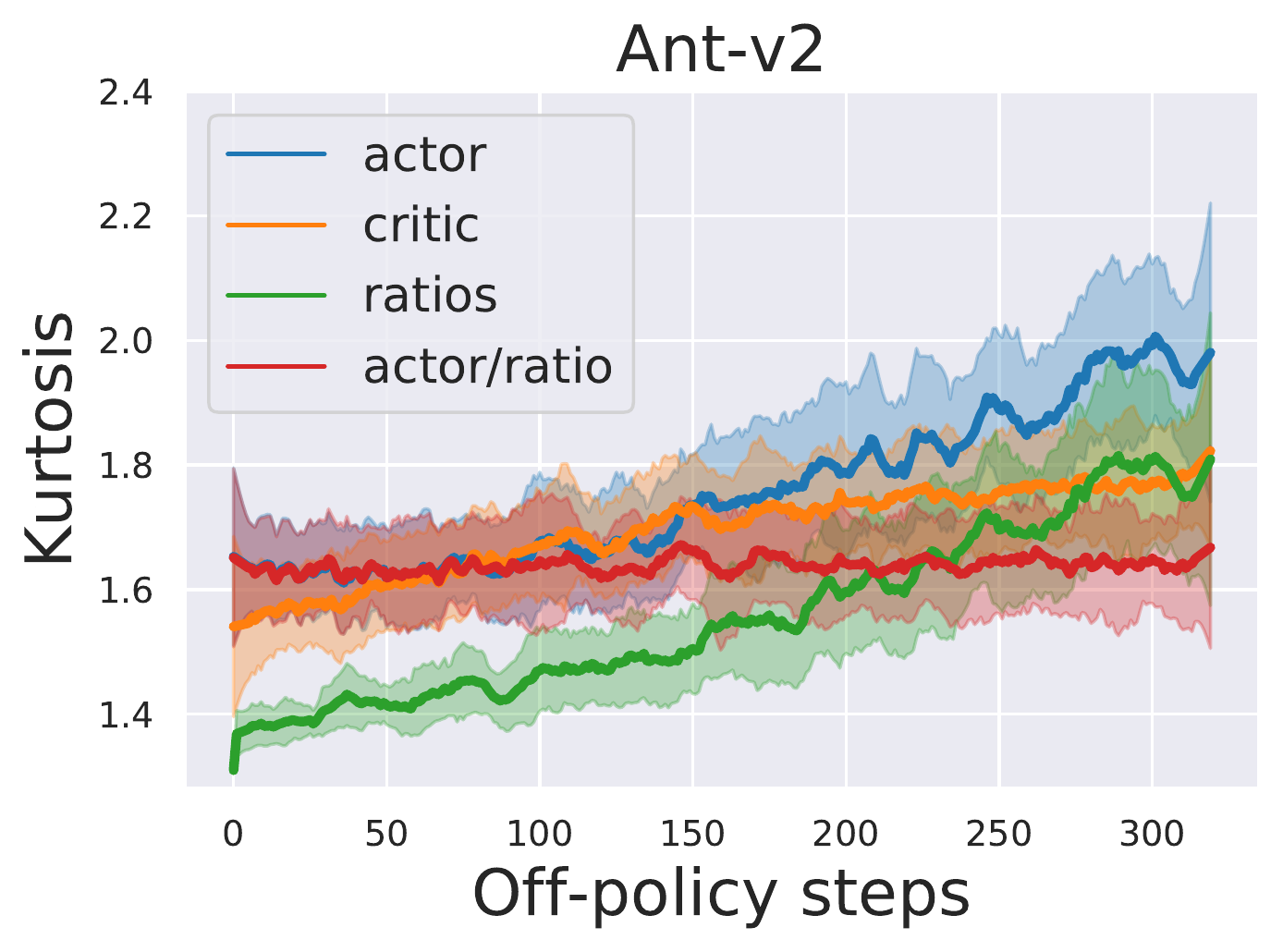}}\hfil
    \subfigure{\includegraphics[width=0.24\linewidth]{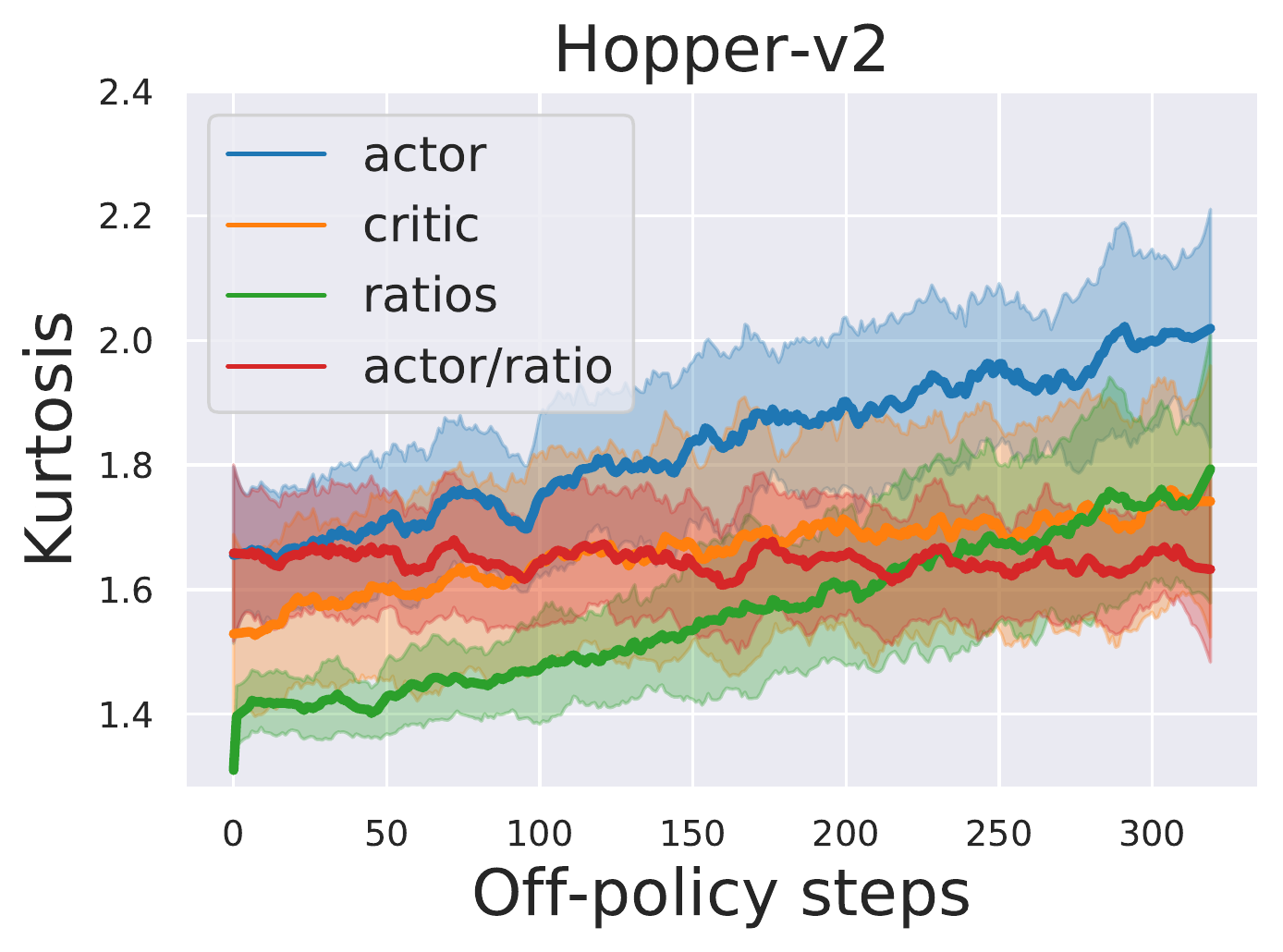}}\hfil
    \subfigure{ \includegraphics[width=0.24\linewidth]{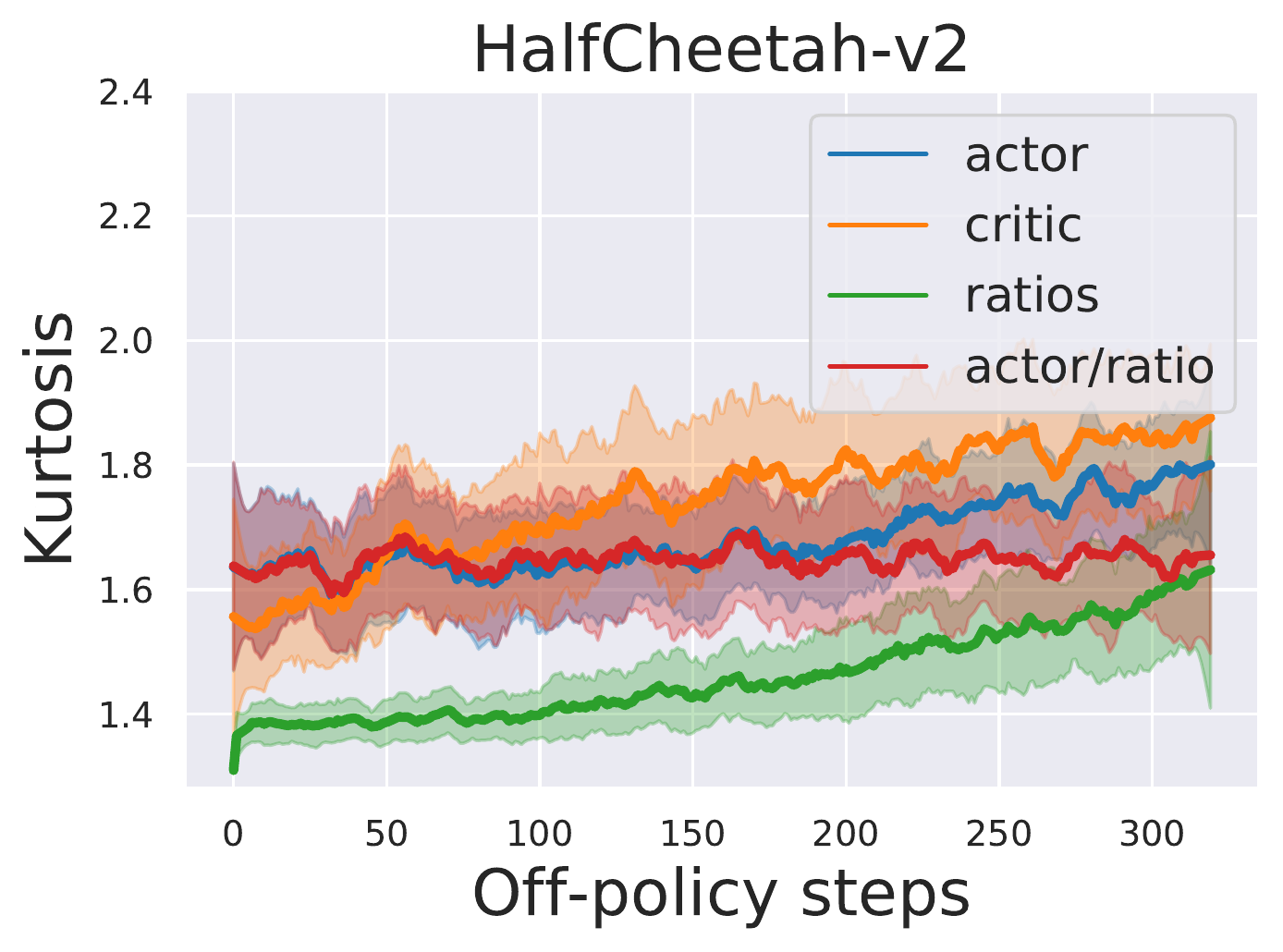}}\hfil
    \subfigure{\includegraphics[width=0.24\linewidth]{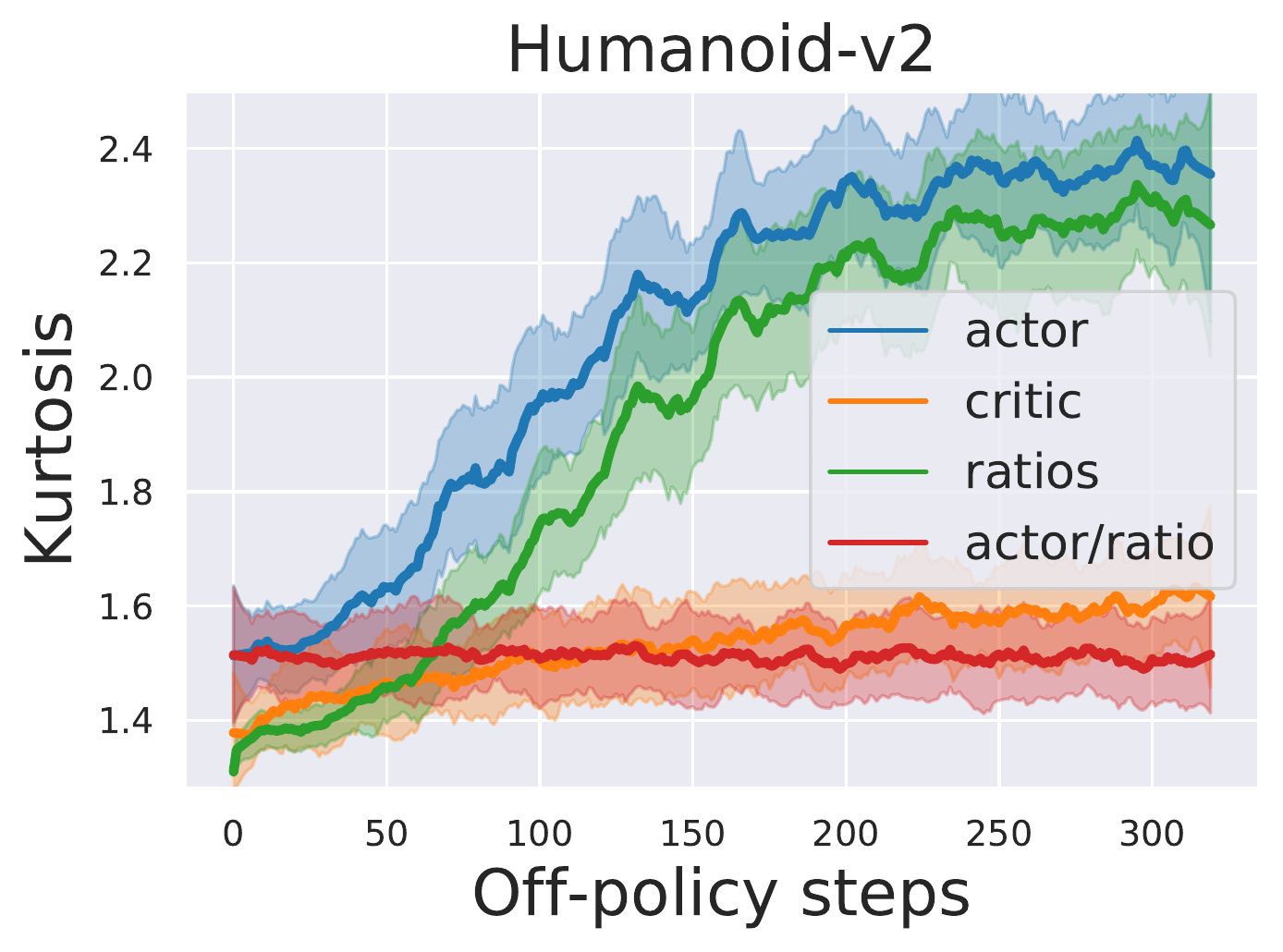}}\hfil
    \par\medskip

    \subfigure{\includegraphics[width=0.24\linewidth]{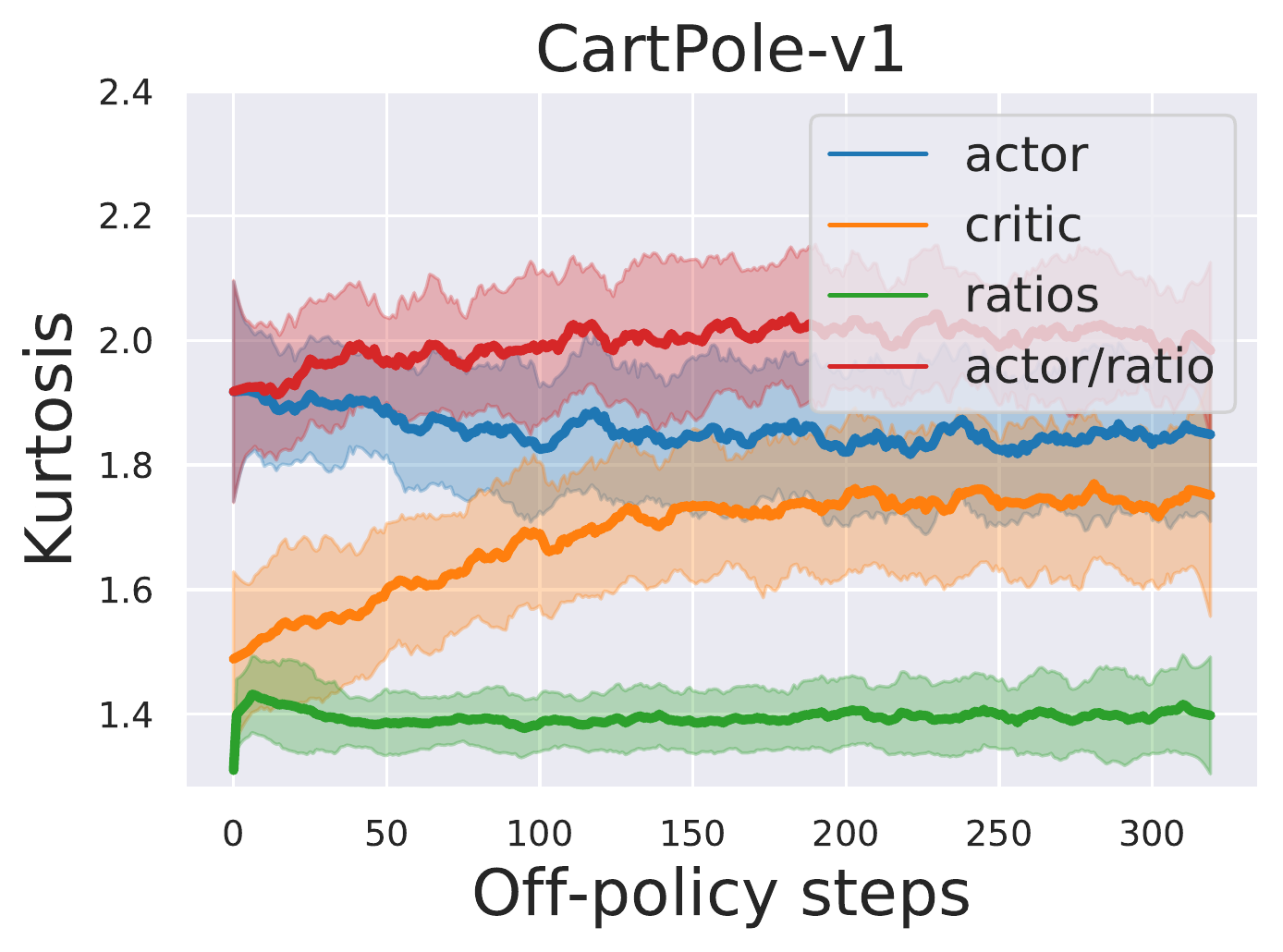}}\hfil
    \subfigure{ \includegraphics[width=0.24\linewidth]{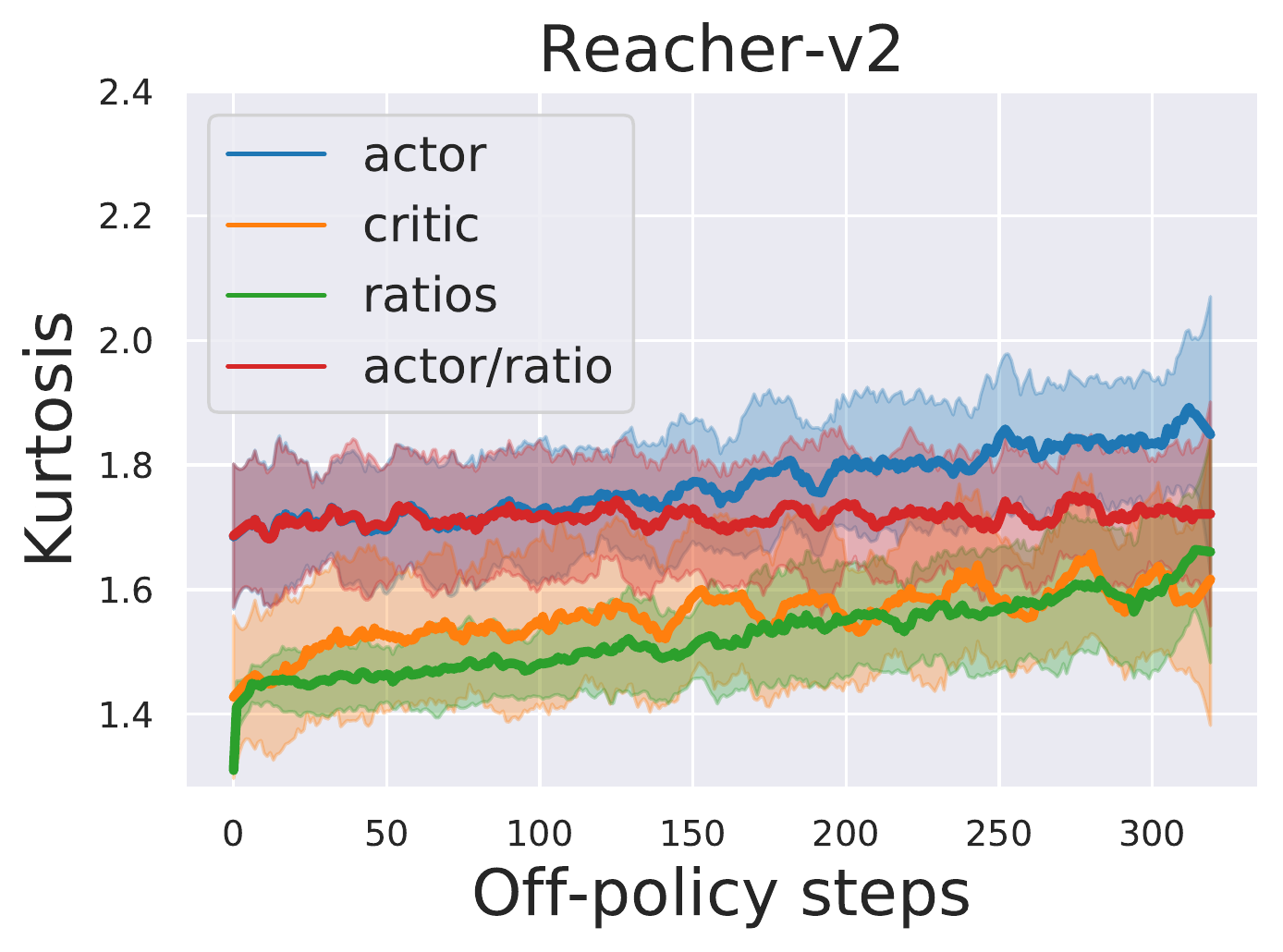}}\hfil
    \subfigure{\includegraphics[width=0.24\linewidth]{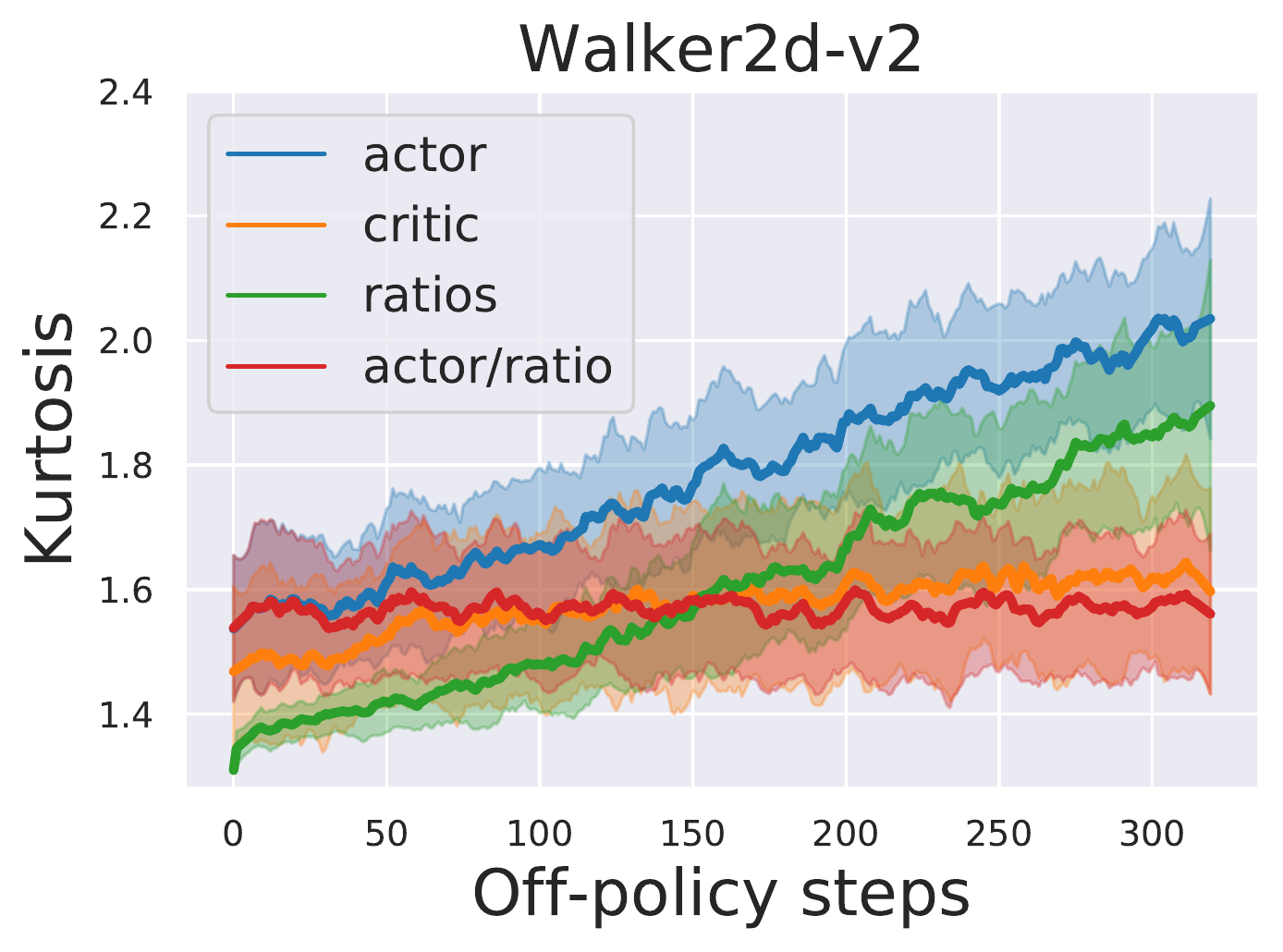}}\hfil
    \subfigure{\includegraphics[width=0.24\linewidth]{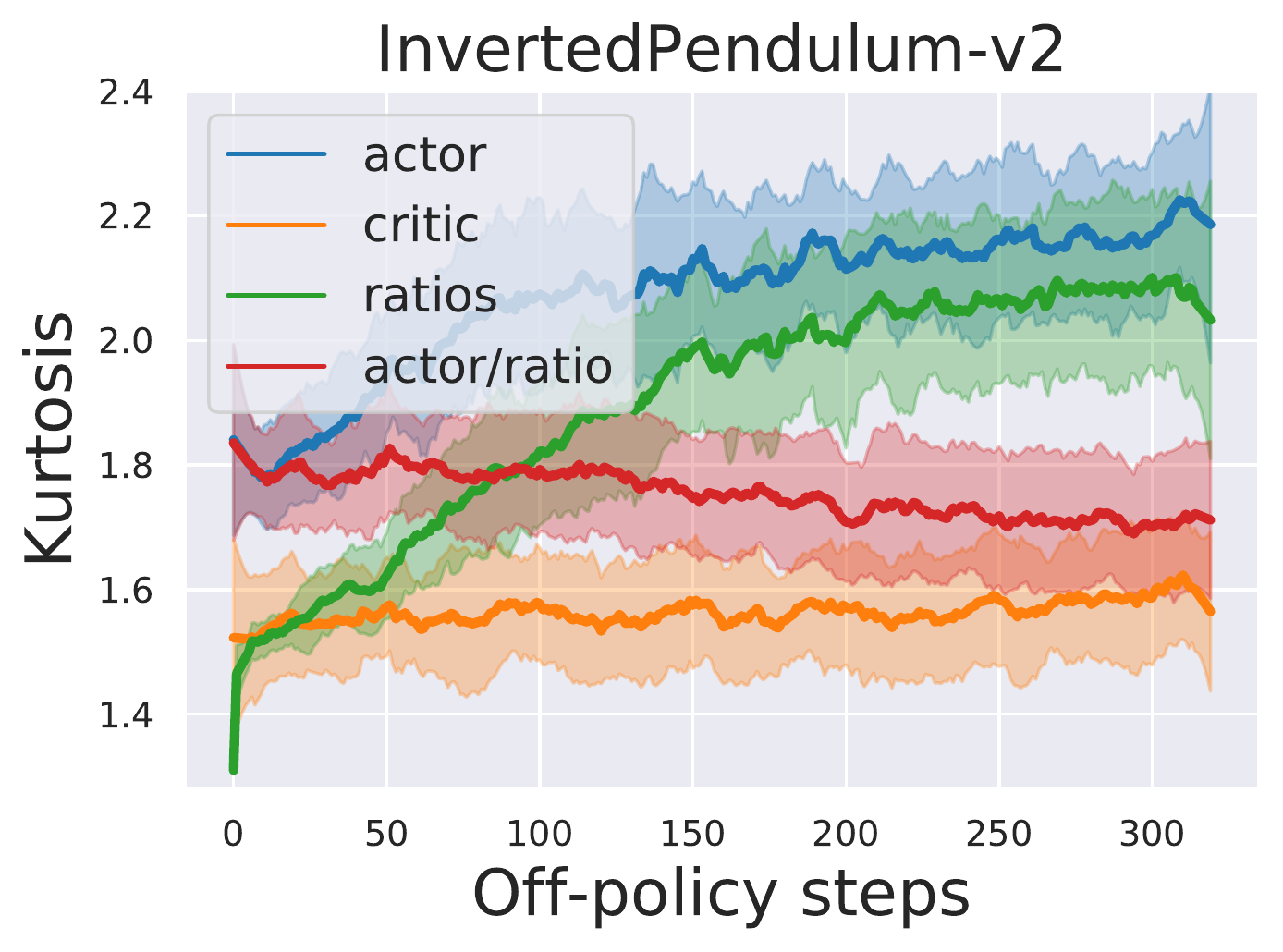}}\hfil
    \par\medskip
       
    \caption{ \update{\textbf{Heavy-tailedness in PPO-\textsc{NoClip} during off-policy steps at Initialization} for 8 MuJoCo environments. All plots show mean and std of kurtosis aggregated over 30 random seeds.  As off-policyness increases, the actor gradients get substantially heavy-tailed. This trend is corroborated by the increase of heavy-tailedness in ratios. Moreover, consistently we observe that the heavy-tailedness in ``actor/ratios'' stays constant. 
    The trend in heavy-tailedness at later training iteration follow similar trends but the increase in heavy-tailedness tapers off. The overall increase across training iterations is explained by the induced heavy-tailedness in the advantage estimates (cf. Sec.~\ref{subsec:on-policy}).}
     }\label{fig:ppo-offpolicy-envs} 
     
\end{figure}

\section{\update{How do heavy-tailed policy-gradients affect training?}} \label{sec:App_ablation}

\subsection{\update{Effect of heavy-tailedness in advantages}} \label{subsec:App_adv}

\begin{figure}[H] 
    \centering

        \subfigure{ \includegraphics[width=0.24\linewidth]{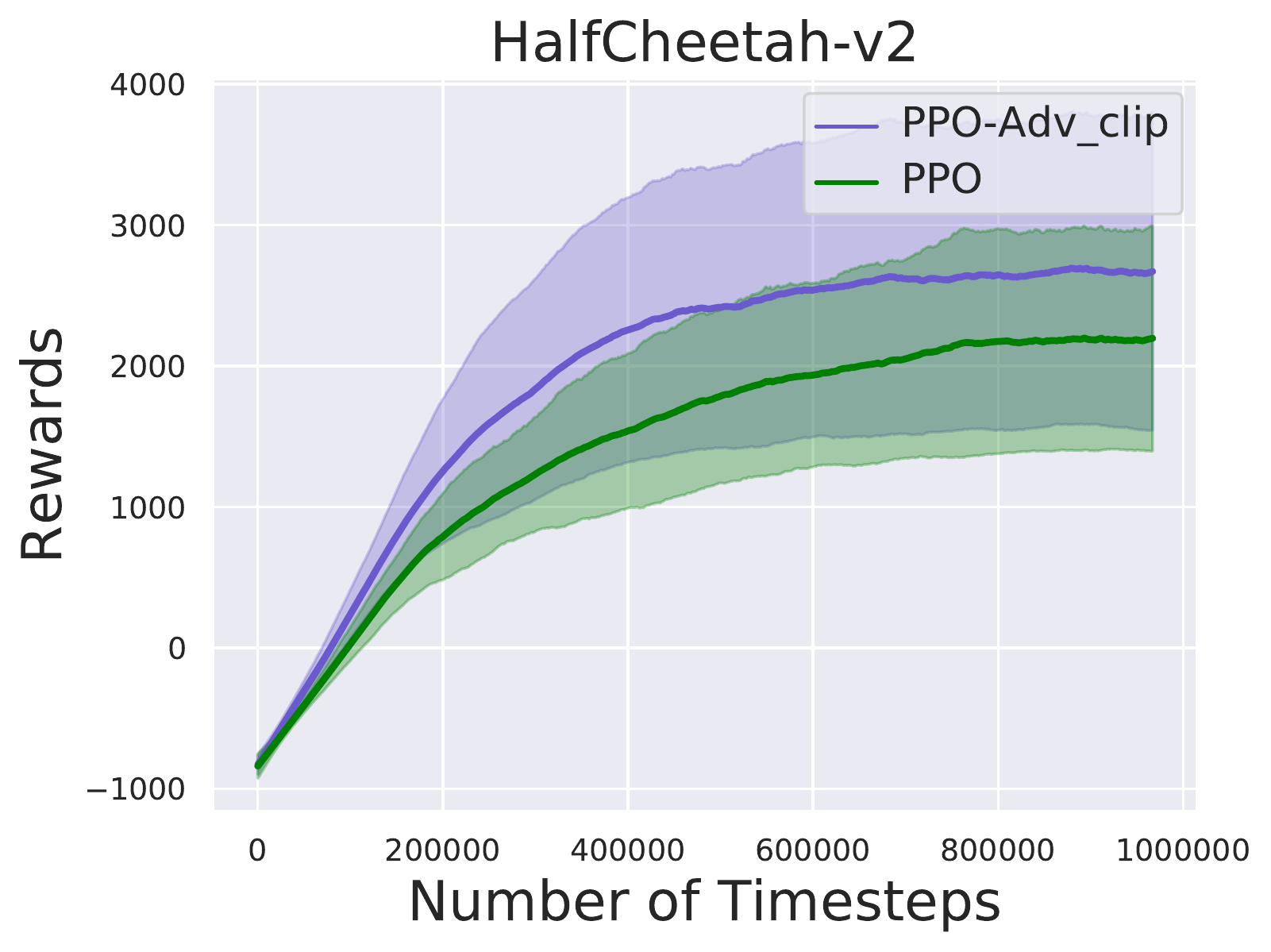}}\hfil
        \subfigure{ \includegraphics[width=0.24\linewidth]{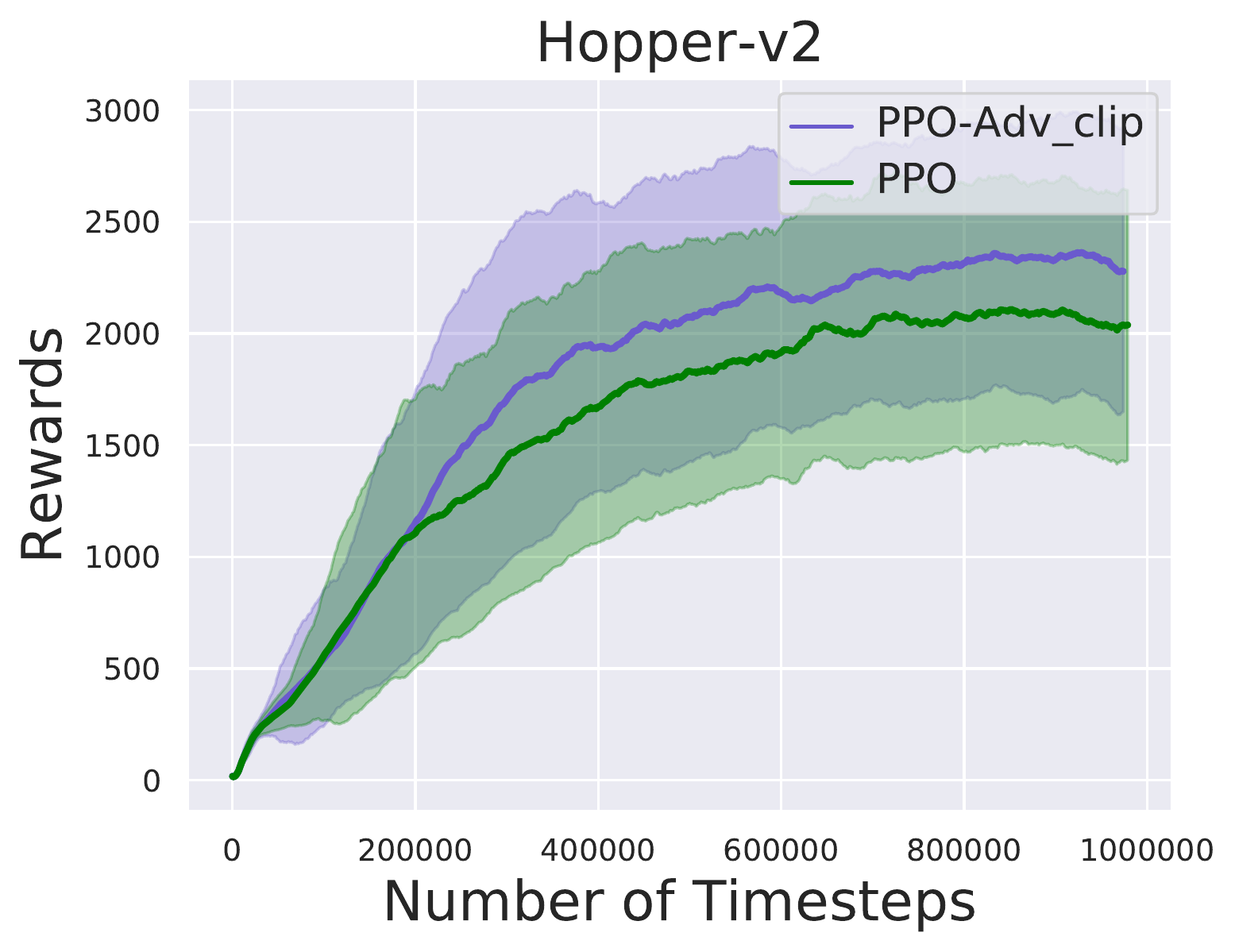}}\hfil
       \subfigure{ \includegraphics[width=0.24\linewidth]{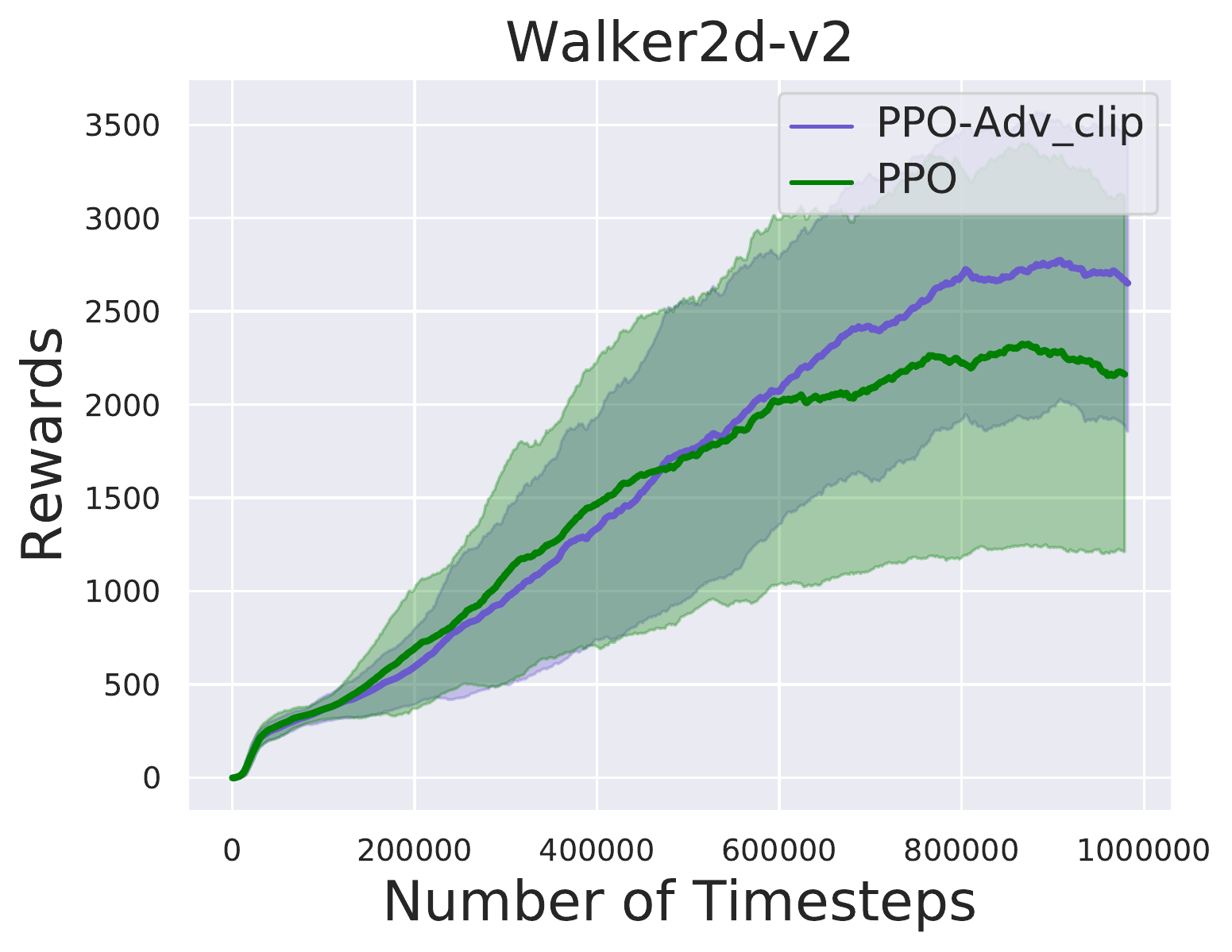}}\hfil
        \subfigure{\includegraphics[width=0.24\linewidth]{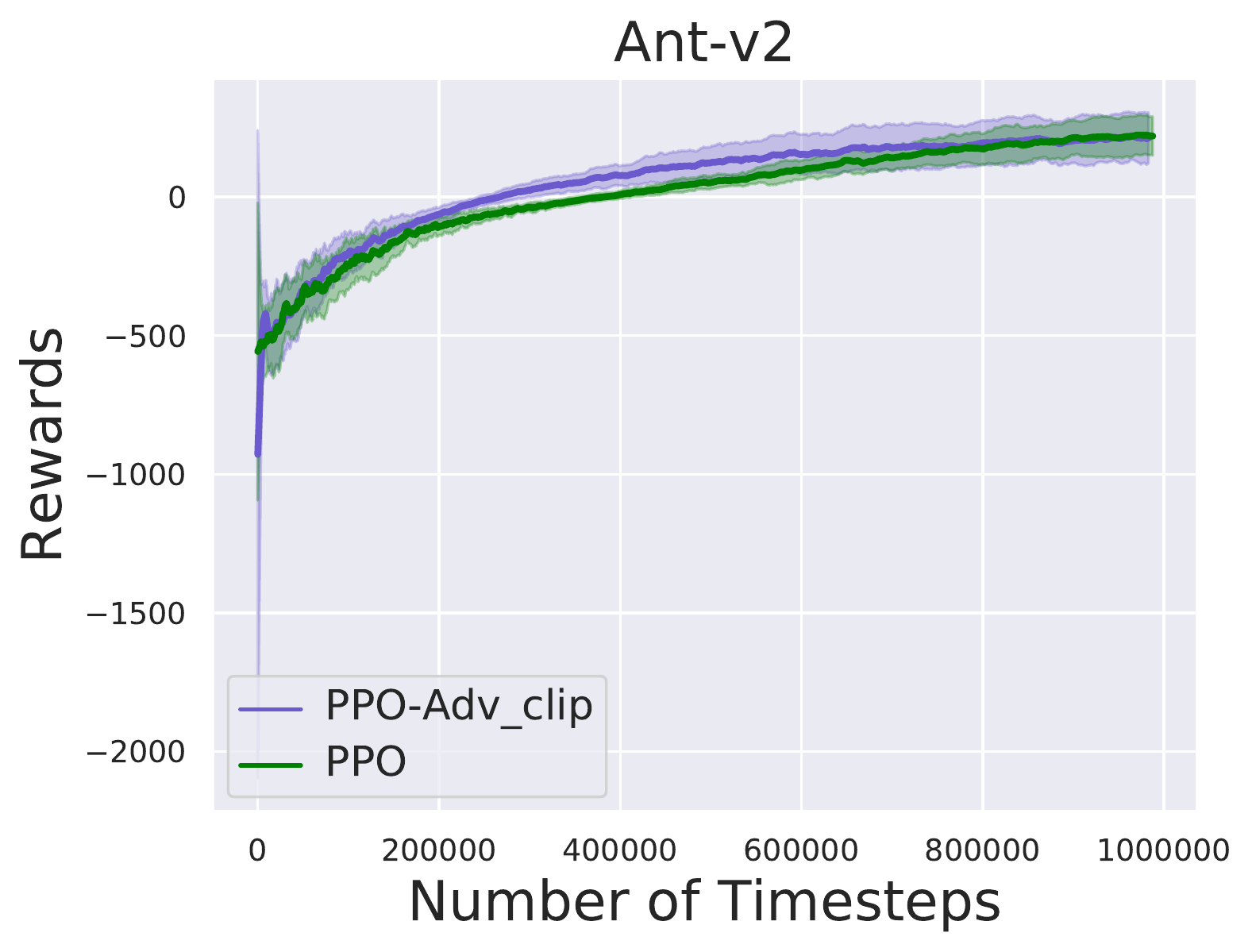}}\hfil
        \par\medskip
        
        \subfigure{ \includegraphics[width=0.24\linewidth]{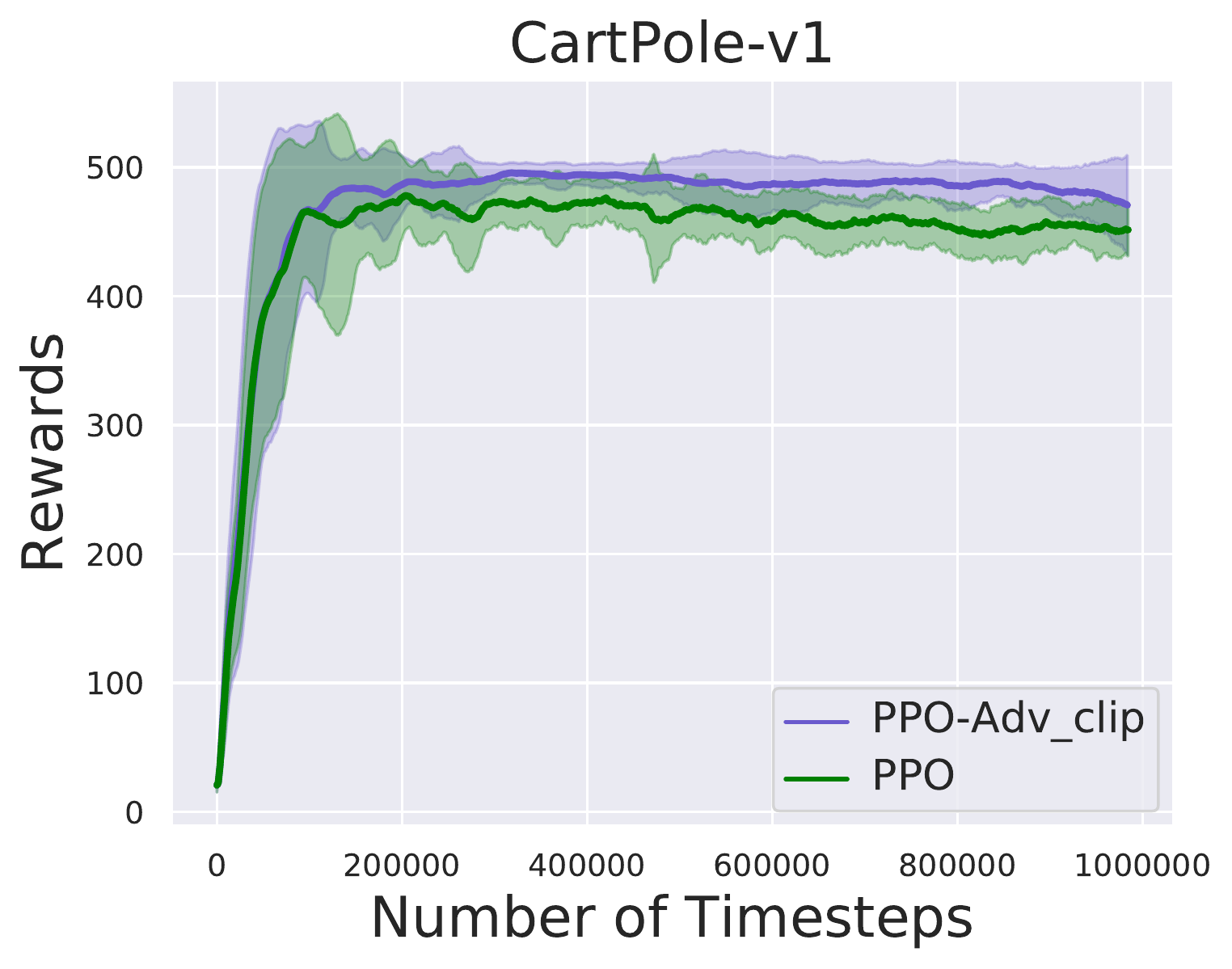}}\hfil
        \subfigure{ \includegraphics[width=0.24\linewidth]{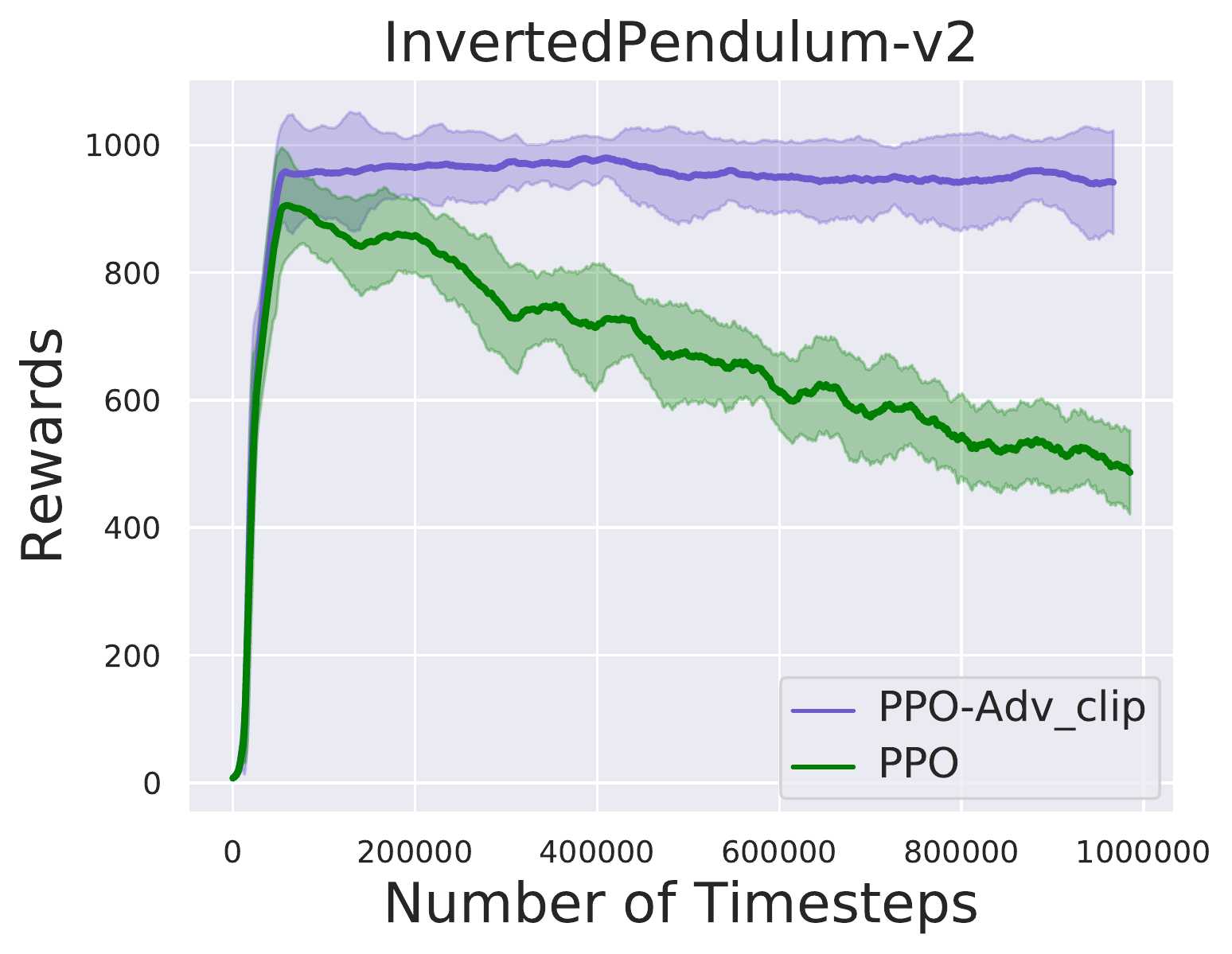}}\hfil
        \subfigure{\includegraphics[width=0.24\linewidth]{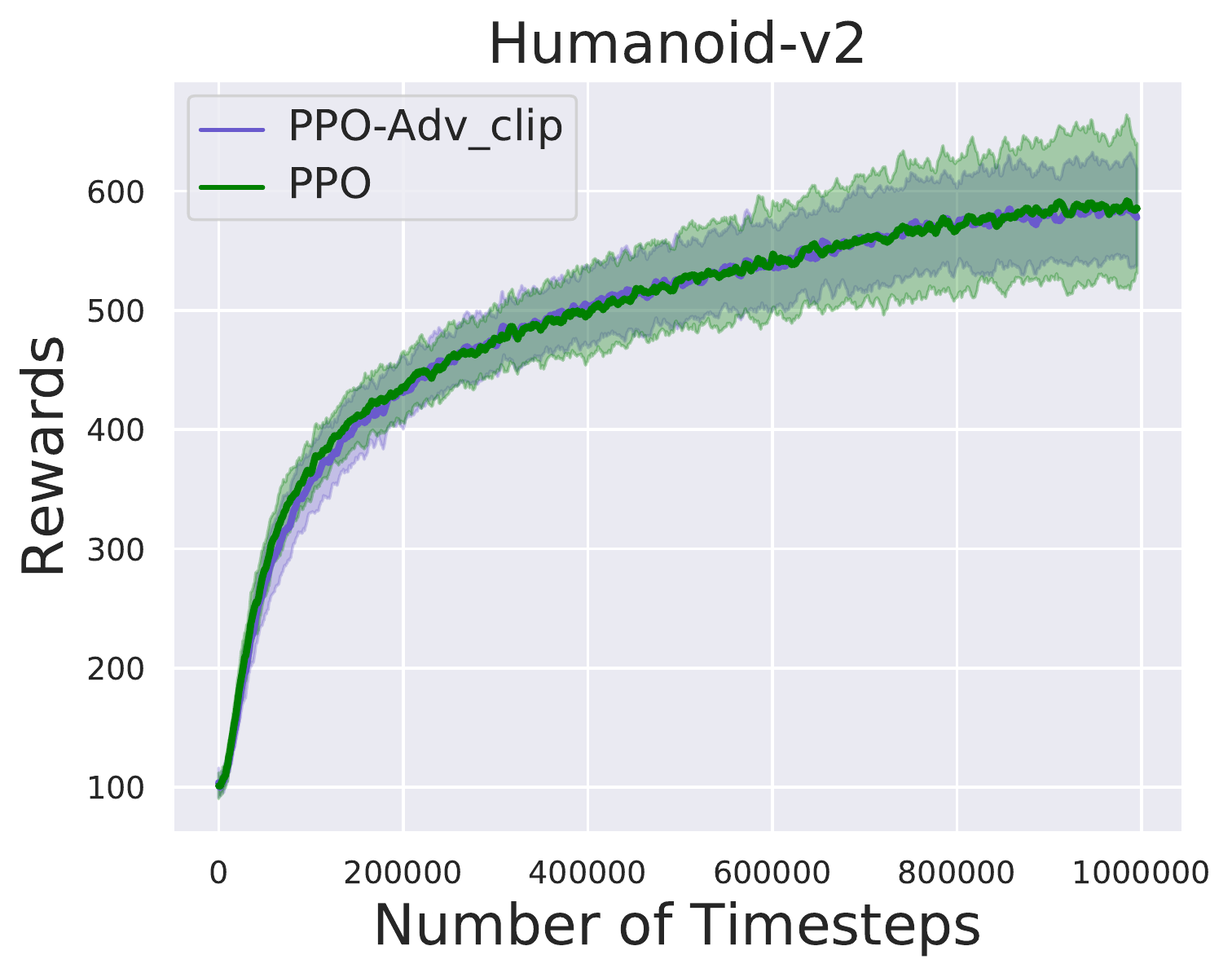}}\hfil
        \subfigure{ \includegraphics[width=0.24\linewidth]{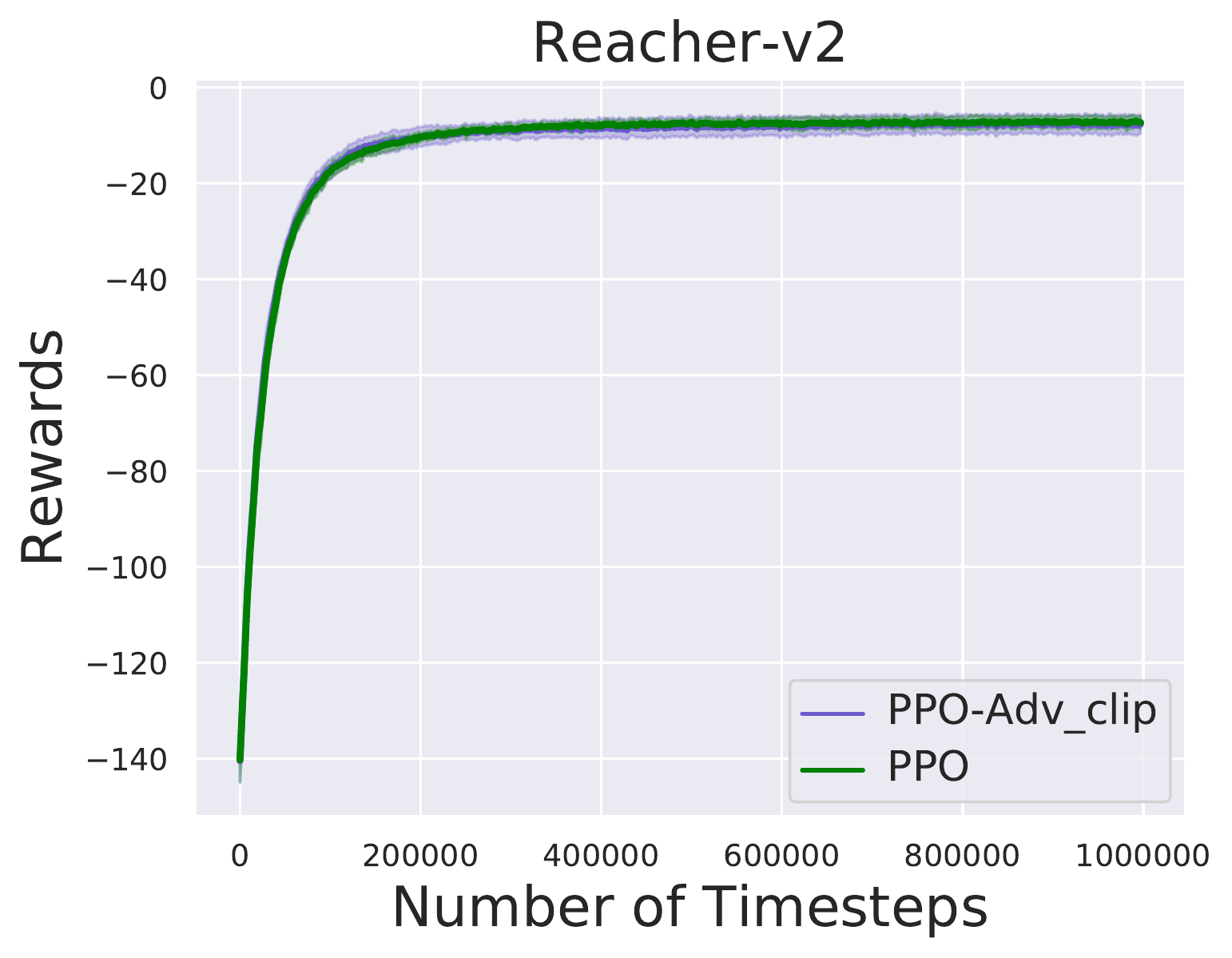}}\hfil

    \caption{  \update{\textbf{Reward curves with advantage clipping in 8 different Mujoco Environments} aggregated across 30 random seeds. The shaded region denotes the one standard deviation across seeds. The clipping threshold is tuned per environment. We observe that by clipping \emph{outlier} advantages, we substantially improve the mean rewards for 5 environments. While for the remaining three environments, we didn't observe any differences in the agent performance.}
    }\label{figure:adv-clip-rewards} 
\end{figure}

\subsection{\update{Effect of heavy-tailedness in likelihood-ratios}}
\label{subsec:App_ratios}

\begin{figure}[H] 
    \centering
        \subfigure{ \includegraphics[width=0.24\linewidth]{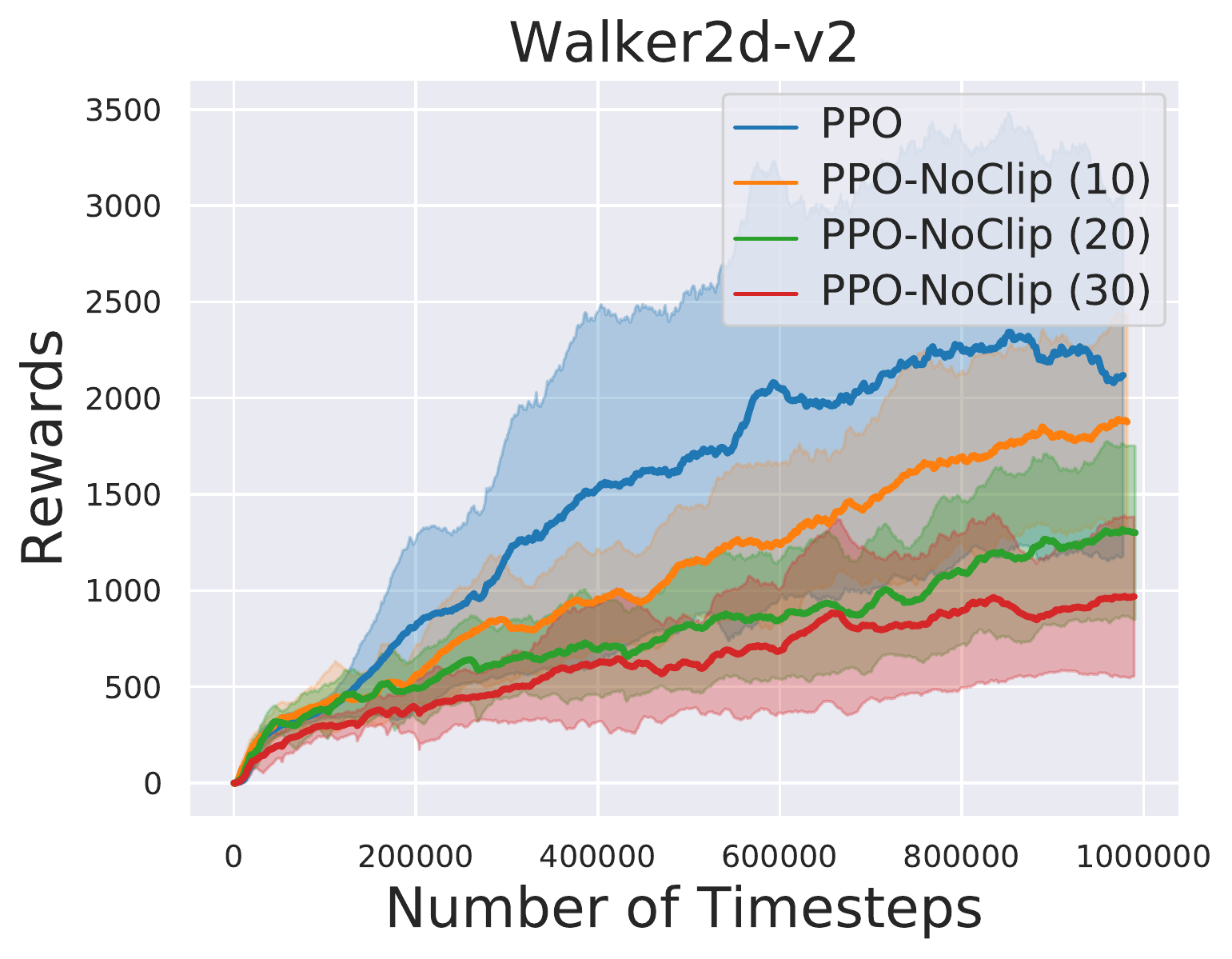}}\hfil
        \subfigure{ \includegraphics[width=0.24\linewidth]{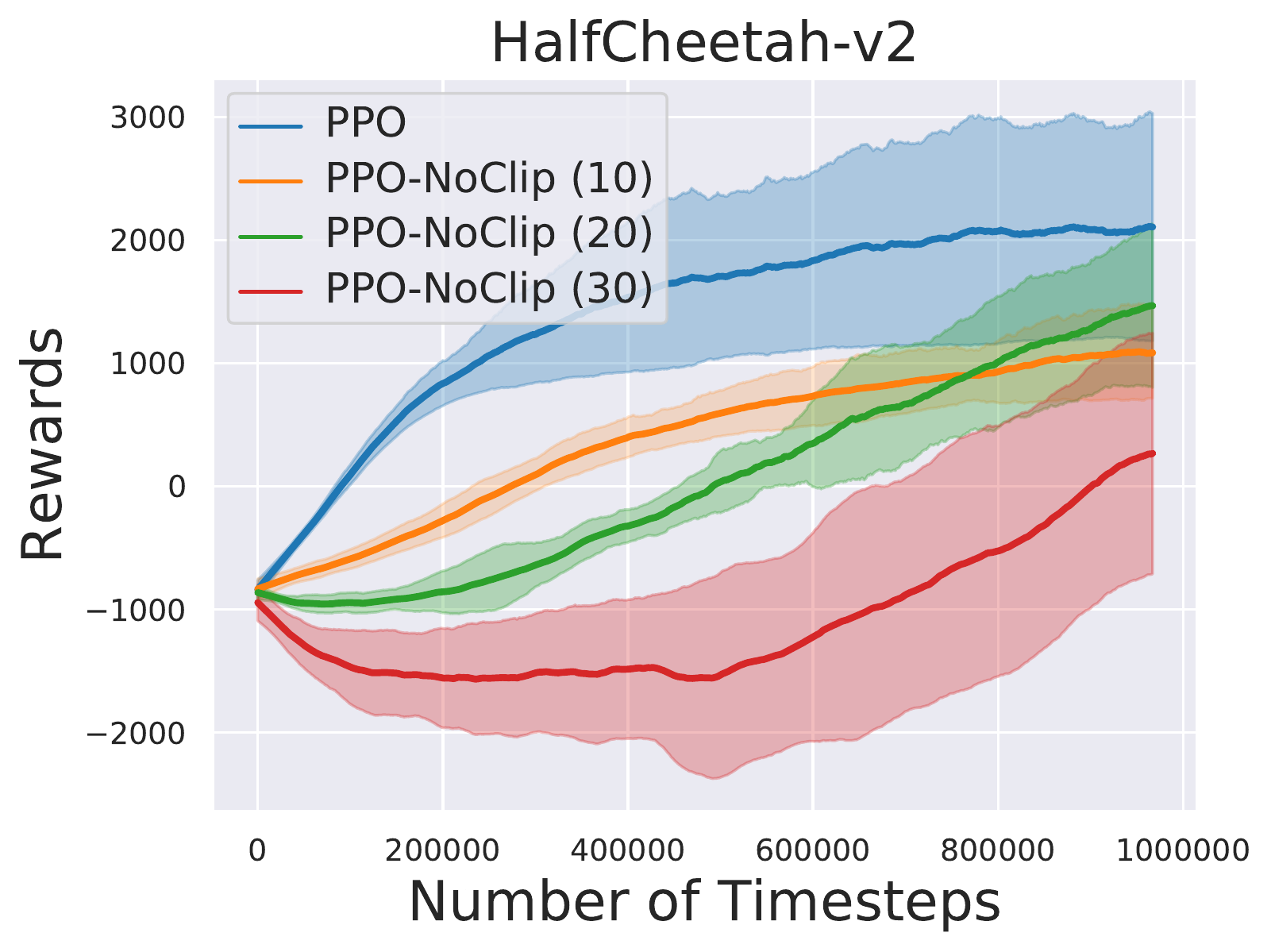}}\hfil
        \subfigure{\includegraphics[width=0.24\linewidth]{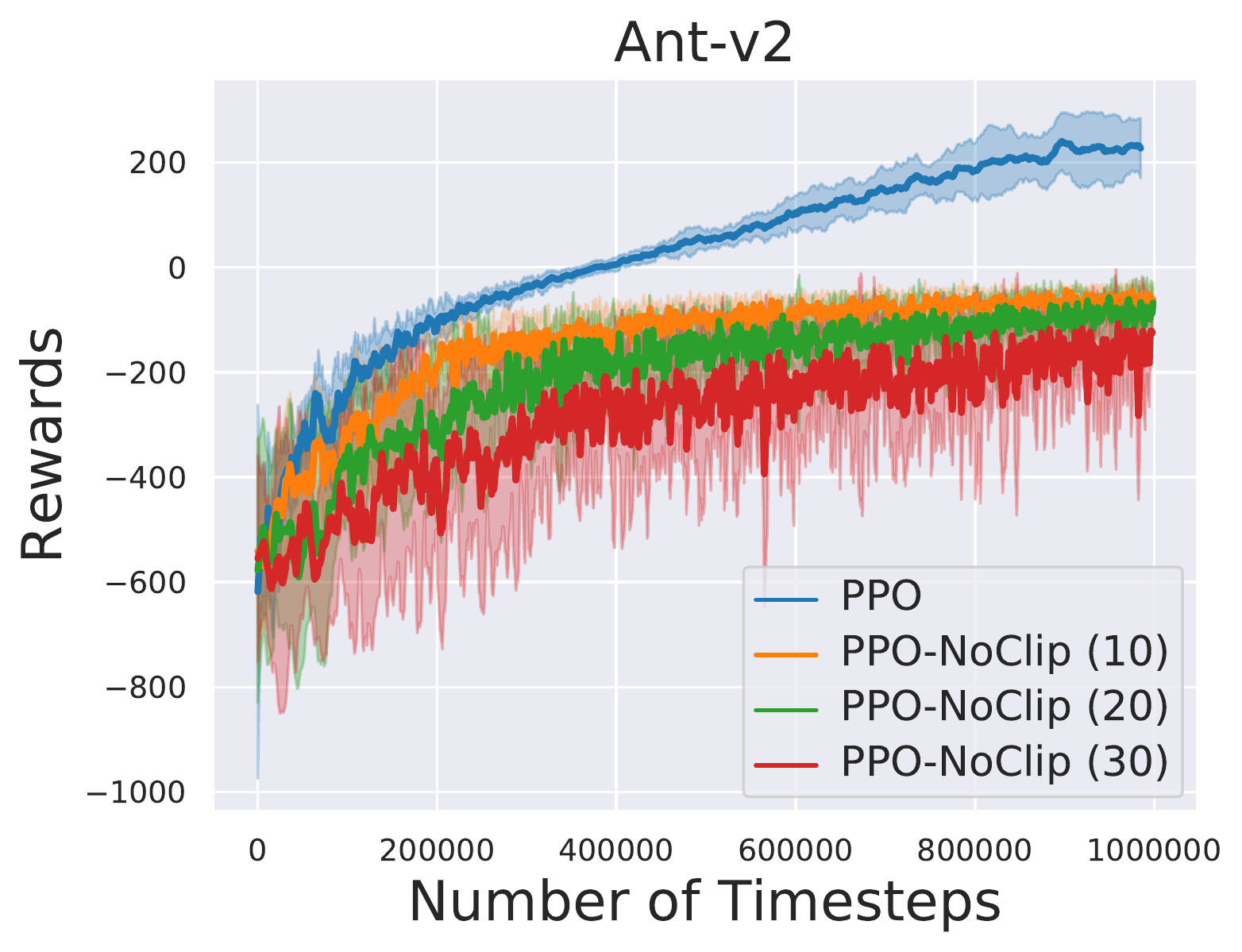}}\hfil
        \subfigure{ \includegraphics[width=0.24\linewidth]{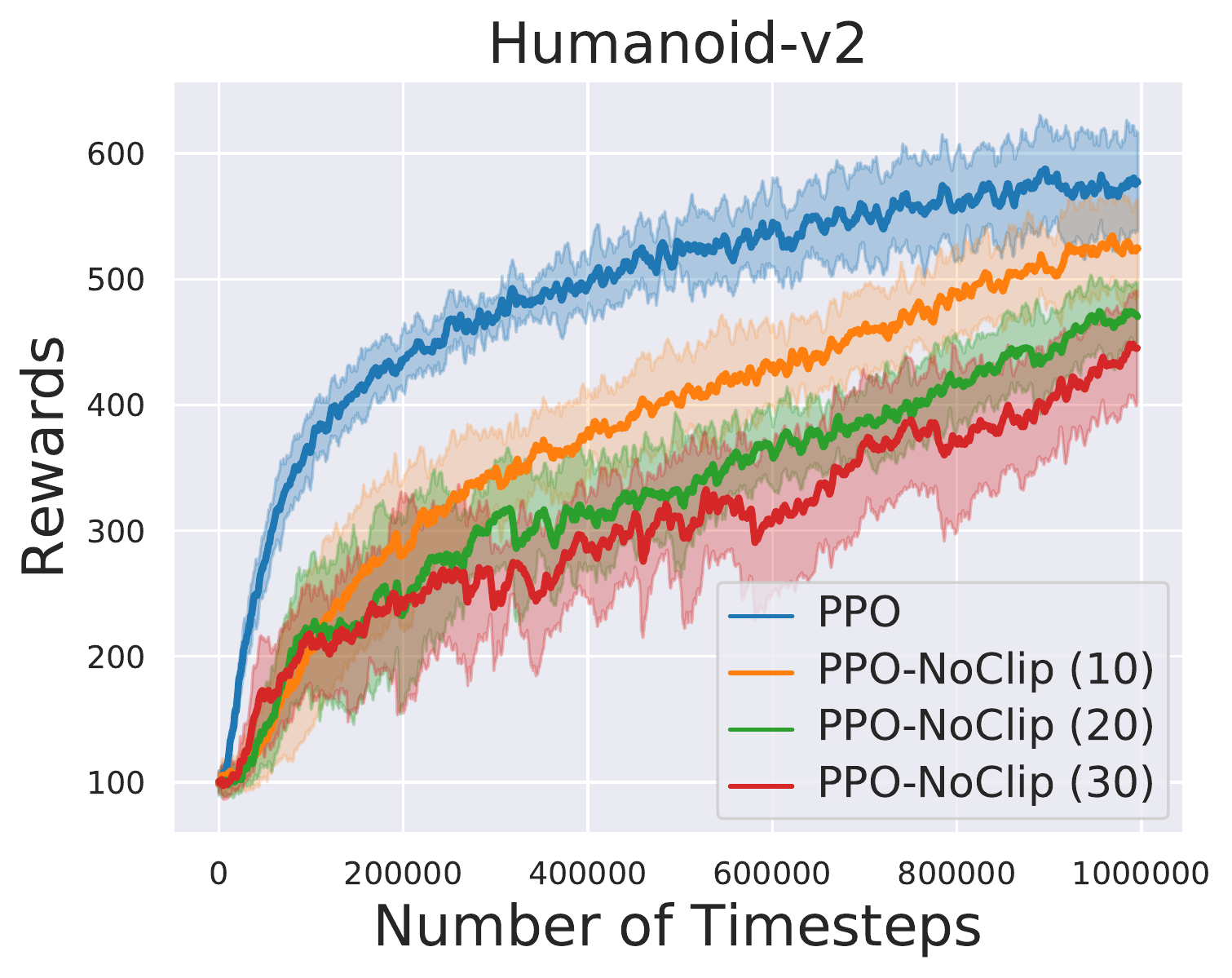}}\hfil
        \par\medskip

        \subfigure{\includegraphics[width=0.24\linewidth]{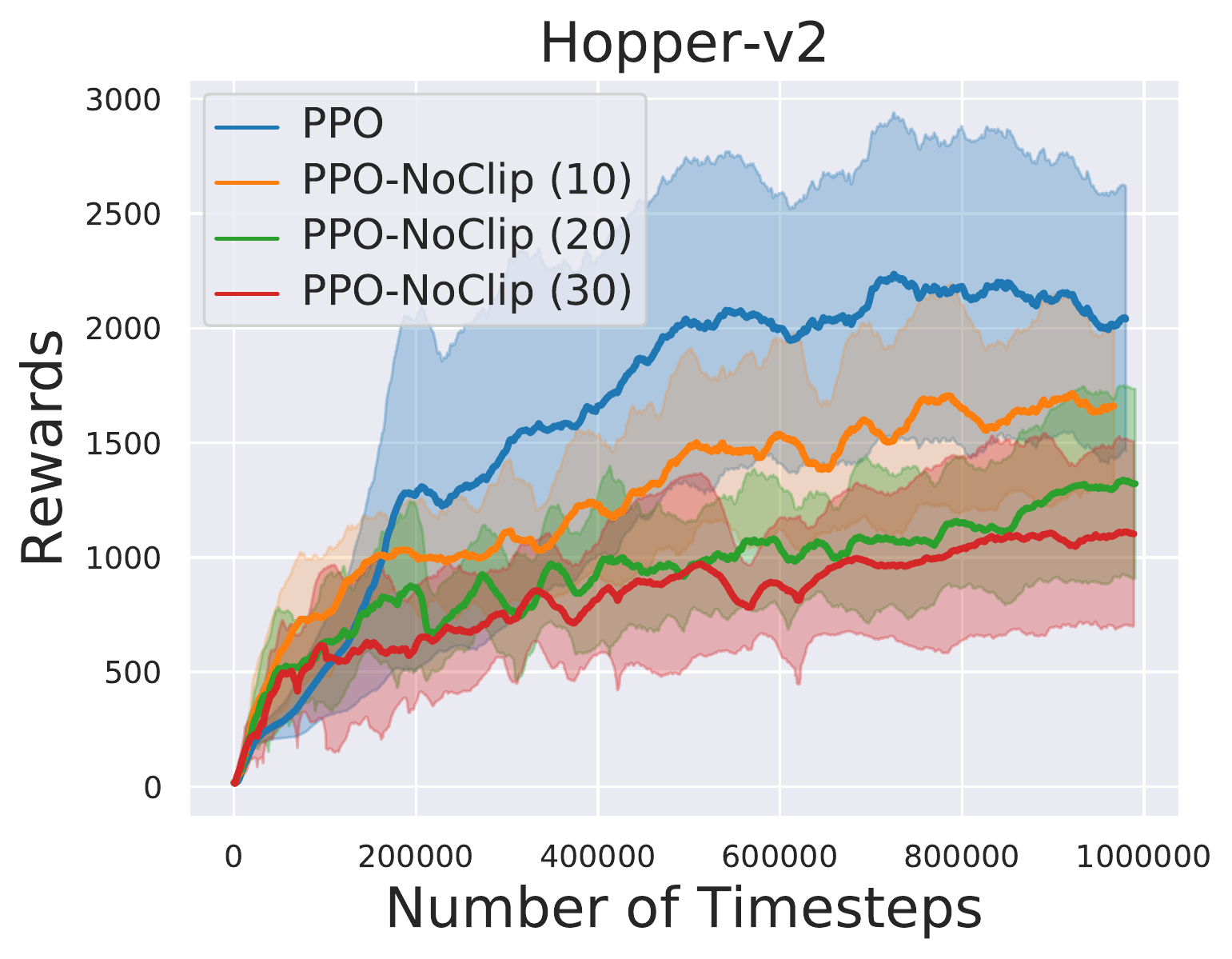}}\hfil        
        \subfigure{ \includegraphics[width=0.24\linewidth]{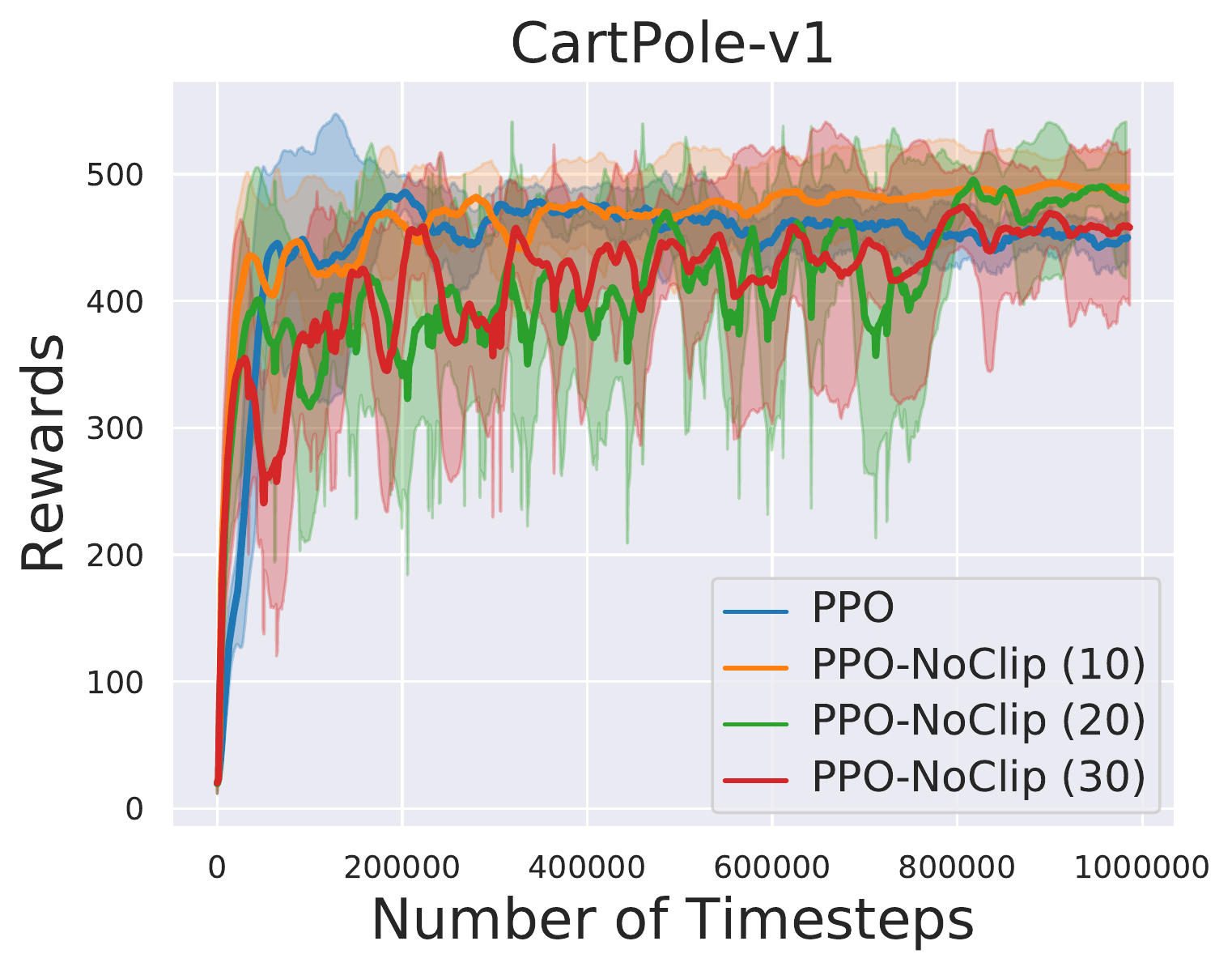}}\hfil
        \subfigure{ \includegraphics[width=0.24\linewidth]{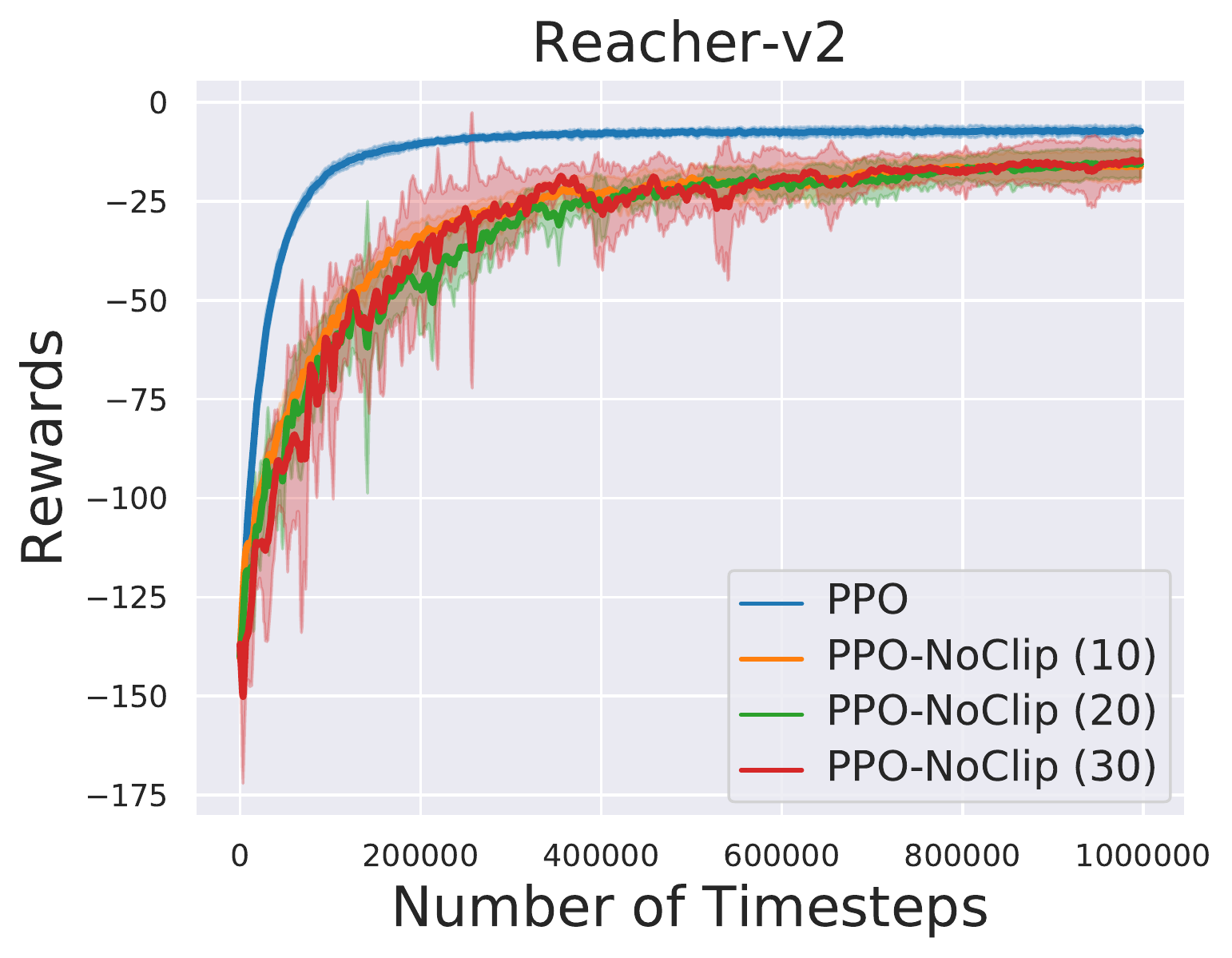}}\hfil
        \subfigure{ \includegraphics[width=0.24\linewidth]{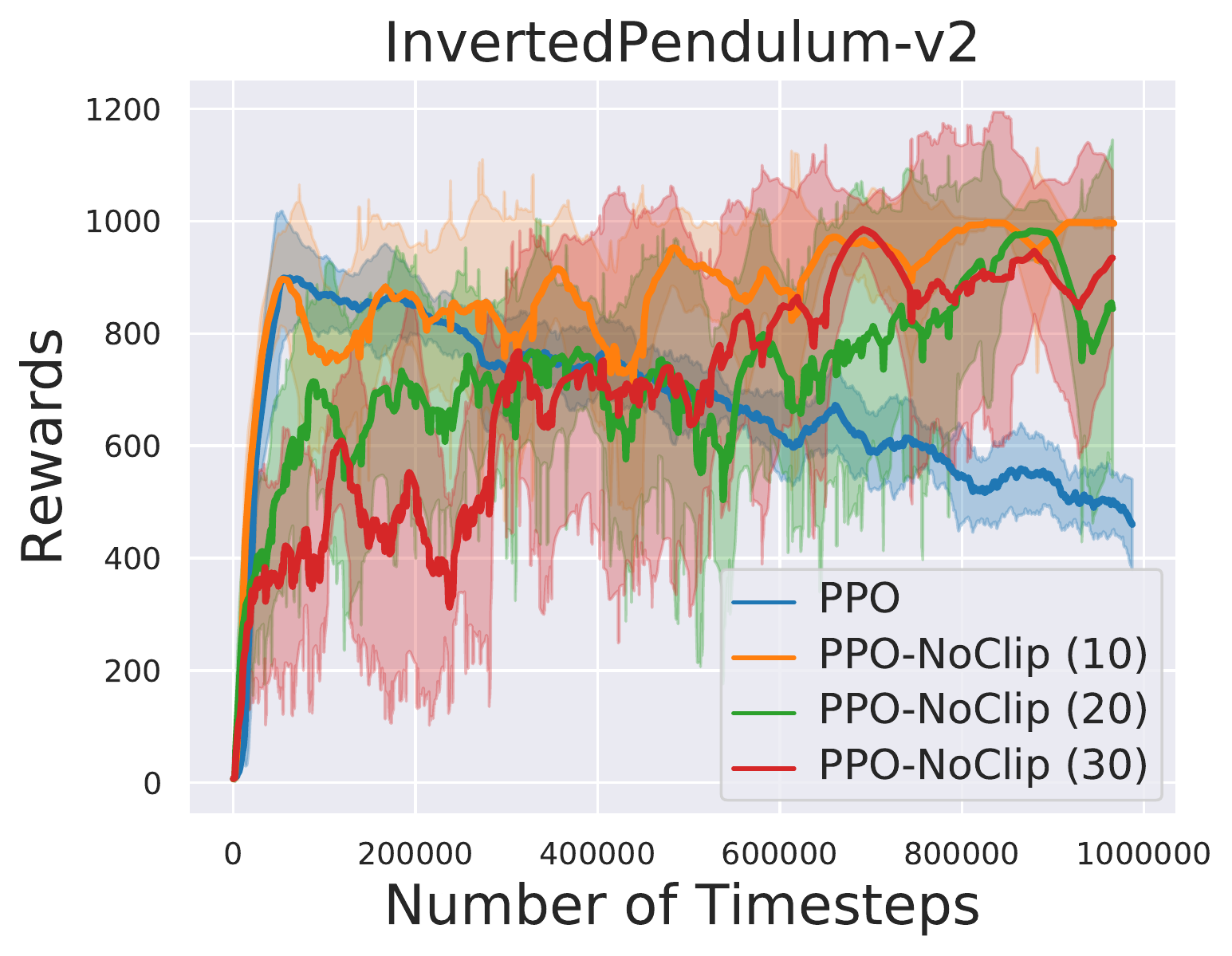}}\hfil
        \par\medskip

        \subfigure{\includegraphics[width=0.24\linewidth]{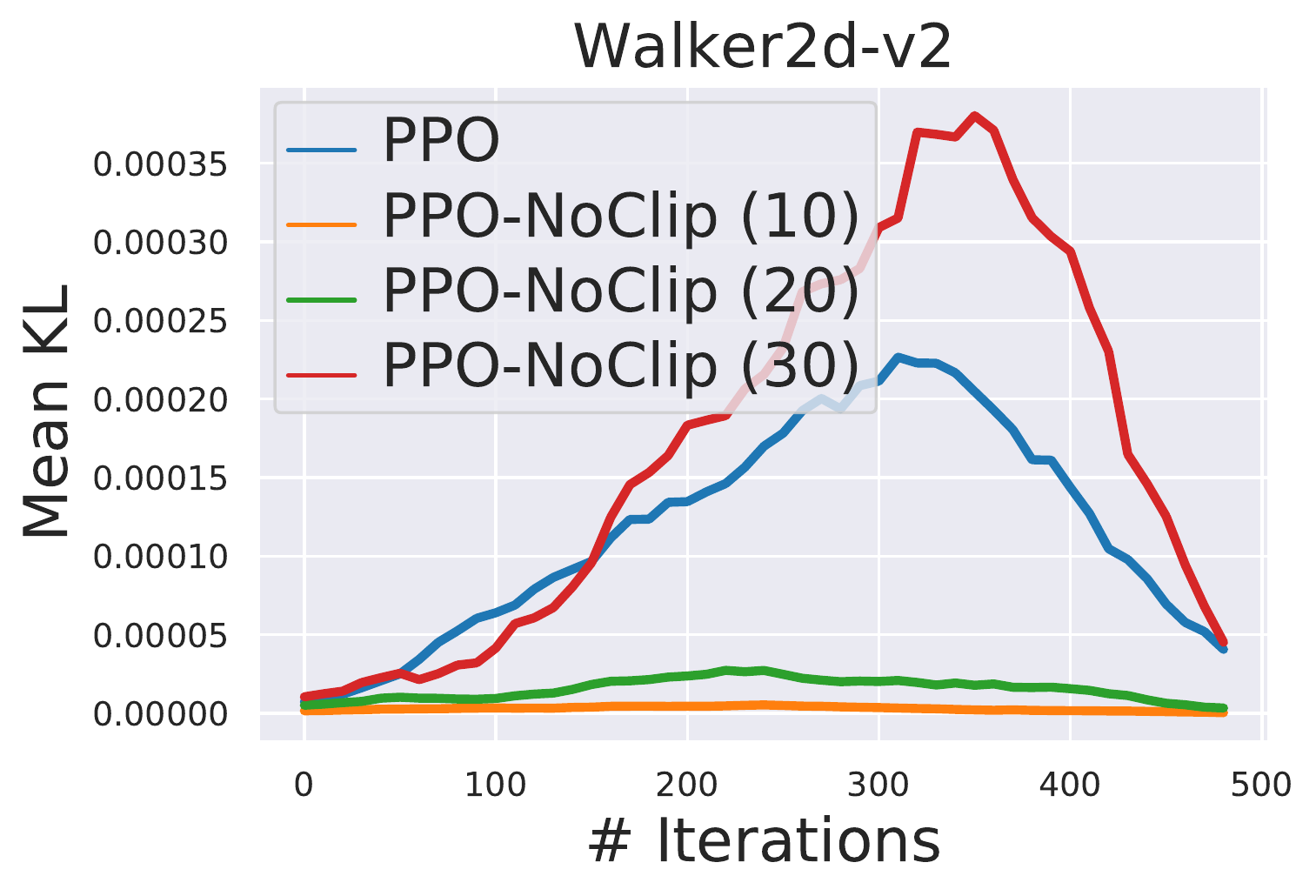}}\hfil
        \subfigure{ \includegraphics[width=0.24\linewidth]{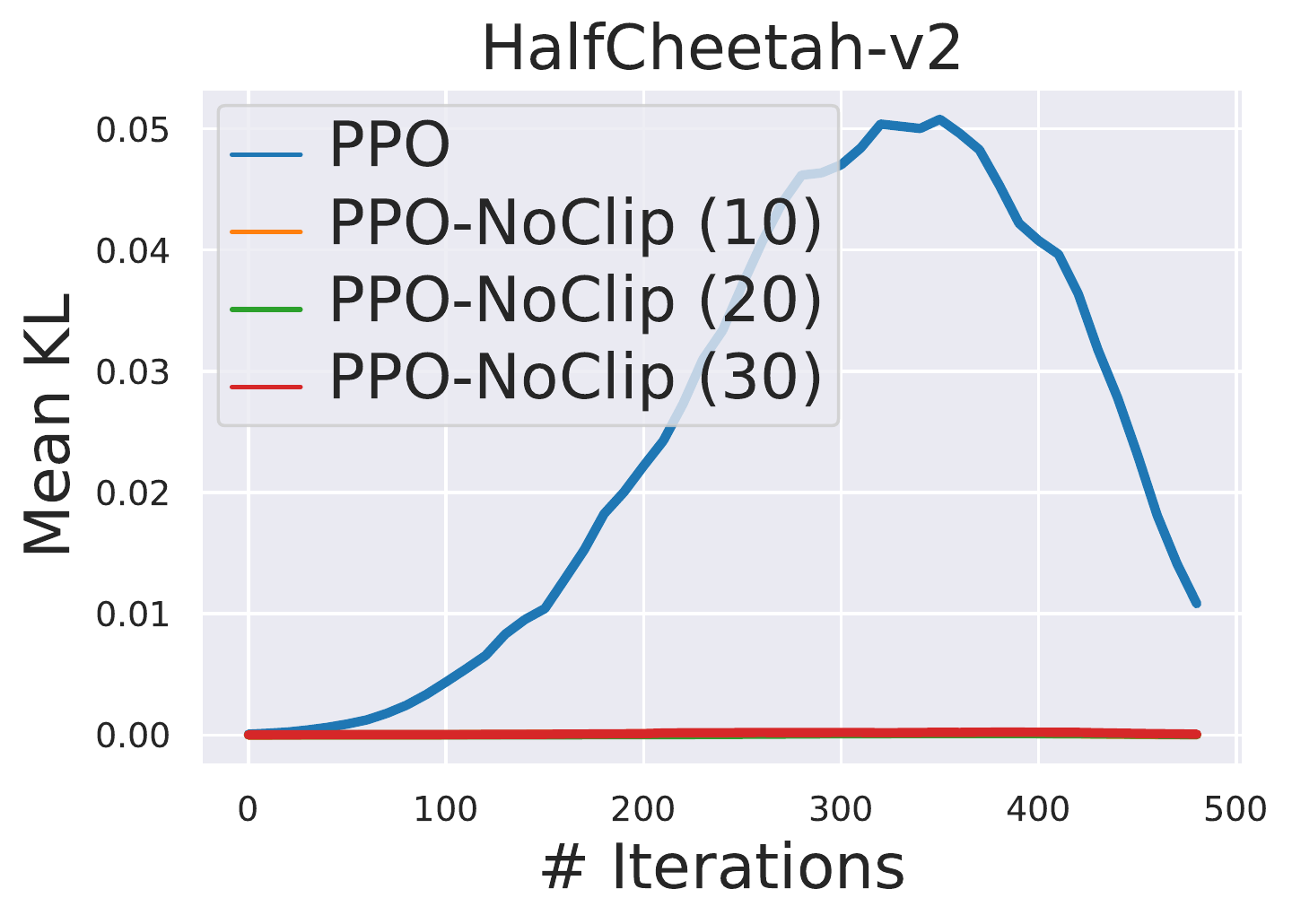}}\hfil
        \subfigure{\includegraphics[width=0.24\linewidth]{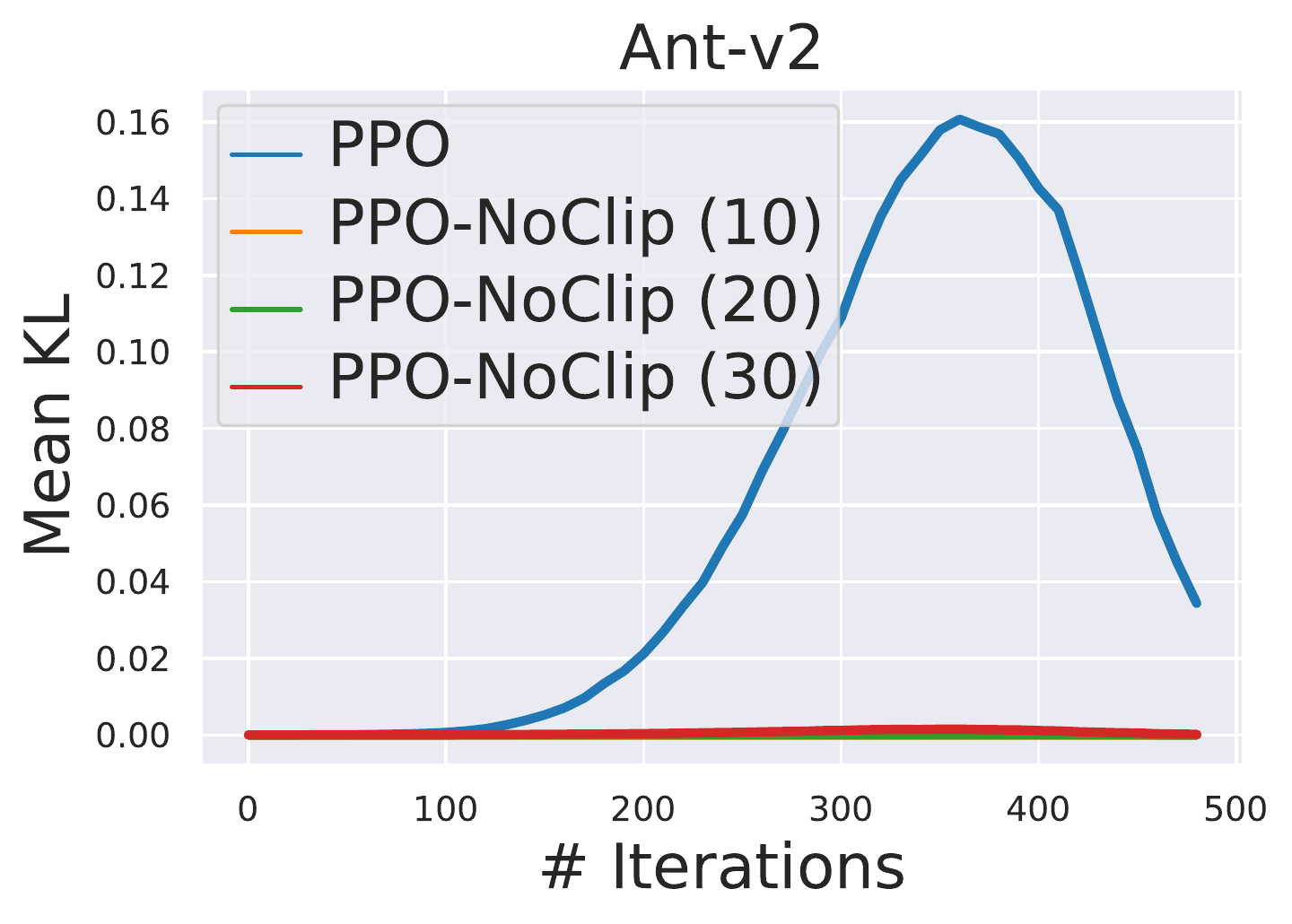}}\hfil
        \subfigure{\includegraphics[width=0.24\linewidth]{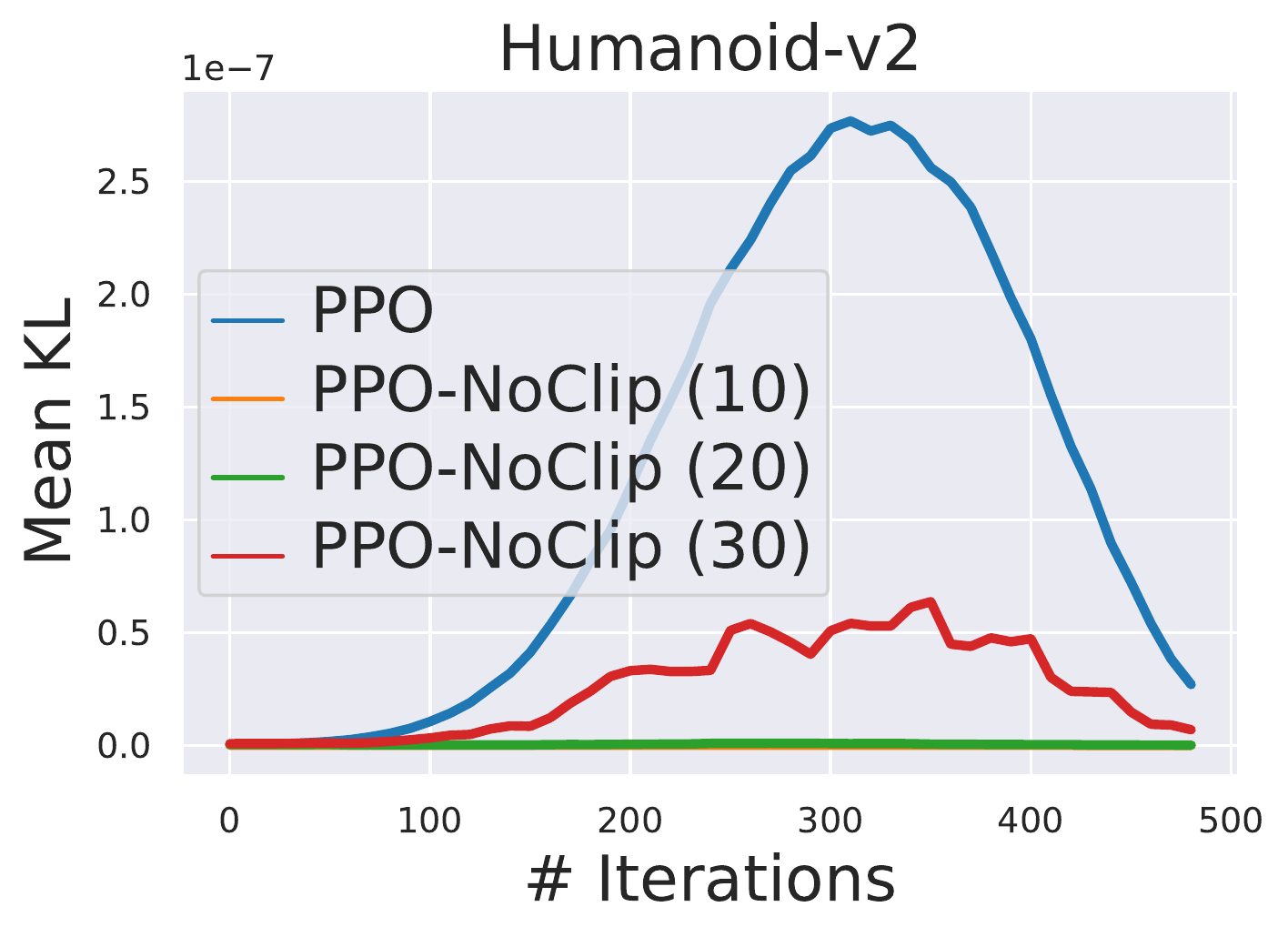}}\hfil
        \par\medskip
        \subfigure{\includegraphics[width=0.24\linewidth]{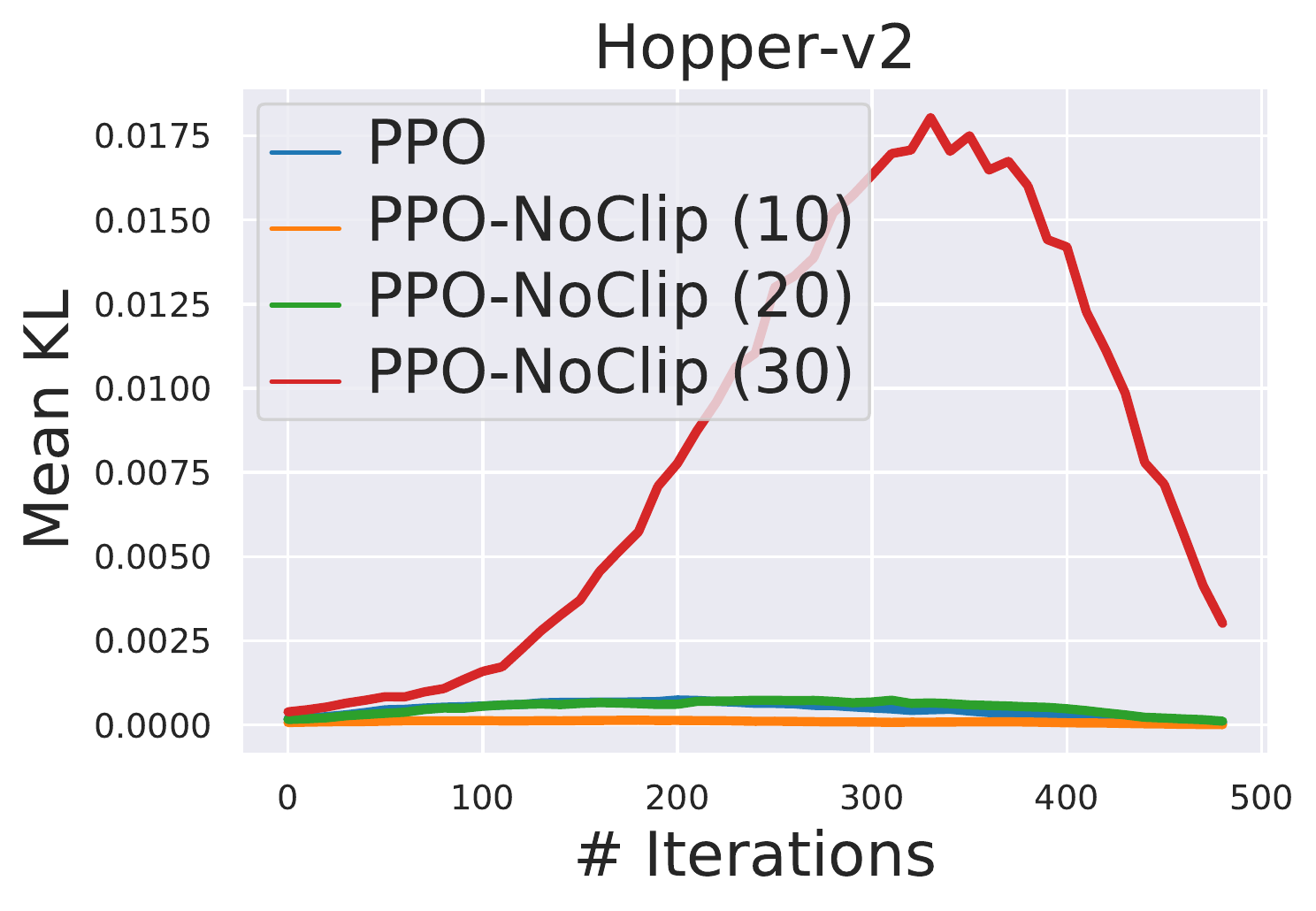}}\hfil     
        \subfigure{\includegraphics[width=0.24\linewidth]{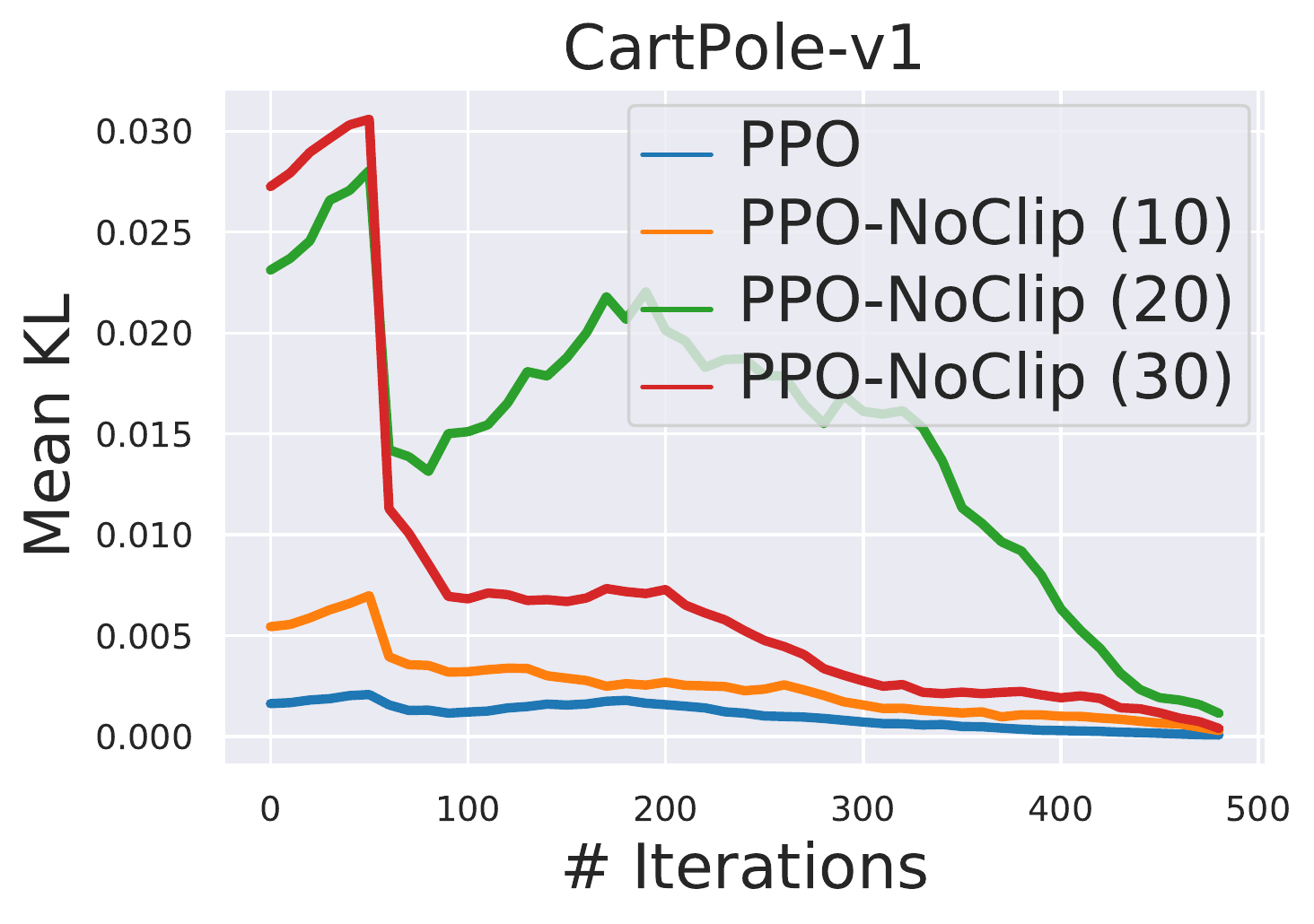}}\hfil
        \subfigure{ \includegraphics[width=0.24\linewidth]{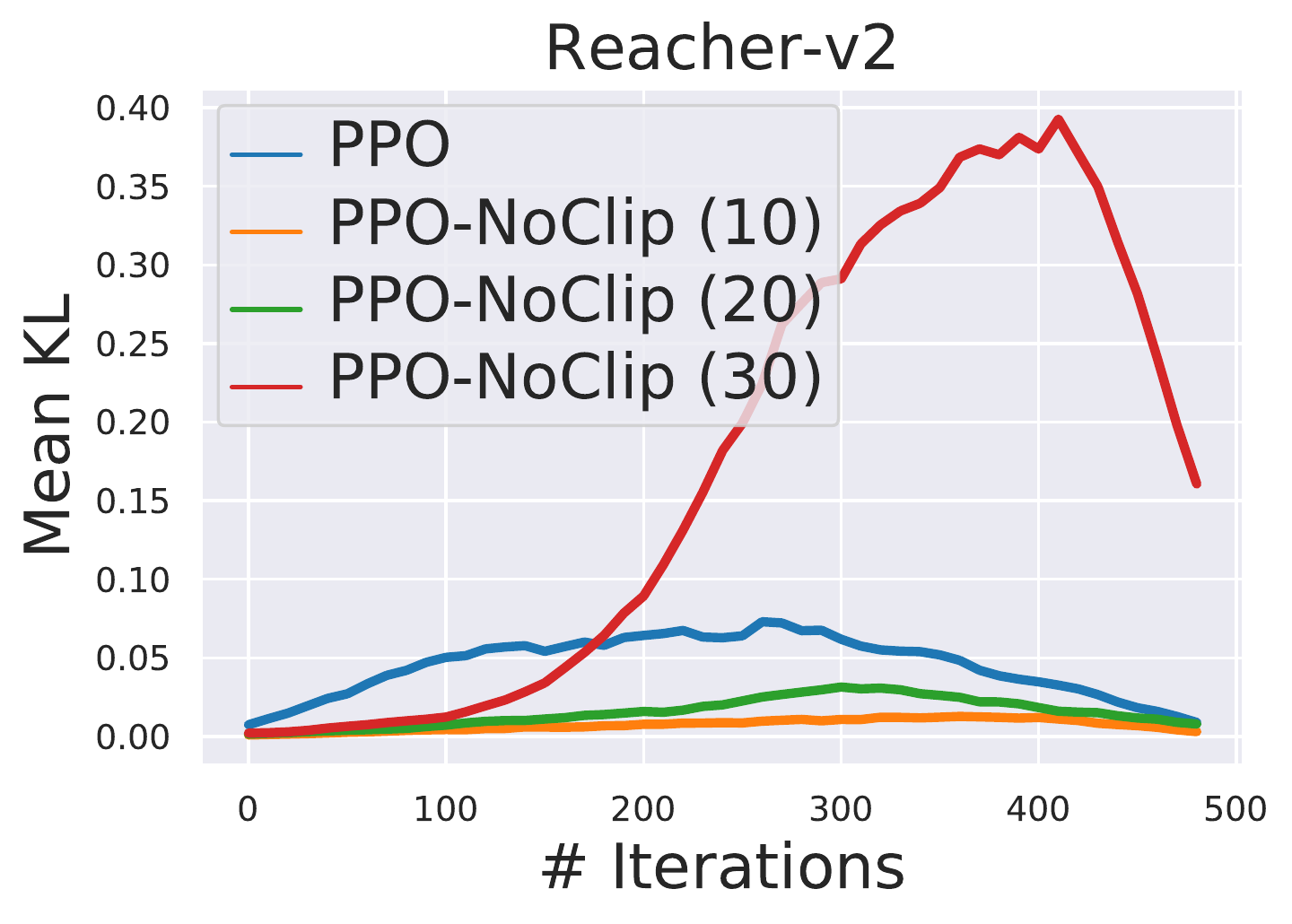}}\hfil
        \subfigure{\includegraphics[width=0.24\linewidth]{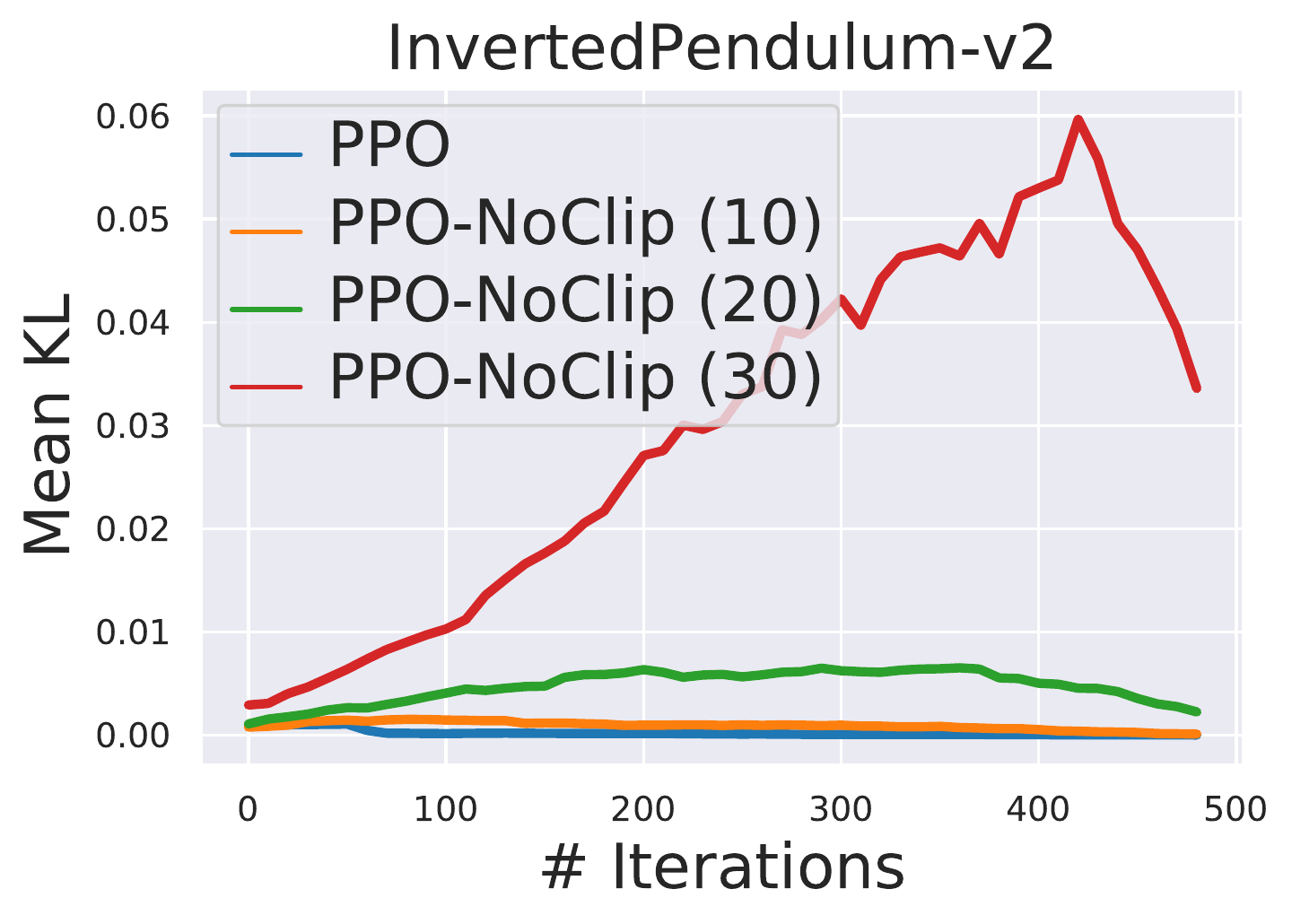}}\hfil
        \par\medskip
    \caption{ \update{\textbf{(Top two rows)} Reward curves with the varying number of offline epochs in 8 different Mujoco Environments aggregated across 10 random seeds. Bracketed quantity in the legend denotes the number of offline epochs used for PPO-\textsc{NoClip} training. Clearly, as the number of offline epochs increases, the performance of the agent drops (consistent behavior across all environments). Furthermore, at $30$ epochs the training also gets unstable. We also show the PPO performance curve for comparison. \textbf{(Bottom two rows)} KL divergence between current and previous policies with the optimal hyperparameters (parameters in Table~\ref{table:hyperparameters}) for PPO and PPO-\textsc{NoClip} with varying number of offline epochs. We measure mean empirical KL divergence between the policy obtained at the end of off-policy training and the sampling policy at the beginning of every training iteration. The quantities are measured over the state-action pairs collected in the training step. We observe that till 30 offline epochs PPO-\textsc{NoClip} maintains a trust-region with mean KL metric.} }\label{fig:offpolicy-kl-reward} 
     
\end{figure}
\newpage

\begin{figure}[H]
        \centering 
        \includegraphics[width=0.35\textwidth]{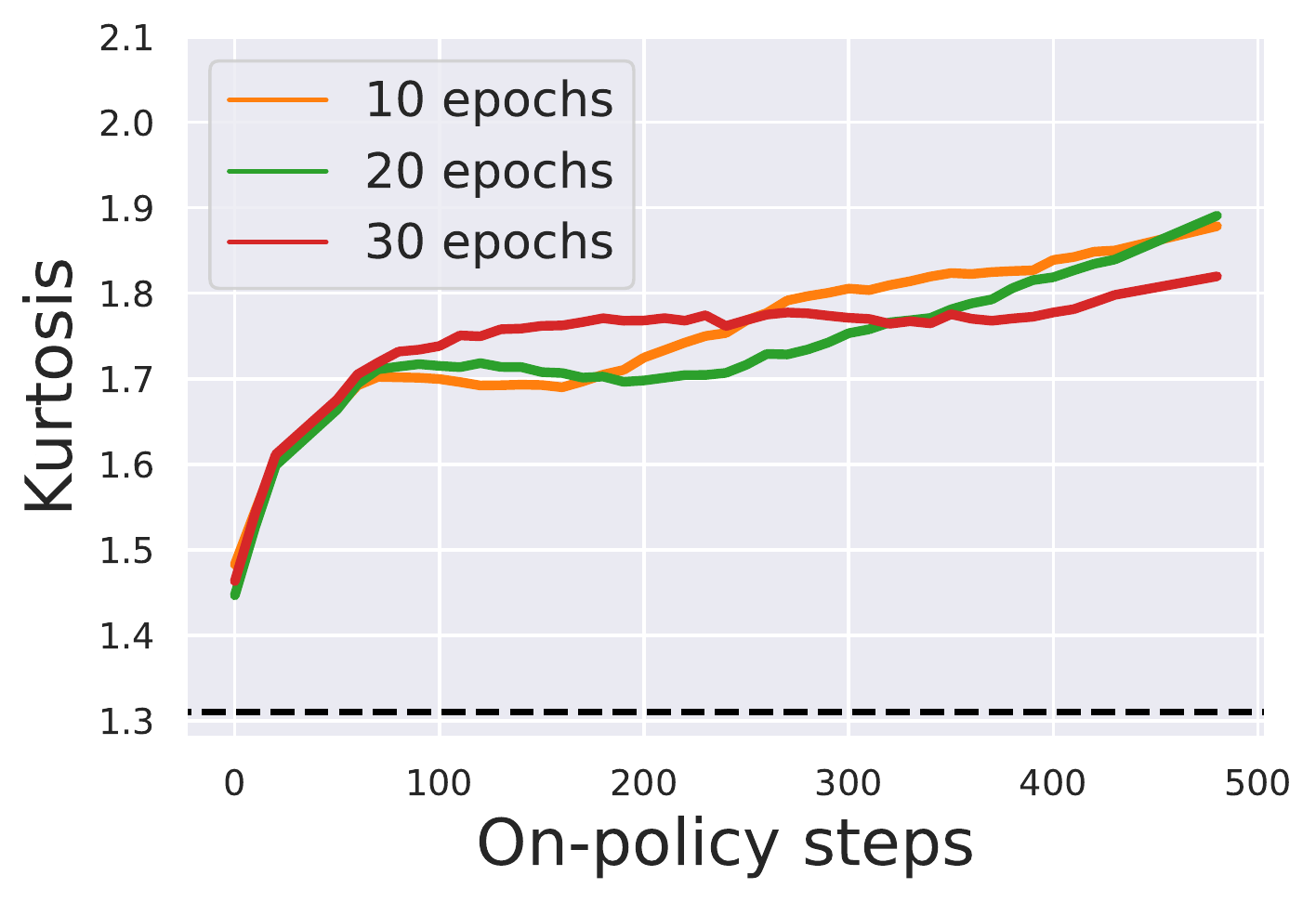} 
        \caption{\update{ \textbf{Heavy-tailedness in PPO-\textsc{NoClip} advantages throughout the training as the degree of off-policyness is varied} in MuJoCo environments. Kurtosis is aggregated over 8 Mujoco environments. We plot kurtosis vs on-policy iterates. As the number of off-policy epochs increases, the heavy-tailedness in advantages remains the same showing an increase in the number of offline epochs has a minor effect on the induced heavy-tailedness in the advantage estimates.} }\label{fig:offpolicy-traininig-adv}
\end{figure}

\end{document}